\documentclass[5p,times,twocolumn]{elsarticle}
\usepackage{graphicx}
\usepackage{soul}
\usepackage{float}
\usepackage{multirow}
\usepackage{booktabs}
\usepackage{amssymb}
\usepackage{amsmath}
\usepackage{longtable}
\usepackage{subfig}
\usepackage{multirow}
\usepackage{comment}
\usepackage{adjustbox}
\usepackage{csquotes}
\usepackage[T1]{fontenc}
\usepackage[table]{xcolor}
\usepackage[pagebackref=true,breaklinks=true,colorlinks,bookmarks=false]{hyperref}
\usepackage{titlesec}

\usepackage[normalem]{ulem}
\titleformat{\paragraph}[runin]{\normalfont\normalsize\bfseries}{\theparagraph}{1em}{}[:]
\titlespacing*{\paragraph}{0pt}{\baselineskip}{\baselineskip}

\begin{document}
\begin{frontmatter}
\title{A Comprehensive Overview of Large Language Models}

\author[1]{Humza Naveed}
\ead{humza_naveed@yahoo.com}

\author[2]{Asad Ullah Khan\corref{cor1}}
\ead{aukhanee@gmail.com}

\author[3]{Shi Qiu\corref{cor1}}
\ead{shiqiu@cse.cuhk.edu.hk}

\author[4,5]{Muhammad Saqib\corref{cor1}}
\ead{muhammad.saqib@data61.csiro.au}

\author[6,7]{Saeed Anwar}
\ead{saeed.anwar@kfupm.edu.sa}

\author[6,7]{Muhammad Usman}
\ead{muhammad.usman@kfupm.edu.sa}

\author[8,10]{Naveed Akhtar}
\ead{naveed.akhtar1@unimelb.edu.au}

\author[9]{Nick Barnes}
\ead{nick.barnes@anu.edu.au}

\author[10]{Ajmal Mian}
\ead{ajmal.mian@uwa.edu.au}

\cortext[cor1]{Equal contribution}

\address[1]{The University of Sydney, Sydney, Australia}
\address[2]{University of Engineering and Technology (UET), Lahore, Pakistan}
\address[3]{The Chinese University of Hong Kong (CUHK), HKSAR, China}
\address[4]{University of Technology Sydney (UTS), Sydney, Australia}
\address[5]{Commonwealth Scientific and Industrial Research Organisation (CSIRO), Sydney, Australia}
\address[6]{King Fahd University of Petroleum and Minerals (KFUPM), Dhahran, Saudi Arabia}
\address[7]{SDAIA-KFUPM Joint Research Center for Artificial Intelligence (JRCAI), Dhahran, Saudi Arabia}
\address[8]{The University of Melbourne (UoM), Melbourne, Australia}
\address[9]{Australian National University (ANU), Canberra, Australia}
\address[10]{The University of Western Australia (UWA), Perth, Australia}

\begin{abstract}
% Large Language Models (LLMs) have shown excellent generalization abilities that lead to the development of numerous models. These models suggest different approaches, for example, increasing model parameters, training data, tweaking architecture, training strategies, and pipelines to outperform baselines. Analyzing these aspects is important to identify changes that bring stability to the training and better generalization for the LLMs. In this paper, we summarize these fine-grained details of LLMs. Initially, we provide a background for the LLMs to discuss basic building blocks followed by a comprehensive overview of LLMs. The end of the paper discusses important findings by the LLMs and summarizes important architectural and training strategies to develop better LLMs. Because of the continuous development in LLMs, we will update this paper regularly with new sections and LLMs.  

Large Language Models (LLMs) have recently demonstrated remarkable capabilities in natural language processing tasks and beyond. This success of LLMs has led to a large influx of research contributions in this direction. These works encompass diverse topics such as architectural innovations, better training strategies, context length improvements, fine-tuning, multi-modal LLMs, robotics, datasets, benchmarking, efficiency, and more. With the rapid development of techniques and regular breakthroughs in LLM research, it has become considerably challenging to perceive the bigger picture of the advances in this direction. Considering the rapidly emerging plethora of literature on LLMs, it is imperative that the research community is able to benefit from a concise yet comprehensive overview of the recent developments in this field. This article provides an overview of the literature on a broad range of LLM-related concepts. Our self-contained comprehensive overview of LLMs discusses relevant background concepts along with covering the advanced topics at the frontier of research in LLMs. This review article is intended to provide not only a systematic survey but also a quick, comprehensive reference for the researchers and practitioners to draw insights from extensive, informative summaries of the existing works to advance the LLM research. 

\end{abstract}
\end{frontmatter}

\begin{figure}[tbp]
\centering
\hspace{-5mm}
\includegraphics[width=1\columnwidth]{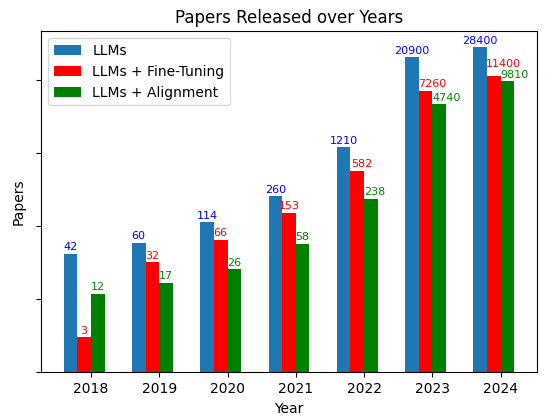}
\caption{The trend of papers released over the years containing
keywords "Large Language Model", "Large Language Model
+ Fine-Tuning", and "Large Language Model + Alignment".}
\label{fig:num_LLMs_barchart}
\end{figure}

\begin{figure*}[t]
\vspace{-1.5cm}
\hspace{-1cm}
\centering
\includegraphics[width=2\columnwidth]{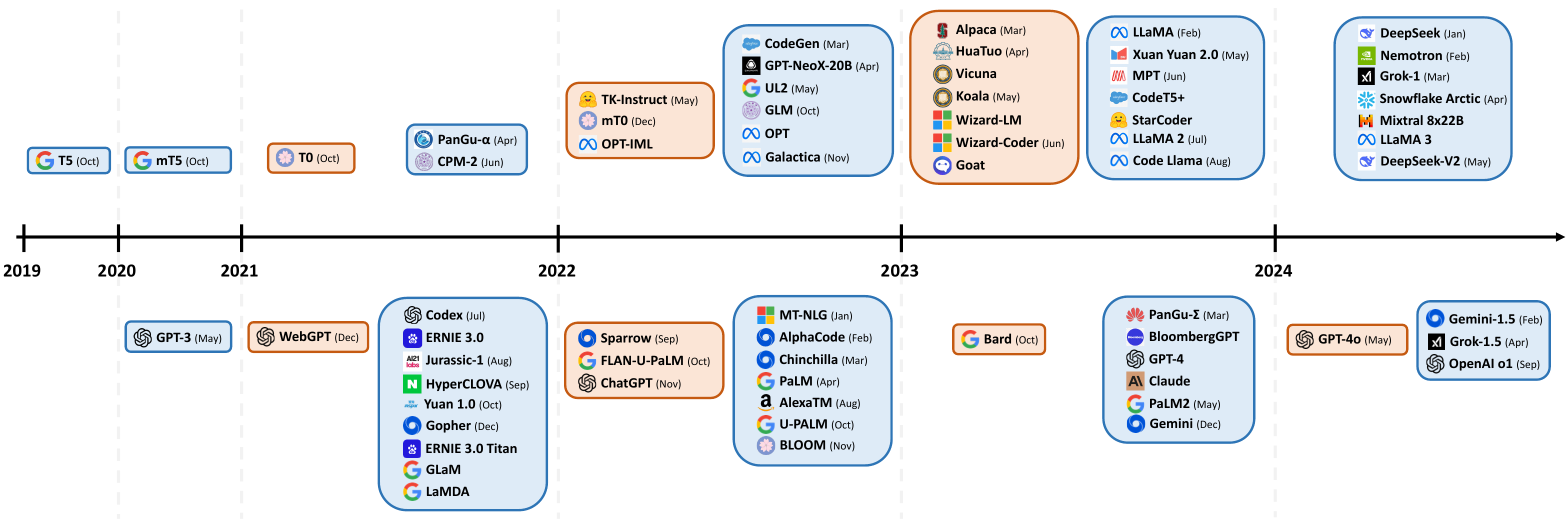}
\caption{Chronological display of LLM releases: blue cards represent `pre-trained' models, while orange cards correspond to `instruction-tuned' models. Models on the upper half signify open-source availability, whereas those on the bottom are closed-source. The chart illustrates the increasing trend towards instruction-tuned and open-source models, highlighting the evolving landscape and trends in natural language processing research.}
%\caption{LLMs introduced over the years.}
\label{fig:LLMs_bubblechart}
\end{figure*}

\section{Introduction}
Language plays a fundamental role in facilitating communication and self-expression for humans and their interaction with machines. The need for generalized models stems from the growing demand for machines to handle complex language tasks, including translation, summarization, information retrieval, conversational interactions, etc. 
Recently, significant breakthroughs have been witnessed in language models, primarily attributed to transformers~\cite{chernyavskiy2021transformers}, increased computational capabilities, and the availability of large-scale training data. These developments have brought about a revolutionary transformation by enabling the creation of LLMs that can approximate human-level performance on various tasks~\cite{wang2019superglue,adiwardana2020towards}. Large Language Models (LLMs) have emerged as cutting-edge artificial intelligence systems that can process and generate text with coherent communication~\cite{y2022large} and generalize to multiple tasks~\cite{GPT-2,GPT-3}.  \\
The historical progress in natural language processing (NLP) evolved from statistical to neural language modeling and then from pre-trained language models (PLMs) to LLMs. While conventional language modeling (LM) trains task-specific models in supervised settings, PLMs are trained in a self-supervised setting on a large corpus of text~\cite{Bert, ELMO, BART} with the aim of learning a generic representation that is shareable among various NLP tasks. After fine-tuning for downstream tasks, PLMs surpass the performance gains of traditional language modeling (LM). The larger PLMs bring more performance gains, which has led to the transitioning of PLMs to LLMs by significantly increasing model parameters (tens to hundreds of billions)~\cite{T5} and training dataset (many GBs and TBs)~\cite{T5,mT5}. Following this development, numerous LLMs have been proposed in the literature~\cite{T5, mT5, CPM-2, GPT-3, BLOOM, OPT, PaLM}. An increasing trend in the number of released LLMs and names of a few significant LLMs proposed over the years are shown in Fig~\ref{fig:num_LLMs_barchart} and Fig~\ref{fig:LLMs_bubblechart}, respectively.\\
The early work on LLMs, such as T5~\cite{T5} and mT5~\cite{mT5} employed transfer learning until GPT-3~\cite{GPT-3} showed LLMs are zero-shot transferable to downstream tasks without fine-tuning. LLMs accurately respond to task queries when prompted with task descriptions and examples. However, pre-trained LLMs fail to follow user intent and perform worse in zero-shot settings than in few-shot. Fine-tuning them with task instructions data~\cite{Flan,T0, Tk-INSTRUCT,ft_self_instruct} and aligning with human preferences~\cite{instructgpt,llama_2} enhances generalization to unseen tasks, improving zero-shot performance significantly and reducing misaligned behavior.   \\ 
In addition to better generalization and domain adaptation, LLMs appear to have emergent abilities, such as reasoning, planning, decision-making, in-context learning, answering in zero-shot settings, etc. These abilities are known to be acquired by them due to their gigantic scale even when the pre-trained LLMs are not trained specifically to possess these attributes~\cite{emergent_abilities, emergent_analogical, emergent_research}. Such abilities have led LLMs to be widely adopted in diverse settings, including multi-modal, robotics, tool manipulation, question answering, autonomous agents, etc. Various improvements have also been suggested in these areas either by task-specific training~\cite{atlas,PaLME,talm,zhang2023large,ye2023mplug, wang2023visionllm, gpt4tools} or better prompting~\cite{prompt_eng_guide}. \\
\begin{figure*}[!t]
\centering
\includegraphics[width=1.8\columnwidth,height=2\columnwidth]{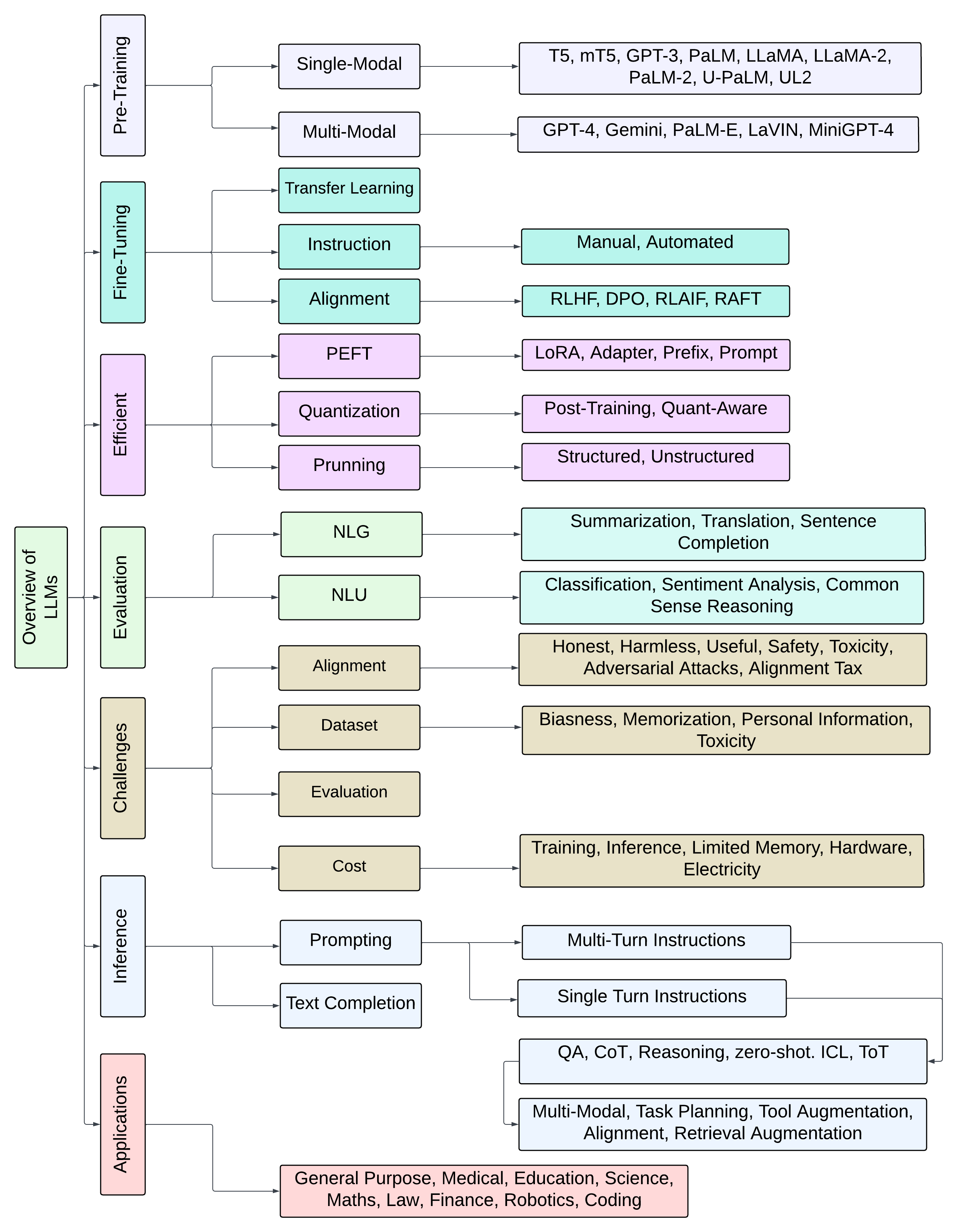}
\caption{A broader overview of LLMs, dividing LLMs into seven branches: 1. Pre-Training 2. Fine-Tuning 3. Efficient 4. Inference 5. Evaluation 6. Applications 7. Challenges}
%\caption{LLMs introduced over the years.}
\label{fig:Overview_LLMs}
\end{figure*}

\vspace{-1em}
\noindent
The LLMs abilities to solve diverse tasks with human-level performance come at the cost of slow training and inference, extensive hardware requirements, and higher running costs. Such requirements have limited their adoption and opened up opportunities to devise better architectures~\cite{PaLM,GLM-130B,codet5+,ernie3titan} and training strategies~\cite{Lib_DeepSpeed,ZeroOpt,llama_2,unified_peft,LMAdapter_3,Prompt_Tuning, Prefix_Tuning}. Parameter efficient tuning~\cite{unified_peft, Prefix_Tuning, Prompt_Tuning}, pruning~\cite{llm_pruner,contrastive_pruning}, quantization~\cite{SmoothQuant,compression_PLM_quant}, knowledge distillation, and context length interpolation~\cite{giraffe,yarn,longt5,pe_extending} among others are some of the methods widely studied for efficient LLM utilization. \\
\noindent
Due to the success of LLMs on a wide variety of tasks, the research literature has recently experienced a large influx of LLM-related contributions. Researchers have organized the LLMs literature in surveys~\cite{Survey_LLM,survey_smaller_LLMs_1,survey_smaller_LLMs_2,survey_1}, and topic-specific surveys in~\cite{survey_incontext_learning,survey_reasoning,survey_aligning,survey_compression, survey_multimodal_llm}.
In contrast to these surveys, our contribution focuses on providing a comprehensive yet concise overview of the general direction of LLM research. This article summarizes architectural and training details of pre-trained LLMs and delves deeper into the details of concepts like fine-tuning, multi-modal LLMs, augmented LLMs, datasets, evaluation, applications, challenges, and others to provide a self-contained comprehensive overview. Our key contributions are summarized as follows.

\begin{itemize}
\item We present a survey on the 
 developments in LLM research, providing a concise, comprehensive overview of the direction. 
\vspace{-0.25cm}
\item We present extensive summaries of pre-trained models that include fine-grained details of architecture and training details.
\vspace{-0.25cm}
\item We summarize major findings of the popular contributions and provide a detailed discussion on the key design and development aspects of LLMs to help practitioners effectively leverage this technology. 
\vspace{-0.25cm}
\item In this self-contained article, we cover a range of concepts to present the general direction of LLMs comprehensively, including background, pre-training, fine-tuning, multi-modal LLMs, augmented LLMs, LLMs-powered agents, datasets, evaluation, etc.
\end{itemize}
\noindent
We loosely follow the existing terminology to ensure a standardized outlook of this research direction. For instance, following  \cite{Survey_LLM}, our survey discusses pre-trained LLMs with 10B parameters or more. We refer the readers interested in smaller pre-trained models to~\cite{survey_smaller_LLMs_1, survey_smaller_LLMs_2, survey_1}.

\noindent
The organization of this paper is as follows. Section~\ref{sec:Background} discusses the background of LLMs. Section~\ref{sec_review} focuses on LLMs overview, architectures, training pipelines and strategies, fine-tuning, and utilization in different domains. Section~\ref{Model_Configurations_} highlights the configuration and parameters that play a crucial role in the functioning of these models. Summary and discussions are presented in section~\ref{sec:summary_and_discussion}. The LLM training and evaluation, datasets, and benchmarks are discussed in section~\ref{Datasets_and_Evaluation_}, followed by challenges and future directions, and conclusion in sections~\ref{sec:Challenges} and~\ref{sec:conclusion}, respectively.
\section{Background}
\label{sec:Background}
We provide the relevant background to understand the fundamentals related to LLMs in this section. We briefly discuss necessary components in LLMs and refer the readers interested in details to the original works.  

\subsection{Tokenization}
\label{ss:tokenization}
Tokenization~\cite{webster1992tokenization} is an essential pre-processing step in LLM training that parses the text into non-decomposing units called tokens. Tokens can be characters, subwords~\cite{unigramLM}, symbols~\cite{bpe}, or words, depending on the tokenization process. Some of the commonly used tokenization schemes in LLMs include wordpiece~\cite{wordpiece}, byte pair encoding (BPE)~\cite{bpe}, and unigramLM~\cite{unigramLM}. Readers are encouraged to refer to~\cite{tokenizationsurvey} for a detailed survey.

% \noindent
% \emph{\textbf{WordPiece~\cite{wordpiece}:}}
% It was introduced as a novel text segmentation technique for Japanese and Korean languages to improve the language model for voice search systems. WordPiece starts with the individual characters in the dataset and merge characters progressively to build a vocabulary that increases the likelihood of training data. \\
% \emph{\textbf{BPE~\cite{bpe}:}}
% Byte Pair Encoding (BPE) initially creates a set of unique words paired with frequency. Using these words, it builds a base vocabulary of characters merged iteratively to generate tokens based on the most frequent adjacent characters in the data until the vocabulary size. \\
% \emph{\textbf{UnigramLM~\cite{unigramLM}:}}
% In this tokenization, a simple unigram language model (LM) is trained using an initial vocabulary of \textit{subword} units. The vocabulary is pruned iteratively by removing the lowest probability items from the list, which are the worst performing on the unigram LM.

\subsection{Encoding Positions}
The transformer processes input sequences in parallel and independently of each other. Moreover, the attention module in the transformer does not capture positional information. As a result, positional encodings were introduced in transformer~\cite{Transformers}, where a positional embedding vector is added to the token embedding. Variants of positional embedding include absolute, relative, or learned positional encodings. Within relative encoding, Alibi and RoPE are two widely used positional embeddings in LLMs. \\
%Below we discuss variants of positional encodings proposed in the literature~\cite{alibi, su2021roformer}. \\
% \emph{\textbf{Absolute:}}
% The absolute positional encoding creates a unique positional embedding vector for each position which is added with the input embeddings. Now, the embeddings forwarded to the model contains position and token information.\\
% \emph{\textbf{Relative:}}
% To pass the information on the relative dependencies of different tokens appearing at different locations in the sequence, a relative positional encoding is calculated by some kind of learning. Two famous types of relative encodings used in LLMs are: \\
\emph{\textbf{Alibi~\cite{alibi}:}} It subtracts a scalar bias from the attention score that increases with the distance between token positions. This favors using recent tokens for attention. \\
\emph{\textbf{RoPE~\cite{su2021roformer}:}} It rotates query and key representations at an angle proportional to the token absolute position in the input sequence, resulting in a relative positional encoding scheme which decays with the distance between the tokens.

\subsection{Attention in LLMs}
\label{ss:llmattention}
Attention assigns weights to input tokens based on importance so that the model gives more emphasis to relevant tokens. Attention in transformers~\cite{Transformers} calculates query, key, and value mappings for input sequences, where the attention score is obtained by multiplying the query and key, and later used to weight values. We discuss different attention strategies used in LLMs below. \\
\emph{\textbf{Self-Attention~\cite{Transformers}:}}
Calculates attention using queries, keys, and values from the same block (encoder or decoder). \\ 
\emph{\textbf{Cross Attention:}}
It is used in encoder-decoder architectures, where encoder outputs are the queries, and key-value pairs come from the decoder.  \\
\emph{\textbf{Sparse Attention~\cite{sparse_transformer}:}}
Self-attention has $O(n^2)$ time complexity which becomes infeasible for large sequences. To speed up the computation, sparse attention~\cite{sparse_transformer} iteratively calculates attention in sliding windows for speed gains. \\
% The self-attention has a time complexity of $O(n^2)$, which becomes prohibitive when scaling the LLMs to large context windows. An approximation to the self-attention was proposed in~\cite{sparse_transformer}, which greatly enhanced the capacity of GPT series LLMs to process a greater number of input tokens in a reasonable time. \\
\emph{\textbf{Flash Attention~\cite{flashattention}:}}
Memory access is the major bottleneck in calculating attention using GPUs. To speed up, flash attention employs input tiling to minimize the memory reads and writes between the GPU high bandwidth memory (HBM) and the on-chip SRAM. \\
%The bottleneck for calculating the attention using GPUs lies in the memory access rather than the computational speed.
\subsection{Activation Functions}
\label{sec:activation functions}
The activation functions serve a crucial role in the curve-fitting abilities of neural networks~\cite{activationfunction}. We discuss activation functions used in LLMs in this section.  \\
\emph{\textbf{ReLU~\cite{relu}:}}
\label{ss:relu}
The Rectified linear unit (ReLU) is defined as:
\begin{equation}
ReLU(x) = max(0,x)    
\label{eq:relu}
\end{equation}
\noindent
\emph{\textbf{GeLU~\cite{gelu}:}}
\label{ss:gelu}
The Gaussian Error Linear Unit (GeLU) is the combination of ReLU,  dropout~\cite{srivastava2014dropout} and zoneout~\cite{krueger2016zoneout}. \\
\noindent
\emph{\textbf{GLU variants~\cite{shazeer2020glu}:}}
\label{ss:gluvariants}
The Gated Linear Unit~\cite{glu} is a neural network layer that is an element-wise product ($\otimes$) of a linear transformation and a sigmoid transformed ($\sigma$) linear projection of the input given as: 
\begin{equation}
GLU(x, W, V, b, c) = (xW + b) \otimes \sigma (xV + c),
\end{equation}
where $X$ is the input of layer and $l$, $W, b, V \textnormal{ and }c$ are learned parameters. Other GLU variants~\cite{shazeer2020glu} used in LLMs are: 
\begin{align*}
ReGLU(x, W, V, b, c) &= max(0, xW + b) \otimes , \\
GEGLU(x, W, V, b, c) &= GELU(xW + b) \otimes (xV + c), \\
SwiGLU(x, W, V, b, c, \beta) &= Swish\beta (xW + b) \otimes (xV + c).          
\end{align*}

\subsection{Layer Normalization}
\label{sec:layernormalization}
Layer normalization leads to faster convergence and is an integrated component of transformers~\cite{Transformers}. In addition to LayerNorm~\cite{layernorm} and RMSNorm~\cite{rmsnorm}, LLMs use pre-layer normalization~\cite{preLN}, applying it before multi-head attention (MHA). Pre-norm is shown to provide training stability in LLMs. Another normalization variant, DeepNorm~\cite{deepnorm} fixes the issue with larger gradients in pre-norm. 

\subsection{Distributed LLM Training}
This section describes distributed LLM training approaches briefly. More details are available in~\cite{BLOOM, ZeroOpt, Lib_Megatron, Lib_Bmtrain}. \\
\emph{\textbf{Data Parallelism:}}
Data parallelism replicates the model on multiple devices where data in a batch gets divided across devices. At the end of each training iteration weights are synchronized across all devices. \\
\emph{\textbf{Tensor Parallelism:}}
Tensor parallelism shards a tensor computation across devices. It is also known as horizontal parallelism or intra-layer model parallelism. \\
\emph{\textbf{Pipeline Parallelism:}}
Pipeline parallelism shards model layers across different devices. This is also known as vertical parallelism. \\
\emph{\textbf{Model Parallelism:}}
A combination of tensor and pipeline parallelism is known as model parallelism. \\
\emph{\textbf{3D Parallelism:}}
A combination of data, tensor, and model parallelism is known as 3D parallelism. \\
\emph{\textbf{Optimizer Parallelism:}}
Optimizer parallelism also known as zero redundancy optimizer~\cite{ZeroOpt} implements optimizer state partitioning, gradient partitioning, and parameter partitioning across devices to reduce memory consumption while keeping the communication costs as low as possible. 

%\subsubsection{Rematerialization}

\subsection{Libraries}
Some commonly used libraries for LLMs training are: 
\emph{\textbf{Transformers~\cite{Lib_Transformers}:}} The library provides access to various pre-trained transformer models with APIs to train, fine-tune, infer, and develop custom models. \\
\emph{\textbf{DeepSpeed~\cite{Lib_DeepSpeed}:}} A library for scalable distributed training and inference of deep learning models. \\
\emph{\textbf{Megatron-LM~\cite{Lib_Megatron}:}} It provides GPU-optimized techniques for large-scale training of LLMs. \\
\emph{\textbf{JAX~\cite{Lib_Jax}:}} A Python library for high-performance numerical computing and scaleable machine learning. It can differentiate native Python and NumPy functions and execute them on GPUs. \\
\emph{\textbf{Colossal-AI~\cite{Lib_Colossal}:}} A collection of components to write distributed deep learning models. \\
\emph{\textbf{BMTrain~\cite{Lib_Bmtrain}:}} A library to write efficient stand-alone LLMs training code. \\ 
\emph{\textbf{FastMoE~\cite{Lib_Fastmoe}:}} Provides API to build mixture-of-experts (MoE) model in PyTorch. \\
\emph{\textbf{MindSpore~\cite{Lib_Mindspore}:}} A deep learning training and inference framework extendable to mobile, edge, and cloud computing. \\
\emph{\textbf{PyTorch~\cite{Lib_Pytorch}:}} A framework developed by Facebook AI Research lab (FAIR) to build deep learning models. The main features of PyTorch include a dynamic computation graph and a pythonic coding style. \\
\emph{\textbf{Tensorflow~\cite{Lib_Tensorflow}:}} A deep learning framework written by Google. The key features of TensorFlow are graph-based computation, eager execution, scalability, etc. \\
\emph{\textbf{MXNet~\cite{Lib_Mxnet}:}} Apache MXNet is a deep learning framework with support to write programs in multiple languages, including, Python, C++, Scala, R, etc. It also provides support for dynamic and static computation graphs.

\subsection{Data PreProcessing}
This section briefly summarizes data preprocessing techniques used in LLMs training. \\
\emph{\textbf{Quality Filtering:}}
For better results, training data quality is essential. Some approaches to filtering data are: 1) classifier-based and 2) heuristics-based. Classifier-based approaches train a classifier on high-quality data and predict the quality of text for filtering, whereas heuristics-based employ some rules for filtering like language, metrics, statistics, and keywords. \\
\emph{\textbf{Data Deduplication:}}
Duplicated data can affect model performance and increase data memorization; therefore, to train LLMs, data deduplication is one of the preprocessing steps. This can be performed at multiple levels, like sentences, documents, and datasets. \\
\emph{\textbf{Privacy Reduction:}}
Most of the training data for LLMs is collected through web sources. This data contains private information; therefore, many LLMs employ heuristics-based methods to filter information such as names, addresses, and phone numbers to avoid learning personal information. \\
\subsection{Architectures}
\begin{figure}[tbp]
\centering
\includegraphics[width=1\columnwidth]{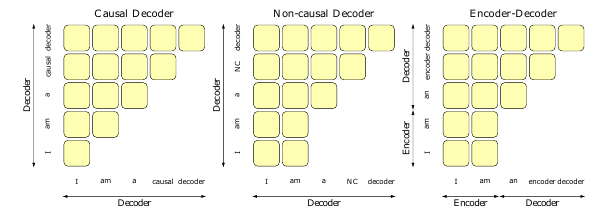}
\caption{An example of attention patterns in language models, image is taken from~\cite{LLM_Objectives}.}
\label{architectures}
\end{figure}

Here we discuss the variants of the transformer architectures used in LLMs. The difference arises due to the application of the attention and the connection of transformer blocks. An illustration of attention patterns of these architectures is shown in Figure~\ref{architectures}. \\
\emph{\textbf{Encoder Decoder:}}
This architecture processes inputs through the encoder and passes the intermediate representation to the decoder to generate the output. Here, the encoder sees the complete sequence utilizing self-attention whereas the decoder processes the sequence one after the other with implementing cross-attention. \\  
\noindent
\emph{\textbf{Causal Decoder:}}
A type of architecture that does not have an encoder and processes and generates output using a decoder, where the predicted token depends only on the previous time steps. \\ 
\noindent
\emph{\textbf{Prefix Decoder:}}
It is also known as a non-causal decoder, where the attention calculation is not strictly dependent on the past information and the attention is bidirectional. An example of a non-causal attention mask is shown in Figure~\ref{architectures}. \\
\emph{\textbf{Mixture-of-Experts:}}
It is a variant of transformer architecture with parallel independent experts and a router to route tokens to experts. These experts are feed-forward layers after the attention block~\cite{fedus2022switch}. Mixture-of-Experts (MoE) is an efficient sparse architecture that offers comparable performance to dense models and allows increasing the model size without increasing the computational cost by activating only a few experts at a time~\cite{du2022glam,PanGu_sigma}.  

\begin{figure}[tbp]
\centering
\includegraphics[width=1\columnwidth]{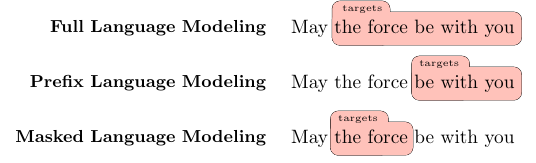}
\caption{An example of language model training objectives, image from~\cite{LLM_Objectives}.}
\label{t_objectives}
\end{figure}

\subsection{Pre-Training Objectives}
\label{sec:pretrainobjectives}
This section describes LLMs pre-training objectives. For more details see the paper~\cite{LLM_Objectives}. 

\begin{figure*}[!t]
\centering
\hspace{-15mm}
\includegraphics[width=2.1\columnwidth]{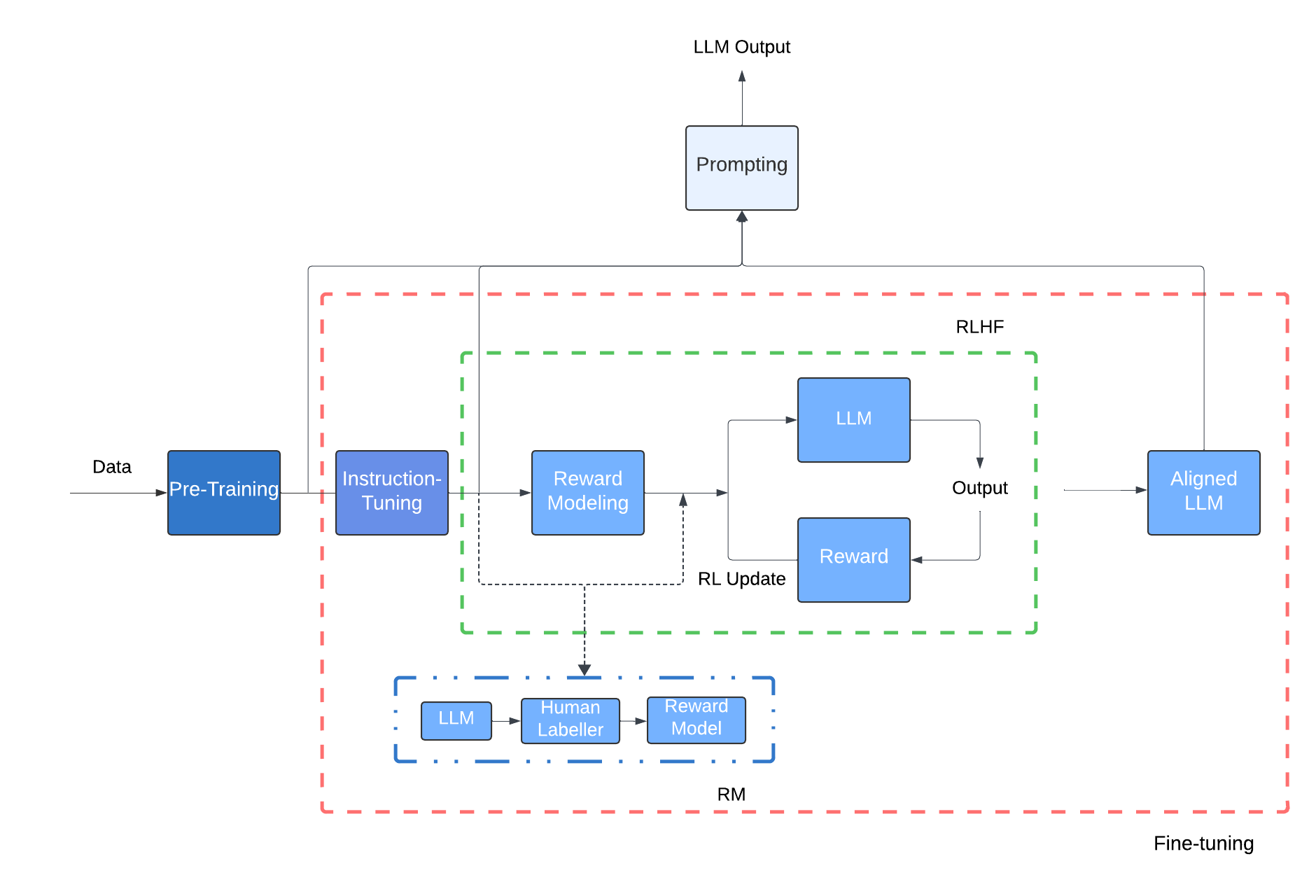}
\caption{A basic flow diagram depicting various stages of LLMs from pre-training to prompting/utilization. Prompting LLMs to generate responses is possible at different training stages like pre-training, instruction-tuning, or alignment tuning. \enquote{RL} stands for reinforcement learning, \enquote{RM} represents reward-modeling, and \enquote{RLHF} represents reinforcement learning with human feedback. }
\label{fig:flow_diagram}
\end{figure*}
\noindent
\emph{\textbf{Full Language Modeling:}}
An autoregressive language modeling objective where the model is asked to predict future tokens given the previous tokens, an example is shown in Figure~\ref{t_objectives}. \\
\emph{\textbf{Prefix Language Modeling:}}
A non-causal training objective, where a prefix is chosen randomly and only remaining target tokens are used to calculate the loss. An example is shown in Figure~\ref{t_objectives}. \\
\emph{\textbf{Masked Language Modeling:}}
In this training objective, tokens or spans (a sequence of tokens) are masked randomly and the model is asked to predict masked tokens given the past and future context. An example is shown in Figure~\ref{t_objectives}.  \\
\emph{\textbf{Unified Language Modeling:}}
Unified language modeling~\cite{Unified_LM} is a combination of causal, non-causal, and masked language training objectives. Here in masked language modeling, the attention is not bidirectional but unidirectional, attending either left-to-right or right-to-left context. \\

\subsection{LLMs Scaling Laws}
Scaling laws study the optimal combination of model parameters, dataset size, and computational resources that predict the improvement in the model performance. It has been shown that the loss scales according to the power-law with model size, dataset size, and compute resources~\cite{kaplan_scaling}. This study suggests larger models are more important than big data for better performance. Another variant of scaling law~\cite{chinchilla} suggests the model size and the number of training tokens should be scaled equally.      
\subsection{LLMs Adaptation Stages}
This section discusses the fundamentals of LLMs adaptation stages, from pre-training to fine-tuning for downstream tasks and utilization. An example of different training stages and inference in LLMs is shown in Figure~\ref{fig:flow_diagram}. In this paper, we refer to alignment-tuning as aligning with human preferences, while occasionally the literature uses the term alignment for different purposes.

\subsubsection{Pre-Training}
In the very first stage, the model is trained in a self-supervised manner on a large corpus to predict the next tokens given the input. The design choices of LLMs vary from encoder-decoder to decoder-only architectures with different building blocks and loss functions in sections~\ref{sec:layernormalization},~\ref{sec:activation functions}, ~\ref{sec:pretrainobjectives}.  

\subsubsection{Fine-Tuning} 
There are different styles to fine-tune an LLM. This section briefly discusses fine-tuning approaches. \\
~\emph{~\textbf{Transfer Learning:}}
The pre-trained LLMs perform well for various tasks~\cite{GPT-3, PaLM}. However, to improve the performance for a downstream task, pre-trained models are fine-tuned with the task-specific data~\cite{T5, mT5}, known as transfer learning. \\ 
~\emph{~\textbf{Instruction-tuning:}}
To enable a model to respond to user queries effectively, the pre-trained model is fine-tuned on instruction formatted data i.e., instruction and an input-output pair. Instructions generally comprise multi-task data in plain natural language, guiding the model to respond according to the prompt and the input. This type of fine-tuning improves zero-shot generalization and downstream task performance. Details on formatting instruction data and its various styles are available in~\cite{Flan, Survey_LLM, OPT_IML}.   \\   
~\emph{~\textbf{Alignment-tuning:}}
LLMs are prone to generating false, biased, and harmful text. To make them helpful, honest, and harmless, models are aligned using human feedback. Alignment involves asking LLMs to generate unexpected responses and then updating their parameters to avoid such responses~\cite{instructgpt, llama_2, self_align}. \\
It ensures LLMs operate according to human intentions and values. A model is defined to be an “aligned” model if the model fulfills three criteria of helpful, honest, and harmless or \enquote{HHH}~\cite{ askell2021general}. \\
Researchers employ reinforcement learning with human feedback (RLHF)~\cite{ziegler2019fine} for model alignment. In RLHF, a fine-tuned model on demonstrations is further trained with reward modeling (RM) and reinforcement learning (RL), shown in Figure~\ref{fig:flow_diagram}. Below we briefly discuss RM and RL pipelines in RLHF.  

% They focus on training the models to understand and adhere to human preferences, ethics, and guidelines, reducing the risk of generating harmful, biased, or misleading content. 

% Proper alignment enhances the accuracy and relevance of LLM responses, making them more reliable and effective in applications like natural language understanding, text generation, and question-answering. 

%The researchers investigated alignment techniques from human feedback, specifically emphasizing reinforcement learning from human feedback (RLHF)~\cite{ziegler2019fine}. 

% The fine-tuning process using RL consist of two phases~\cite{ouyang2022training}.

% RLHF has demonstrated promising outcomes in applications such as text summarization~\cite{wu2021recursively}, dialogue~\cite{ jaques2019way}, review generation~\cite{cho2018towards} and translation~\cite{kreutzer2018can} prompting further exploration and advancement to optimize model performance and better align it with human preferences and expectations.
% In RL with human feedback for fine-tuning LLMs, the LLM takes actions (outputs text) based on the input it receives. The reward function evaluates the LLM's output and assigns a reward based on its quality. The LLM then uses this reward as feedback to adjust and improve its behavior, making it better at generating accurate and relevant text. This process is repeated iteratively to train the LLM effectively, using the rewards to update its policies and optimize its performance. 

%\textbf{\textit{- Reward modeling:}}
\noindent
\textit{Reward modeling:} trains a model to rank generated responses according to human preferences using a classification objective. To train the classifier humans annotate LLMs generated responses based on the HHH criteria. \\
%\indent
%\textbf{\textit{- Reinforcement Learning:}}
% \subsubsubsection{Reinforcement Learning}
\textit{Reinforcement learning:} in combination with the reward model is used for alignment in the next stage. The previously trained reward model ranks LLM-generated responses into preferred vs. non-preferred, which is used to align the model with proximal policy optimization (PPO). This process repeats iteratively until convergence.   
%In the previous stage, a reward model assigns a high score to completions, and therefore the associated token in this stage will receive reinforcement, increasing their probabilities for future occurrences. On the other hand, when the reward model assigns a low score will have lower probabilities for the future. This iterative process is repeated across multiple prompts and batches, leading to the development of a policy that generates excellent outputs. This process represents the RLHF pipeline, which culminates in the deployment of a model. For example, ChatGPT is an example of an RLHF model, while models like Vicuna-13B are categorized as SFT models.

% \emph{\textbf{CodeRL:}}~\cite{le2022coderl} is a framework to improve pre-trained LLM for program synthesis through reinforcement learning. The framework uses an actor-critic approach in which the LLMs model act as an actor to generate a sample of code, and the critic model is trained as a predictor to assess the functional correctness of the code and provide feedback to the actor model. The model employs a novel generation method utilizing critical sampling in the inference phase. This approach enables the model to automatically create new programs by incorporating feedback from example, unit tests and critic scores.
\subsubsection{Prompting/Utilization}
Prompting is a method to query trained LLMs for generating responses, as illustrated in Figure~\ref{fig:flow_diagram}. LLMs can be prompted in various prompt setups, where they can be adapted to the instructions without fine-tuning and in other cases with fine-tuning on data containing different prompt styles~\cite{Flan, cot_collection, zero_to_hero}. A good guide on prompt engineering is available at~\cite{prompt_eng_guide}. Below, we will discuss various widely used prompt setups.\\
~\emph{\textbf{Zero-Shot Prompting:}}
LLMs are zero-shot learners and capable of answering queries never seen before. This style of prompting requires LLMs to answer user questions without seeing any examples in the prompt.  \\
~\emph{\textbf{In-context Learning:}}
Also known as few-shot learning, here, multiple input-output demonstration pairs are shown to the model to generate the desired response. This adaptation style is also called few-shot learning. A discussion on formatting in-context learning (ICL) templates is available in~\cite{survey_incontext_learning,Survey_LLM, Tk-INSTRUCT, Flan}. \\
~\emph{~\textbf{Reasoning in LLMs:}}
LLMs are zero-shot reasoners and can be provoked to generate answers to logical problems, task planning, critical thinking, etc.~with reasoning. Generating reasons is possible only by using different prompting styles, whereas to improve LLMs further on reasoning tasks many methods~\cite{Flan, OPT_IML} train them on reasoning datasets. We discuss various prompting techniques for reasoning below. \\ 
\noindent
\textit{Chain-of-Thought (CoT):}
A special case of prompting where demonstrations contain reasoning information aggregated with inputs and outputs so that the model generates outcomes with step-by-step reasoning. More details on CoT prompts are available in~\cite{survey_reasoning,wei2022chain, cot_collection}. \\
\noindent
\textit{Self-Consistency:} Improves CoT performance by generating multiple responses and selecting the most frequent answer~\cite{self_consistency}.\\
\noindent
\textit{Tree-of-Thought (ToT):} Explores multiple reasoning paths with possibilities to look ahead and backtrack for problem-solving~\cite{ToT}. \\
\noindent
\emph{\textbf{Single-Turn Instructions:}}
In this prompting setup, LLMs are queried only once with all the relevant information in the prompt. LLMs generate responses by understanding the context either in a zero-shot or few-shot setting.  \\
\noindent
\emph{\textbf{Multi-Turn Instructions:}}
Solving a complex task requires multiple interactions with LLMs, where feedback and responses from the other tools are given as input to the LLM for the next rounds. This style of using LLMs in the loop is common in autonomous agents. 

%~\emph{~\textbf{Algorithm-of-Thoughts:}}

\section{Large Language Models}
\label{sec_review}
This section reviews LLMs, briefly describing their architectures, training objectives, pipelines, datasets, and fine-tuning details.

\subsection{Pre-Trained  LLMs}
Here, we provide summaries of various well-known pre-trained LLMs with significant discoveries, changing the course of research and development in NLP. These LLMs have considerably improved the performance in NLU and NLG domains, and are widely fine-tuned for downstream tasks. Moreover, We also identify key findings and insights of pre-trained LLMs in Table~\ref{tab:pre_trained_findings}~and~\ref{tab:instruction_tuned_findings} that improve their performance. 

\subsubsection{General Purpose}
\begin{figure}[tbp]
\centering
\includegraphics[width=1\columnwidth]{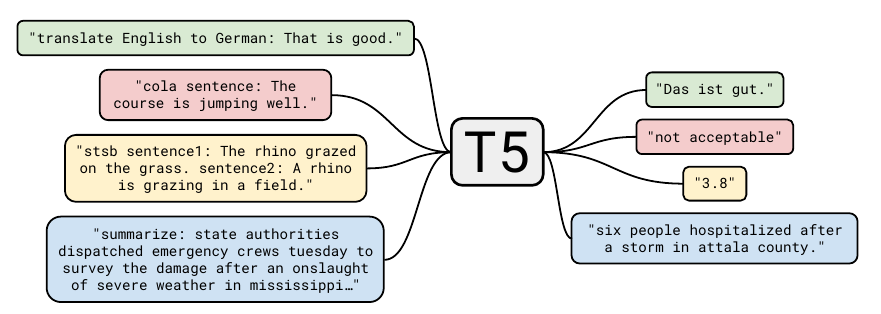}
\caption{Unified text-to-text training example, source image from~\cite{T5}.}
\label{t5_image}
\end{figure}

\noindent
~\emph{~\textbf{T5~\cite{T5}:}}
An encoder-decoder model employing a unified text-to-text training for all NLP problems is shown in Figure~\ref{t5_image}. T5 places layer normalization outside the residual path in a conventional transformer model~\cite{Transformers}. It uses masked language modeling as a pre-training objective where spans (consecutive tokens) are replaced with a single mask instead of separate masks for each token. This type of masking speeds up the training as it produces shorter sequences. After pre-training, the model is fine-tuned using adapter layers~\cite{LMAdapter} for downstream tasks.

\noindent
~\emph{~\textbf{GPT-3~\cite{GPT-3}:}}
The GPT-3 architecture is the same as the GPT-2~\cite{GPT-2} but with dense and sparse attention in transformer layers similar to the Sparse Transformer~\cite{sparse_transformer}. It shows that large models can train on larger batch sizes with a lower learning rate to decide the batch size during training, GPT-3 uses the gradient noise scale as in ~\cite{batch_size_selec}. Overall, GPT-3 increases model parameters to 175B showing that the performance of large language models improves with the scale and is competitive with the fine-tuned models.

\noindent
~\emph{~\textbf{mT5~\cite{mT5}:}}
A multilingual T5 model~\cite{T5} trained on the mC4 dataset with 101 languages. The dataset is extracted from the public common crawl scrape. The model uses a larger vocabulary size of 250,000 to cover multiple languages. To avoid over-fitting or under-fitting for a language, mT5 employs a data sampling procedure to select samples from all languages. The paper suggests using a small amount of pre-training datasets, including all languages when fine-tuning for a task using English language data. This allows the model to generate correct non-English outputs.  
%$p(L) \propto \mid L \mid^\alpha$, where $p(L)$ is sampling probability, $L$ is language sampling count, and $\alpha$ controls the sampling probability

\begin{figure}[tbp]
\centering
\includegraphics[width=1\columnwidth]{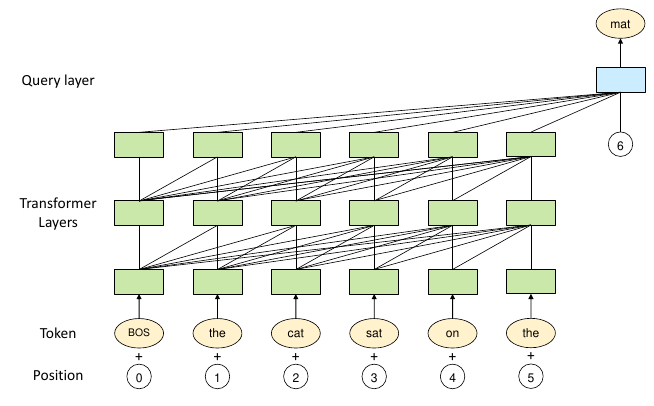}
\caption{The image is the article of~\cite{PanGU_alpha}, showing an example of PanGu-$\alpha$ architecture.}
\label{pangu_alpha_image}
\end{figure}

\noindent
~\emph{~\textbf{PanGu-$\alpha$~\cite{PanGU_alpha}:}}
An autoregressive model that has a query layer at the end of standard transformer layers, example shown in Figure~\ref{pangu_alpha_image}, to predict the next token. Its structure is similar to the transformer layer but with an additional embedding for the next position in the attention mechanism, given in Eq.~\ref{PanGu_alpha_eq}.
\begin{equation}
a = p_nW_h^qW_h^kTH_L^T
\label{PanGu_alpha_eq}
\end{equation}

\noindent
~\emph{~\textbf{CPM-2~\cite{CPM-2}:}}
Cost-efficient Pre-trained language Models (CPM-2) pre-trains bilingual (English and Chinese) 11B and 198B mixture-of-experts (MoE) models on the WuDaoCorpus~\cite{WuDaoCorpus} dataset. The tokenization process removes \enquote{\_} white space tokens in the sentencepiece tokenizer. The models are trained with knowledge inheritance, starting with only the Chinese language in the first stage and then adding English and Chinese data. This trained model gets duplicated multiple times to initialize the 198B MoE model. Moreover, to use the model for downstream tasks, CPM-2 experimented with both complete fine-tuning and prompt fine-tuning as in~\cite{Prompt_Tuning} where only prompt-related parameters are updated by inserting prompts at various positions, front, middle, and back. CPM-2 also proposes the INFMOE, a memory-efficient framework with a strategy to dynamically offload parameters to the CPU for inference at a 100B scale. It overlaps data movement with inference computation for lower inference time. 

%It has an encoder-decoder architecture with a bidirectional encoder and a unidirectional decoder. 

\noindent
~\emph{~\textbf{ERNIE 3.0~\cite{ernie3}:}}
ERNIE 3.0 takes inspiration from multi-task learning to build a modular architecture using Transformer-XL~\cite{dai2019transformer} as the backbone. The universal representation module is shared by all the tasks, which serve as the basic block for task-specific representation modules, which are all trained jointly for natural language understanding, natural language generation, and knowledge extraction. This LLM is primarily focused on the Chinese language. It claims to train on the largest Chinese text corpora for LLM training, and achieved state-of-the-art in 54 Chinese NLP tasks.

\noindent
~\emph{~\textbf{Jurassic-1~\cite{lieber2021jurassic}:}}
A pair of auto-regressive language models, including a 7B-parameter J1-Large model and a 178B-parameter J1-Jumbo model. The training vocabulary of Jurassic-1 comprise word pieces, complete words, and multi-word expressions without any word boundaries, where possible out-of-vocabulary instances are interpreted as Unicode bytes. Compared to the GPT-3 counterparts, the Jurassic-1 models apply a more balanced depth-to-width self-attention architecture~\cite{limit_to_depth} and an improved tokenizer for a faster prediction based on broader resources, achieving a comparable performance in zero-shot learning tasks and a superior performance in few-shot learning tasks given the ability to feed more examples as a prompt.
%The Jurassic-1 models are mainly structured on the Transformer decoder module~\cite{Transformers}, while the architecture modifications proposed by GPT-2~\cite{GPT-2} are also incorporated.   The Jurassic-1 models are mainly structured on the Transformer decoder module~\cite{Transformers}, while the architecture modifications proposed by GPT-2~\cite{GPT-2} are also incorporated.

\noindent
~\emph{~\textbf{HyperCLOVA~\cite{hyperclova}:}}
A Korean language model with GPT-3 architecture.
%The architecture is the same as that of GPT3~\cite{GPT-3} with morphene aware byte level encoding tokenization step. A large Korean-centric corpus gathered from various sources (see table for details) is trained using Megatron LM. Prompt-based tuning is also applied to enhance performance on downstream tasks. The main objective of training this model is to see how the non-English language model fares compared to universally found English-based LMs.

\noindent
~\emph{~\textbf{Yuan 1.0~\cite{wu2021yuan}:}}
Trained on a Chinese corpus with 5TB of high-quality text collected from the Internet. A Massive Data Filtering System (MDFS) built on Spark is developed to process the raw data via coarse and fine filtering techniques. To speed up the training of Yuan 1.0 to save energy expenses and carbon emissions, various factors that improve the performance of distributed training are incorporated in architecture and training: like increasing the hidden state size improves pipeline and tensor parallelism performance, larger micro batches improve pipeline parallelism performance, and larger global batch size improve data parallelism performance. In practice, the Yuan 1.0 model performs well on text classification, Winograd Schema, natural language inference, and reading comprehension tasks.

\noindent
~\emph{~\textbf{Gopher~\cite{gopher}:}}
The Gopher family of models ranges from 44M to 280B parameters in size to study the effect of \textit{scale} on the LLMs performance. The 280B model beats GPT-3~\cite{GPT-3}, Jurrasic-1~\cite{lieber2021jurassic}, MT-NLG~\cite{mtnlg}, and others on 81\% of the evaluated tasks.
%It is the largest of six causal decoder LLMs trained on the subsets of MassiveWeb, Books, C4, News, GitHub, and Wikipedia samples from high-quality curated MassiveText. The model is a modified version of Transformer architecture used in~\cite{GPT-2}. 

\noindent
~\emph{~\textbf{ERNIE 3.0 TITAN~\cite{ernie3titan}:}}
ERNIE 3.0 Titan extends ERNIE 3.0 by training a larger model with 26x the number of parameters of the latter. This bigger model outperformed other state-of-the-art models in 68 NLP tasks. LLMs produce text with incorrect facts. In order to have control of the generated text with factual consistency, ERNIE 3.0 Titan adds another task, \textit{Credible and Controllable Generations}, to its multi-task learning setup. It introduces additional self-supervised adversarial and controllable language modeling losses to the pre-training step, which enables ERNIE 3.0 Titan to beat other LLMs in their manually selected Factual QA task set evaluations.

\noindent
~\emph{~\textbf{GPT-NeoX-20B~\cite{GPT_NeoX}:}}
An auto-regressive model that largely follows GPT-3 with a few deviations in architecture design, trained on the Pile dataset without any data deduplication. GPT-NeoX has parallel attention and feed-forward layers in a transformer block, given in Eq.~\ref{GPT-NeoX-20B_eq}, that increases throughput by 15\%. It uses rotary positional embedding~\cite{su2021roformer}, applying it to only 25\% of embedding vector dimension as in~\cite{GPT_J_6B}. This reduces the computation without performance degradation. As opposed to GPT-3, which uses dense and sparse layers, GPT-NeoX-20B uses only dense layers. The hyperparameter tuning at this scale is difficult; therefore, the model chooses hyperparameters from the method~\cite{GPT-3} and interpolates values between 13B and 175B models for the 20B model. The model training is distributed among GPUs using both tensor and pipeline parallelism.   
\begin{equation}
x + Attn(LN_1(x)) + FF(LN_2(x))
\label{GPT-NeoX-20B_eq}
\end{equation}

\noindent
~\emph{~\textbf{OPT~\cite{OPT}:}}
It is a clone of GPT-3, developed to open-source a model that replicates GPT-3 performance. Training of OPT employs dynamic loss scaling ~\cite{Mixed_Precision} and restarts from an earlier checkpoint with a lower learning rate whenever loss divergence is observed. Overall, the performance of OPT-175B models is comparable to the GPT3-175B model.

\begin{figure}[tbp]
\centering
\includegraphics[width=1\columnwidth]{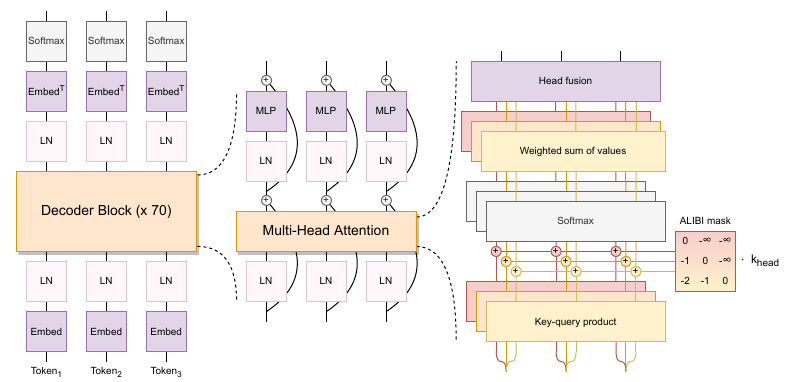}
\caption{The BLOOM architecture example sourced from~\cite{BLOOM}.}
\label{bloom_image}
\end{figure}

\noindent
~\emph{~\textbf{BLOOM~\cite{BLOOM}:}}
A causal decoder model trained on the ROOTS corpus to open-source an LLM. The architecture of BLOOM is shown in Figure~\ref{bloom_image}, with differences like ALiBi positional embedding, an additional normalization layer after the embedding layer as suggested by the bitsandbytes\footnote{https://github.com/TimDettmers/bitsandbytes} library. These changes stabilize training with improved downstream performance.

\noindent
~\emph{~\textbf{GLaM~\cite{du2022glam}:}}
Generalist Language Model (GLaM) represents a family of language models using a sparsely activated decoder-only mixture-of-experts (MoE) structure~\cite{shazeer2017outrageously,fedus2022switch}. To gain more model capacity while reducing computation, the experts are sparsely activated where only the best two experts are used to process each input token. The largest GLaM model, GLaM (64B/64E), is about 7$\times$ larger than GPT-3~\cite{GPT-3}, while only part of the parameters are activated per input token. The largest GLaM (64B/64E) model achieves better overall results as compared to GPT-3 while consuming only one-third of GPT-3's training energy.

\noindent
~\emph{~\textbf{MT-NLG~\cite{mtnlg}:}}
A 530B causal decoder based on the GPT-2 architecture that has roughly 3$\times$ GPT-3 model parameters. MT-NLG is trained on filtered high-quality data collected from various public datasets and blends various types of datasets in a single batch, which beats GPT-3 on several evaluations.

\noindent
~\emph{~\textbf{Chinchilla~\cite{chinchilla}:}}
A causal decoder trained on the same dataset as the Gopher~\cite{gopher} but with a little different data sampling distribution (sampled from MassiveText). The model architecture is similar to the one used for Gopher, with the exception of AdamW optimizer instead of Adam. Chinchilla identifies the relationship that model size should be doubled for every doubling of training tokens. Over 400 language models ranging from 70 million to over 16 billion parameters on 5 to 500 billion tokens are trained to get the estimates for compute-optimal training under a given budget. The authors train a 70B model with the same compute budget as Gopher (280B) but with 4 times more data. It outperforms Gopher~\cite{gopher}, GPT-3~\cite{GPT-3}, and others on various downstream tasks, after fine-tuning.

\noindent
~\emph{~\textbf{AlexaTM~\cite{soltan2022alexatm}:}}
An encoder-decoder model, where encoder weights and decoder embeddings are initialized with a pre-trained encoder to speed up training. The encoder stays frozen for the initial 100k steps and is later unfrozen for end-to-end training. The model is trained on a combination of denoising and causal language modeling (CLM) objectives, concatenating a $[CLM]$ token at the beginning for mode switching. During training, the CLM task is applied for 20\% of the time, which improves the in-context learning performance.

\noindent
~\emph{~\textbf{PaLM~\cite{PaLM}:}}
A causal decoder with parallel attention and feed-forward layers similar to Eq.~\ref{GPT-NeoX-20B_eq}, speeding up training by a factor of 15. Additional changes to the conventional transformer model include SwiGLU activation, RoPE embeddings, multi-query attention that saves computation cost during decoding, and shared input-output embeddings. During training, loss spiking was observed, and to fix it, model training was restarted from a 100-step earlier checkpoint by skipping 200-500 batches around the spike. Moreover, the model was found to memorize around 2.4\% of the training data at the 540B model scale, whereas this number was lower for smaller models.

\noindent
~\emph{~\textbf{PaLM-2}~\cite{palm_2}:}
A smaller multi-lingual variant of PaLM, trained for larger iterations on a better quality dataset. PaLM-2 shows significant improvements over PaLM, while reducing training and inference costs due to its smaller size. To lessen toxicity and memorization, it appends special tokens with a fraction of pre-training data, which shows a reduction in generating harmful responses.

\noindent
~\emph{~\textbf{U-PaLM~\cite{U-PaLM}:}}
This method trains PaLM for 0.1\% additional compute with the UL2 (also named as UL2Restore) objective~\cite{UL2}, using the same dataset it outperforms the baseline significantly on various NLP tasks, including zero-shot, few-shot, commonsense reasoning, CoT, etc. Training with UL2R involves converting a causal decoder PaLM to a non-causal decoder PaLM and employing 50\% sequential denoising, 25\% regular denoising, and 25\% extreme denoising loss functions.

\noindent
~\emph{~\textbf{UL2~\cite{UL2}:}}
An encoder-decoder architecture trained using a mixture of denoisers (MoD) objective. Denoisers include 1) R-Denoiser: a regular span masking, 2) S-Denoiser: which corrupts consecutive tokens of a large sequence and 3) X-Denoiser: which corrupts a large number of tokens randomly. During pre-training, UL2 includes a denoiser token from ${R, S, X}$ to represent a denoising setup. It helps improve fine-tuning performance for downstream tasks that bind the task to one of the upstream training modes. This MoD style of training outperforms the T5 model on many benchmarks.

\noindent
~\emph{~\textbf{GLM-130B~\cite{GLM-130B}:}}
GLM-130B is a bilingual (English and Chinese) model trained using an auto-regressive mask infilling pre-training objective similar to the GLM~\cite{GLM}. This training style makes the model bidirectional as compared to GPT-3, which is unidirectional. As opposed to GLM, the training of GLM-130B includes a small amount of multi-task instruction pre-training data (5\% of the total data) along with self-supervised mask infilling. To stabilize the training, it applies embedding layer gradient shrink.

\noindent
~\emph{~\textbf{LLaMA~\cite{touvron2023llama, llama_2}:}}
A set of decoder-only language models varying from 7B to 70B parameters. LLaMA models series is the most famous among the community for parameter efficiency and instruction tuning.

\noindent
~\emph{~\textbf{LLaMA-1}~\cite{touvron2023llama}:} Implements efficient causal attention~\cite{self_att_reduced_mem} by not storing and computing masked attention weights and key/query scores. Another optimization is reducing the number of activations recomputed in the backward pass, as in ~\cite{reducing_act_recompute}.

\noindent
~\emph{~\textbf{LLaMA-2}~\cite{llama_2}:} This work is more focused on fine-tuning a safer and better LLaMA-2-Chat model for dialogue generation. The pre-trained model has 40\% more training data with a larger context length and grouped-query attention.

\noindent
~\emph{~\textbf{LLaMA-3/3.1}~\cite{llama3}:} A collection of models trained on a seven times larger dataset as compared to LLaMA-2 with double the context length, outperforming its previous variants and other models. 

\begin{figure}[tbp]
\centering
\includegraphics[width=1\columnwidth]{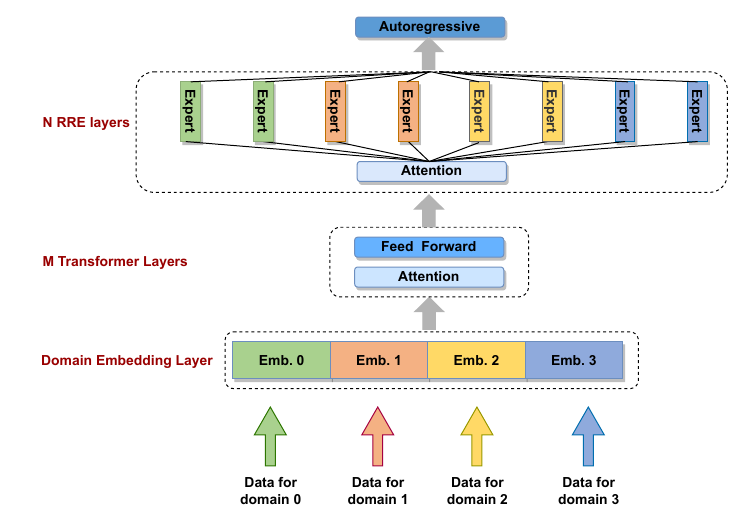}
\caption{This example illustrates the PanGu-$\sum$ architecture, as depicted in the image sourced from~\cite{PanGu_sigma}.}
\label{pangu_sigma_image}
\end{figure}

\noindent
~\emph{~\textbf{PanGu-$\Sigma$~\cite{PanGu_sigma}:}}
An autoregressive model with parameters copied from PanGu-$\alpha$ and extended to a trillion scale with Random Routed Experts (RRE), the architectural diagram is shown in Figure~\ref{pangu_sigma_image}. RRE is similar to the MoE architecture, with distinctions at the second level, where tokens are randomly routed to experts in a domain instead of using a learnable gating method. The model has bottom layers densely activated and shared across all domains, whereas top layers are sparsely activated according to the domain. This training style allows for extracting task-specific models and reduces catastrophic forgetting effects in the case of continual learning. 

\noindent
~\emph{~\textbf{Mixtral8x22b}~\cite{mixtral}:} A mixture-of-experts (MoE) model with eight distinct experts routes each token to two experts at each layer and combines the outputs additively.   

\noindent
~\emph{~\textbf{Snowflake Arctic}~\cite{snowflake_arctic}:}
Arctic LLM is a hybrid of dense and mixture-of-experts (MoE) architecture. The MoE (128$\times$3.66B MLP experts) is parallel to the dense transformer (10B) with only two experts activated. The model has many experts, compared to other MoE LLMs~\cite{mixtral,grok_1}, to increase the model capacity and provide an opportunity to choose among many experts for a diverse configuration. The model has 480B parameters, and only 17B are active during a forward pass, reducing the computation significantly.

\noindent
~\emph{~\textbf{Grok}~\cite{grok_1,grok_15}:}
Grok is a family of LLMs including Grok-1 and Grok-1.5, released by XAI. 

\noindent
~\emph{~\textbf{Grok-1}~\cite{grok_1}:}
Grok-1 is a 314B parameters language MoE model (eight experts), where two experts are activated per token. 

\noindent
~\emph{~\textbf{Grok-1.5}~\cite{grok_15}:}
Grok-1.5 is a multi-modal LLM with a larger context length and improved performance.

\noindent
~\emph{~\textbf{Gemini}~\cite{gemini,gemini_15}:}
Gemini replaces Bard (based on PaLM) with multi-modal capabilities and significant language modeling performance improvements.  

\noindent
~\emph{~\textbf{Gemini-1}~\cite{gemini}:}
The first-ever auto-regressive model to achieve human-level capabilities on the MMLU benchmark.   

\noindent
~\emph{~\textbf{Gemini-1.5}~\cite{gemini_15}:}
A multi-modal LLM with MoE architecture builds on the findings of Gemini-1. The model has a 2M context window and can reason over information up to 10M tokens. Such large context windows were never achieved previously and shown to have a huge impact on performance gain.

\noindent
~\emph{~\textbf{Nemotron-4 340B}~\cite{nemotron4}:}
A decoder-only model that has been aligned on 98\% synthetic data and only 2\% manually annotated data. Utilizing synthetic data at a large proportion improves the model performance significantly. The paper suggested introducing alignment data with a smaller subset of previously seen data during the late stage of the model pre-training, enabling the smooth transition from the pre-trained stage to the final training stage.  To train better instruction-following models, weaker models are trained into stronger models iteratively. The synthetic data generated by the weaker instruction-tuned model is used to train a base model which is later supervised fine-tuned outperforming the weaker model.  

\noindent
~\emph{~\textbf{DeepSeek}~\cite{deepseek}:}
DeepSeek studies the LLMs scaling laws in detail to determine the optimal non-embedding model size and training data. The experiments were performed for 8 budgets ranging from 1e$^{17}$ to 3e$^{20}$ training FLOPs. Each compute budget was tested against ten different models/data scales. The batch size and learning rates were also fitted for the given compute budget finding that the batch size should increase with the increased compute budget while decreasing the learning rate. Following are the equations for the optimal batch-size ($B$), learning rate ($\eta$), model size ($M$), and data ($D$):
\begin{equation}
\begin{gathered}
    B_{opt} = 0.2920.C^{0.3271}  \\
    \eta_{opt} = 0.3118.C^{-0.1250}   \\
    M_{opt} = M_{base}.C^a  \\
    D_{opt} = D_{base}.C^b \\
    M_{base} = 0.1715,  D_{base} = 5.8316, a = 0.5243, b = 0.4757 \\
\end{gathered}
\end{equation}

\noindent
~\emph{~\textbf{DeepSeek-v2}~\cite{deepseek_v2}:}
An MoE model that introduces multi-head latent attention (MLA) to reduce inference costs, by compressing Key-Value (KV) cache into a latent vector. MLA achieves better performance than multi-head attention (MHA), and other efficient attention mechanisms such as grouped query attention (GQA), multi-query attention (MQA), etc. Because of MLA, DeepSeek-v2 achieves 5.76 times faster inference throughput as compared to DeepSeek~\cite{deepseek}.
\subsubsection{Coding}
\label{sec:coding}

\noindent
~\emph{~\textbf{CodeGen~\cite{CodeGen}:}}
CodeGen has a similar architecture to PaLM~\cite{PaLM}, i.e., parallel attention, MLP layers, and RoPE embeddings. The model is trained on both natural language and programming language data sequentially (trained on the first dataset, then the second, and so on) on the following datasets 1) PILE, 2) BIGQUERY, and 3) BIGPYTHON. CodeGen proposed a multi-step approach to synthesizing code. The purpose is to simplify the generation of long sequences where the previous prompt and generated code are given as input with the next prompt to generate the next code sequence. CodeGen opensource a Multi-Turn Programming Benchmark (MTPB) to evaluate multi-step program synthesis.

\noindent
~\emph{~\textbf{Codex~\cite{codex}:}}
This LLM is trained on a subset of public Python Github repositories to generate code from docstrings. Computer programming is an iterative process where the programs are often debugged and updated before fulfilling the requirements. Similarly, Codex generates 100 versions of a program by repetitive sampling for a given description, which produces a working solution for 77.5\% of the problems passing unit tests.  Its powerful version powers Github Copilot\footnote{https://github.com/features/copilot}.

\noindent
~\emph{~\textbf{AlphaCode~\cite{li2022competition}:}}
A set of large language models, ranging from 300M to 41B parameters, designed for competition-level code generation tasks. It uses the multi-query attention~\cite{shazeer2019fast} to reduce memory and cache costs. Since competitive programming problems highly require deep reasoning and an understanding of complex natural language algorithms, the AlphaCode models are pre-trained on filtered GitHub code in popular languages and then fine-tuned on a new competitive programming dataset named CodeContests. The CodeContests dataset mainly contains problems, solutions, and test cases collected from the Codeforces platform\footnote{https://codeforces.com/}. The pre-training employs standard language modeling objectives, while GOLD~\cite{pang2020text} with tempering~\cite{dabre2020softmax} serves as the training objective for the fine-tuning on CodeContests data. To evaluate the performance of AlphaCode, simulated programming competitions are hosted on the Codeforces platform: overall, AlphaCode ranks at the top 54.3\% among over 5000 competitors, where its Codeforces rating is within the top 28\% of recently participated users.

\noindent
~\emph{~\textbf{CodeT5+~\cite{codet5+}:}}
CodeT5+ is based on CodeT5~\cite{codet5}, with shallow encoder and deep decoder, trained in multiple stages initially unimodal data (code) and later bimodal data (text-code pairs). Each training stage has different training objectives and activates different model blocks encoder, decoder, or both according to the task. The unimodal pre-training includes span denoising and CLM objectives, whereas bimodal pre-training objectives contain contrastive learning, matching, and CLM for text-code pairs. CodeT5+ adds special tokens with the text to enable task modes, for example, $[CLS]$ for contrastive loss, $[Match]$ for text-code matching, etc.

\noindent
~\emph{~\textbf{StarCoder~\cite{starcoder}:}}
A decoder-only model with the SantaCoder architecture, employing Flash attention to scale up the context length to 8k. The StarCoder trains an encoder to filter names, emails, and other personal data from the training data. Its fine-tuned variant outperforms PaLM, LLaMA, and LAMDA on HumanEval and MBPP benchmarks.

\subsubsection{Scientific Knowledge}

\noindent
~\emph{~\textbf{Galactica~\cite{galactica}:}}
A large curated corpus of human scientific knowledge with 48 million papers, textbooks, lecture notes, millions of compounds and proteins, scientific websites, encyclopedias, and more are trained using the metaseq library3, which is built on PyTorch and fairscale~\cite{fairscale}. The model wraps reasoning datasets with the $<work>$ token to provide step-by-step reasoning context to the model, which has been shown to improve the performance on reasoning tasks.

\subsubsection{Dialog}

\noindent
~\emph{~\textbf{LaMDA~\cite{thoppilan2022lamda}:}}
A decoder-only model pre-trained on public dialog data, public dialog utterances, and public web documents, where more than 90\% of the pre-training data is in English. LaMDA is trained with the objective of producing responses that exhibit high levels of quality, safety, and groundedness. To achieve this, discriminative and generative fine-tuning techniques are incorporated to enhance the model's safety and quality aspects. As a result, the LaMDA models can be utilized as a general language model performing various tasks.

\subsubsection{Finance}

\noindent
~\emph{~\textbf{BloombergGPT~\cite{bloomberggpt}:}}
A non-causal decoder model trained using both financial (\enquote{FINPILE} from the Bloomberg archive) and general-purpose datasets. The model's architecture is similar to the BLOOM~\cite{BLOOM} and OPT~\cite{OPT}. It allocates 50B parameters to different blocks of the model using the approach~\cite{limit_to_depth}. For effective training, BloombergGPT packs documents together with $<|endoftext|>$ to use the maximum sequence length, uses warmup batch size starting from 1024 to 2048, and manually reduces the learning rate multiple times during the training.

\noindent
~\emph{~\textbf{Xuan Yuan 2.0~\cite{xuanyuan}:}}
A Chinese financial chat model with BLOOM's~\cite{BLOOM} architecture trained on a combination of general purpose, financial, general purpose instructions, and financial institutions datasets. Xuan Yuan 2.0 combined the pre-training and fine-tuning stages to avoid catastrophic forgetting.

\begin{table*}[!tbhp]
\vspace{-6mm}
\caption{Noteworthy findings and insights of \emph{pre-trained} Large Language Models.}
%\caption{Important findings of pre-trained LLMs}

\begin{tabular}{lc}
\hline \hline
\rowcolor{gray!50} Models & Findings \& Insights\\ \hline \hline
\vspace{-3mm}
T5 & \begin{tabular}{c} 
\multicolumn{1}{p{14.5cm}}{\begin{itemize} 
\vspace{-3mm}
\item Encoder and decoder with shared parameters perform equivalently when parameters are not shared
\vspace{-3mm}
\item Fine-tuning model layers (adapter layers) work better than the conventional way of training on only classification layers
\end{itemize}}
\end{tabular}  \\ \cline{2-2}%\hline

\vspace{-3mm}
GPT-3 & \begin{tabular}{c}
\multicolumn{1}{p{14.5cm}}{\begin{itemize} 
\vspace{-3mm}
\item Few-shot performance of LLMs is better than the zero-shot, suggesting that LLMs are meta-learners
\end{itemize}}
\end{tabular}    \\ \cline{2-2}%\hline

\vspace{-3mm}
mT5 & \begin{tabular}{c}
\multicolumn{1}{p{14.5cm}}{\begin{itemize} 
\vspace{-3mm}
\item Large multi-lingual models perform equivalently to single language models on downstream tasks. However, smaller multi-lingual models perform worse
\end{itemize}}
\end{tabular}    \\ \cline{2-2}%\hline

\vspace{-3mm}
PanGu-$\alpha$ & \begin{tabular}{c}
\multicolumn{1}{p{14.5cm}}{\begin{itemize} 
\vspace{-3mm}
\item LLMs have good few shot capabilities 
\end{itemize}}
\end{tabular} \\ \cline{2-2}%\hline

\vspace{-3mm}
CPM-2 & \begin{tabular}{c}
\multicolumn{1}{p{14.5cm}}{\begin{itemize} 
\vspace{-3mm}
\item Prompt fine-tuning requires updating very few parameters while achieving performance comparable to full model fine-tuning 
\vspace{-3mm}
\item Prompt fine-tuning takes more time to converge as compared to full model fine-tuning 
\vspace{-3mm}
\item Inserting prompt tokens in-between sentences can allow the model to understand relations between sentences and long sequences
\vspace{-3mm}
\item In an analysis, CPM-2 finds that prompts work as a provider (additional context) and aggregator (aggregate information with the input text) for the model
\end{itemize}}
\end{tabular}  \\ \cline{2-2}%\hline

\vspace{-3mm}
ERNIE 3.0 & \begin{tabular}{c}
\multicolumn{1}{p{14.5cm}}{\begin{itemize} 
\vspace{-3mm}
\item A modular LLM architecture with a universal representation module and task-specific representation module helps in the finetuning phase
\vspace{-3mm}
\item Optimizing the parameters of a task-specific representation network during the fine-tuning phase is an efficient way to take advantage of the powerful pre-trained model
\end{itemize}}
\end{tabular}    \\ \cline{2-2}%\hline

\vspace{-3mm}
Jurassic-1 & \begin{tabular}{c}
\multicolumn{1}{p{14.5cm}}{\begin{itemize} 
\vspace{-3mm}
\item The performance of LLM is highly related to the network size
\vspace{-3mm}
\item To improve runtime performance, more operations can be performed
in parallel (width) rather than sequential (depth)
\vspace{-3mm}
\item To efficiently represent and fit more text in the same context length, the model uses a larger vocabulary to train a SentencePiece tokenizer without restricting it to word boundaries. This further benefits in few-shot learning tasks
\end{itemize}}

\end{tabular}    \\ \cline{2-2}%\hline

\vspace{-2mm}
HyperCLOVA & \begin{tabular}{c}
\multicolumn{1}{p{14.5cm}}{\begin{itemize} \vspace{-2mm}
\item By employing prompt-based tuning, the performances of models can be improved, often surpassing those of state-of-the-art models when the backward gradients of inputs are accessible
\end{itemize}}
\end{tabular}    \\ \cline{2-2}%\hline

\vspace{-2mm}
Yuan 1.0 & \begin{tabular}{c}
\multicolumn{1}{p{14.5cm}}{\begin{itemize} \vspace{-2mm}
\item The model architecture that excels in pre-training and fine-tuning cases may exhibit contrasting behavior in zero-shot and few-shot learning
\end{itemize}}
\end{tabular}    \\ \cline{2-2}%\hline

\vspace{-3mm}
Gopher & \begin{tabular}{c}
\multicolumn{1}{p{14.5cm}}{\begin{itemize} 
\vspace{-3mm}
\item Relative encodings enable the model to evaluate for longer sequences than training.
\end{itemize}}
\end{tabular}    \\ \cline{2-2}%\hline

\vspace{-3mm}
ERNIE 3.0 Titan & \begin{tabular}{c}
\multicolumn{1}{p{14.5cm}}{\begin{itemize} 
\vspace{-3mm}
\item Additional self-supervised adversarial loss to distinguish between real and generated text improves the model performance as compared to ERNIE 3.0
\end{itemize}}
\end{tabular}    \\ \cline{2-2}%\hline

\vspace{-3mm}
GPT-NeoX-20B & \begin{tabular}{c}
\multicolumn{1}{p{14.5cm}}{\begin{itemize} 
\vspace{-3mm}
\item Parallel attention + FF layers speed-up training 15\% with the same performance as with cascaded layers
\vspace{-3mm}
\item Initializing feed-forward output layers before residuals with scheme in~\cite{Mesh_Transformer_JAX} avoids activations from growing with increasing depth and width
\vspace{-3mm}
\item Training on Pile outperforms GPT-3 on five-shot 
\end{itemize}}
\end{tabular}    \\ \cline{2-2}%\hline

%\\ \hline
\end{tabular}%

\vspace{-3mm}
\begin{flushright}
Table Continued on Next Page 
\end{flushright}
\label{tab:pre_trained_findings}
\end{table*}

\begin{table*}[!tbhp]
%\ContinuedFloat
%\caption{(Continued on next page)}
\begin{tabular}{lc}
\hline \hline
\rowcolor{gray!50}Models & Findings  \& Insights\\ \hline \hline

\vspace{-3mm}
OPT & \begin{tabular}{c}
\multicolumn{1}{p{14.5cm}}{\begin{itemize} 
\vspace{-3mm}
\item Restart training from an earlier checkpoint with a lower learning rate if loss diverges
\vspace{-3mm}
\item Model is prone to generate repetitive text and stuck in a loop
\end{itemize}}
\end{tabular}    \\ \cline{2-2}%\hline

\vspace{-3mm}
Galactica & \begin{tabular}{c}
\multicolumn{1}{p{14.5cm}}{\begin{itemize} 
\vspace{-3mm}
\item Galactica's performance has continued to improve across validation set, in-domain, and out-of-domain benchmarks, even with multiple repetitions of the corpus, which is superior to existing research on LLMs
\vspace{-3mm}
\item A working memory token approach can achieve strong performance over existing methods on mathematical MMLU and MATH benchmarks. It sets a new state-of-the-art on several downstream tasks such as PubMedQA (77.6\%) and MedMCQA dev (52.9\%)
\end{itemize}}
\end{tabular}    \\ \cline{2-2}%\hline

\vspace{-3mm}
GLaM & \begin{tabular}{c}
\multicolumn{1}{p{14.5cm}}{\begin{itemize} 
\vspace{-3mm}
\item The model capacity can be maintained at reduced computation by replacing the feed-forward layer in each transformer layer with a mixture-of-experts (MoE)
%\vspace{-3mm}
%\item By leveraging sparsity, we can make significant strides toward developing high-quality NLP models while simultaneously reducing energy consumption. Consequently, MoE emerges as a robust candidate for future scaling endeavors
\vspace{-3mm}
\item The model trained on filtered data shows consistently better performances on both NLG and NLU tasks, where the effect of filtering is more significant on the former tasks
\vspace{-3mm}
\item Filtered pretraining corpora play a crucial role in the generation capability of LLMs, especially for the downstream tasks
\vspace{-3mm}
\item The scaling of GLaM MoE models can be achieved by increasing the size or number of experts in the MoE layer. Given a fixed budget of computation, more experts contribute to a better performance
\end{itemize}}
\end{tabular}    \\ \cline{2-2}% \hline

\vspace{-3mm}
LaMDA & \begin{tabular}{c}
\multicolumn{1}{p{14.5cm}}{\begin{itemize} 
\vspace{-3mm}
\item The model can be fine-tuned to learn to call different external information resources and tools
\end{itemize}}
\end{tabular}    \\ \cline{2-2}%\hline

\vspace{-3mm}
AlphaCode & \begin{tabular}{c}
\multicolumn{1}{p{14.5cm}}{\begin{itemize} 
\vspace{-3mm}
\item For higher effectiveness and efficiency, a transformer model can be asymmetrically constructed with a shallower encoder and a deeper decoder
\vspace{-3mm}
\item To achieve better performances, it is necessary to employ strategies such as massively scaling upsampling, followed by the filtering and clustering of samples into a compact set
\vspace{-3mm}
\item The utilization of novel sampling-efficient transformer architectures designed to facilitate large-scale sampling is crucial
\vspace{-3mm}
\item Simplifying problem descriptions can effectively improve the model's performance
\end{itemize}}
\end{tabular}\\ \cline{2-2} %\hline

\vspace{-3mm}
Chinchilla & \begin{tabular}{c}
\multicolumn{1}{p{14.5cm}}{\begin{itemize} 
\vspace{-3mm}
\item The model size and the number of training tokens should be scaled proportionately: for each doubling of the model size, the number of training tokens should be doubled as well
\end{itemize}}
\end{tabular}    \\  \cline{2-2} %\hline

\vspace{-3mm}
PaLM &  \begin{tabular}{c}
\multicolumn{1}{p{14.5cm}}{\begin{itemize} 
\vspace{-3mm}
\item English-centric models produce better translations when translating to English as compared to non-English
\vspace{-3mm}
\item Generalized models can have equivalent performance for language translation to specialized small models
\vspace{-3mm}
\item Larger models have a higher percentage of training data memorization
\vspace{-3mm}
\item Performance has not yet saturated even at 540B scale, which means larger models are likely to perform better
\end{itemize}}
\end{tabular}   \\ \cline{2-2}%\hline

\vspace{-3mm}
AlexaTM & \begin{tabular}{c}
\multicolumn{1}{p{14.5cm}}{\begin{itemize} 
\vspace{-3mm}
\item Encoder-decoder architecture is more suitable to train LLMs given bidirectional attention to the context than decoder-only 
\vspace{-3mm}
\item Causal Language Modeling (CLM) task can be added to benefit the model with efficient in-context learning
%\vspace{-3mm}
%\item The key to training powerful seq2seq-based LLMs lies in mixed pre-training, rather than additional multitask training
\vspace{-3mm}
\item Placing layer norm at the beginning of each transformer layer improves the training stability
\end{itemize}}
\end{tabular}    \\ \cline{2-2}%\hline

%\\ \hline
\end{tabular}%

\vspace{-3mm}
\begin{flushright}
Table Continued on Next Page 
\end{flushright}
\end{table*}

%\clearpage
%\pagebreak

\begin{table*}[!tbhp]
%\ContinuedFloat
%\caption{(Continued on next page)}
\begin{tabular}{lc}
\hline \hline
\rowcolor{gray!50}Models & Findings  \& Insights\\ \hline \hline

\vspace{-3mm}
U-PaLM & \begin{tabular}{c}
\multicolumn{1}{p{14.5cm}}{\begin{itemize} 
\vspace{-3mm}
\item Training with a mixture of denoisers outperforms PaLM when trained further for a few more FLOPs 
\vspace{-3mm}
\item Training with a mixture of denoisers improves the infilling ability and open-ended text generation diversity
\end{itemize}}
\end{tabular}\\\cline{2-2}% \hline

\vspace{-3mm}
UL2 & \begin{tabular}{c}
\multicolumn{1}{p{14.5cm}}{\begin{itemize} 
\vspace{-3mm}
\item Mode switching training enables better performance on downstream tasks
\vspace{-3mm}
\item CoT prompting outperforms standard prompting for UL2  
\end{itemize}}
\end{tabular}    \\ \cline{2-2}%\hline

\vspace{-3mm}
GLM-130B &  \begin{tabular}{c}
\multicolumn{1}{p{14.5cm}}{\begin{itemize} 
\vspace{-3mm}
\item Pre-training data with a small proportion of multi-task instruction data improves the overall model performance
\end{itemize}}
\end{tabular}    \\ \cline{2-2}%\hline

\vspace{-3mm}
CodeGen & \begin{tabular}{c}
\multicolumn{1}{p{14.5cm}}{\begin{itemize} 
\vspace{-3mm}
\item Multi-step prompting for code synthesis leads to a better user intent understanding and code generation
\end{itemize}}
\end{tabular}    \\ \cline{2-2}%\hline

\vspace{-3mm}
LLaMA & \begin{tabular}{c}
\multicolumn{1}{p{14.5cm}}{\begin{itemize} 
\vspace{-3mm}
\item A constant performance improvement is observed when scaling the model
\vspace{-3mm}
\item Smaller models can achieve good performances with more training data and computing time
\end{itemize}}
\end{tabular}    \\ \cline{2-2}%\hline

\vspace{-3mm}
PanGu-$\Sigma$ & \begin{tabular}{c}
\multicolumn{1}{p{14.5cm}}{\begin{itemize} 
\vspace{-3mm}
\item Sparse models provide the benefits of large models at a lower computation cost
\vspace{-1mm}
\item Randomly Routed Experts reduces catastrophic forgetting effects which in turn is essential for continual learning 
\vspace{-3mm}
\item Randomly Routed Experts allow extracting a domain-specific sub-model in deployment which is cost-efficient while maintaining a performance similar to the original
\end{itemize}}
\end{tabular} \\\cline{2-2}    

\vspace{-3mm}
BloombergGPT & \begin{tabular}{c}
\multicolumn{1}{p{14.5cm}}{\begin{itemize} 
\vspace{-3mm}
\item Pre-training with general-purpose and task-specific data improves task performance without hurting other model capabilities     
\end{itemize}}
\end{tabular} \\ \cline{2-2}

\vspace{-3mm}
XuanYuan 2.0 & \begin{tabular}{c}
\multicolumn{1}{p{14.5cm}}{\begin{itemize} 
\vspace{-3mm}
\item Combining pre-training and fine-tuning stages in single training avoids catastrophic forgetting     
\end{itemize}}
\end{tabular} \\ \cline{2-2}

\vspace{-3mm}
CodeT5+ & \begin{tabular}{c}
\multicolumn{1}{p{14.5cm}}{\begin{itemize} 
\vspace{-3mm}
\item Causal LM is crucial for a model's generation capability in encoder-decoder architectures   
\vspace{-1mm}
\item Multiple training objectives like span corruption, Causal LM, matching, etc complement each other for better performance
\end{itemize}}
\end{tabular} \\ \cline{2-2}

\vspace{-3mm}
StarCoder & \begin{tabular}{c}
\multicolumn{1}{p{14.5cm}}{\begin{itemize} 
\vspace{-3mm}
\item HHH prompt by Anthropic allows the model to follow instructions without fine-tuning  
\end{itemize}}
\end{tabular} \\ \cline{2-2} %\hline

\vspace{-3mm}
LLaMA-2 & \begin{tabular}{c}
\multicolumn{1}{p{14.5cm}}{\begin{itemize} 
\vspace{-3mm}
\item Model trained on unfiltered data is more toxic but may perform better on downstream tasks after fine-tuning
\vspace{-3mm}
\item Model trained on unfiltered data requires fewer samples for safety alignment 
\end{itemize}}
\end{tabular} \\ \cline{2-2} %\hline

\vspace{-3mm}
PaLM-2 & \begin{tabular}{c}
\multicolumn{1}{p{14.5cm}}{\begin{itemize} 
\vspace{-3mm}
\item Data quality is important to train better models
\vspace{-3mm}
\item Model and data size should be scaled with 1:1 proportions
\vspace{-3mm}
\item Smaller models trained for larger iterations outperform larger models
\end{itemize}}
\end{tabular} \\ \cline{2-2}

\vspace{-3mm}
LLaMA-3/3.1 & \begin{tabular}{c}
\multicolumn{1}{p{14.5cm}}{\begin{itemize} 
\vspace{-3mm}
\item Increasing batch size gradually stabilizes the training without loss spikes
\vspace{-3mm}
\item High-quality data at the final stages of training improves the model performance
\vspace{-3mm}
\item Increasing model context length windows step-wise allows it to better adapt to various sequence lengths  
\end{itemize}}
\end{tabular} \\ \cline{2-2}
%\\ \hline

\vspace{-3mm}
Nemotron-40B & \begin{tabular}{c}
\multicolumn{1}{p{14.5cm}}{\begin{itemize} 
\vspace{-3mm}
\item Model aligned iteratively on synthetic data with data generated from the previously aligned model achieves competitive performance  
\end{itemize}}
\end{tabular} \\ \cline{2-2}

\vspace{-3mm}
DeepSeek & \begin{tabular}{c}
\multicolumn{1}{p{14.5cm}}{\begin{itemize} 
\vspace{-3mm}
\item Batch size should increase with the increase in compute budget while decreasing the learning rate
\end{itemize}}
\end{tabular} \\ \cline{2-2}

\vspace{-3mm}
DeepSeek-v2 & \begin{tabular}{c}
\multicolumn{1}{p{14.5cm}}{\begin{itemize} 
\vspace{-3mm}
\item Mult-head latent attention (MLA) performs better than multi-head attention (MHA) while requiring a significantly smaller KV cache, therefore achieving faster data generation 
\end{itemize}}
\end{tabular} \\ \hline

\vspace{-3mm}
\end{tabular}
\end{table*}

%\clearpage
%\pagebreak

%InstructGPT & \begin{tabular}{c}
%\multicolumn{1}{p{14.5cm}}{\begin{itemize}
%\item InstructGPT is trained using a novel RLHF technique which produces outputs that are preferred to the outputs from its base model of GPT-3 (175B), despite having 100x fewer parameters.
%\item The human feedback step-trained models show improvements in truthfulness and reductions in toxic output generation while having minimal performance regressions on public NLP datasets
%\end{itemize}}
%\end{tabular}    \\ \hline %\hline

\begin{table*}[!tbhp]
\caption{Key insights and findings from the study of \emph{instruction-tuned} Large Language Models. }
%\caption{Important findings of instruction-tuned LLMs}

\begin{tabular}{lccl}
\hline \hline
\rowcolor{gray!50}Models & Findings \& Insights \\ \hline \hline
\vspace{-3mm}
T0 & \begin{tabular}{c}
\multicolumn{1}{p{14.5cm}}{\begin{itemize}
\vspace{-3mm}
\item Multi-task prompting enables zero-shot generalization and outperforms baselines    
\vspace{-3mm}
\item Even a single prompt per dataset task is enough to improve performance
\end{itemize}
}
\end{tabular}\\ \cline{2-2}%\hline
\vspace{-3mm}
WebGPT & \begin{tabular}{c}
\multicolumn{1}{p{14.5cm}}{\begin{itemize}
\vspace{-3mm}
\item To aid the model in effectively filtering and utilizing relevant information, human labelers play a crucial role in answering questions regarding the usefulness of the retrieved documents
\vspace{-3mm}
\item Interacting a fine-tuned language model with a text-based web-browsing environment can improve end-to-end retrieval and synthesis via imitation learning and reinforcement learning
\vspace{-3mm}
\item Generating answers with references can make labelers easily judge the factual accuracy of answers
\end{itemize}}
\end{tabular}    \\ \cline{2-2}%\hline
\vspace{-3mm}
Tk-INSTRUCT & \begin{tabular}{c}
\multicolumn{1}{p{14.5cm}}{\begin{itemize}
\vspace{-3mm}
\item Instruction tuning leads to a stronger generalization of unseen tasks 
\vspace{-3mm}
\item More tasks improve generalization whereas only increasing task instances does not help 
\vspace{-3mm}
\item Supervised trained models are better than generalized models 
\vspace{-3mm}
\item Models pre-trained with instructions and examples perform well for different types of inputs 
\end{itemize}}
\end{tabular}    \\ \cline{2-2}%\hline

\vspace{-3mm}
mT0 and BLOOMZ & \begin{tabular}{c}
\multicolumn{1}{p{14.5cm}}{\begin{itemize} 
\vspace{-3mm}
\item Instruction tuning enables zero-shot generalization to tasks never seen before    
\vspace{-3mm}
\item Multi-lingual training leads to even better zero-shot generalization for both English and non-English
\vspace{-3mm}
\item Training on machine-translated prompts improves performance for held-out tasks with non-English prompts  
\vspace{-3mm}
\item English only fine-tuning on multilingual pre-trained language model is enough to generalize to other pre-trained language tasks
\end{itemize}}
\end{tabular}    \\ \cline{2-2} 

\vspace{-3mm}
OPT-IML & \begin{tabular}{c}
\multicolumn{1}{p{14.5cm}}{\begin{itemize}
\vspace{-3mm}
\item Creating a batch with multiple task examples is important for better performance 
\vspace{-3mm}
\item Only example proportional sampling is not enough, training datasets should also be proportional for better generalization/performance
\vspace{-3mm}
\item Fully held-out and partially supervised tasks performance improves by scaling tasks or categories whereas fully supervised tasks have no effect
\vspace{-3mm}
\item Including small amounts i.e.~5\% of pretraining data during fine-tuning is effective 
\vspace{-3mm}
\item Only 1\% reasoning data improves the performance, adding more deteriorates performance
\vspace{-3mm}
\item Adding dialogue data makes the performance worse

%\item Meta training for in-context learning degrades performance??? It improves performance for other LLMs though.
\end{itemize}}
\end{tabular}    \\ \cline{2-2}%\hline

\vspace{-3mm}
Sparrow & \begin{tabular}{c}
\multicolumn{1}{p{14.5cm}}{\begin{itemize} 
\vspace{-3mm}
\item Labelers' judgment and well-defined alignment rules help the model generate better responses
\vspace{-3mm}
\item Good dialogue goals can be broken down into detailed natural language rules for the agent and the raters
\vspace{-3mm}
\item The combination of reinforcement learning (RL) with reranking yields optimal performance in terms of preference win rates and resilience against adversarial probing
\end{itemize}}
\end{tabular}    \\ \cline{2-2}

\vspace{-3mm}
Flan & \begin{tabular}{c}
\multicolumn{1}{p{14.5cm}}{\begin{itemize}
\vspace{-3mm}
\item Finetuning with CoT improves performance on held-out tasks 
\vspace{-3mm}
\item Fine-tuning along with CoT data improves reasoning abilities 
\vspace{-3mm}
\item CoT tuning improves zero-shot reasoning 
\vspace{-3mm}
\item Performance improves with more tasks 
\vspace{-3mm}
\item Instruction fine-tuning improves usability which otherwise is challenging for pre-trained models
\vspace{-3mm}
\item Improving the model's performance with instruction tuning is compute-efficient 
\vspace{-3mm}
\item Multitask prompting enables zero-shot generalization abilities in LLM
\end{itemize}}
\end{tabular}\\ \cline{2-2}

\vspace{-3mm}
WizardCoder & \begin{tabular}{c}
\multicolumn{1}{p{14.5cm}}{\begin{itemize}
\vspace{-3mm}
\item Fine-tuning with re-written instruction-tuning data into a complex set improves performance   
\end{itemize}}
\end{tabular}\\ \cline{2-2}

\vspace{-3mm}
LLaMA-2-Chat & \begin{tabular}{c}
\multicolumn{1}{p{14.5cm}}{\begin{itemize}
\vspace{-3mm}
\item Model learns to write safe responses with fine-tuning on safe demonstrations, while additional RLHF step further improves model safety and make it less prone to jailbreak attacks       
\end{itemize}}
\end{tabular}\\ \cline{2-2}

\vspace{-3mm}
LIMA & \begin{tabular}{c}
\multicolumn{1}{p{14.5cm}}{\begin{itemize}
\vspace{-3mm}
\item Less high quality data is enough for fine-tuned model generalization
\end{itemize}}
\end{tabular}\\ \hline %\cline{2-2}

%\\ \hline
\end{tabular}%

 \label{tab:instruction_tuned_findings}
\end{table*}

%\clearpage
%\pagebreak

\subsection{Fine-Tuned LLMs}
Pre-trained LLMs have excellent generalization abilities to unseen tasks. However, because they are generally trained with the objective of next token prediction, LLMs have limited capacity to follow user intent and are prone to generate unethical, toxic or inaccurate responses~\cite{instructgpt}. For their effective utilization, LLMs are fine-tuned to follow instructions~\cite{Flan, T0, OPT_IML} and generate safe responses~\cite{instructgpt}, which also results in increasing zero-shot, few-shot, and cross-task generalization~\cite{OPT_IML, Flan, Tk-INSTRUCT}, with minimal compute increment, e.g., 0.2\% of the total pre-training for PaLM 540B~\cite{Flan}. \\
We review various fine-tuned LLMs and strategies for effective fine-tuning in this section.
\subsubsection{Instruction-Tuning with Manually Created Datasets}
Numerous hand-crafted instruction-tuning datasets with different design choices are proposed in the literature to instruction-tune LLMs. The performance of fine-tuned LLMs depends on multiple factors, such as dataset, instruction  diversity, prompting templates, model size, and training objectives. Keeping this in view, diverse fine-tuned models have emerged in the literature using manually created datasets. \\
The models T0~\cite{T0} and mT0 (multi-lingual)~\cite{mT0andBLOOMZ} employ templates to convert existing datasets into prompt datasets. They have shown improvements in generalization to zero-shot and held-out tasks. Tk-Instruct~\cite{Tk-INSTRUCT} fine-tuned the T5 model with in-context instructions to study generalization on unseen tasks when given in-context instructions during test time. The model outperformed Instruct-GPT, despite being smaller in size, i.e., 11B parameters as compared to 175B of GPT-3.  \\
\emph{\textbf{Increasing Tasks and Prompt Setups:}} 
Zero-shot and few-shot performance improves significantly by expanding task collection and prompt styles. OPT-IML~\cite{OPT_IML} and Flan~\cite{Flan} curated larger 2k and 1.8k task datasets, respectively. While increasing task size alone is not enough, OPT-IML and Flan add more prompting setups in their datasets, zero-shot, few-shot, and CoT. In continuation, CoT Collection~\cite{cot_collection} fine-tunes Flan-T5 further on 1.88M CoT samples. Another method~\cite{zero_to_hero} uses symbolic tasks with tasks in T0, Flan, etc.   \\

\begin{figure}[!tbp]
\centering
\includegraphics[width=1\columnwidth]{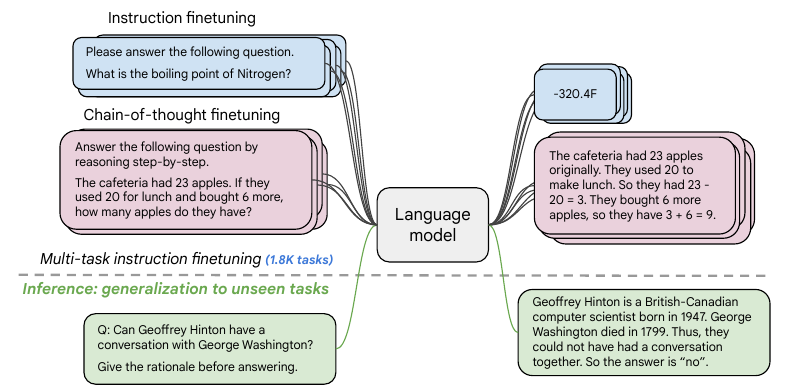}
\caption{An example image shows an instance of the Flan training paradigm, taken from~\cite{Flan}.}
\label{flan_image}
\end{figure}
%\paragraph{OPT-IML~\cite{OPT_IML}}
%An instruction-tuned OPT model, trained on instruction meta-learning benchmark of 2000 NLP tasks that is a combination of 8 meta-datasets including, Super-NaturalInstructions, PromptSource, FLAN, and others as given in Table~\ref{datasets}. For computational efficiency, OPT-IML utilizes the maximum sequence length of 2048 tokens by packing multiple instances together, separated by the $<eos>$ token during training. It employs a masking mechanism to separate instances in a sequence to avoid attending tokens from different instances. Overall, OPT-IML outperforms baseline model OPT with instruction-finetuning on zero and few-shot generalization abilities.

%\paragraph{Flan~\cite{Flan}}
%Fine-tuning language models (Flan), fine-tunes T5, PaLM, and UPaLM with 1836 instruction tasks taken from Muffin (80 tasks), T0-SF (193 tasks), NIV2 (1554 tasks), and CoT (taken from nine datasets), as shown in Figure~\ref{flan_image}. Instruction fine-tuning improves the model performance significantly with minimal computing, only 0.2\% of the total pre-training compute in the case of PaLM 540B. Flan also suggests that adding more instruction fine-tuning tasks with CoT reasoning data will likely improve the performance further.   

\subsubsection{Instruction-Tuning with LLMs Generated Datasets}
Generating an instruction-tuning dataset requires carefully writing instructions and input-output pairs, which are often written by humans, smaller in size, and less diverse. To overcome this, self-instruct~\cite{ft_self_instruct} proposed an approach to prompt available LLMs to generate instruction-tuning datasets. Self-instruct outperformed models trained on manually created dataset SUPER-NATURALINSTRUCTIONS (a dataset with 1600+ tasks)~\cite{Tk-INSTRUCT} by 33\%. It starts with a seed of 175 tasks, 1 instruction, and 1 sample per task and iteratively generates new instructions (52k) and instances (82k input-output pairs) using GPT-3~\cite{GPT-3}. Contrary to this, Dynosaur~\cite{dynosaur} uses the meta-data of datasets on Huggingface to prompt LLMs to generate multiple task instruction-tuning datasets. \\ 
\emph{\textbf{LLaMA Tuned:}} Various models in the literature instruction-tune LLaMA~\cite{gao2023llama} with GPT-3~\cite{GPT-3} or GPT-4~\cite{GPT-4} generated datasets.  Among these, Alpaca~\cite{alpaca}, Vicuna~\cite{vicuna}, and LLaMA-GPT-4~\cite{llama_gpt_4} are a few general-purpose fine-tuned models, where Alpaca is trained on 52k samples from text-davinci-003, Vicuna on 70k samples from ShareGPT.com, and LLaMA-GPT-4 by re-creating Alpaca instructions from GPT-4. Goat~\cite{goat} fine-tunes LLaMA for arithmetic tasks (1 million samples) by generating data from ChatGPT and outperforms GPT-4, PaLM, BLOOM, OPT, etc., attributing its success to the LLaMA's consistent tokenization of numbers. HuaTuo~\cite{huatuo} is a medical knowledge model, fine-tuned with a generated QA dataset of 8k instructions.  \\
\emph{\textbf{Complex Instructions:}} Evol-Instruct~\cite{wizardlm, wizardcoder} prompts LLMs to convert given instructions into a more complex set. The instructions are iteratively evolved with re-writing instructions in complex wording and creating new instructions. With this style of automated instruction generation, WizardLM~\cite{wizardlm} (fine-tuned LLaMA on 250k instructions), outperforms Vicuna and Alpaca, and WizardCoder~\cite{wizardcoder} (fine-tuned StarCoder) beats Claude-Plus, Bard, and others.

~\subsubsection{Aligning with Human Preferences}
Incorporating human preferences into LLMs presents a significant advantage in mitigating undesirable behaviors and ensuring accurate outputs. The initial work on alignment, such as InstructGPT~\cite{instructgpt} aligns GPT-3 using a 3-step approach, instruction-tuning, reward modeling, and fine-tuning with reinforcement learning (RL). The supervised fine-tuned GPT-3 on demonstrations is queried to generate responses, which human labelers rank according to human values, and a reward model is trained on the ranked data. Lastly, the GPT-3 is trained with proximal policy optimization (PPO) using rewards on the generated data from the reward model. LLaMA 2-Chat~\cite{llama_2} improves alignment by dividing reward modeling into helpfulness and safety rewards and using rejection sampling in addition to PPO. The initial four versions of LLaMA 2-Chat are fine-tuned with rejection sampling and then with PPO on top of rejection sampling. \\ 
~\emph{~\textbf{Aligning with Supported Evidence:}} This style of alignment allows the model to generate responses with proofs and facts, reduces hallucination, and assists humans more effectively, which increases trust in the model's output. Similar to the RLHF training style, a reward model is trained to rank generated responses containing web citations in answers to questions, which is later used to train the model, as in GopherCite~\cite{gopher_cite}, WebGPT~\cite{nakano2021webgpt}, and Sparrow~\cite{sparrow}. The ranking model in Sparrow~\cite{sparrow} is divided into two branches, preference reward and rule reward, where human annotators adversarial probe the model to break a rule. These two rewards together rank a response to train with RL. \\
~\emph{~\textbf{Aligning Directly with SFT:}} The PPO in the RLHF pipeline is complex, memory-intensive, and unstable, requiring multiple models, reward, value, policy, and reference models. Avoiding this sophisticated alignment pipeline is possible by incorporating minimal changes in the supervised fine-tuning (SFT) pipeline 
as in~\cite{DPO,raft,rrhf}, with better or comparable performance to PPO. Direct preference optimization (DPO)~\cite{DPO} trains a model directly on the human-preferred responses to maximize the likelihood of preferred against unpreferred responses, with per-sample importance weight. Reward ranked fine-tuning RAFT~\cite{raft} fine-tunes the model on ranked responses by the reward model. Preference ranking optimization (PRO)~\cite{PRO} and RRHF~\cite{rrhf} penalize the model to rank responses with human preferences and supervised loss. On the other hand, chain-of-hindsight (CoH)~\cite{CoH} provides feedback to the model in language rather than reward, to learn good versus bad responses.\\
~\emph{\textbf{Aligning with Synthetic Feedback:}} Aligning LLMs with human feedback is slow and costly. The literature suggests a semi-automated process to align LLMs by prompting LLMs to generate helpful, honest, and ethical responses to the queries, and fine-tuning using the newly created dataset. Constitutional AI~\cite{constitutional_ai} replaces human feedback in RLHF with AI, calling it RL from AI feedback (RLAIF). AlpacaFarm~\cite{alpacafarm} designs prompts to imitate human feedback using LLMs APIs. Opposite to constitutional AI, AlpacaFarm injects noise in feedback to replicate human mistakes. Self-Align~\cite{self_align} prompts the LLM with ICL examples, instructing the LLM about what the response should contain to be considered useful and ethical. The same LLM is later fine-tuned with the new dataset. \\
~\emph{~\textbf{Aligning with Prompts:}} LLMs can be steered with prompts to generate desirable responses without training~\cite{align_with_prompts, aling_with_prompts_2}. The self-correction prompting in~\cite{aling_with_prompts_2} concatenates instructions and CoT with questions, guiding the model to answer its instruction following a strategy to ensure moral safety before the actual answer. This strategy is shown to reduce the harm in generated responses significantly. \\
~\emph{~\textbf{Red-Teaming/Jailbreaking/Adversarial Attacks:}} LLMs exhibit harmful behaviors, hallucinations, leaking personal information, and other shortcomings through adversarial probing. The models are susceptible to generating harmful responses even though they are aligned for safety~\cite{jailbroken, red_team_lessons_learned}. Red-teaming is a common approach to address illicit outputs, where the LLMs are prompted to generate harmful outputs~\cite{red_team_lessons_learned, red_team_explore}. The dataset collected through red-teaming is used to fine-tune models for safety. While red-teaming largely relies on human annotators, another work~\cite{red_team_language_models} red-team LLMs to find prompts that lead to harmful outputs for other LLMs.  \\

\subsubsection{Continue Pre-Training}
Although fine-tuning boosts a model's performance, it leads to catastrophic forgetting of previously learned information. Concatenating fine-tuning data with a few randomly selected pre-training samples in every iteration avoids network forgetting~\cite{cont_learn, xuanyuan}. This is also effective in adapting LLMs for cases where fine-tuning data is small and the original capacity is to be maintained. Prompt-based continued pre-training (PCP)~\cite{cont_learn_dont_stop_pretrain} trains the model with text and instructions related to tasks and then finally instruction-tunes the model for downstream tasks.
\subsubsection{Sample Efficiency}
While fine-tuning data is generally many-fold smaller than the pre-training data, it still has to be large enough for acceptable performance~\cite{Flan,OPT_IML, Tk-INSTRUCT} and requires proportional computing resources. Studying the effects on performance with less data, existing literature~\cite{sample_eff_inst_quick_learner, sample_eff_maybe_less_data} finds that models trained on less data can outperform models trained with more data. In~\cite{sample_eff_inst_quick_learner}, 25\% of the total downstream data is found enough for state-of-the-art performance. 
Selecting coreset-based 0.5\% of the total instruction-tuning data improves the model performance by 2\%  in~\cite{sample_eff_maybe_less_data}, as compared to the complete data tuning. %This is not making sense, please rephrase.
Less is more for alignment (LIMA)~\cite{lima} uses only 1000 carefully created demonstrations to fine-tune the model and has achieved comparable performance to GPT-4.

\subsection{Increasing Context Window}
LLMs are trained with limited context windows due to expensive attention and high memory requirements. A model trained on limited sequence lengths fails to generalize to unseen lengths at inference time~\cite{wt_lm_infinite, pe_extending}. Alternatively, LLMs with ALiBi~\cite{alibi} positional encodings can perform zero-shot length extrapolation. However, ALiBi has less expressive power~\cite{su2021roformer} and inferior performance on multiple benchmarks~\cite{giraffe}, and many LLMs use RoPE positional embedding that is unable to perform zero-shot extrapolation. A larger context length has benefits such as a better understanding of longer documents, more samples in in-context learning, execution of bigger reasoning processes, etc. Expanding context length during fine-tuning is slow, inefficient, and computationally expensive~\cite{pe_extending}. Therefore, researchers employ various context window extrapolation techniques discussed below. \\
\emph{\textbf{Position Interpolation:}}
Rather than extrapolating, \cite{pe_extending} shows that interpolating position encodings within the pre-trained context window are more effective. The work demonstrates that only 1000 steps of fine-tuning are enough to achieve better results on larger windows without reducing performance compared to the original context size. Giraffe~\cite{giraffe} uses power scaling in RoPE, and YaRN~\cite{yarn} proposed NTK-aware interpolation.   \\ 
\emph{\textbf{Efficient Attention Mechanism:}}
Dense global attention is one of the major constraints in training larger context window LLMs. Using efficient attention variants, such as local, sparse, and dilated attention, reduces the computation cost significantly. LongT5~\cite{longt5} proposes transient global attention (TGlobal), applying attention to local and global tokens (windowed token averaging). The model replaces attention in T5~\cite{T5} with TGlobal attention, pre-trains the model on 4098 sequence length, fine-tunes on larger window sizes, as large as 16k, and improves task performance on longer inputs. This shows the extrapolation ability of TGlobal attention with only fine-tuning. COLT5~\cite{colt5} uses two branches, one with lightweight and the other with heavyweight attention and feed-forward layers. All tokens are processed from the lightweight branch, and only important tokens are routed to the heavyweight branch. LongNet~\cite{longnet} replaces standard attention with dilated attention, expanding sequence length to 1 billion tokens. LongLoRA~\cite{longlora} proposes shift-short attention, used during fine-tuning to reduce dense attention costs. However, the model during inference uses dense attention and achieves similar performance as full attention fine-tuning.\\
\emph{\textbf{Extrapolation without Training:}}
LM-Infinite~\cite{wt_lm_infinite} and parallel context windows (PCW)~\cite{PCW} show length extrapolation is possible using pre-trained LLMs. LM-Infinite suggested $\Lambda$-shaped attention applied within the original context window limits. Likewise, PCW chunks larger inputs into the pre-trained context lengths and applies the same positional encodings to each chunk.

\subsection{Augmented LLMs} 
LLMs are capable of learning from the examples concatenated with the input, known as context augmentation, in-context learning (ICL), or few-shot prompting. They show excellent generalization to unseen tasks with few-shot prompting, enabling LLMs to answer queries beyond the capacity acquired during training~\cite{GPT-3,survey_reasoning}. These emergent abilities allow for adapting the model without fine-tuning---a costly process. Aside from this, hallucination, producing inaccurate, unsafe, or factually incorrect responses, is common for LLMs, which is avoided by augmenting contextual data. While the user can provide in-context samples in the query~\cite{survey_incontext_learning,prompt_eng_guide}, here we specifically refer to the methods that access external storage programmatically, calling them augmented LLMs.\\
The literature suggests various external memory designs to augment LLMs, long-term~\cite{mem_augmenting_long, mem_augmenting_long_conver,retro, memorybank}, short-term~\cite{reflexion}, symbolic~\cite{chatdb}, and non-symbolic~\cite{flare,in_context_ralm}. The memory can be maintained in different formats such as documents, vectors, or databases. A few systems maintain intermediate memory representations to retain information across multiple iterations~\cite{memorybank, mem_augmenting_long_conver}, while others extract important information from the datasets and save it in memory for recall~\cite{mot}. The memory read and write operations are performed either with or without LLMs cooperation~\cite{mem_augmenting_long_conver,rw_memory, memorybank,retLLM}, acting as a feedback signal in~\cite{reflexion}. 
We discuss different types of augmented LLMs below.
\begin{figure}[t!]
\centering
\includegraphics[width=1\columnwidth]{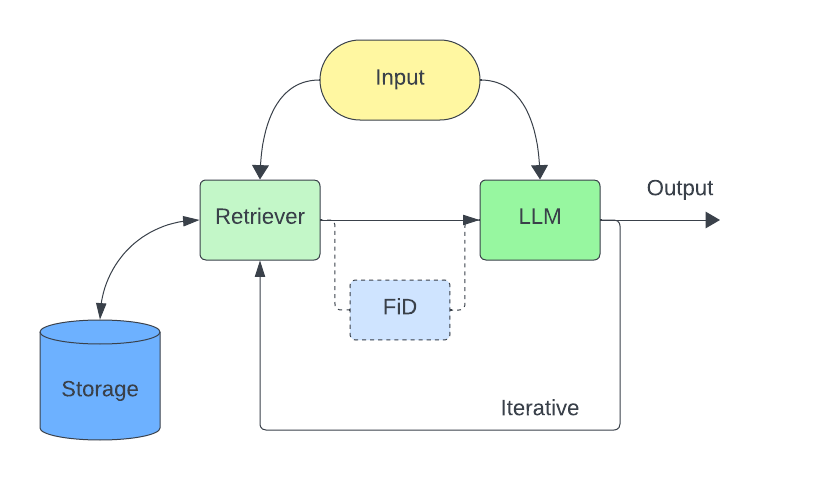}
\caption{A flow diagram of Retrieval Augmented \textcolor{black}{LLMs.} The retriever extracts a similar context to the input and forwards it to the LLM either in simple language or encoded through Fusion-in-Decoder (FiD). Depending on the task, retrieval and generation may repeat multiple times. %NA: Better make caption a little more self-contained (more explanatory) because in long papers, figures are viewed independent of the text.
}
\label{rag_image}
\end{figure}

\subsubsection{Retrieval Augmented LLMs}
LLMs may have limited memory and outdated information, leading to inaccurate responses. Retrieving relevant information from external up-to-date storage enables the LLMs to accurately answer with references and utilize more information. With retrieval augmentation, smaller models have been shown to perform at par with larger models. For instance, the 11B model can become competitive to 540B PaLM in~\cite{atlas} and 7.5B to 280B Gopher in~\cite{retro}. Retrieval augmented language modeling (RALM) has two major components, shown in Figure~\ref{rag_image}, namely: 1) retriever and 2) language model. In RALM, the retriever plays a crucial role in driving LLM response, where incorrect information can steer LLMs to false behavior. This leads to the development of various methods to retrieve accurate information and fuse with the query for better performance.\\
\emph{\textbf{Zero-Shot Retrieval Augmentation:}}
This kind of augmentation keeps the original LLM architecture and weights unchanged and uses BM25~\cite{BM25}, nearest neighbors, or frozen pre-trained models like Bert~\cite{Bert} as a retriever. The retrieved information is provided as input to the model for response generation, shown to improve performance over LLMs without retrieval~\cite{in_context_ralm, rationale_aug}. 
In some scenarios, multiple retrieval iterations are required to complete the task. The output generated in the first iteration is forwarded to the retriever to fetch similar documents. Forward-looking active retrieval (FLARE)~\cite{flare} initially generates the response and corrects the output by retrieving relevant documents if the response contains low-confidence tokens. Similarly, RepoCoder~\cite{repocoder} fetches code snippets recursively for code completion. \\
\emph{\textbf{Training with Retrieval Augmentation:}}
To reduce failures in retrieval augmentation generation (RAG), researchers train or fine-tune retrievers and LLMs with a retrieval augmentation pipeline. We discuss the literature below based on their focus on the respective training processes of the pipeline. \\
\textit{Training LLM:} Retrieval-enhanced transformer (RETRO)~\cite{retro} shows pre-training smaller LLMs with RAG pipeline outperforms larger LLMs, such as GPT-3 trained without RAG. RETRO uses a 2-trillion token subset of MassiveText as a database. The retrieval pipeline divides the input query into subsets and retrieves relevant chunks from the database for each subset, encoded together with input intermediate representations for generating tokens. It uses cross-chunked attention to attend to previous chunks auto-regressively. A study on RETRO~\cite{retro_study} shows models pre-trained without RAG but fine-tuned using RAG lack the performance gains obtained by pre-training with RAG. \\
\textit{Training Retriever:} Quality of responses generated by LLMs is highly dependent on the in-context examples. Therefore,~\cite{retrieve_icl_train, retrieve_icl_GPT_3, retrieve_prompts_learning, replug} train retrievers to retrieve accurate few-shot samples while keeping the LLM frozen for generation. Retrieved samples are ranked to build ground-truth data to train retrievers with contrastive learning in~\cite{retrieve_icl_train, retrieve_prompts_learning}. RoBERTa is trained for downstream tasks in~\cite{retrieve_icl_GPT_3} for ICL samples retrieval. REPLUG~\cite{replug} trains the retriever with supervised signals from the frozen LLM-generated outputs.   \\
\textit{Training Retriever and LLM:} Further benefits are achieved by training both the retriever and the model in~\cite{atlas, rag_train_long,realm}. In this case, the error propagates back to the retriever, updating both the language model and the retriever. While masked language modeling (MLM) is a common pre-training objective~\cite{atlas, realm}, retrieval pre-trained transformer (RPT)~\cite{rag_train_long} used document chunk prediction as a pre-training objective for long text modeling. \\
\emph{\textbf{Encoded Context Augmentation:}} 
Concatenating retrieved documents with the query becomes infeasible as the sequence length and sample size grow. Encoding the context and fusing it with the decoder (Fusion-in-Decoder) using cross-attention makes it possible to augment more samples without increasing computation costs significantly~\cite{fid,retro, rag_train_long, atlas}. \\
\emph{\textbf{Web Augmented:}} 
Locally stored memory, but external to LLM, has limited information. However, a large amount of information is available on the internet, which gets updated regularly. Rather than storing information locally, various methods retrieve query-related context through a web search and forward it to LLMs~\cite{internet_aug, internet_aug_2, nakano2021webgpt}.  \\

\subsubsection{Tool Augmented LLMs}
\begin{figure}[t]
\centering
\includegraphics[width=1\columnwidth]{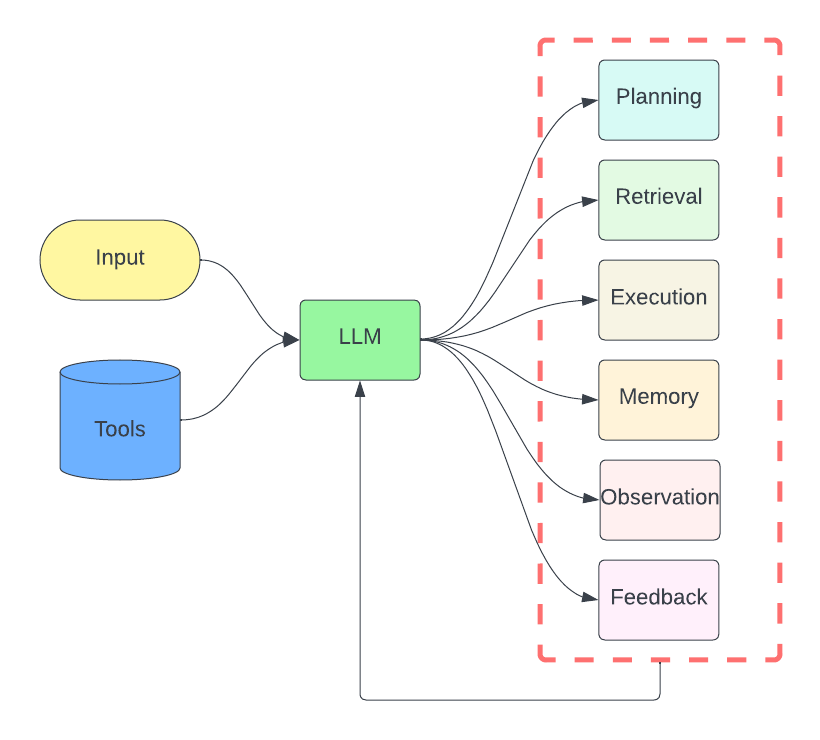}
\caption{A basic flow diagram of tool augmented \textcolor{black}{LLMs.} Given an input and a set of available tools, the model generates a plan to complete the task. The tool augmented LLMs utilize different modules iteratively, such as retriever, tool execution, read-write to memory, feedback, etc., depending on the task.  %NA: The figure is not telling much, especially considering the details in the text that refers to it. Perhaps it needs re-drawing and a more detailed caption.
}
\label{tallm_image}
\end{figure}

While RAG relies on the retriever to provide context to the LLM to answer queries, tool augmented LLMs capitalize on the reasoning abilities of LLMs to iteratively plan by dividing tasks into sub-tasks, selecting necessary tools, and taking actions to complete the task~\cite{assistgpt,chameleon,ART,talm}. A generic pipeline of tool-augmented LLMs is shown in Figure~\ref{tallm_image}, where different modules in Figure~\ref{tallm_image} are selected in a loop until the task completion.
%NA: We need some references here. There are none in the whole para.  \\ 

\noindent\emph{\textbf{Zero-Shot Tool Augmentation:}} LLMs in-context learning and reasoning abilities enable them to interact with tools without training. Automatic reasoning and tool-use (ART)~\cite{ART} builds a task library with demonstrations of reasoning steps and calling external tools. It retrieves similar task examples and provides the context to the LLM for inference. Aside from this,~\cite{tool_aug_doc} shows tool documentation is enough to teach LLMs to use tools without demonstrations. RestGPT~\cite{RestGPT} integrates LLMs with RESTful APIs by decomposing tasks into planning and API selection steps. The API selector understands the API documentation to select a suitable API for the task and plan the execution. ToolkenGPT~\cite{ToolkenGPT} uses tools as tokens by concatenating tool embeddings with other token embeddings. During inference, the LLM generates the tool tokens representing the tool call, stops text generation, and restarts using the tool execution output.   \\
\emph{\textbf{Training with Tool Augmentation:}} LLMs are trained to interact with diverse tools, enhancing planning abilities to overcome the limitations of zero-shot tool augmentation~\cite{gorilla, talm, tool_manipulation_cap, toolllm}. Gorilla~\cite{gorilla} instruction-tunes LLaMA with information retrieval from API documentation. It uses the self-instruct~\cite{ft_self_instruct} data generation pipeline with GPT-4 by providing in-context examples retrieved from API documentation. Tool augmented language model (TALM)~\cite{talm} fine-tunes T5~\cite{T5} for tool use with a self-play approach, where it iteratively completes tool manipulation tasks and includes them back in the training set. ToolLLM~\cite{toolllm} collects 16k APIs from RapidAPI. It samples APIs from the list to generate an instruction-tuning dataset using ChatGPT in single-tool and multi-tool scenarios. For high-quality datasets, ToolLLM suggested a depth-first search-based decision tree (DFSDT) method to generate ground-truths with diverse reasoning and planning. \\
\emph{\textbf{Multimodal Tool Augmentation:}} The compositional reasoning capacity of LLMs allows them to manipulate tools in multimodal settings~\cite{assistgpt,chameleon,hugginggpt}. Following the pipeline shown in Figure~\ref{tallm_image}, the LLM outlines a plan, generally executing in a sequence: Plan $\,\to\,$ Tool selection $\,\to\,$ Execute $\,\to\,$ Inspect $\,\to\,$ Generate, to respond to the user query. Here, the database of tools is rich in modalities, including text, images, etc. Many of the multimodal tool augmentation systems employ multimodal LLMs~\cite{gpt4tools,taskmatrix, hugginggpt, chameleon}, while others utilize single modality LLMs and generate a plan on using different modality tools to solve multimodal queries~\cite{vipergpt}.

\subsection{LLMs-Powered Agents}
AI agents are autonomous entities, capable of planning, decision-making, and performing actions to achieve complex goals. In the early days, AI agents were rule-based, designed for narrow tasks, and had limited capabilities, such as Clippy~\cite{clippy} and Deep Blue~\cite{deepblue}. In contrast to this, LLMs abilities to respond to dynamic scenarios have made it possible to incorporate them in diverse applications, including LLMs-powered agents~\cite{hugginggpt,chameleon}, where LLMs behave as the brain of agents. LLMs have been incorporated in web agents~\cite{nakano2021webgpt, sparrow}, coding agents~\cite{metagpt}, tool agents~\cite{talm,toolllm}, embodied agents~\cite{PaLME}, and conversational agents~\cite{reflexion}, requiring minimal to no fine-tuning". Below we summarize the research in LLMs-based autonomous agents. For a more detailed discussion, please refer to~\cite{survey_agents,survey_agents_2}. \\
\emph{\textbf{LLMs Steering Autonomous Agents:}} 
LLMs are the cognitive controllers of the autonomous agents. They generate plans, reason about tasks, incorporate memory to complete tasks, and adapt the outline depending on the feedback from the environment. Depending on the acquired capabilities of LLMs, many methods fine-tune, propose a better prompting approach, or utilize different modules to enhance agents' performance. Modules and strategies employed in autonomous agents are briefly discussed below. \\
\emph{Planning and Reasoning:} Completing a complex task requires human-like logical thinking, planning necessary steps, and reasoning current and future directions. Prompting methods like chain-of-thoughts~\cite{wei2022chain}, tree-of-thoughts~\cite{ToT}, and self-consistency~\cite{self_consistency} are central to agents, eliciting LLMs to reason its actions and choose among different paths for task completion. When LLMs are prompted with a task description and a sequence of actions, they can accurately generate plan actions without any fine-tuning~\cite{LLMs_zs_planners}. Reasoning via planning (RAP)~\cite{RAP} incorporates a re-purposed LLM as a world model to reason about future outcomes and explore alternative paths for task completion. Retroformer~\cite{retroformer} uses a retrospective LLM to improve main LLM planning and reasoning capabilities by providing helpful task cues. \\ 
\emph{Feedback:} LLMs in open-loop systems generate plans and assume that the agent will complete them successfully. However, the actual scenario is different with failures and variable responses from the environment. To correctly complete tasks, many methods use LLMs in a closed-loop where the action response is provided as feedback to the LLMs to re-assess and update the plan as required~\cite{inner_monologue,alphablock,progprompt,reflexion}. Another direction of research exploits LLMs as reward functions to train reinforcement learning (RL) policies instead of humans~\cite{LLM_rewards}.     \\
\emph{Memory:} LLMs can learn from the context provided in the prompt. In addition to internal memory, various systems employ external memory to save the response history. Reflexion~\cite{reflexion} maintains an episodic memory to use previous responses as feedback to improve future decision-making. Retroformer~\cite{retroformer} improves its responses by employing short-term and long-term memory, where short-term memory contains recent responses and long-term memory keeps summarized failed attempts to add in the prompt as reflection.  \\
\emph{Multi-Agents Systems:} LLMs can play user-defined roles and behave like a specific domain expert. In multi-agent systems, each LLM is assigned a unique role, simulating human behavior and collaborating with other agents to complete a complex task~\cite{metagpt,medagents}.     \\
\emph{\textbf{LLMs in Physical Environment:}}
\label{sec:robotics}
LLMs are good at instruction-following, however, utilizing them for physically grounded tasks requires adaptation, as they lack real-world knowledge. This could lead to generating illogical responses for a particular physical situation~\cite{saycan,PaLME}. SayCan~\cite{saycan} make LLMs aware of the available low-level task operations. LLM (Say) builds a high-level plan to complete the task and a learned affordance function (Can) explores the possibility of executing the plan in the real world. SayCan uses RL to train the language-conditioned affordance function. PaLM-E enables the LLM to solve grounded tasks by training multi-modal LLM feeding inputs directly from the sensors.  \\   
\emph{Manipulation:} In the area of manipulation~\cite{alphablock, ha2023scaling}, LLMs enhance a robot's dexterity and adaptability, excelling in tasks like object recognition, grasping, and collaboration. They analyze visual and spatial information to determine the most effective approach to interact with objects. \\
% , proving invaluable in operations requiring precision and flexibility, such as surgical procedures or assembly line tasks. \\
% They also enable the integration of sensor inputs and linguistic cues in an embodied framework, enhancing decision-making in real-world scenarios. It enhances the model's performance across various embodied tasks by allowing it to gather insights and generalize from diverse training data spanning language and vision domains. \\
\emph{Navigation:} LLMs enhance a robot's ability to navigate complex environments with precision and adaptability~\cite{rajvanshi2023saynav, song2022llm, dorbala2023can,huang2023visual}. They generate feasible paths and trajectories for robots, accounting for intricate environmental details~\cite{ding2023task}. This ability is valuable in scenarios requiring precise and dynamically adaptable navigation in environments like warehouses, transport, healthcare facilities, and residences.

\begin{figure*}[t]
\centering
\includegraphics[width=2\columnwidth]{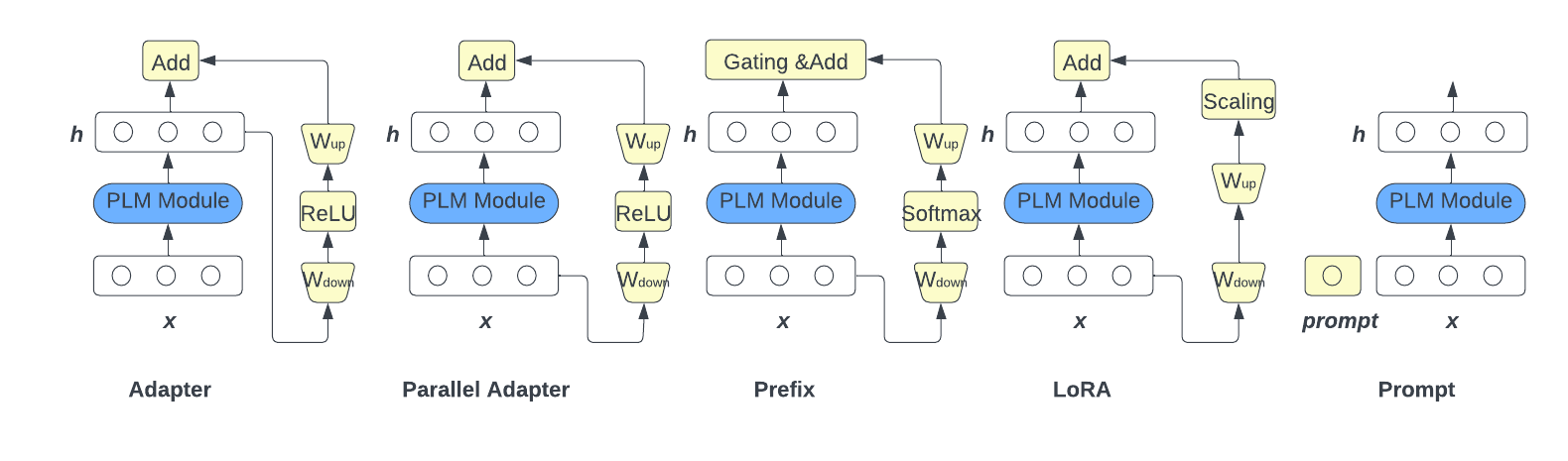}
\caption{Illustration of parameter-efficient fine-tuning paradigms, where $x$ is input and $h$ is hidden state, figure courtesy~\cite{unified_peft}. Parallel adapter and LoRA fall in the adapter tuning category.
}
\label{peft_image}
\end{figure*}

\subsection{Efficient LLMs}
Deploying LLMs in production is expensive. Reducing their running costs while preserving performance is an appealing area of research. This section summarizes the approaches suggested to enhance LLMs' efficiency.
% \textcolor{red}{Write a couple of sentences introducing the content of this section. Currently, it starts abruptly. }

\subsubsection{Parameter Efficient Fine-Tuning}
Fine-tuning LLMs with tens or hundreds of billions of parameters, such as GPT-3 (175B), BLOOM (176B), MT-NLG (540B), etc., is computationally intensive and time-consuming. To avoid complete model fine-tuning, numerous parameter-efficient fine-tuning (PEFT) techniques~\cite{Prompt_Tuning, Prompt_Tuning_2, Prefix_Tuning,unified_peft,LMAdapter_3}  try to achieve acceptable model fine-tuning performance at reduced costs. As compared to full fine-tuning~\cite{revisiting_peft}, PEFT performs better in low-resource setups, achieves comparable performance on medium-resource scenarios, and performs worse than full fine-tuning under high-resource availability. An overview of different PEFT approaches is shown in Figure~\ref{peft_image}. 
%We discuss these approaches below.  \\
% \textcolor{red}{Can the types in Fig 14 and paragraph titles match? If not, do we really need Fig 14?}.  A. Mentioned parallel adapter and LoRA fall in the adapter tuning category. Yes for three of them headings match, while fo the rest of the two we mentioned them in caption\\

\noindent\emph{\textbf{Adapter Tuning:}} Adds a few trainable parameters within the transformer block. The adapter layer is a sequence of feature downscaling, non-linearity, and upscaling~\cite{LMAdapter}. Variants of adapter tuning inject adapter layers sequentially~\cite{LMAdapter} and in parallel~\cite{unified_peft}, whereas the mixture of adapter (AdaMix)~\cite{adamix} employs multiple adapter modules in a single layer. AdaMix routes input instances randomly to one of the multiple downscale and upscale modules. The mixture of adapters is averaged out for inference to avoid additional latency. Low-Rank Adaptation (LoRA)~\cite{hu2021lora} learns low-rank decomposed matrices to freeze original weights. The learned weights are fused with the original weights for inference,   avoiding latency. \\
\noindent
\emph{\textbf{Prompt Tuning:}} Prompting is an effective way to adapt a pre-trained LLM for the downstream task. However, manual prompts bring uncertainty in the model's prediction, where a change in a single word drops the performance~\cite{Prompt_Tuning_2}. Prompt tuning alleviates this problem by fine-tuning only 0.001\%-3\% additional parameters~\cite{p_tuning_v2}. It concatenates trainable prompt parameters with the model embeddings~\cite{Prompt_Tuning_2, Prompt_Tuning, p_tuning_v2}. Task-specific fixed discrete prompts are concatenated with input embeddings in~\cite{Prompt_Tuning}. As discrete prompts bring instability, prompts are encoded through a learnable mapping in P-Tuning~\cite{Prompt_Tuning_2}, naming continuous prompts, which are appended with the discrete prompts. Only the prompt encoder is trainable in the model. In an extension of P-Tuning, continuous prompts are concatenated with each layer of the network in~\cite{p_tuning_v2}. Progressive prompts~\cite{progressive_prompts} avoid catastrophic forgetting and transfer previously learned knowledge by sequentially adding trainable prompt embeddings to the previously frozen task embeddings.         

\noindent
\emph{\textbf{Prefix Tuning:}} A set of trainable task-specific prefix vectors are appended to the frozen transformer layers in prefix tuning~\cite{Prefix_Tuning}. The prefix vectors are virtual tokens attended by the context tokens on the right. In addition, adaptive prefix tuning~\cite{adaptive_prefix} applies a gating mechanism to control the information from the prefix and actual tokens. \\
\noindent
\emph{\textbf{Bias Tuning:}} Fine-tuning only bias terms in small to medium training data has been found effective in BitFit~\cite{bitfit}. This method achieves full fine-tuning performance for tasks with less training data and comparable performance with more training data.

\subsubsection{Quantization}
LLMs require extensive computing and memory for inference. Deploying a 175B parameter GPT-3  model needs at least five 80GB A100 GPUs and 350GB of memory to store in FP16 format~\cite{SmoothQuant}. Such demanding requirements for deploying LLMs make it harder for smaller organizations to utilize them. Model compression is an effective solution but comes at the cost of degraded performance, especially at large scales greater than 6B. These models exhibit very large magnitude outliers that do not exist in smaller models~\cite{8_bit_LLM}, making it challenging and requiring specialized methods for quantizing LLMs~\cite{SmoothQuant,OPTQ}. \\ 
\emph{\textbf{Post-Training Quantization:}}
Minimal or no training is required in this type of quantization,  without significantly compromising the model performance. LLM-8-bit~\cite{8_bit_LLM} uses full-precision matrix multiplication for weights associated with outlier features and 8-bit for remaining features. The lower precision multiplication outputs are converted to FP-16 and concatenated with others. The quantized models have homogenous word embeddings, which may degrade their performance. To fix this, token-level knowledge distillation is employed in~\cite{compression_PLM_quant} along with independent quantization scaling factors for each module due to varying weight distribution. Feature distributions are asymmetric and appear in different channels; outlier suppression~\cite{outlier_suppression} shifts and scales per-channel activation distributions for effective quantization. SmoothQuant~\cite{SmoothQuant} quantizes activations and weights to INT8 format by smoothing activations and migrating the quantization difficulty toward weights. It multiplies the inverse of the smoothing factor with weights, which introduces a few outliers in the weights but is easier to quantify than unsmoothed activations. OPTQ~\cite{OPTQ} uses the optimal brain compression (OBC)~\cite{OBC} algorithm to quantize the model layer-by-layer and update weights to compensate for quantization error. To improve speed and performance, OPTQ updates weights in arbitrary order, employs lazy updates, and uses better Cholesky kernels. Outlier-aware weight quantization (OWQ)~\cite{OWQ} uses the OPTQ algorithm for quantization but assigns higher precision to vulnerable weights, causing outliers and lower precision for others.\\ 
\emph{\textbf{Quantization-Aware Training:}}
To compensate for performance degradation, a quantized model is fine-tuned in quantization-aware training (QAT)~\cite{alphatuning,qlora,llm_qat}. Alpha Tuning quantizes the model using binary coding quantization (BCQ)~\cite{BCQ} and fine-tunes only quantization scaling factors. This approach improves performance over parameter-efficient fine-tuning of the pre-trained model. Similarly, parameter-efficient and quantization-aware adaptation (PEQA)~\cite{PEQA} reduces the precision of fully-connected layers and fine-tunes only quantization scaling parameters. LLM-QAT~\cite{llm_qat} generates training data from the pre-trained network and trains a quantized student model with knowledge distillation. QLoRA~\cite{qlora} fine-tunes 4-bit quantized pre-trained LLM with LoRA~\cite{hu2021lora} using a 4-bit normal float, which shows better performance over a 4-bit integer and float. 
\subsubsection{Pruning}
Pruning is an alternative approach to quantization to compress model size, thereby reducing LLMs deployment costs significantly. Compared to task-agnostic pruning, task-specific pruning is easily achievable with good performance, where a model is fine-tuned on the downstream task and pruned for faster inference.  
It is possible to prune LLMs for individual tasks, but the cost of pruning and deploying task-specific models is high. To overcome this, many structured and unstructured pruning methods for LLMs have been proposed to maintain reasonable performance across all tasks while shrinking the model size~\cite{wanda,llm_pruner,structured_p_llms}.\\
\emph{\textbf{Unstructured Pruning:}} This kind of pruning removes less important weights without maintaining any structure. Existing LLM pruning methods take advantage of the unique characteristics of LLMs, uncommon for smaller models, where a small subset of hidden states are activated with large magnitude~\cite{8_bit_LLM}. Pruning by weights and activations (Wanda)~\cite{wanda} prunes weights in every row based on importance, calculated by multiplying the weights with the norm of input. The pruned model does not require fine-tuning, thereby saving computational costs. Outlier weighed layerwise sparsity (OWL)~\cite{OWL} extends Wanda with non-uniform layer pruning. It shows that the number of outliers varies for different layers; therefore, the model should have variable pruning ratios for better performance for every layer. Contrastive pruning (CAP)~\cite{contrastive_pruning} iteratively prunes the model by training the sparse model using contrastive loss between pre-trained, fine-tuned, and snapshots of previous sparse models to learn task-specific and task-agnostic knowledge. \\
\emph{\textbf{Structured Pruning:}} Here, the parameters are removed in groups, rows, columns, or matrices, which speeds up the inference because of effective hardware tensor core utilization~\cite{wanda}. LLM-Pruner~\cite{llm_pruner} employs a 3-stage structured pruning strategy, identifying the groups of hidden states causing each other to activate during the forward-pass, keeping important groups and removing less important ones, and fine-tuning the pruned model with LoRA. Sparsity-induced mask learning (SIMPLE)~\cite{SIMPLE} prunes the network using learnable masks. Similarly, another method prunes LLMs by learning masks and removing unimportant rank-1 components of the factorized weight matrix~\cite{structured_p_llms}.

\subsection{Multimodal LLMs}
Inspired by the success of LLMs in natural language processing applications, an increasing number of research works are now facilitating LLMs to perceive different modalities of information like image~\cite{alayrac2022flamingo, li2023blip, liu2023visual}, video~\cite{li2023videochat, maaz2023video, zhang2023video}, audio~\cite{mei2023wavcaps, zhang2023video, lyu2023macaw}, etc.  Multimodal LLMs (MLLMs) present substantial benefits compared to standard LLMs that process only text. By incorporating information from various modalities, MLLMs can achieve a deeper understanding of context, leading to more intelligent responses infused with a variety of expressions. Importantly, MLLMs align closely with human perceptual experiences, leveraging the synergistic nature of our multisensory inputs to form a comprehensive understanding of the world~\cite{lyu2023macaw, PaLME}. Coupled with a user-friendly interface, MLLMs can offer intuitive, flexible, and adaptable interactions, allowing users to engage with intelligent assistants through a spectrum of input methods. According to the ways of constructing models, current MLLMs can be generally divided into three streams: pre-training, fine-tuning, and prompting. In this section, we will discuss more details of these main streams, as well as the important application of MLLMs in visual reasoning. \\ 
\emph{\textbf{Pre-training:}} This stream of MLLMs intends to support different modalities using unified end-to-end models. For instance, Flamingo~\cite{alayrac2022flamingo} applies gated cross-attention to fuse vision and language modalities, which are collected from pre-trained and frozen visual encoder and LLM, respectively. Moreover, BLIP-2~\cite{li2023blip} proposes a two-stage strategy to pre-train a Querying Transformer (Q-Former) for the alignment between vision and language modalities: in the first stage, vision-language representation learning is bootstrapped from a frozen visual encoder; and in the second stage, a frozen LLM bootstraps vision-to-language generative learning for zero-shot image-to-text generation. Similarly, MiniGPT-4~\cite{zhu2023minigpt} deploys pre-trained and frozen ViT~\cite{dosovitskiy2020image}, Q-Former and Vicuna LLM~\cite{vicuna}, only training the linear projection layer for vision and language modalities alignment. \\
\emph{\textbf{Fine-tuning:}} Derived from instruction tuning~\cite{Flan} for NLP tasks~\cite{instructgpt, Flan, OPT_IML}, researchers are fine-tune pre-trained LLMs using multimodal instructions. Following this method, LLMs can be easily and effectively extended as multimodal chatbots~\cite{zhu2023minigpt, liu2023visual, ye2023mplug} and multimodal task solvers~\cite{dai2023instructblip, wang2023visionllm, xu2022multiinstruct}. The key issue of this stream of MLLMs is to collect multimodal instruction-following data for fine-tuning~\cite{survey_multimodal_llm}. To address this issue, the solutions of benchmark adaptation~\cite{dai2023instructblip, zhao2023chatbridge, li2023m}, self-instruction~\cite{ft_self_instruct, gpt4tools, pi2023detgpt}, and hybrid composition~\cite{luo2023cheap, xu2022multiinstruct} are employed, respectively. To mitigate the gap between the original language modality and additional modalities, the learnable interface is introduced to connect different modalities from frozen pre-trained models. Particularly, the learnable interface is expected to work in a parameter-efficient tuning manner: e.g., LLaMA-Adapter~\cite{zhang2023llama} applies an efficient transformer-based adapter module for training, and LaVIN~\cite{luo2023cheap} dynamically learns the multimodal feature weights using a mixture-of-modality adapter. Different from the learnable interface, the expert models can directly convert multimodalities into language: e.g., VideoChat-Text~\cite{li2023videochat} incorporates Whisper~\cite{radford2023robust}, a speech recognition expert model, to generate the captions of given videos for the understanding of following LLMs. \\ 
\emph{\textbf{Prompting:}} Different from the fine-tuning technique that directly updates the model parameters given task-specific datasets, the prompting technique provides certain context, examples, or instructions to the model, fulfilling specialized tasks without changing the model parameters. Since prompting can significantly reduce the need for large-scale multimodal data, this technique is widely used to construct MLLMs. Particularly, to solve multimodal Chain of Thought (CoT) problems~\cite{wei2022chain}, LLMs are prompted to generate both the reasoning process and the answer given multimodal inputs~\cite{zhang2023multimodal}. On this front, different learning paradigms are exploited in practice: for example, Multimodal-CoT~\cite{zhang2023multimodal} involves two stages of rationale generation and answer inference, where the input of the second stage is a combination of the original input and the output of the first stage; and CoT-PT~\cite{ge2023chain} applies both prompt tuning and specific visual bias to generate a chain of reasoning implicitly. In addition to CoT problems, LLMs can also be prompted with multimodal descriptions and tools, effectively dividing complex tasks into sub-tasks~\cite{wu2023visual, yang2023mm}. \\
\emph{\textbf{Visual Reasoning Application:}} Recent visual reasoning systems~\cite{wang2023caption, zhu2022pointclip, chameleon, gupta2023visual} tend to apply LLMs for better visual information analysis and visual-language integration. Different from previous works~\cite{gao2019dynamic, yu2019deep} that rely on limited VQA datasets and small-scale neural networks, current LLM-aided methods offer benefits of stronger generalization ability, emergent ability, and interactivity~\cite{survey_multimodal_llm}. To realize visual reasoning with the help of LLMs, prompting and fine-tuning techniques can also be utilized: for example, PointClip V2~\cite{zhu2022pointclip} applies LLMs to generate 3D-specific prompts, which are encoded as textual features and then combined with visual features for 3D recognition; and GPT4Tools~\cite{gpt4tools} employs LoRA~\cite{hu2021lora} to fine-tune LLMs following tool-related instructions. Serving as a controller~\cite{gupta2023visual}, decision maker~\cite{you2023idealgpt}, or semantics refiner~\cite{wang2023caption, zhang2023prompt}, LLMs significantly facilitates the progress of visual reasoning research.

\subsection{Summary and Discussion}
\label{sec:summary_and_discussion}
\subsubsection{Architecture}
Due to the gigantic scale of LLMs, minor changes in architecture and training strategies have a big impact on performance and stability. Here, we summarize key architectural modules used in various LLMs, leading to better performance, reduced training time and memory, and better training stability. \\
\emph{\textbf{Layer Normalization:}} The performance and training stability of LLMs are affected significantly by layer normalization. Pre-norm, that is normalizing inputs rather than outputs, is more common among LLMs stabilizing the training~\cite{GPT-3, touvron2023llama, PanGU_alpha}. BLOOM~\cite{BLOOM} and AlexaTM~\cite{soltan2022alexatm} utilize an additional layer normalization before embedding layer to stabilize the training of large-scale models, while the model's zero-shot generalization ability can be negatively impacted~\cite{BLOOM}. However, another study~\cite{GLM-130B} finds that pre-norm degrades fine-tuned model performance as compared to post-norm, and there are no stability benefits of pre-norm beyond the 100B scale. Therefore, GLM-130B~\cite{GLM-130B} used deep-norm which is a variant of post-norm for better downstream task performance after fine-tuning.   \\
\emph{\textbf{Positional Encoding:}} Like other building blocks of the model, positional encoding also affects the performance and training stability of LLMs. BLOOM~\cite{BLOOM} finds ALiBi outperforms learned and rotary positional encodings. Contrary to this, GLM-130B~\cite{GLM-130B} identifies rotary positional encoding as being better than ALiBi. So, there is no conclusion in the literature about positional encodings yet. \\
\emph{\textbf{Parallel Attention:}} In this type of attention, feed-forward and attention layers are parallel to each other rather than sequential in a transformer block. It has been shown to reduce training time by 15\%. There is no evidence of performance drop due to this change in the literature and it is used by the models PaLM~\cite{PaLM}, GPT-NeoX~\cite{GPT_NeoX}, and CodeGen~\cite{CodeGen}.  \\
\emph{\textbf{Multi-Query Attention}} It has shared key and value attention heads in a transformer block while query attention heads are projected as usual. This reduces memory usage and speeds up sampling in autoregressive decoding. No performance degradation has been observed with this change and it makes the training efficient allowing larger batch sizes. Multi-query attention is used in~\cite{PaLM, li2022competition}.    \\
\emph{\textbf{Mixture of Experts:}} This type of architecture enables easily scaling models to trillions of parameters~\cite{PanGu_sigma, du2022glam}. Only a few experts are activated during the computation making them compute-efficient. The performance of MoE models is better than dense models for the same amount of data and requires less computation during fine-tuning to achieve performance similar to dense models as discussed in~\cite{du2022glam}. MoE architectures are less prone to catastrophic forgetting, therefore are more suited for continual learning~\cite{PanGu_sigma}. Extracting smaller sub-models for downstream tasks is possible without losing any performance, making MoE architecture hardware-friendly~\cite {PanGu_sigma}.    \\
\emph{\textbf{Sparse vs Dense Activated:}}
GPT-3~\cite{GPT-3} uses sparse transformers~\cite{sparse_transformer} whereas GLaM~\cite{du2022glam} and PanGu-$\sum$~\cite{PanGu_sigma} use MoE~\cite{shazeer2017outrageously} architectures to lower computational costs and increase the model size and capacity. According to the literature, sparse modules do not degrade the model's performance~\cite{sparse_transformer}. However, more experiments are required to verify this statement.

\subsubsection{Training Strategies}
Training models at a huge scale require tricks to reduce training costs, avoid loss divergence, and achieve better performance. We summarize and discuss some of these key tricks used in different LLMs. \\
\emph{\textbf{Mixed Precision:}} It is a famous method for LLMs to reduce memory usage and improve training efficiency. In mixed precision, forward and backward passes are performed in FP16 format whereas optimizer states and master weights are kept in FP32 format~\cite{Mixed_Precision}. A drawback associated with this format change is training instability due to a smaller value range resulting in loss spikes~\cite{GLM-130B}. An alternative to FP16 is BF16 which has a comparatively larger range and performs precision-sensitive operations like gradient accumulation and softmax in FP32~\cite{BLOOM}. BF16 has better performance and training stability but uses more memory and is supported on specific hardware, for example, A100 GPUs. Therefore, its adoption in LLMs is limited. \\ %\textcolor{blue}{for instance, the GLaM~\cite{du2022glam} model adopts this method for training, where FP32 is used for model weights and BF16 is used for activations.}  \\
\emph{\textbf{Training Instability:}} Loss divergence or spiking is a common issue in LLMs that occurs multiple times during training. This happens in the presence of gradient clipping~\cite{PaLM}. To mitigate this problem, many approaches suggest restarting training from an earlier checkpoint~\cite{PaLM, GLM-130B, du2022glam}, skipping 200-500 earlier data batches at the point of divergence in~\cite{PaLM} and re-shuffling batches in~\cite{du2022glam}. The embedding layer gradient shrink proves to further stabilize the training as its gradient norm is significantly larger than the other layers~\cite{GLM-130B}. Another suggestion to improve training stability for larger models is not to use \textbf{biases} in dense and norm layers as in~\cite{PaLM}.    \\
\emph{\textbf{Weight Initialization:}} It plays a significant role in model convergence and training stability. GPT-NeoX~\cite{GPT_NeoX} initializes feed-forward layers before residuals with $\frac{2}{L\sqrt{d}}$ as in~\cite{Mesh_Transformer_JAX} and other layers with the small initialization scheme~\cite{small_init}. This avoids activations growing exponentially with increasing depth. MT-NLG~\cite{mtnlg} found higher variance for weight initialization leads to unstable training, hence validating small initialization scheme~\cite{small_init}. Various models perform random weight initialization which can cause bad initialization, Galactica~\cite{galactica} suggests a longer warmup to negate the effect.  \\
\emph{\textbf{Learning Rate:}} A suitable learning rate is important for stable training. It is suggested to use a lower value~\cite{BLOOM, PaLM, U-PaLM} with warmup and decay (cosine or linear). Usually, the learning rate is within the range $1e^{-4}$ to $8e^{-4}$. Moreover, MT-NLG (530B)~\cite{mtnlg} and GPT-NeoX (20B)~\cite{GPT_NeoX} suggest interpolating learning rates based on the model size using the GPT-3~\cite{GPT-3} models ranging between 13B and 175B. This avoids tuning the learning rate hyperparameter.  \\ 
\emph{\textbf{Training Parallelism:}} 3D parallelism, a combination of data, pipeline, and tensor parallelism, is the most utilized training parallelism approach in LLMs~\cite{GLM-130B, PaLM, OPT, BLOOM, mtnlg, wu2021yuan,lieber2021jurassic}. In addition to 3D parallelism, BLOOM~\cite{BLOOM} uses a zero optimizer~\cite{ZeroOpt} to shard optimizer states. PanGu-$\alpha$~\cite{PanGU_alpha} and PanGu-$\Sigma$~\cite{PanGu_sigma} go beyond 3D parallelism and apply 5D parallelism which additionally contains optimizer parallelism and rematerialization.     \\
\emph{\textbf{Mode Switching:}} It adds task-related tokens at the beginning of the text during training. These tokens refer to the natural language understanding and natural language generation tasks which are shown to improve downstream task performance in~\cite{UL2, U-PaLM, soltan2022alexatm}. During fine-tuning and inference, tokens are appended based on the downstream tasks.  \\
\emph{\textbf{Controllable Text Generation:}} Generating credible and controlled text from a pre-trained model is challenging. GPT-3~\cite{GPT-3} and other LLMs use in-context learning to control generated text. While in-context learning helps in controlling the generated text, ERNIE 3.0 Titan~\cite{ernie3titan} suggests using adversarial loss to rank its generated text for credibility and soft prompts such as genre, topic, keywords, sentiment, and length for better control on generated text.

\subsubsection{Supervised Models vs Generalized Models}
Although generalized models are capable of performing diverse tasks with good performance they have not yet outperformed models trained in supervised settings. The supervised trained models are still state-of-the-art in various NLP tasks by a large margin as shown in~\cite{GPT-3, PaLM, Tk-INSTRUCT}.   
\subsubsection{Zero-Shot vs Few-Shot}
LLMs perform well in zero-shot and few-shot settings. But the performance difference between zero-shot and few-shot is large for pre-trained models~\cite{GPT-3, PaLM}, naming LLMs as meta-learners~\cite{GPT-3}. LLMs zero-shot evaluations underperform unsupervised methods in neural machine translation~\cite{GPT-3}. The literature shows pre-training is not enough for good zero-shot performance~\cite{PaLM, Flan}. To improve the zero-shot performance the literature suggests using instruction fine-tuning that improves the zero-shot performance significantly and outperforms baselines. Instruction fine-tuning has also been shown to improve zero-shot generalization to unseen tasks. Another model, Flan-PaLM~\cite{Flan}, unlocks zero-shot reasoning with CoT training. 

\subsubsection{Encoder vs Decoder vs Encoder-Decoder}
Traditionally, these architectures perform well for different tasks, for example, encoder-only for NLU tasks, decoder-only for NLG, and encoder-decoder for sequence2sequence modeling. Encoder-only models are famous for smaller models such as Bert~\cite{Bert}, RoBERTa~\cite{roberta}, etc., whereas LLMs are either decoder-only~\cite{GPT-3, GPT_NeoX, BLOOM} or encoder-decoder~\cite{T5, mT5, soltan2022alexatm}. While decoder-only models are good at NLG tasks, various LLMs, PaLM~\cite{PaLM}, OPT~\cite{OPT}, GPT-3~\cite{GPT-3}, BLOOM~\cite{BLOOM}, LLaMA~\cite{gao2023llama}, are decoder-only models with significant performance gains on both NLU and NLG tasks. In contradiction to this, T5~\cite{T5} and UL2~\cite{UL2} identify encoder-decoder models out-performing decoder-only models. In another study, PaLM~\cite{PaLM} finds increasing the size of decoder-only models can reduce the performance gap between decoder-only and encoder-decoder architectures. \\
Although decoder-only architectures have become a trend for LLMs, many recently proposed approaches~\cite{UL2, soltan2022alexatm} use mode-switching tokens in text with encoder-decoder architectures to enable task-specific modes. Similarly, CodeT5+~\cite{codet5+} uses an encoder-decoder architecture with multiple training objectives for different tasks, activating the encoder, decoder, or both according to the tasks. These variations in architecture and training objectives allow a model to perform well in different settings. Because of this dynamic configuration, the future of LLMs can be attributed to encoder-decoder architectures.

\begin{table*}[!t]
\rowcolors{2}{gray!25}{white}
\begin{center}
\caption{Summary of pre-trained LLMs (>10B). Only the LLMs discussed individually in the previous sections are summarized. \enquote{Data/Tokens} is the model's pre-training data, which is either the number of tokens or data size. 
%\enquote{No.~of Params} is the size of the model's parameters,
\enquote{Data Cleaning} indicates whether data cleaning is performed or not. This includes heuristics (Heur), deduplication (Dedup), quality filtering (QF), and privacy filtering (PF), \enquote{Cost} is the calculated training cost obtained by multiplying the GPUs/TPUs hourly rate with the number of GPUs and the training time. The actual cost may vary due to many reasons such as using in-house GPUs or getting a discounted rate, re-training, number of employees working on the problem, etc. \enquote{Training Parallelism} indicates distributed training using data parallelism (D), tensor parallelism (T), pipeline parallelism (P), context parallelism (C), model parallelism (M), optimizer parallelism (OP), and rematerialization (R), where for \enquote{Library} column, \enquote{DS} is a short form for Deep Speed. In column \enquote{Commercial Use}, we assumed a model is for non-commercial purposes if its license is unavailable.  
}
\label{tab:statistics_pt}
%\footnotesize
\renewcommand\tabcolsep{-0.05pt}{
%\hspace*{-5em}
\resizebox{\linewidth}{!}{
\begin{tabular}{llcccccccccccccc}
\toprule
\rowcolor{gray!50}
  &  &   &  & & &   &    & &  &  & &   \\
\rowcolor{gray!50}
\multirow{-2}{*}{\textbf{Models}} & \multirow{-2}{*}{\textbf{\begin{tabular}[c]{@{}c@{}}Publication\\ Venue\end{tabular}}} & \multirow{-2}{*}{\textbf{\begin{tabular}[c]{@{}c@{}}License\\ Type\end{tabular}}} & \multirow{-2}{*}{\textbf{\begin{tabular}[c]{@{}c@{}}Model\\ Creators\end{tabular}}} & \textbf{Purpose~~ } & \multicolumn{1}{c}{\multirow{-2}{*}{\textbf{\begin{tabular}[c]{@{}c@{}}No. of\\Params\end{tabular}}}} & \multirow{-2}{*}{\textbf{\begin{tabular}[c]{@{}c@{}} 
 ~~Commercial\\ Use\end{tabular}}} & \multirow{-2}{*}{\textbf{\begin{tabular}[c]{@{}c@{}}Steps\\ Trained~~ \end{tabular}}} & \multirow{-2}{*}{\textbf{\begin{tabular}[c]{@{}c@{}}Data/\\ Tokens\end{tabular}}} & \multirow{-2}{*}{\textbf{\begin{tabular}[c]{@{}c@{}}Data\\ Cleaning\end{tabular}}} & \multirow{-2}{*}{\textbf{\begin{tabular}[c]{@{}c@{}}No. of\\ Processing Units~~ \end{tabular}}} & \multirow{-2}{*}{\textbf{\begin{tabular}[c]{@{}c@{}}Processing\\ Unit Type~~ \end{tabular}}} & \multirow{-2}{*}{\textbf{\begin{tabular}[c]{@{}c@{}}Training~~ \\ Time\end{tabular}}} &  \multirow{-2}{*}{\textbf{\begin{tabular}[c]{@{}c@{}}Calculated\\ Train. Cost\end{tabular}}} & \multirow{-2}{*}{\textbf{\begin{tabular}[c]{@{}c@{}}Training\\ Parallelism\end{tabular}}} & \multirow{-2}{*}{\textbf{\begin{tabular}[c]{@{}c@{}}\\ Library\end{tabular}}} \\
\midrule
T5~\cite{T5}   & JMLR\textquotesingle20 & Apache-2.0 & Google & General & 11B   & $\checkmark$   & 1M & 1T & Heur+Dedup & 1024 & TPU v3 & -& - & D+M & Mesh TensorFlow  \\

GPT-3~\cite{GPT-3}    & NeurIPS\textquotesingle20   & - & OpenAI & General & 175B   & $\times$ & - & 300B & Dedup+QF  &  -  & V100 & - & - & M & -       \\

mT5~\cite{mT5}  & NAACL\textquotesingle21  & Apache-2.0  & Google & General & 13B    & $\checkmark$  & 1M & 1T & - & -  & - & - & -  & - & - \\

{PanGu-$\alpha$}~\cite{PanGU_alpha}   & arXiv\textquotesingle21    &  Apache-2.0  & Huawei & General & 200B  &  $\checkmark$  &  260k  & 1.1TB  & Heur+Dedup &  2048 & Ascend 910 & - & - & D+OP+P+O+R & MindSpore\\

CPM-2~\cite{CPM-2}    & AI Open\textquotesingle21  & MIT & Tsinghua & General & 198B   & $\checkmark$ & 1M & 2.6TB  & Dedup & - & - & - & - & D+M  & JAXFormer   \\

Codex~\cite{codex}    & arXiv\textquotesingle21 & -  & OpenAI & Coding & 12B    & $\times$ & - & 100B & Heur & - & -  & - & - & -    & - \\

ERNIE 3.0~\cite{ernie3}    & arXiv\textquotesingle21  & -  & Baidu & General & 10B  & $\times$ & 120k$^*$ & 375B & Heur+Dedup & 384 & V100   & -  & - & M$^*$ & PaddlePaddle    \\

Jurassic-1~\cite{lieber2021jurassic}   & White-Paper\textquotesingle21 & Apache-2.0  & AI21 & General & 178B      & $\checkmark$ & - & 300B   & -  & 800 & GPU & -  & - & D+M+P    & Megatron+DS       \\

HyperCLOVA~\cite{hyperclova}   & EMNLP\textquotesingle21 & -   & Naver & General & 82B & $\times$  & - &  300B  & Clf+Dedup+PF & 1024 & A100   & 321h  & 1.32 Mil & M  & Megatron      \\

Yuan 1.0~\cite{wu2021yuan}   & arXiv\textquotesingle21  & Apache-2.0  & - & General & 245B   & $\checkmark$  & 26k$^*$ &  180B  &  Heur+Clf+Dedup  & 2128 & GPU  &  -  & - & D+T+P & - \\

% WebGPT~\cite{nakano2021webgpt}   & Dec-2021  &$\times$  & 175   & GPT-3   & - & -  & - & -  & -   & -  & -   \\

Gopher~\cite{gopher}   & arXiv\textquotesingle21 &  - & Google & General & 280B & $\times$ & - & 300B   & QF+Dedup  & 4096 & TPU v3 & 920h  & 13.19 Mil & D+M    & JAX+Haiku        \\

ERNIE 3.0 Titan~\cite{ernie3titan}   & arXiv\textquotesingle21 & -   & Baidu & General & 260B  & $\times$ & - &  300B  & Heur+Dedup &  - & Ascend 910  & -  & - & D+M+P+D* & PaddlePaddle  \\

GPT-NeoX-20B~\cite{GPT_NeoX}~ & BigScience\textquotesingle22    & Apache-2.0    & EleutherAI & General & 20B   & $\checkmark$ & 150k & 825GB & None & 96 & 40G A100 & - & - & M & Megatron+DS+PyTorch     \\

OPT~\cite{OPT}  & arXiv\textquotesingle22    & MIT   & Meta & General & 175B   & $\checkmark$ &  150k  & 180B  &  Dedup   & 992 & 80G A100  & -  & - & D+T & Megatron   \\

BLOOM~\cite{BLOOM}    & arXiv\textquotesingle22  & RAIL-1.0  & BigScience & General & 176B   & $\checkmark$   & - & 366B & Dedup+PR  & 384 & 80G A100    & 2520h  & 3.87 Mil & D+T+P  & Megatron+DS    \\

Galactica~\cite{galactica} & arXiv\textquotesingle22    & Apache-2.0 & Meta & Science & 120B   & $\times$ & 225k & 106B & Dedup  &  128 & 80GB A100   & -  & - & - & Metaseq    \\

GLaM~\cite{du2022glam} & ICML\textquotesingle22  &  -  & Google & General & 1.2T  & $\times$   & 600k$^*$  & 600B   & Clf  & 1024 & TPU v4 & -  & - & M & GSPMD       \\

LaMDA~\cite{thoppilan2022lamda}    & arXiv\textquotesingle22   & - & Google & Dialog & 137B      & $\times$ & 3M & 2.81T  & Filtered  & 1024 & TPU v3 & 1384h & 4.96 Mil & D+M    & Lingvo    \\

MT-NLG~\cite{mtnlg}   & arXiv\textquotesingle22    &  Apache-v2.0  & MS.+Nvidia & General & 530B   & $\times$ & - & 270B & - & 4480 & 80G A100   & -  & - & D+T+P    & Megatron+DS       \\

AlphaCode~\cite{li2022competition}    & Science\textquotesingle22  & Apache-v2.0  & Google & Coding & 41B    & $\checkmark$  & 205k & 967B & Heur+Dedup & - & TPU v4  & -  & - & M   & JAX+Haiku       \\

Chinchilla~\cite{chinchilla}   & arXiv\textquotesingle22    &  -  & Google & General & 70B   & $\times$ & - & 1.4T  & QF+Dedup  & - & TPUv4   & -  & -  & -  & JAX+Haiku        \\

PaLM~\cite{PaLM} & arXiv\textquotesingle22   & - & Google & General & 540B & $\times$ &  255k  & 780B  & Heur & 6144 & TPU v4  & - & - & D+M & JAX+T5X\\

AlexaTM~\cite{soltan2022alexatm}  & arXiv\textquotesingle22 & Apache v2.0   & Amazon & General & 20B  & $\times$  & 500k  & 1.1T   & Filtered  & 128 & A100    & 2880h  & 1.47 Mil & M & DS      \\

U-PaLM~\cite{U-PaLM}   & arXiv\textquotesingle22    & - & Google & General & 540B   & $\times$ & 20k & - & -   & 512 & TPU v4  & 120h    & 0.25 Mil & - & - \\

UL2~\cite{UL2}  & ICLR\textquotesingle23  &  Apache-2.0  & Google & General & 20B   & $\checkmark$ & 2M & 1T & - & 512 & TPU v4  & -  & - & M & JAX+T5X    \\

GLM~\cite{GLM-130B}  & ICLR\textquotesingle23 & Apache-2.0  & Multiple & General & 130B   & $\times$ & - & 400B   & -  & 768 & 40G A100    & 1440h  & 3.37 Mil & M & -    \\

CodeGen~\cite{CodeGen}  & ICLR\textquotesingle23   &  Apache-2.0  & Salesforce  & Coding & 16B   & $\checkmark$  & 650k & 577B  & Heur+Dedup  & - & TPU v4 & - & - & D+M & JAXFormer  \\

LLaMA~\cite{touvron2023llama}    & arXiv\textquotesingle23 &  -  & Meta & General & 65B      & $\times$ & 350k & 1.4T & Clf+Heur+Dedup   & 2048 & 80G A100 & 504h   &  4.12 Mil& D+M & xFormers     \\

PanGu$\Sigma$~\cite{PanGu_sigma}    & arXiv\textquotesingle23  &  - & Huawei & General & 1.085T & $\times$ & -  & 329B & -  & 512 & Ascend 910   & 2400h  & - & D+OP+P+O+R & MindSpore  \\
BloombergGPT~\cite{bloomberggpt}~ & arXiv23  & -  & Bloomberg & Finance & 50B & $\times$ & 139k  & 569B & Dedup  & 512 & 40G A100   & 1272h  & 1.97 Mil & M & PyTorch
\\
Xuan Yuan 2.0~\cite{xuanyuan}~ & arXiv23 & RAIL-1.0 & Du Xiaoman & Finance & 176B & $\checkmark$ & -  & 366B & Filtered & - & 80GB A100 & - & - & P & DS \\

CodeT5+~\cite{codet5+}  & arXiv\textquotesingle23 & BSD-3 & Salesforce & Coding & 16B & $\checkmark$ &  110k &51.5B  & Dedup & 16 & 40G A100 & -  & - & - & DS \\

StarCoder~\cite{starcoder}    & arXiv\textquotesingle23 & OpenRAIL-M & BigCode & Coding & 15.5B & $\checkmark$ & 250k & 1T  & Dedup+QF+PF & 512 & 80G A100 & 624h  & 1.28 Mil & D+T+P & Megatron-LM  \\

LLaMA-2~\cite{llama_2}    & arXiv\textquotesingle23 &  LLaMA-2.0  & Meta & General & 70B      & $\checkmark$ & 500k & 2T & Minimal Filtering   & - & 80G A100 & 1.7Mh   &  - & - & -     \\
PaLM-2~\cite{palm_2}    & arXiv\textquotesingle23 &  -  & Google & General & -  & $\times$ & - & - & Ddedup+PF+QF  & - & - & -   &  - & - & -     \\

LLaMA-3.1~\cite{llama3}    & arXiv\textquotesingle24 &  LLaMA-3.0  & Meta & General & 405B      & $\checkmark$ & 1.2M & 15T & Dedup+QF   & 16k & 80G H100 & 30.84Mh   &  - & D+T+P+C & PyTorch    \\

Mixtral 8x22B~\cite{mixtral}   & web\textquotesingle24 &  Apache-2.0  & Mistral AI & General & 141B      & $\checkmark$ & - & - & -   & - & - & -  &  - & - & -     \\

Snowflake Arctic~\cite{snowflake_arctic}  & web\textquotesingle24 &  Apache-2.0  & Snowflake & General &  480B     & $\checkmark$ & - & 3.5T & -   & - &  & -  &  - & T+P & DS     \\

Nemotron-4 340B~\cite{nemotron4}  & web\textquotesingle24 &  Nvidia  & Nvidia & General &  340B     & $\checkmark$ & - & 9T & -   &  6144 & 80G H100 & -  &  - & D+T+P & -     \\

DeepSeek~\cite{deepseek}  & arXiv\textquotesingle24 &  MIT  & DeepSeek & General &  67B     & $\checkmark$ & - & 2T & Dedup+QF   & - & - & 300.6Kh &  - & D+T+P & DS     \\

DeepSeek-v2~\cite{deepseek_v2}  & arXiv\textquotesingle24 &  MIT  & DeepSeek & General &  67B     & $\checkmark$ & - & 8.1T & QF   & - & H800 & 172.8Kh  &  - & D+P & HAI-LLM    \\

%\multirow{-25}{*}{\begin{tabular}[c]{@{}c@{}}Closed\\ Source\end{tabular}} & \multirow{-18}{*}{\begin{tabular}[c]{@{}c@{}}Publicly\\ Available\end{tabular}} &
\bottomrule
\end{tabular}
}
}
\end{center}
\end{table*}

\begin{table*}[!tbp]
\rowcolors{2}{gray!25}{white}
\centering
\caption{Summary of instruction tuned LLMs (>10B). All abbreviations are the same as Table~\ref{tab:statistics_pt}. Entries in \enquote{Data/Tokens} starting with \enquote{S-} represent the number of training samples.}
%\footnotesize
\renewcommand\tabcolsep{-0.05pt}{
\resizebox{\linewidth}{!}{
\begin{tabular}{llcccccccccccccc}
\toprule
%\rowcolor{gray!50}

%\rowcolor{gray!50}& \textbf{Publication} & \textbf{License} & \multicolumn{2}{c|}{\textbf{Models}} & \textbf{No. of} & \textbf{Commercial} & \textbf{Pre-trained} & \textbf{Steps} &\textbf{Data/} & \textbf{No. of} & \textbf{Processing}  & \multicolumn{3}{c|}{\textbf{Training}}  &   \\

%\rowcolor{gray!50}\textbf{Models} & \textbf{Venue} &  \textbf{Type} & \textbf{Creators} & \textbf{Purpose} & \textbf{Params} &  \textbf{Use} & \textbf{Models}& \textbf{Trained} &  \textbf{Tokens} & \textbf{Processing Units} &  \textbf{Unit Type} &\textbf{Time} & \textbf{Calcu. Cost} &  \textbf{Parallelism} & \textbf{Library}   \\

\rowcolor{gray!50}
  &  &   &  & & &   &    & &  &  & &   \\
\rowcolor{gray!50} \multirow{-2}{*}{\textbf{Models}} & \multirow{-2}{*}{\textbf{\begin{tabular}[c]{@{}c@{}}Publication\\ Venue\end{tabular}}} & \multirow{-2}{*}{\textbf{\begin{tabular}[c]{@{}c@{}}License\\ Type\end{tabular}}} & \multirow{-2}{*}{\textbf{\begin{tabular}[c]{@{}c@{}}Model\\ Creators\end{tabular}}} & \textbf{Purpose~ } &\multicolumn{1}{c}{\multirow{-2}{*}{\textbf{\begin{tabular}[c]{@{}c@{}}No. of~~ \\Params\end{tabular}}}} & \multirow{-2}{*}{\textbf{\begin{tabular}[c]{@{}c@{}}Commercial~~ \\ Use\end{tabular}}} & \multirow{-2}{*}{\textbf{\begin{tabular}[c]{@{}c@{}}Pre-trained\\ Models\end{tabular}}} & \multirow{-2}{*}{\textbf{\begin{tabular}[c]{@{}c@{}}Steps\\ Trained~ \end{tabular}}} & \multirow{-2}{*}{\textbf{\begin{tabular}[c]{@{}c@{}}Data/\\ Tokens~~ \end{tabular}}} & \multirow{-2}{*}{\textbf{\begin{tabular}[c]{@{}c@{}}No. of\\Processing Units~~ \end{tabular}}} & \multirow{-2}{*}{\textbf{\begin{tabular}[c]{@{}c@{}}Processing~  \\ Unit Type~ \end{tabular}}} &\multirow{-2}{*}{\textbf{\begin{tabular}[c]{@{}c@{}} Train.~~ \\ Time~ \end{tabular}}} & \multirow{-2}{*}{\textbf{\begin{tabular}[c]{@{}c@{}}Calculated\\ Train. Cost~~  \end{tabular}}} & \multirow{-2}{*}{\textbf{\begin{tabular}[c]{@{}c@{}}Train.\\ Parallelism~ \end{tabular}}} & \multirow{-2}{*}{\textbf{\begin{tabular}[c]{@{}c@{}}\\ Library\end{tabular}}}   \\

\midrule

WebGPT~\cite{nakano2021webgpt}   & arXiv\textquotesingle21  & -  & OpenAI & General & 175B & $\times$ & GPT-3   & -   & -  & -& -   & -  &  - &  - & - \\

T0~\cite{T0}   & ICLR\textquotesingle22  & Apache-2.0 &  BigScience & General & 11B   & $\checkmark$ & T5 & -  & 250B  & 512 & TPU v3  & 270h  & 0.48 Mil & - & - \\

Tk-Instruct~\cite{Tk-INSTRUCT}  & EMNLP\textquotesingle22  & MIT & AI2+ & General & 11B    & $\checkmark$ & T5  & 1000   & - & 256 & TPU v3   & 4h  & 0.0036 Mil  & - & Google T5   \\

OPT-IML~\cite{OPT_IML}  & arXiv\textquotesingle22  & - & Meta & General & 175B & $\times$   & OPT & 8k  & 2B & 128 & 40G A100   & - & - & D+T & Megatron    \\

Flan-U-PaLM~\cite{Flan}  & ICLR\textquotesingle22  & Apache-2.0 & Google & General & 540B & $\checkmark$  & U-PaLM  & 30k   & -  & 512 & TPU v4 & - & - & - & JAX+T5X  \\

mT0~\cite{mT0andBLOOMZ}  & ACL\textquotesingle23  & Apache-2.0  & ~HuggingFace+~ & General & 13B  & $\checkmark$ & mT5 & -  & - & -  & - & -  & -  & - & -\\

Sparrow~\cite{sparrow}  & arXiv\textquotesingle22 & -   & Google & Dialog & 70B    & $\times$   & Chinchilla & -  & -  & 64 & TPU v3   & -  & - & M    & -       \\

WizardCoder~\cite{wizardcoder}~ & arXiv\textquotesingle23 & Apache-2.0 & HK Bapt. & Coding & 15B & $\times$ & StarCoder & 200 & S-78k  & - & - & -  & - & - & -\\

Alpaca~\cite{alpaca} & Github\textquotesingle23 & Apache-2.0 & Stanford & General & 13B & $\checkmark$ & LLaMA & 3-Epoch & S-52k  & 8 & 80G A100 & 3h  & 600 & FSDP & PyTorch\\

Vicuna~\cite{vicuna} & Github\textquotesingle23 & Apache-2.0 & LMSYS & General & 13B & $\checkmark$ & LLaMA & 3-Epoch & S-125k  & - & - & -  & - & FSDP & PyTorch\\

LIMA~\cite{lima}& arXiv\textquotesingle23 & - & Meta+ & General & 65B & - & LLaMA & 15-Epoch & S-1000  & - & - & -  & - & - & -\\

Koala~\cite{koala}& Github\textquotesingle23 & Apache-2.0 & UC-Berkley & General & 13B & $\times$ & LLaMA & 2-Epoch & S-472k  & 8 & A100 & 6h & 100 & - & JAX/FLAX\\

%\multirow{-25}{*}{\begin{tabular}[c]{@{}c@{}}Closed\\ Source\end{tabular}} & \multirow{-18}{*}{\begin{tabular}[c]{@{}c@{}}Publicly\\ Available\end{tabular}} &
\bottomrule
\end{tabular}
}
}
\label{tab:statistics_it}
\end{table*}

% \section{Findings \& Insights}
% \label{sec:Findings}
% Training a billion-scale model is difficult as compared to a smaller model. LLMs are prone to various instabilities during training, such as hardware failure and instability. Other than this, LLMs exhibit different behaviors such as emergent abilities, improved zero-shot, few-shot, and reasoning abilities. Researchers report these essential details in their papers for results reproduction and field progress. We identify critical information in Table~\ref{tab:pre_trained_findings}~and~\ref{tab:instruction_tuned_findings} such as architecture, training strategies, and pipelines that improve LLMs' performance or other abilities acquired because of changes mentioned in section~\ref{sec_review}.  
\section{Model Configurations}
\label{Model_Configurations_}
We provide different statistics of pre-trained and instruction-tuned models in this section. This includes information such as publication venue, license type, model creators, steps trained, parallelism, etc in Table~\ref{tab:statistics_pt} and Table~\ref{tab:statistics_it}. Architecture details of pre-trained LLMs are available in Table~\ref{tab:arch_details}. Providing these details for instruction-tuned models is unnecessary because it fine-tunes pre-trained models for instruction datasets. Hence, architectural details are the same as the baselines. Moreover, optimization settings for various LLMs are available in Table~\ref{tab:opt_details} and Table~\ref{tab:opt_details_it}. We do not include details on precision, warmup, and weight decay in Table~\ref{tab:opt_details_it}. These details are not as important as others to mention for instruction-tuned models, and are not provided by the papers. \\

\begin{table*}[!t]
\rowcolors{2}{gray!25}{white}

    \centering
    \caption{Architecture details of LLMs. Here, \enquote{PE} is the positional embedding, \enquote{nL} is the number of layers, \enquote{nH} is the number of attention heads, \enquote{HS} is the size of hidden states.}
    \label{tab:arch_details}
\resizebox{\linewidth}{!}{
    \begin{tabular}{lcrccccccrrrr}
    \toprule
 \rowcolor{gray!50}
        \textbf{Models}&\textbf{Type}&\begin{tabular}[c]{@{}c@{}}\textbf{Training}\\ \textbf{Objective}\end{tabular}&\textbf{Attention}&\textbf{Vocab}&\textbf{Tokenizer}&\textbf{Norm}&\textbf{PE}&\textbf{Activation}&\textbf{Bias}& \textbf{nL}& \textbf{nH}&\textbf{HS} \\
\midrule

T5~(11B)&Enc-Dec&Span Corruption&Standard&32k&SentencePiece&Pre-RMS &Relative&ReLU&$\times$&24&128&1024\\

GPT3~(175B) & Causal-Dec & Next Token & Dense+Sparse &- & - & Layer & Learned & GeLU & \checkmark & 96 & 96 & 12288\\

mT5~(13B)&Enc-Dec&Span Corruption&Standard&250k&SentencePiece&Pre-RMS &Relative&ReLU& - & - & - &-\\

PanGu-$\alpha$~(200B)&Causal-Dec&Next Token&Standard&40k&BPE&Layer&-&-&-&64&128&16384\\

CPM-2~(198B)&Enc-Dec&Span Corruption&Standard&250k&SentencePiece&Pre-RMS &Relative&ReLU& - & 24 & 64 &-\\

Codex~(12B)&Causal-Dec&Next Token&Standard& - &BPE+ &Pre-Layer&Learned&GeLU&-&96&96&12288\\

ERNIE 3.0~(10B)&Causal-Dec&Next Token&Standard& - &WordPiece &Post-Layer&Relative&GeLU&-&48&64&4096\\

Jurassic-1~(178B)&Causal-Dec&Next Token &Standard&256k &SentencePiece$^*$ &Pre-Layer&Learned&GeLU &$\checkmark$ &76 &96&13824 \\

HyperCLOVA~(82B)&Causal-Dec&Next Token&Dense+Sparse& - &BPE* &Pre-Layer&Learned&GeLU&-&64&80&10240\\

Yuan 1.0~(245B)&Causal-Dec&Next Token& Standard & - &- &-&-&-&-&76&-&16384\\

Gopher~(280B)&Causal-Dec&Next Token&Standard&32k&SentencePiece&Pre-RMS&Relative&GeLU&$\checkmark$& 80&128&16384 \\

ERNIE 3.0 Titan~(260B)&Causal-Dec&Next Token& Standard&- &WordPiece &Post-Layer&Relative&GeLU&-&48&192&12288\\

GPT-NeoX-20B & Causal-Dec & Next Token & Parallel & 50k & BPE & Layer & Rotary & GeLU & $\checkmark$ & 44 & 64 & - \\

OPT~(175B)&Causal-Dec&Next Token&Standard&-&BPE&-&-&ReLU&$\checkmark$&96&96&-\\

BLOOM~(176B)&Causal-Dec&Next Token&Standard&250k&BPE&Layer&ALiBi&GeLU&$\checkmark$&70&112&14336\\

Galactica~(120B)&Causal-Dec&Next Token&Standard&50k&BPE+custom&Layer&Learned&GeLU&$\times$&96&80&10240\\

GLaM~(1.2T)&MoE-Dec&Next Token& Standard&256k &SentencePiece &Layer&Relative&GeLU&$\checkmark$&64&128&32768\\

LaMDA~(137B)&Causal-Dec&Next Token &Standard&32k &BPE &Layer &Relative &GeGLU&-&64&128&8192\\

MT-NLG~(530B)&Causal-Dec&Next Token&Standard&50k&BPE&Pre-Layer&Learned&GeLU&$\checkmark$&105&128&20480\\

AlphaCode~(41B)&Enc-Dec&Next Token%+ Masked language modeling loss
& Multi-query&8k &SentencePiece &-&-&-&-&64&128&6144\\

Chinchilla~(70B)&Causal-Dec&Next Token&Standard&32k&SentencePiece-NFKC&Pre-RMS&Relative&GeLU&$\checkmark$& 80&64&8192\\

PaLM~(540B)&Causal-Dec&Next Token&Parallel+Multi-query&256k&SentencePiece&Layer&RoPE&SwiGLU&$\times$&118&48&18432\\

AlexaTM~(20B)&Enc-Dec&Denoising& Standard&150k &SentencePiece &Pre-Layer&Learned&GeLU&$\checkmark$&78&32&4096\\

Sparrow~(70B)&Causal-Dec&Pref.\&Rule RM&-&32k&SentencePiece-NFKC&Pre-RMS&Relative&GeLU&$\checkmark$&16$^*$&64&8192\\

U-PaLM~(540B)&Non-Causal-Dec&MoD&Parallel+Multi-query&256k&SentencePiece&Layer&RoPE&SwiGLU&$\times$&118&48&18432\\

UL2~(20B) & Enc-Dec & MoD &Standard& 32k & SentencePiece & - & - & - & - & 64 & 16 & 4096\\

GLM (130B) & Non-Causal-Dec & AR Blank Infilling &Standard& 130k & SentencePiece & Deep & RoPE & GeGLU & \checkmark & 70 & 96 & 12288\\

CodeGen~(16B)&Causal-Dec&Next Token&Parallel&-&BPE&Layer&RoPE&-&-&34&24&-\\

LLaMA~(65B)&Causal-Dec&Next Token& Standard&32k &BPE &Pre-RMS&RoPE&SwiGLU&-&80&64&8192\\

PanGu-$\Sigma$~(1085B) & Causal-Dec & Next Token &Standard& - & BPE & Fused Layer & - & FastGeLU & - & 40 & 40 & 5120 \\

%CPM-2~(198B)&Enc-Dec&Span Corruption&250k&SentencePiece&Pre-RMS &Relative&ReLU& - & 24 & 64 &-\\

%WebGPT~(175B)&Causal-Dec&BC+RM &- &BPE &Pre-Layer & Learned & GeLU & \checkmark & 96 & 96 & 12288\\
%InstructGPT~(6B)&Causal-Dec&RM+PPO& - &BPE &Pre-Layer&Learned&GeLU&-&32&32&4096\\
BloombergGPT~(50B)&Causal-Dec&Next Token&Standard&131k&Unigram&Layer&ALiBi&GeLU&$\checkmark$&70&40&7680\\
Xuan Yuan 2.0~(176B)& Causal-Dec & Next Token &Self&250k & BPE & Layer & ALiBi & GeLU & $\checkmark$ & 70 & 112 & 14336 \\
CodeT5+~(16B)& Enc-Dec & SC+NT+Cont.+Match &Standard&- & Code-Specific & - & - & - & - & - & - & - \\
StarCoder~(15.5B)& Causal-Dec & FIM & Multi-query&49k & BPE & - & Learned & -  & -  & 40 & 48 & 6144 \\
LLaMA-2~(70B)& Causal-Dec & Next Token & Grouped-query & 32k & BPE & Pre-RMS & RoPE  & SwiGLUE  & - & - & - & -\\
PaLM-2 & - & MoD & Parallel & - & - & - & -  & -  & - & - & - & - \\

LLaMA-3.1~(405B) & Causal-Dec & Next Token & Grouped-query & 128k & BPE & Pre-RMS & RoPE  & SwiGLU  & - & 126 & 128 & 16384 \\

Nemotron-4~(340B) & Causal-Dec & Next Token & Standard & 256k & SentencePiece  & - & RoPE  & ReLU  & $\times$ & 96 & 96 & 18432 \\

DeepSeek~(67B) & Causal-Dec & Next Token & Grouped-query & 100k & BBPE & Pre-RMS & RoPE & SwiGLU & - & 95 & 64 & 8192 \\

DeepSeek-v2~(67B) & MoE-Dec & Next Token & Multi-Head Latent & 100k & BBPE & Pre-RMS & RoPE & SwiGLU & - & 60 & 128 & 5120 \\

\bottomrule
\end{tabular}}
\end{table*}

\begin{table*}[!t]
\rowcolors{2}{gray!25}{white}
\centering
\caption{Summary of optimization settings used for pre-trained LLMs. The values for weight decay, gradient clipping, and dropout are 0.1, 1.0, and 0.1, respectively, for most of the LLMs.}
\label{tab:opt_details}
\resizebox{\linewidth}{!}{
\begin{tabular}{lrrccc|p{1.05cm}p{0.6cm}p{0.85cm}|p{0.5cm}p{0.5cm}p{0.6cm}|ccc}
\toprule
\rowcolor{gray!50} &&\textbf{Sequence}&& & \textbf{LR}& \multicolumn{3}{c|}{\textbf{Optimizers}}& \multicolumn{3}{c|}{\textbf{Precision}}&\textbf{Weight}&\textbf{Grad}&\\ 

\rowcolor{gray!50} \textbf{Models}&\textbf{Batch Size}&\textbf{Length}& \textbf{LR}&\textbf{Warmup}&\textbf{Decay}&\textbf{AdaFactor}&\textbf{Adam}&\textbf{AdamW}&\textbf{FP16}&\textbf{BF16}&\textbf{Mixed}&\textbf{Decay}&\textbf{Clip}& \textbf{Dropout}\\\midrule

T5~(11B) & 2$^{11}$ & 512 & 0.01 & $\times$  & inverse square root     & $\checkmark$ & & & -&- & - & - & - & $\checkmark$              \\

GPT3~(175B) & 32K  & - & 6e-5 & $\checkmark$ & cosine    & & $\checkmark$ & & $\checkmark$ & & & $\checkmark$ & $\checkmark$  & -                \\

mT5~(13B) & 1024& 1024 & 0.01 & -  & inverse square root  & $\checkmark$ & &  &- & - & - & - & - & $\checkmark$              \\

PanGu-$\alpha$~(200B)   & - & 1024 & 2e-5 & - & -  & - & - & - &- & $\checkmark$ & - & - &- &-              \\
CPM-2~(198B) & 1024& 1024 & 0.001 & -  & -    & $\checkmark$ & & &  -&  -& - & - & - & $\checkmark$              \\

Codex~(12B) & -  & - & 6e-5 & $\checkmark$ & cosine    & & $\checkmark$ & & $\checkmark$ & & & $\checkmark$ & -  & -                \\

ERNIE 3.0~(12B) & 6144  & 512 & 1e-4 & $\checkmark$ & linear    & & $\checkmark$ &  &  -& -& - & $\checkmark$ & -  & -                \\

Jurassic-1~(178B) & 3.2M &2048 & 6e-5 & $\checkmark$ & cosine & & $\checkmark$ &  & $\checkmark$ & & & $\checkmark$ & $\checkmark$ & -                \\

HyperCLOVA~(82B) & 1024  & - & 6e-5 & - & cosine    & & &$\checkmark$  &  -&  -&- & $\checkmark$ & -  & -                \\

Yuan 1.0~(245B) & $<$10M &2048 & 1.6e-4 & $\checkmark$ & cosine decay to 10\%    & & $\checkmark$ &  & - & - & -  & $\checkmark$ & - & -                \\

Gopher~(280B) & 3M  & 2048 & 4e-5 & $\checkmark$ & cosine decay to 10\%    & & $\checkmark$ &   & &$\checkmark$  &  & -   & $\checkmark$  & -                \\

ERNIE 3.0 Titan~(260B)  & -  & 512 & 1e-4 & $\checkmark$ & linear    & & $\checkmark$ &  & $\checkmark$ & & & $\checkmark$ & $\checkmark$  & -      \\

GPT-NeoX-20B & 1538  & 2048 & 0.97e-5 & $\checkmark$ & cosine & & &$\checkmark$  & $\checkmark$ & &  & $\checkmark$ & $\checkmark$ & $\times$              \\
OPT~(175B) & 2M  & 2048 & 1.2e-4 & - & linear & & &$\checkmark$  & $\checkmark$ & & & $\checkmark$  & $\checkmark$ & $\checkmark$              \\

BLOOM~(176B) & 2048  & 2048 & 6e-5 & $\checkmark$  & cosine    & & $\checkmark$ &   & &$\checkmark$  & & $\checkmark$ & $\checkmark$ & $\times$              \\

Galactica~(120B) & 2M & 2048 & 7e-6 & $\checkmark$ & linear decay to 10\%    & & &$\checkmark$  & - & - & - & $\checkmark$  & $\checkmark$ & $\checkmark$              \\

GLaM~(1.2T) & 1M &1024 & 0.01 & - & inverse square root   & $\checkmark$ & &  & \multicolumn{3}{c|}{FP32 + $\checkmark$}  & - & $\checkmark$ & $\times$                \\

LaMDA~(137B) & 256K  & - & - & - & -  & - & - & - & - &- & - &- & - &-              \\

MT-NLG~(530B) & 1920  & 2048 & 5e-5 & $\checkmark$ & cosine decay to 10\%   & & $\checkmark$ &  & &$\checkmark$  & & $\checkmark$ & $\checkmark$  & - \\

AlphaCode~(41B) & 2048 &1536+768 & 1e-4 &$\checkmark$& cosine decay to 10\%   & & &$\checkmark$  & &$\checkmark$  &  &$\checkmark$ & $\checkmark$ &-               \\

Chinchilla~(70B) & 1.5M & 2048 & 1e-4 & $\checkmark$ & cosine decay to 10\% & & &$\checkmark$  & &$\checkmark$  & & - & - & -                \\

PaLM~(540B) & 2048 & 2048 & 0.01 & - & inverse square root & $\checkmark$ & &  & - & - & -  & $\checkmark$ & $\checkmark$ & $\times$             \\

AlexaTM~(20B) & 2M &1024 & 1e-4 &- & linear decay to 5\%   & & $\checkmark$ &  & &$\checkmark$  &  &$\checkmark$ & - &$\checkmark$              \\

U-PaLM~(540B) & 32 & 2048 & 1e-4 & - & cosine & $\checkmark$ & & & - & -  & -  & - & - & -              \\

UL2~(20B) & 1024  & 1024 & - & - & inverse square root & - & - & - & -  & - & - & $\times$  & - & -            \\

GLM~(130B) & 4224  & 2048 & 8e-5  & $\checkmark$ & cosine    & & &$\checkmark$  & $\checkmark$ & & & $\checkmark$ & $\checkmark$ & $\checkmark$              \\

CodeGen~(16B) & 2M& 2048 & 5e-5 & $\checkmark$  & cosine    & & $\checkmark$ &  & - & - & - & $\checkmark$ & $\checkmark$ & -              \\

LLaMA~(65B) & 4M Tokens &2048 & 1.5e-4 & $\checkmark$ & cosine decay to 10\%    & & &$\checkmark$  & - & - &  -  & $\checkmark$ & $\checkmark$ & -                \\

PanGu-$\Sigma$~(1.085T) & 512  & 1024 & 2e-5 & $\checkmark$ & - & & $\checkmark$ &  & &&$\checkmark$   & - & - & -                \\
BloombergGPT~(50B) & 2048  & 2048 & 6e-5 & $\checkmark$ & cosine & & &$\checkmark$  & &&$\checkmark$   & $\checkmark$ & $\checkmark$ & $\times$               \\
Xuan Yuan 2.0~(176B) & 2048  & 2048 & 6e-5 & $\checkmark$ & cosine & & $\checkmark$ &  & $\checkmark$ & & & $\checkmark$ & $\checkmark$ & -                \\
CodeT5+~(16B) & 2048  & 1024 & 2e-4 & - & linear & & &$\checkmark$  & &&$\checkmark$   & $\checkmark$ & - & -                \\
StarCoder~(15.5B) & 512  & 8k & 3e-4 & $\checkmark$ & cosine & & $\checkmark$ &  & &$\checkmark$  & & $\checkmark$ & - & -                \\
LLaMA-2~(70B) & 4M Tokens  & 4k & 1.5e-4 & $\checkmark$ & cosine & &  & $\checkmark$ & &$\checkmark$  & & $\checkmark$ & $\checkmark$ & -                \\

LLaMA-3.1~(405B) & 16M  & 8192 & 8e-5 & $\checkmark$ & linear+cosine & &  & $\checkmark$ & &$\checkmark$  & & - & - & -                \\

Nemotron-4~(340B) & 2304  & 4096 & - & - & linear & - & - & - & &$\checkmark$  & & - & - & $\times$                \\

DeepSeek~(67B) & 4608  & 4096 & 3.2e-4 & $\checkmark$ & cosine & &  & $\checkmark$ & &$\checkmark$  & & $\checkmark$ & $\checkmark$ & -                \\

DeepSeek-v2~(67B) & 9216  & 4k & 2.4e-4 & $\checkmark$ & step-decay & &  & $\checkmark$ & -&-  & -& $\checkmark$ & $\checkmark$ & -                \\

\bottomrule
\end{tabular}
}
\end{table*}

\begin{table*}[!tbp]
\rowcolors{2}{gray!25}{white}

\centering
\caption{Summary of optimization settings used for instruction-tuned LLMs. Values for gradient clipping and dropout are the same as the pre-trained models, while no model uses weight decay for instruction tuning.}
\label{tab:opt_details_it}
\resizebox{\linewidth}{!}{

%\begin{tabular}{lrrccc|ccc|cc}
%\toprule
%\rowcolor{gray!50}         &  & \textbf{Sequence}&   &    &  & \multicolumn{2}{c}{\textbf{Optimizers}}           &    & \textbf{Grad} & \\ 
%\rowcolor{gray!50} \textbf{Models}          & \textbf{Batch Size} & \textbf{Length}& \textbf{LR}   & \textbf{Warmup} & \textbf{LR\_Decay}   & 
% \textbf{AdaFactor} &      \textbf{Adam}  & \textbf{AdamW}     &  \textbf{Clip} & \textbf{Dropout} \\ \midrule

\begin{tabular}{lrrccc|ccc|cc}
\toprule
\rowcolor{gray!50} & & \textbf{Sequence} & & & & \multicolumn{3}{c}{\textbf{Optimizers}} & \textbf{Grad} & \\ 
\rowcolor{gray!50} \textbf{Models} & \textbf{Batch Size} & \textbf{Length} & \textbf{LR} & \textbf{Warmup} & \textbf{LR\_Decay} & \textbf{AdaFactor} & \textbf{Adam} & \textbf{AdamW} & \textbf{Clip} & \textbf{Dropout} \\ \midrule

WebGPT~(175B) & BC:512, RM:32  & - & 6e-5 & - & -    & & $\checkmark$  & & -  & -                \\

T0~(11B) & 1024 & 1280 & 1e-3 & -  & -   & $\checkmark$ & & & - & $\checkmark$             \\

Tk-Instruct~(11B) & 1024 & - & 1e-5  & -  & constant  & -& - & - &  - & -             \\

OPT-IML~(175B) & 128 & 2048 & 5e-5 & $\times$  & linear   & &$\checkmark$  & &  $\checkmark$ & $\checkmark$             \\
%InstructGPT~(6B) & SFT:32  & 2000 & 9.65e-6 & no & cosine decay to 10\%   & Adam & FP16 & - & -  & 0.2 \\

Flan-U-PaLM~(540B) & 32 & - & 1e-3 & -  & constant   & $\checkmark$ & & & -  & $\checkmark$             \\

Sparrow~(70B)& RM: 8+16, RL:16 &- & 2e-6 &$\checkmark$ & cosine decay to 10\%   & $\checkmark$ &  & & $\checkmark$ &$\times$ \\

WizardCoder~(15B) & 512 & 2048 & 2e-5 & $\checkmark$  & cosine   & - & - & - & -  & -             \\

Alpaca~(13B) & 128 & 512 & 1e-5 & $\checkmark$  & cosine   & - & - & $\checkmark$ & $\checkmark$  & $\times$             \\

Vicuna~(13B) & 128 & -2048& 2e-5 & $\checkmark$  & cosine   &  &  & $\checkmark$ & -  & $\times$             \\

LIMA~(65B) & 32 & 2048 & 1e-5 & $\times$  &  linear  &  &  & $\checkmark$& -  & $\checkmark$             \\

\bottomrule
\end{tabular}
}
\end{table*}

\section{Datasets and Evaluation}
\label{Datasets_and_Evaluation_}
Generating training and evaluation datasets is expensive because of the large-scale data demand of LLMs. Hence, datasets for training and benchmarking these models are topics of key importance. A summary of datasets commonly used by LLMs is provided next.    
% In Fig.~\ref{fig:dataDist}, we show the distribution of the existing datasets for various NLP tasks. We restrict our distribution to only the most important tasks in the literature by including tasks with at least 20 datasets. LLMs can directly benefit from these datasets for training and evaluation. 
%In general, the performance of LLMs greatly depends on the training dataset. A model trained on good-quality data will likely perform better on evaluation benchmarks. Specific training and evaluation datasets LLMs use are summarized in Table~\ref{tab:datasets} and~\ref{tab:datasets_inst_tuned}. }
\subsection{Training Datasets}
The performance of LLMs largely depends on the training data's quality, size, and diversity. Preparing training datasets of high quality at a large scale is laborious. Researchers have suggested various pre-training and fine-tuning datasets to enhance LLMs capabilities. We summarize these efforts in Table~\ref{tab:datasets}. While numerous training datasets are available in the literature, we cover the most widely used ones in our summary.        
\subsection{Evaluation Datasets and Tasks}
\label{ss:evaluation_tasks}
The evaluation of LLMs is important in gauging their proficiency and limitations. This process measures the model's ability to comprehend, generate, and interact with human language across a spectrum of tasks. Evaluating a language model (LM) is divided into two broader categories: 1) natural language understanding (NLU) and 2) natural language generation (NLG). It is emphasized that tasks in NLU and NLG are softly categorized and are often used interchangeably in the literature. \\
\noindent
~\emph{~\textbf{Natural Language Understanding:}} It measures the language understanding capacity of LMs. It encompasses multiple tasks, including sentiment analysis, text classification, natural language inference (NLI), question answering (QA), commonsense reasoning (CR), mathematical reasoning (MR), reading comprehension (RC), etc. \\
\noindent
~\emph{~\textbf{Natural Language Generation:}} It assesses the language generation capabilities of LLMs by understanding the provided input context. It includes tasks such as summarization, sentence completion, machine translation (MT), dialogue generation, etc. \\
\noindent
Numerous datasets are proposed for each task, evaluating LLMs against different characteristics. To provide an overview of evaluation datasets, we briefly discuss a few famous datasets within each category and offer a comprehensive list of datasets in Table~\ref{tab:evaluation_datasets}. Moreover, we show a detailed overview of the training datasets and evaluation tasks and benchmarks used by various pre-trained LLMs in Table~\ref{tab:illustration_datasets} and fine-tuned LLMs in Table~\ref{tab:illustration_fine_tuned_datasets}. We also compare the top-performing LLMs in various NLP tasks in Table~\ref{tab:performance_comparison}.    

% In contrast, Natural Language Generation (NLG) tasks include text summarization, translation, and more. These tasks form part of established benchmarks that facilitate the comparison of different models. % While the quality of the training dataset significantly impacts the performance of LLMs, the evaluation process also takes into account the model's ability to generalize to new tasks. The following subsections delve deeper into the various aspects of evaluating LLMs, providing a comprehensive understanding of their performance across NLU and NLG tasks.

\begin{table*}[!t]
\caption{Details of various well-known pre-training and fine-tuning datasets. Here, alignment means aligning with human preferences.}
\label{tab:datasets}
\resizebox{1\textwidth}{!}{% 
\begin{tabular}{l|c|c|c|c|c|p{5.8cm}} \hline

\rowcolor{gray!50}Dataset & Type  & Size/Samples & Tasks & Source & Creation & Comments \\ \hline \hline

C4~\cite{T5} & Pretrain & 806GB & - & Common Crawl & Automated & A clean, multilingual dataset with billions of tokens \\ \hline

mC4~\cite{mT5} & Pretrain & 38.49TB & - & Common Crawl & Automated & A multilingual extension of the C4 dataset, mC4 identifies over 100 languages using cld3 from 71 monthly web scrapes of Common Crawl. \\ \hline

% Common Crawl & Automated & \begin{tabular}[c]{@{}c@{}}A multilingual extension of the C4 dataset,\\ mC4 identifies over 100 languages using cld3\\ from 71 monthly web scrapes of Common Crawl. \end{tabular}

PILE~\cite{gao2020pile} & Pretrain & 825GB & - &  \begin{tabular}[c]{@{}c@{}} Common Crawl, PubMed Central, \\OpenWebText2, ArXiv, GitHub,\\Books3, and others\end{tabular} & Automated & A massive dataset comprised of 22 constituent sub-datasets\\ \hline

ROOTs~\cite{laurenccon2022bigscience} & Pretrain & 1.61TB & - & 498 Hugging Face datasets & Automated & 46 natural and 13 programming languages \\ \hline

MassiveText~\cite{gopher} & Pretrain & 10.5TB & - & \begin{tabular}[c]{@{}c@{}}MassiveWeb, Books, News, \\Wikipedia, Github, C4\end{tabular} & Automated & 99\% of the data is in English \\ \hline

Wikipedia~\cite{Wikipedia} & Pretrain & - & - & Wikipedia & Automated & Dump of wikipedia \\ \hline

RedPajama~\cite{redpajama} & Pretrain & 5TB & - & \begin{tabular}[c]{@{}c@{}}CommonCrawl, C4, Wikipedia, \\Github, Books, StackExchange\end{tabular} & Automated & Open-source replica of LLaMA dataset  \\ \hline

PushShift.io Reddit & Pretrain & 21.1GB & - & Reddit & Automated & Submissions and comments on Reddit from 2005 to 2019 \\ \hline

BigPython~\cite{CodeGen} & Pretrain & 5.5TB & Coding & GitHub & Automated & - \\ \hline

Pool of Prompt (P3)~\cite{T0} & Instructions & 12M & 62 & PromptSource & Manual & A Subset of PromptSource, created from 177 datasets including summarization, QA, classification, etc.  \\ \hline

xP3~\cite{mT0andBLOOMZ} & Instructions & 81M & 71 & P3+Multilingual datasets & Manual & Extending P3 to total 46 languages     \\ \hline

Super-NaturalInstructions (SNI)~\cite{Tk-INSTRUCT} & Instructions & 12.4M & 1616 & Multiple datasets & Manual & Extending P3 with additional multi-lingual datasets, total 46 languages    \\ \hline

Flan~\cite{Flan} & Instructions & 15M & 1836 & Muffin+T0-SF+NIV2 & Manual & Total 60 languages   \\ \hline

OPT-IML~\cite{OPT_IML} & Instructions & 18.1M & 1667 & - & Manual & -    \\ \hline

Self-Instruct~\cite{ft_self_instruct} & Instructions & 82k & 175 & - & Automated & Generated 52k instructions with 82k samples from 175 seed tasks using GPT-3   \\ \hline

Alpaca~\cite{alpaca} & Instructions & 52k & - & - & Automated & Employed self-instruct method to generate data from text-davinci-003  \\ \hline

Vicuna~\cite{vicuna} & Instructions & 125k & - & ShareGPT & Automated & Conversations shared by users on ShareGPT using public APIs    \\ \hline

LLaMA-GPT-4~\cite{llama_gpt_4} & Instructions & 52k & - & Alpaca & Automated & Recreated Alpaca dataset with GPT-4 in English and Chinese   \\ \hline

Unnatural Instructions~\cite{unnatural_inst} & Instructions & 68k & - & 15-Seeds (SNI) & Automated & -    \\ \hline

LIMA~\cite{lima} & Instructions & 1k & - & Multiple datasets & Manual & Carefully created samples to test performance with fine-tuning on less data    \\ \hline

Anthropic-HH-RLHF~\cite{hh_rlhf} & Alignment & 142k & - & - & Manual &     \\ \hline

Anthropic-HH-RLHF-2~\cite{red_team_lessons_learned} & Alignment & 39k & - & - & Manual &     \\ \hline

\end{tabular}}
\vspace{2mm}
\end{table*}

% \begin{figure*}[htbp]
%     \centering
%     \includegraphics[width = 0.9\textwidth]{Figure/Datasets_for_Tasks.png}
%     \caption{A distribution of datasets proposed for different NLP tasks. We include only the tasks for which at least 20 datasets have already been proposed.}
%     \label{fig:dataDist}
% \end{figure*}

\begin{table*}[!t]
\caption{Categorized evaluation datasets used in evaluating LLMs.}
\label{tab:evaluation_datasets}
\resizebox{1\textwidth}{!}{%
\begin{tabular}{l|p{15cm}}
\hline
\textbf{Type} &
  \textbf{Datasets/Benchmarks} \\ \hline
Multi-Task &
  MMLU~\cite{hendrycks2020measuring}, SuperGLUE~\cite{wang2019superglue}, BIG-bench~\cite{srivastava2022beyond}, GLUE~\cite{wang2018glue}, BBH~\cite{srivastava2022beyond}, CUGE~\cite{yao2021cuge}, ZeroCLUE~\cite{xu2020clue}, FewCLUE~\cite{xu2021fewclue}, Blended Skill Talk~\cite{smith2020can}, HELM~\cite{HELM}, KLUE-STS~\cite{park2021klue} \\ \hline
Language Understanding &
  CoQA~\cite{reddy2019coqa}, WiC~\cite{pilehvar2018wic}, Wikitext103~\cite{merity2016pointer}, PG19~\cite{rae2019compressive}, LCQMC~\cite{liu2018lcqmc}, QQP~\cite{QQP}, WinoGender~\cite{rudinger2018gender}, CB~\cite{de2019commitmentbank}, FinRE~\cite{li2019chinese}, SanWen~\cite{xu2017discourse}, AFQMC~\cite{xu2020clue}, BQ Corpus~\cite{chen2018bq}, CNSS~\cite{liu2018matching}, CKBQA 13~\cite{li2016dataset}, CLUENER~\cite{xu2020clue}, Weibo~\cite{peng2015named}, AQuA~\cite{ling2017program}, OntoNotes~\cite{weischedel2011ontonotes},  HeadQA~\cite{vilares2019head},  Twitter Dataset~\cite{blodgett2016demographic}\\ \hline
\begin{tabular}[c]{@{}l@{}}Story Cloze and \\ Sentence Completion\end{tabular} &
  StoryCloze~\cite{mostafazadeh2016corpus}, LAMBADA~\cite{paperno2016lambada}, LCSTS~\cite{hu2015lcsts}, AdGen~\cite{shao2019long}, E2E~\cite{novikova2017e2e}, CHID~\cite{zheng2019chid}, CHID-FC~\cite{xu2021fewclue} \\ \hline
\begin{tabular}[c]{@{}l@{}}Physical Knowledge and \\ World Understanding\end{tabular} &
  PIQA~\cite{bisk2020piqa}, TriviaQA~\cite{joshi2017triviaqa}, ARC~\cite{clark2018think}, ARC-Easy~\cite{clark2018think}, ARC-Challenge~\cite{clark2018think}, PROST~\cite{aroca2021prost}, OpenBookQA~\cite{mihaylov2018can}, WebNLG~\cite{ferreira20202020}, DogWhistle Insider \& Outsider~\cite{xu2021blow} \\ \hline
\begin{tabular}[c]{@{}l@{}}Contextual Language \\ Understanding\end{tabular} &
  \begin{tabular}[c]{@{}l@{}}RACE~\cite{lai2017race}, RACE-Middle~\cite{lai2017race}, RACE-High~\cite{lai2017race}, QuAC~\cite{choi2018quac}, StrategyQA~\cite{geva2021did}, Quiz Bowl~\cite{boyd2012besting}, \\cMedQA~\cite{zhang2017chinese},cMedQA2~\cite{zhang2018multi}, MATINF-QA~\cite{xu2020matinf}\end{tabular} \\ \hline
Commonsense Reasoning &
  WinoGrande~\cite{sakaguchi2021winogrande}, HellaSwag~\cite{zellers2019hellaswag}, COPA~\cite{roemmele2011choice}, WSC~\cite{levesque2012winograd}, CSQA~\cite{talmor2018commonsenseqa}, SIQA~\cite{sap2019socialiqa}, C\textsuperscript{3}~\cite{sun2020investigating}, CLUEWSC2020~\cite{xu2020clue}, CLUEWSC~\cite{xu2020clue}, CLUEWSC-FC~\cite{xu2021fewclue}, ReCoRD~\cite{zhang2018record} \\ \hline
Reading Comprehension &
  SQuAD~\cite{rajpurkar2016squad}, BoolQ~\cite{clark2019boolq}, SQUADv2~\cite{rajpurkar2018know}, DROP~\cite{dua2019drop}, RTE~\cite{dagan2005pascal}, WebQA~\cite{chang2022webqa}, CMRC2017~\cite{cui2017cmrc17}, CMRC2018~\cite{cui2018span}, CMRC2019~\cite{cui2020sentence}, COTE-BD~\cite{li2018character}, COTE-DP~\cite{li2018character}, COTE-MFW~\cite{li2018character}, MultiRC~\cite{khashabi2018looking}, Natural Questions~\cite{kwiatkowski2019natural}, CNSE~\cite{liu2018matching}, DRCD~\cite{shao2018drcd}, DuReader~\cite{he2017dureader}, Dureader\textsubscript{robust}~\cite{tang2020dureaderrobust}, DuReader-QG~\cite{he2017dureader}, SciQ~\cite{welbl2017crowdsourcing}, Sogou-log~\cite{xiong2017end}, Dureader\textsubscript{robust}-QG~\cite{tang2020dureaderrobust}, QA4MRE~\cite{penas2013qa4mre}, KorQuAD 1.0~\cite{lim2019korquad1}, CAIL2018-Task1 \& Task2~\cite{xiao2018cail2018} \\ \hline
Mathematical Reasoning &
  MATH~\cite{hendrycks2021measuring}, Math23k~\cite{wang2017deep}, GSM8K~\cite{cobbe2021training}, MathQA~\cite{MathQA}, MGSM~\cite{shi2022language}, MultiArith~\cite{roy2016solving}, ASDiv~\cite{ASDiv}, MAWPS~\cite{koncel2016mawps}, SVAMP~\cite{patel2021nlp} \\ \hline
Problem Solving &
  HumanEval~\cite{codex}, DS-1000~\cite{ds_1000}, MBPP~\cite{austin2021program}, APPS~\cite{hendrycks2021measuring}, CodeContests~\cite{li2022competition} \\ \hline
\begin{tabular}[c]{@{}l@{}}Natural Language Inference \\\& Logical Reasoning\end{tabular} &
  ANLI~\cite{nie2019adversarial}, MNLI-m~\cite{williams2017broad}, MNLI-mm~\cite{williams2017broad},QNLI~\cite{rajpurkar2016squad}, WNLI~\cite{levesque2012winograd}, OCNLI~\cite{xu2020clue}, CMNLI~\cite{xu2020clue}, ANLI R1~\cite{nie2019adversarial}, ANLI R2~\cite{nie2019adversarial}, ANLI R3~\cite{nie2019adversarial}, HANS~\cite{mccoy2019right}, OCNLI-FC~\cite{xu2021fewclue}, LogiQA~\cite{liu2020logiqa}, StrategyQA~\cite{geva2021did} \\ \hline
Cross-Lingual Understanding &
  MLQA~\cite{lewis2019mlqa}, XNLI~\cite{conneau2018xnli}, PAWS-X~\cite{yang2019paws}, XSum~\cite{narayan1808don}, XCOPA~\cite{ponti2020xcopa}, XWinograd~\cite{tikhonov2021s}, TyDiQA-GoldP~\cite{clark2020tydi}, MLSum~\cite{scialom2020mlsum} \\ \hline
Truthfulness and Fact Checking &
  TruthfulQA~\cite{lin2021truthfulqa}, MultiFC~\cite{augenstein2019multifc}, Fact Checking on Fever~\cite{thorne2018fever} \\ \hline
Biases and Ethics in AI &
  ETHOS~\cite{mollas2020ethos}, StereoSet~\cite{nadeem2020stereoset}, BBQ~\cite{parrish2021bbq}, Winobias~\cite{zhao2018gender}, CrowS-Pairs~\cite{nangia2020crows} \\ \hline
Toxicity &
  RealToxicityPrompts~\cite{gehman2020realtoxicityprompts}, CivilComments toxicity classification~\cite{borkan2019nuanced} \\ \hline
Language Translation &
  WMT~\cite{bojar2016findings}, WMT20~\cite{loic2020findings}, WMT20-enzh~\cite{loic2020findings}, EPRSTMT~\cite{xu2021fewclue}, CCPM~\cite{li2021ccpm} \\ \hline
Scientific Knowledge &
  AminoProbe~\cite{galactica}, BioLAMA~\cite{galactica}, Chemical Reactions~\cite{galactica}, Galaxy Clusters~\cite{galactica}, Mineral Groups~\cite{galactica} \\ \hline
Dialogue &
  Wizard of Wikipedia~\cite{dinan2018wizard}, Empathetic Dialogues~\cite{rashkin2018towards}, DPC-generated~\cite{chinchilla} dialogues, ConvAI2~\cite{dinan2020second}, KdConv~\cite{zhou2020kdconv} \\ \hline
Topic Classification &
  TNEWS-FC~\cite{xu2021fewclue}, YNAT~\cite{park2021klue}, KLUE-TC~\cite{park2021klue}, CSL~\cite{xu2020clue}, CSL-FC~\cite{xu2021fewclue}, IFLYTEK~\cite{co2019iflytek} \\ \hline
%Others (without references) &
%  \begin{tabular}[c]{@{}l@{}}CNNDM, EnDe, ENFr, EnRo, WebQS, LPCC2014-SC, SE-ABSA16\_PHNS, SE-ABSA16\_CAME, \\ BDCI2019, Dureader\textsubscript{checklist}, Dureader\textsubscript{yesno}, KBQG, Knowledge probes, Latex equations, Codeforces competitions, \\ WMT language pairs, Financial Data, Multi-Language Evaluation, AGI Eval, CCKS2019, CCKS2020, TNEWS, \\ NLPCC-DBQA, CHIP2019, IFLYTEK-FC, CSLDCP, NLPCC-DBQA, CHIP2019, NLPCC2014-SC, SE-ABSA16\_PHNS, \\ SE-ABSA16\_CAME, BDCI2019, NSMC: a movie review dataset from NAVER movie, , Korean ML dataset, \\ AI Hub Korean-English, BUSTM, TNEWS, THUCNEWS, WPLC, Dureader\textsubscript{checklist}, Dureader\textsubscript{yesno}, PD\&CFT\end{tabular} \\ \hline
%Others (Ins Tuned) &
%  \begin{tabular}[c]{@{}l@{}}Wino (XL)~\cite{sakaguchi2021winogrande}, ELI5~\cite{fan2019eli5}, SUP-NATINST~\cite{Tk-INSTRUCT}, PromptSource~\cite{T0}, FLAN~\cite{Flan}, Super-NaturalInstructions~\cite{wang2022benchmarking}, \\ UnifiedSKG~\cite{xie2022unifiedskg}, CrossFit~\cite{ye2021crossfit}, ExMix~\cite{aribandi2021ext5}, T5~\cite{T5}, Reasoning, RAFT~\cite{alex2021raft}\end{tabular} \\ \hline
\end{tabular}}
\end{table*}

\begin{table*}[!tbhp]
\caption{An illustration of training datasets and evaluation tasks employed by pre-trained LLMs. Here, \enquote{QA} is question-answering, \enquote{Clf} is classification, \enquote{NLI} is natural language inference, \enquote{MT} is machine translation, \enquote{RC} is reading comprehension, \enquote{CR} is commonsense reasoning, \enquote{MR} is mathematical reasoning, \enquote{Mem.} is memorization.}
\label{tab:illustration_datasets}
\resizebox{\textwidth}{!}{ 
\begin{tabular}{l|p{6cm}|c|c|c|c|c|c|c|c|c|c|c|c|c}
\hline \hline
\rowcolor{gray!50} & & \multicolumn{3}{c}{\textbf{Benchmark}} & & & &  & & &&&& \\ 
\rowcolor{gray!50}\textbf{Models} & \begin{tabular}[c]{@{}c@{}}\textbf{Training Dataset} \end{tabular} & \multicolumn{1}{c}{\begin{tabular}[c]{@{}c@{}}BIG-\\bench\end{tabular}} & \multicolumn{1}{c}{MMLU} & \begin{tabular}[c]{@{}c@{}}Super\\GLUE\end{tabular} & \textbf{QA} & \textbf{Clf} & \textbf{NLI} & \textbf{MT} & \begin{tabular}[c]{@{}c@{}}\textbf{Cloze/}\\\textbf{Completion}\end{tabular} & \textbf{RC} & \textbf{CR} & \textbf{MR} & \textbf{Coding} & \begin{tabular}[c]{@{}c@{}}\textbf{Truthful/}\\\textbf{Bias/}\\\textbf{Toxicity/}\\\textbf{Mem.}\end{tabular} 
\\ \hline \hline

T5 &  C4~\cite{T5} & &&$\checkmark$&$\checkmark$ & & $\checkmark$ & $\checkmark$ & $\checkmark$ & $\checkmark$ & $\checkmark$ & $\checkmark$& &   \\ \hline

GPT-3 & Common Crawl, WebText, Books Corpora, Wikipedia &&&$\checkmark$&$\checkmark$ &  & & $\checkmark$& $\checkmark$& $\checkmark$ & & & & $\checkmark$ \\ \hline

mT5 & mC4~\cite{mT5} &    &     &       & $\checkmark$    &    & $\checkmark$ &  $\checkmark$  &        &   &   &  & \\ \hline

%                     Big  mmlu   sglue  qa     lu   nli  mt    cloze   comp
PanGu-$\alpha$ & 1.1TB Chinese Text Corpus & & & & $\checkmark$ & & $\checkmark$ & & $\checkmark$& $\checkmark$ & $\checkmark$& & &\\ \hline

CPM-2 & WuDaoCorpus~\cite{WuDaoCorpus} & & & &  & & & & & $\checkmark$& & $\checkmark$& & \\ \hline

Codex & 54 million public repositories from Github& & & &  & & & & & & & & $\checkmark$&  \\ \hline

ERNIE-3.0 & Chinese text corpora, Baidu Search, Web text, QA-long, QA-short, Poetry and Couplet Domain-specific data from medical, law, and financial area Baidu knowledge graph with more than 50 million facts& & & $\checkmark$& $\checkmark$ & $\checkmark$&$\checkmark$ & $\checkmark$&$\checkmark$ &$\checkmark$ & & $\checkmark$& &  \\ \hline

Jurassic-1 & Wikipedia, OWT, Books, C4,
Pile~\cite{gao2020pile}, arXiv, GitHub& & & & $\checkmark$ & & $\checkmark$& & $\checkmark$ &$\checkmark$ & & & & \\ \hline

HyperCLOVA & Korean blogs, Community sites, News, KiN Korean Wikipedia, Wikipedia (English and Japanese), Modu-Corpus: Messenger, News, Spoken and written language corpus, Web corpus& & & &  & & & $\checkmark$ & & & & & & \\ \hline

Yuan 1.0 & Common Crawl, SogouT, Sogou News,
Baidu Baike, Wikipedia, Books& & & & $\checkmark$ & $\checkmark$&$\checkmark$ & & & $\checkmark$& & & & \\ \hline

Gopher &subsets of MassiveWeb
Books, C4, News, GitHub and
Wikipedia samples from MassiveText & $\checkmark$&$\checkmark$ & $\checkmark$& $\checkmark$ & & & & & &$\checkmark$ & $\checkmark$& & $\checkmark$ \\ \hline

ERNIE-3.0 TITAN & Same as ERNIE 3.0 and ERNIE 3.0 adversarial dataset, ERNIE 3.0 controllable dataset& & & & $\checkmark$ & $\checkmark$&$\checkmark$ & &$\checkmark$ & $\checkmark$& & & & \\ \hline

GPT-NeoX-20B & Pile~\cite{gao2020pile}& & & $\checkmark$&$\checkmark$  & & $\checkmark$& &$\checkmark$ & & $\checkmark$& $\checkmark$& & \\ \hline

OPT &RoBERTa~\cite{roberta}, Pile~\cite{gao2020pile},
PushShift.io Reddit~\cite{baumgartner2020pushshift} & & & & $\checkmark$ & $\checkmark$& & & & & $\checkmark$& & & $\checkmark$\\ \hline

BLOOM & ROOTs~\cite{BLOOM} & & &$\checkmark$ & & &$\checkmark$ &$\checkmark$ & $\checkmark$& & & &$\checkmark$ &$\checkmark$ \\ \hline

Galactica & arXiv, PMC, Semantic Scholar, Wikipedia, StackExchange, LibreText, Open Textbooks, RefSeq Genome, OEIS, LIPID MAPS, NASAExoplanet, Common Crawl, ScientificCC, AcademicCC, GitHub repositories
Khan Problems, GSM8K, OneSmallStep& $\checkmark$& $\checkmark$& &$\checkmark$ & & & & & & & $\checkmark$& &$\checkmark$ \\ \hline

GLaM & Filtered Webpages, Social media conversations Wikipedia, Forums, Books, News& & & & $\checkmark$ &  & $\checkmark$ & & $\checkmark$ &$\checkmark$ & $\checkmark$ & & & \\ \hline

LaMDA &Infiniset : Public documents, Dialogs, Utterances & & & &  & & & & & & & & &$\checkmark$ \\ \hline

MT-NLG & Two snapshots of Common Crawl and Books3, OpenWebText2, Stack Exchange, PubMed Abstracts, Wikipedia, PG-19 [242], BookCorpus2, NIH ExPorter, Pile, CC-Stories, RealNews& & & &  & & $\checkmark$& & $\checkmark$&$\checkmark$ &$\checkmark$ & & & $\checkmark$\\ \hline

AlphaCode & Selected GitHub repositories, CodeContests: Codeforces, Description2Code, CodeNet& & & &  & & & & & & & & $\checkmark$& \\ \hline

Chinchilla & MassiveWeb, MassiveText Books, C4, News, GitHub, Wikipedia& $\checkmark$&$\checkmark$ & &$\checkmark$ & & & & & $\checkmark$& $\checkmark$& & &$\checkmark$ \\ \hline

PaLM &webpages, books, Wikipedia, news, articles, source code, social media conversations & $\checkmark$& & &  $\checkmark$& & & $\checkmark$& & & $\checkmark$& &$\checkmark$ & $\checkmark$\\ \hline

AlexaTM &Wikipedia, mC4 & & &$\checkmark$ &  & &$\checkmark$ & $\checkmark$& & & $\checkmark$& & &$\checkmark$ \\ \hline

%Sparrow & Human data for rule violations and per-turn response preferences,
%Self-play data accumulated through training, GopherCite, FilteredELI5& & & &  & & & & & & & & & \\ \hline

U-PaLM & Same as PaLM& $\checkmark$& &$\checkmark$ &$\checkmark$  & & $\checkmark$& & $\checkmark$& $\checkmark$& $\checkmark$& & & \\ \hline

UL2 & - & & &$\checkmark$ & $\checkmark$ & $\checkmark$&$\checkmark$ & & & & & $\checkmark$& &$\checkmark$ \\ \hline

GLM-130B & -  &$\checkmark$ &$\checkmark$ & &  & & & & $\checkmark$& & & & & \\ \hline

CodeGen &Pile, BigQuery, BigPython  & & & &  & & & & & & & &$\checkmark$ & \\ \hline

LLaMA & CommonCrawl, C4, Github, Wikipedia, Books, arXiv, StackExchange & & $\checkmark$& & $\checkmark$ & & & & & $\checkmark$& $\checkmark$&$\checkmark$ &$\checkmark$ & $\checkmark$\\ \hline

PanGu-$\Sigma$ & WuDaoCorpora, CLUE, Pile, C4, Python code & & & & $\checkmark$ & $\checkmark$&$\checkmark$ & $\checkmark$& $\checkmark$& & & & $\checkmark$& \\ \hline

BloombergGPT &inPile, Pile, C4, Wikipedia &$\checkmark$ &$\checkmark$ & &  & & $\checkmark$& &$\checkmark$ &$\checkmark$ &$\checkmark$ & & & $\checkmark$\\ \hline

%XuanYuan 2.0 &Internet & & & &  & & & & & & & & & \\ \hline

CodeT5+ & CodeSearchNet, Github Code & & & &  & & & & & & & $\checkmark$& $\checkmark$& \\ \hline

StarCoder &The Stack v1.2  & &$\checkmark$ & &  & & & & & & & $\checkmark$&$\checkmark$ &$\checkmark$ \\ \hline

LLaMA-2 & $\checkmark$&$\checkmark$ & &$\checkmark$ &  & & & & & $\checkmark$& $\checkmark$& $\checkmark$& $\checkmark$& \\ \hline

PaLM-2 & Web documents, Code, Books, Maths, Conversation& & & $\checkmark$& $\checkmark$ & $\checkmark$& $\checkmark$& $\checkmark$&$\checkmark$ & $\checkmark$&$\checkmark$ &$\checkmark$ &$\checkmark$ & $\checkmark$\\ \hline

\end{tabular}}
\vspace{2mm}
\label{datasets}
\end{table*}

\begin{table*}[!tbhp]
\centering
\caption{An illustration of training datasets and evaluation benchmarks used in fine-tuned LLMs. \enquote{SNI} is a short of Super-NaturalInsturctions.}
\label{tab:illustration_fine_tuned_datasets}

\resizebox{\textwidth}{!}{ 
\begin{tabular}{l|p{5cm}|c|c|c|c|c|c|c|c|c|c|c}
\hline \hline

\rowcolor{gray!50}\textbf{Models} & \begin{tabular}[c]{@{}c@{}}\textbf{Training Dataset} \end{tabular} & \begin{tabular}[c]{@{}c@{}}\textbf{BIG-}\\\textbf{bench}\end{tabular} & \textbf{MMLU} & \textbf{BBH} & \textbf{RAFT} & \textbf{FLAN} & \textbf{SNI} & \textbf{PromptSource} & \textbf{TyDiQA} & \textbf{HumanEval} & \textbf{MBPP} & \begin{tabular}[c]{@{}c@{}}\textbf{Truthful/}\\\textbf{Bias/}\\\textbf{Toxicity}\end{tabular} 
\\ \hline \hline

T0 & Pool of Prompts & $\checkmark$&  &&&&&&&&& \\ \hline
%\begin{tabular}[c]{@{}c@{}}NLI: ANLI~\cite{nie2019adversarial}, CB~\cite{de2019commitmentbank}, RTE~\cite{dagan2005pascal};\\ SC: COPA~\cite{roemmele2011choice}, HellaSwag~\cite{zellers2019hellaswag} StoryCloze~\cite{mostafazadeh2016corpus};\\ WSD: WiC~\cite{pilehvar2018wic}; CorefR: WSC~\cite{levesque2012winograd}, Wino (XL)~\cite{sakaguchi2021winogrande}\end{tabular} \\ \hline

WebGPT      & ELI5~\cite{fan2019eli5}, ELI5 fact-check~\cite{nakano2021webgpt}, TriviaQA~\cite{joshi2017triviaqa}, ARC-Challenge~\cite{clark2018think}, ARC-Easy~\cite{clark2018think}, Hand-written data, Demonstrations of humans, Comparisons between model-generated answers  &   &&&&&&&&&&$\checkmark$ \\ \hline

Tk-INSTRUCT & SNI~\cite{Tk-INSTRUCT} &  &&&&&$\checkmark$&&&&&    \\ \hline

mT0 & xP3~\cite{mT0andBLOOMZ} &  &&&&&&&&&& \\ \hline

OPT-IML & PromptSource~\cite{T0}, FLAN~\cite{Flan}, SNI~\cite{wang2022benchmarking}, UnifiedSKG~\cite{xie2022unifiedskg}, CrossFit~\cite{ye2021crossfit}, ExMix~\cite{aribandi2021ext5}, T5~\cite{T5}, Reasoning &  &$\checkmark$&$\checkmark$&$\checkmark$&$\checkmark$&$\checkmark$&$\checkmark$&&&&  \\ \hline

%PromptSource, FLAN, Super-NaturalInstructions,  MMLU~\cite{hendrycks2020measuring}, BBH~\cite{srivastava2022beyond}, RAFT~\cite{alex2021raft}

Flan & Muffin, T0-SF, NIv2, CoT &   &$\checkmark$&$\checkmark$&&&&&$\checkmark$&&& \\ \hline

%MMLU~\cite{hendrycks2020measuring}, BBH~\cite{srivastava2022beyond}, TyDiQA~\cite{clark2020tydi}, MGSM~\cite{shi2022language}  \\ \hline

WizardCoder & Code Alpaca &   &&&&&&&&$\checkmark$&$\checkmark$& \\ \hline

%HumanEval~\cite{chen2021evaluating}, MBPP~\cite{austin2021program}, DS-1000~\cite{ds_1000}\\ \hline

%InstructGPT & Text prompts submitted to theOPENAI API & RealToxicityPrompts~\cite{gehman2020realtoxicityprompts}   \\ \hline

\end{tabular}
}
\end{table*}

%\subsection{Evaluation Datasets}
%\label{ss:evaluation_datasets}
%The role of specific datasets, particularly those commonly used, is fundamental in the evaluation of Large Language Models. These datasets, each with its unique design and set of challenges, serve as the basis for assessing the capabilities of LLMs. They offer a comprehensive measure of performance across a variety of tasks, providing insights into the models' proficiency. In the following discussion, we provide a concise overview of a selection of these key datasets. While the Tables~\ref{tab:datasets} and~\ref{tab:datasets_inst_tuned} include a larger set of datasets, we focus on the most commonly used ones in the evaluation of LLMs. Each dataset description encapsulates the core aspects it evaluates in an LLM, offering a snapshot of the model's potential strengths and limitations.

\subsubsection{Multi-task}
\label{sss:evalc-multitask}

\noindent
~\emph{~\textbf{MMLU~\cite{hendrycks2020measuring}:}} 
\label{par:mmlu} 
A benchmark that measures the knowledge acquired by models during pretraining and evaluates models in zero-shot and few-shot settings across 57 subjects, testing both world knowledge and problem-solving ability.

\noindent
~\emph{~\textbf{SuperGLUE~\cite{wang2019superglue}:}} 
\label{par:superglue} 
A more challenging and diverse successor to the GLUE~\cite{wang2018glue} benchmark, SuperGLUE includes a variety of language understanding tasks, such as question answering, natural language inference, and co-reference resolution. It is designed to provide a rigorous test of language understanding and requires significant progress in areas like sample-efficient, transfer, multi-task, and unsupervised or self-supervised learning.

\noindent
~\emph{~\textbf{BIG-bench~\cite{srivastava2022beyond}:}} 
\label{par:big-bench} 
The BIG-bench (Behavior of Intelligent Generative Models Benchmark) is a large-scale benchmark designed to test the abilities of LLMs across a wide range of tasks, including reasoning, creativity, ethics, and understanding of specific domains.

\noindent
~\emph{~\textbf{GLUE~\cite{wang2018glue}:}} 
\label{par:glue} 
The General Language Understanding Evaluation (GLUE) benchmark is a collection of resources for training, evaluating, and analyzing natural language understanding systems. It includes a variety of tasks that test a wide range of linguistic phenomena, making it a comprehensive tool for evaluating language understanding in AI.

\subsubsection{Language Understanding}
\label{sss:evalc-langunderstanding}

\noindent
~\emph{~\textbf{WinoGrande~\cite{sakaguchi2021winogrande}:}} 
\label{par:winogrande} 
A large-scale dataset inspired by the original Winograd~\cite{levesque2012winograd} Schema Challenge tests models on their ability to resolve pronoun ambiguity and encourages the development of models that understand the broad context in natural language text.

\noindent
~\emph{~\textbf{CoQA~\cite{reddy2019coqa}:}} 
\label{par:coqa} 
A conversational question-answering dataset, CoQA challenges models with questions that rely on conversation history and require free-form text answers. Its diverse content from seven domains makes it a rigorous test for models' ability to handle a wide range of topics and conversational contexts.

\noindent
~\emph{~\textbf{WiC~\cite{pilehvar2018wic}:}} 
\label{par:wic} This dataset assesses a model's ability to discern word meanings based on context, aiding in tasks related to Word Sense Disambiguation.

\noindent
~\emph{~\textbf{Wikitext103~\cite{merity2016pointer}:}} 
\label{par:wikitext103} With over 100 million tokens from Wikipedia's top articles, this dataset is a rich resource for tasks that require understanding long-term dependencies, such as language modeling and translation.

\noindent
~\emph{~\textbf{PG19~\cite{rae2019compressive}:}} 
\label{par:pg19} This is a digital library of diverse books from Project Gutenberg. It is specifically designed to facilitate research in unsupervised learning and language modeling, with a special focus on long-form content.

\noindent
~\emph{~\textbf{C4~\cite{T5}:}} 
\label{par:c4} A clean, multilingual dataset, C4 offers billions of tokens from web-crawled data. It is a comprehensive resource for training advanced Transformer models on various languages.

\noindent
~\emph{~\textbf{LCQMC~\cite{liu2018lcqmc}:}} 
\label{par:lcqmc} 
The Large-scale Chinese Question Matching Corpus (LCQMC) is a dataset for evaluating the performance of models in semantic matching tasks. It contains pairs of questions in Chinese and their matching status, making it a valuable resource for research in Chinese language understanding.

\subsubsection{Story Cloze and Sentence Completion}
\label{sss:evalc-storycomprehension}

\noindent
~\emph{~\textbf{StoryCloze~\cite{mostafazadeh2016corpus}:}} 
\label{par:storycloze} 
It introduces a new \enquote{StoryCloze Test}, a commonsense reasoning framework for evaluating story understanding, generation, and script learning. It considers a model's ability to understand and generate coherent and sensible stories.

\noindent
~\emph{~\textbf{LAMBADA~\cite{paperno2016lambada}:}} 
\label{par:lambada} 
This dataset evaluates contextual text understanding through a word prediction task. Models must predict the last word of a passage, which is easy for humans when given the whole passage, but not when given only the last sentence.

\subsubsection{Physical Knowledge and World Understanding}
\label{sss:evalc-physicalknowledge}

\noindent
~\emph{~\textbf{PIQA~\cite{bisk2020piqa}:}} 
\label{par:piqa} 
A dataset that probes the physical knowledge of models, aiming to understand how well they are learning about the real world.

\noindent
~\emph{~\textbf{TriviaQA~\cite{joshi2017triviaqa}:}} 
\label{par:triviaqa} 
A dataset that tests models on reading comprehension and open domain question answering (QA) tasks, with a focus on Information Retrieval (IR)-style QA.

\noindent
~\emph{~\textbf{ARC~\cite{clark2018think}:}} 
\label{par:arc} 
A larger version of the ARC-Challenge, this dataset contains both easy and challenging grade-school level, multiple-choice science questions. It is a comprehensive test of a model's ability to understand and answer complex questions.

\noindent
~\emph{~\textbf{ARC-Easy~\cite{clark2018think}:}} 
\label{par:arc-easy} 
A subset of the ARC dataset, ARC-Easy, contains questions that are answered correctly by either a retrieval-based algorithm or a word co-occurrence algorithm. It is a great starting point for models beginning to explore advanced question-answering.

\noindent
~\emph{~\textbf{ARC-Challenge~\cite{clark2018think}:}} 
\label{par:arc-challenge} 
A rigorous question-answering dataset, ARC-Challenge includes complex, grade-school level questions that demand reasoning beyond simple retrieval, testing the true comprehension capabilities of models.

\subsubsection{Contextual Language Understanding}
\label{sss:evalc-contextlu}

\noindent
~\emph{~\textbf{RACE~\cite{lai2017race}:}} 
\label{par:race} 
The RACE dataset is a reading comprehension dataset collected from English examinations in China, which benchmarks AI models for understanding and answering questions on long and complex passages, simulating the challenge of a real-world examination.

\noindent
~\emph{~\textbf{RACE-Middle~\cite{lai2017race}:}} 
\label{par:race-middle}
Another subset of the RACE~\cite{lai2017race} dataset, RACE-Middle, contains middle school-level English exam questions. It offers a slightly less challenging but academically oriented evaluation of a model's comprehension skills.

\noindent
~\emph{~\textbf{RACE-High~\cite{lai2017race}:}} 
\label{par:race-high} 
A subset of the RACE~\cite{lai2017race} dataset, RACE-High consists of high school-level English exam questions. It is designed to evaluate the comprehension ability of models in a more academic and challenging context.

\noindent
~\emph{~\textbf{QuAC~\cite{choi2018quac}:}} 
\label{par:quac} 
This dataset simulates an information-seeking dialog between students and teachers using hidden Wikipedia text. It introduces unique challenges not found in machine comprehension datasets, making it a valuable resource for advancing dialog systems.

\subsubsection{Commonsense Reasoning}
\label{sss:evalc-commonsense}

\noindent
~\emph{~\textbf{HellaSwag~\cite{zellers2019hellaswag}:}} 
\label{par:hellaswag} 
A dataset that challenges models to pick the best ending to a context uses Adversarial Filtering to create a `Goldilocks' zone of complexity, where generated text is absurd to humans but often misclassified by models.

\noindent
~\emph{~\textbf{COPA~\cite{ponti2020xcopa}:}} 
\label{par:copa} 
This dataset evaluates a model's progress in open-domain commonsense causal reasoning. Each question comprises a premise and two alternatives, and the model must select the more plausible alternative, testing a model's ability to understand and reason about cause and effect.

\noindent
~\emph{~\textbf{WSC~\cite{levesque2012winograd}:}} 
\label{par:wsc} 
The Winograd Schema Challenge (WSC) is a reading comprehension task in which a system must resolve references in a text, often requiring world knowledge and reasoning about the text.

\noindent
~\emph{~\textbf{CSQA~\cite{talmor2018commonsenseqa}:}} 
\label{par:csqa} 
The CommonsenseQA is a question-answering dataset that requires commonsense knowledge to evaluate the ability of AI models to understand and answer questions.

\subsubsection{Reading Comprehension}
\label{sss:readingcomprehension}

\noindent
~\emph{~\textbf{BoolQ~\cite{clark2019boolq}:}} 
\label{par:boolq} 
A dataset derived from Google search queries, BoolQ challenges models to answer binary (yes/no) questions. The questions are naturally occurring and are paired with a paragraph from a Wikipedia article containing the answer. It is a test of reading comprehension and reasoning.

\noindent
~\emph{~\textbf{SQUADv2~\cite{rajpurkar2018know}:}} 
\label{par:squadv2} 
The Stanford Question Answering Dataset (SQuAD)~\cite{rajpurkar2016squad} is a collection of questions posed by crowd workers on a set of Wikipedia articles, where the answer to every question is a segment of text from the corresponding reading passage. SQuADv2 combines the original SQuAD1.1 dataset with over 50,000 unanswerable questions. The aim is to evaluate a model's ability to understand and answer questions based on a given context and to determine when a question is unanswerable.

\noindent
~\emph{~\textbf{DROP~\cite{dua2019drop}:}} 
\label{par:drop} 
DROP, or Discrete Reasoning Over the content of Paragraphs, is designed to test a model's ability to understand a wide variety of reading phenomena. It encourages comprehensive and reliable evaluation of reading comprehension capabilities.

\noindent
~\emph{~\textbf{RTE~\cite{dagan2005pascal}:}} 
\label{par:rte} 
The Recognizing Textual Entailment (RTE) datasets come from a series of annual competitions on textual entailment, predicting whether a given sentence logically follows from another and evaluating a model's understanding of logical relationships in a text.

\noindent
~\emph{~\textbf{WebQA~\cite{chang2022webqa}:}} 
\label{par:webqa} A dataset for open-domain question answering, WebQA offers a large collection of web-based question-answer pairs. It is designed to assess the ability of AI models to understand and answer questions based on web content.

\noindent
~\emph{~\textbf{CMRC2018~\cite{cui2018span}:}} 
\label{par:cmrc2018} 
This dataset is a test of Chinese language models' ability to reason comprehensively and is designed with a challenging span-extraction format that pushes the boundaries of machine performance.

\subsubsection{Mathematical Reasoning}
\label{sss:evalc-mathreason}

\noindent
~\emph{~\textbf{MATH~\cite{hendrycks2021measuring}:}} 
\label{par:math}
This dataset is a platform for evaluating the mathematical problem-solving abilities of AI models. It contains a diverse set of math problems, ranging from arithmetic to calculus, and is designed to test the model's ability to understand and solve complex mathematical problems.

\noindent
~\emph{~\textbf{Math23k~\cite{wang2017deep}:}} 
\label{par:math23k} This one challenges a model's ability to understand and solve mathematical word problems. It contains 23,000 Chinese arithmetic word problems that require models to perform reasoning and computation based on the problem description.

\noindent
~\emph{~\textbf{GSM8K~\cite{cobbe2021training}:}} 
\label{par:gsm8k} 
A dataset of diverse grade school math word problems, testing a model's ability to perform multi-step mathematical reasoning.

\subsubsection{Problem Solving and Logical Reasoning}
\label{sss:evalc-probsolve}

\noindent
~\emph{~\textbf{ANLI~\cite{nie2019adversarial}:}} 
\label{par:anli} 
A large-scale dataset designed to test the robustness of machine learning models in Natural Language Inference (NLI) is created through an iterative, adversarial process where humans try to generate examples that models cannot correctly classify.

\noindent
~\emph{~\textbf{HumanEval~\cite{codex}:}} 
\label{par:humaneval} 
A dataset for evaluating the problem-solving ability of AI models, which includes a diverse set of tasks that require various cognitive abilities, making it a comprehensive tool for assessing general intelligence in AI.

\noindent
~\emph{~\textbf{StrategyQA~\cite{geva2021did}:}} 
\label{par:strategyqa} 
A question-answering dataset that requires reasoning over multiple pieces of evidence to evaluate the strategic reasoning ability of AI models, pushing the boundaries of what machines can understand and answer.

\subsubsection{Cross-Lingual Understanding}
\label{sss:evalc-crosslinglu}

\noindent
~\emph{~\textbf{XNLI~\cite{conneau2018xnli}:}} 
\label{par:xnli} 
A cross-lingual benchmark, XNLI extends the MultiNLI~\cite{williams2018multinli} corpus to 15 languages, including low-resource ones like Urdu. It tests models on cross-lingual sentence understanding, with 112,500 annotated pairs across three categories: entailment, contradiction, and neutral.

\noindent
~\emph{~\textbf{PAWS-X~\cite{yang2019paws}:}} 
\label{par:paws-x} 
PAWS-X, or Cross-lingual Paraphrase Adversaries from Word Scrambling, is a multilingual version of the PAWS~\cite{zhang2019pawss} dataset for paraphrase identification. It includes examples in seven languages and is designed to evaluate the performance of cross-lingual paraphrase identification models.

\subsubsection{Truthfulness}
\label{sss:evalc-truth}

\noindent
~\emph{~\textbf{Truthful-QA~\cite{lin2021truthfulqa}:}} 
\label{par:truthful-qa} 
A unique benchmark that measures a language model's truthfulness when generating answers. The dataset includes questions across various categories like health, law, and politics, some designed to test the model against common human misconceptions.

\subsubsection{Biases and Ethics in AI}
\label{sss:evalc-biases}

\noindent
~\emph{~\textbf{ETHOS~\cite{mollas2020ethos}:}} 
\label{par:ethos} 
ETHOS is a hate speech detection dataset built from YouTube and Reddit comments. It is a tool in the fight against online hate speech, offering binary and multi-label variants for robust content moderation.

\noindent
~\emph{~\textbf{StereoSet~\cite{nadeem2020stereoset}:}} 
\label{par:stereoset} 
StereoSet is a comprehensive dataset designed to measure and evaluate the presence of stereotypical biases in language models. It focuses on four key domains: gender, profession, race, and religion. Contrasting stereotypical bias against language modeling ability provides a valuable tool for understanding and mitigating biases in large language models.

\begin{table*}[t]
    \centering
    \caption{Performance comparison of top performing LLMs across various NLU and NLG tasks. Here, \enquote{N-Shots} indicate the number of example prompts provided to the model during the evaluation, representing its capability in few-shot or zero-shot learning settings, \enquote{f} represents the fine-tuned version, and \enquote{B} represents the benchmark.}
    \label{tab:performance_comparison}
    \resizebox{\textwidth}{!}{% 

    \begin{tabular}{l|l|cc|cc|cc}
    \hline
    \multirow{2}{*}{\textbf{Task}} & \multirow{2}{*}{\textbf{Dataset/Benchmark}} & \multicolumn{2}{c|}{\textbf{Top-1}} & \multicolumn{2}{c|}{\textbf{Top-2}} & \multicolumn{2}{c}{\textbf{Top-3}} \\ \cline{3-8} 
      &  & \multicolumn{1}{c|}{\textbf{Model (Size)}} & \textbf{Score (N-shots)} & \multicolumn{1}{c|}{\textbf{Model (Size)}} & \textbf{Score (N-shots)} & \multicolumn{1}{c|}{\textbf{Model (Size)}} & \textbf{Score (N-shots)} \\ \hline
    \multirow{2}{*}{Multi-Task} & BIG-bench (B) & \multicolumn{1}{c|}{Chinchilla (70B)} & 65.1 (5-shot) & \multicolumn{1}{c|}{Gopher (280B)} & 53.97 (5-shot) & \multicolumn{1}{c|}{PaLM (540B)} & 53.7 (5-shot) \\ \cline{2-8} 
     & MMLU (B) & \multicolumn{1}{c|}{GPT-4 (-)} & 86.4 (5-shot) & \multicolumn{1}{c|}{Gemini (Ultra)} & 83.7 (5-shot) & \multicolumn{1}{c|}{Flan-PaLM-2$_{(f)}$ (Large)} & 81.2 (5-shot) \\ \hline
    Language Understanding & SuperGLUE (B) & \multicolumn{1}{c|}{ERNIE 3.0 (12B)} & 90.6 (-) & \multicolumn{1}{c|}{PaLM$_{(f)}$ (540B)} & 90.4 (-) & \multicolumn{1}{c|}{T5 (11B)} & 88.9 (-) \\ \hline
    \multirow{2}{*}{\begin{tabular}[c]{@{}l@{}}Story Comprehension and \\Generation\end{tabular}} & HellaSwag & \multicolumn{1}{c|}{GPT-4 (-)} & 95.3 (10-shot) & \multicolumn{1}{c|}{Gemini (Ultra)} & 87.8 (10-shot) & \multicolumn{1}{c|}{PaLM-2 (Large)} & 86.8 (one shot) \\ \cline{2-8} 
     & StoryCloze & \multicolumn{1}{c|}{GPT3 (175B)} & 87.7 (few shot) & \multicolumn{1}{c|}{PaLM-2 (Large)} & 87.4 (one shot) & \multicolumn{1}{c|}{OPT (175B)} & 79.82 (-) \\ \hline
    \multirow{2}{*}{\begin{tabular}[c]{@{}l@{}}Physical Knowledge and\\ World Understanding\end{tabular}} & PIQA & \multicolumn{1}{c|}{PaLM-2 (Large)} & 85.0 (one shot) & \multicolumn{1}{c|}{LLaMa (65B)} & 82.8 (zero shot) & \multicolumn{1}{c|}{MT-NLG (530B)} & 81.99 (zero shot) \\ \cline{2-8} 
     & TriviaQA & \multicolumn{1}{c|}{PaLM-2 (Large)} & 86.1 (one shot) & \multicolumn{1}{c|}{LLaMA-2 (70B)} & 85.0 (one shot) & \multicolumn{1}{c|}{PaLM (540B)} & 81.4 (one shot) \\ \hline
    \begin{tabular}[c]{@{}l@{}}Contextual Language \\Understanding\end{tabular} & LAMBADA & \multicolumn{1}{c|}{PaLM (540B)} & 89.7 (few shot) & \multicolumn{1}{c|}{MT-NLG (530B)} & 87.15 (few shot) & \multicolumn{1}{c|}{PaLM-2 (Large)} & 86.9 (one shot) \\ \hline
    \multirow{2}{*}{Commonsense Reasoning} & WinoGrande & \multicolumn{1}{c|}{GPT-4 (-)} & 87.5 (5-shot) & \multicolumn{1}{c|}{PaLM-2 (Large)} & 83.0 (one shot) & \multicolumn{1}{c|}{PaLM (540B)} & 81.1 (zero shot) \\ \cline{2-8} 
     & SIQA & \multicolumn{1}{c|}{LLaMA (65B)} & 52.3 (zero shot) & \multicolumn{1}{c|}{Chinchilla (70B)} & 51.3 (zero shot) & \multicolumn{1}{c|}{Gopher (280B)} & 50.6 (zero shot) \\ \hline
    Reading Comprehension & BoolQ & \multicolumn{1}{c|}{PaLM$_{(f)}$ (540B)} & 92.2 (-) & \multicolumn{1}{c|}{T5 (11B)} & 91.2 (-) & \multicolumn{1}{c|}{PaLM-2 (Large)} & 90.9 (one shot) \\ \hline
    Truthfulness & Truthful-QA & \multicolumn{1}{c|}{LLaMA (65B)} & 57 (-) & \multicolumn{1}{c|}{} &  & \multicolumn{1}{c|}{} &  \\ \hline
    \multirow{2}{*}{Mathematical Reasoning} & MATH & \multicolumn{1}{c|}{Gemini (Ultra)} & 53.2 (4-shot) & \multicolumn{1}{c|}{PaLM-2 (Large)} & 34.3 (4-shot) & \multicolumn{1}{c|}{LLaMa-2 (65B)} & 13.5 (4-shot) \\ \cline{2-8} 
     & GSM8K & \multicolumn{1}{c|}{GPT-4 (-)} & 92.0 (5-shot) & \multicolumn{1}{c|}{PaLM-2 (Large)} & 80.7 (8-shot) & \multicolumn{1}{c|}{U-PaLM (540B)} & 58.5 (-) \\ \hline
    \begin{tabular}[c]{@{}l@{}}Problem Solving and \\Logical Reasoning\end{tabular} & HumanEval & \multicolumn{1}{c|}{Gemini$_{(f)}$ (Ultra)} & 74.4 (zero shot) & \multicolumn{1}{c|}{GPT-4 (-)} & 67.0 (zero shot) & \multicolumn{1}{c|}{Code Llama (34B)} & 48.8 (zero shot) \\ \hline
    \end{tabular}}
\end{table*}

\section{Applications}

\label{sec:Applications}
Applying Large Language Models (LLMs) to a variety of downstream tasks has become a popular trend in both AI-related research communities and industries, with many emerging uses being discovered and explored daily. LLMs, which are capable of understanding and generating human-like text, have found meaningful applications across a variety of fields. This section provides an overview of LLM applications in medicine, education, science, mathematics, law, finance, robotics, and coding. While each of these domains pose different challenges, LLMs open up opportunities to make significant contributions to these domains through their generalizability. \\
\noindent
\emph{\textbf{{General Purpose:}}} 
LLMs are being widely considered as general-purpose tools for a wide variety of tasks~\cite{qin2023is}. This is due to their inherent ability to understand, generate, and manipulate human-like text in a contextually relevant manner. This allows them to perform tasks ranging from simple language translation and question-answering to more complex tasks like summarization, text generation, and even programming help~\cite{hadi2023large}. The utility of LLMs is further enhanced by their ability to adapt to the specific style and tone of the text they are processing, making the outputs more user-friendly and context-aware. In everyday applications, LLMs can be used as personal assistants, helping users draft emails or schedule appointments~\cite{dong2023towards}; they can also be deployed in customer service to handle common questions or applied to generate content for digital platforms like websites by creating human-like text based on given prompts~\cite{pandya2023automating}. Moreover, LLMs play a crucial role in data analysis, where they can filter large volumes of text data, summarize key points, and find patterns that would take humans much longer to identify~\cite{li2023can}. Despite their wide-ranging applications, it is essential to remember that LLMs, similar to any AI system, are only as good as the data they have been trained on. \\
\noindent
\emph{\textbf{{Medicine:}}}
The application of LLMs in the field of medicine is reshaping healthcare delivery and research. For example, LLMs are increasingly used in clinical decision support systems to provide physicians with evidence-based treatment recommendations~\cite{rao2023evaluating, benary2023leveraging, chiesa2023exploring}. By analyzing patient data and medical literature, they can help identify potential diagnoses, suggest appropriate tests, and recommend optimal treatment strategies. Moreover, LLMs can also enhance patient interactions with healthcare systems; e.g., they can be used in chatbot applications~\cite{montagna2023data, bill2023fine, abbasian2023conversational} to answer patient queries about symptoms or medications, schedule appointments, and even provide essential health advice. For medical research, LLMs are used to extract and filter information from a considerable amount of medical literature, identify relevant studies, summarize findings, and even predict future research trends~\cite{lemley2023does, pal2023domain, du2023calla}. For medical education, LLMs can help create training materials, generate exam questions, provide detailed explanations of complex medical topics, and offer personalized feedback to students~\cite{abd2023large, mbakwe2023chatgpt, ahn2023impending, waisberg2023large}. They can also simulate patient interactions, enabling students to practice and improve their clinical skills. At a broader level, LLMs can assist in public health initiatives by analyzing media data to detect disease outbreaks, monitor public sentiment towards health policies, and disseminate health information in a clear and understandable manner~\cite{deiana2023artificial}. LLMs can be employed to support public health initiatives, addressing related issues such as data privacy, the necessity for explainability, and the potential risk of propagating biases~\cite{de2023chatgpt, rane2023contribution}. \\
\noindent
\emph{\textbf{{Education:}}}
The integration of LLMs into the educational sector offers opportunities to enhance learning experiences, teacher support, and educational content development. For students, by analyzing their learning styles, performance, and preferences, LLMs can provide customized study materials and practice questions to develop personalized learning experiences~\cite{dai2023can}. For teachers, LLMs can help to create lesson plans and grade assignments and generate diverse and inclusive educational content, significantly saving more time for teaching and student interaction~\cite{kasneci2023chatgpt, rane2023enhancing}. In language learning, LLMs serve as advanced conversational partners capable of simulating conversations in multiple languages, correcting grammar, enhancing vocabulary, and aiding pronunciation for the needs of fluency in practice~\cite{young2023investigating}. Furthermore, LLMs improve accessibility in education by providing support for students with disabilities. They can generate real-time transcriptions for the hearing impaired, offer reading assistance for the visually impaired, and simplify complex texts for those with learning disabilities~\cite{rane2023contribution}. As LLMs continue to evolve, their applications in education can benefit more students and teachers from different perspectives in practice. \\
\noindent
\emph{\textbf{{Science:}}}
Similar to medical applications, LLMs can expedite the research process by quickly analyzing and summarizing scientific literature. By briefing comprehensible and accessible research summaries, LLMs can assist researchers in staying up-to-date with the latest findings, even in fields outside their area of expertise~\cite{irons2023exploring, schmidt2023using}. In addition, LLMs can aid scientists in formulating new hypotheses and research questions since their ability to process large-scale datasets allows them to unveil insights that might not be immediately apparent to human researchers~\cite{zheng2023large}. Moreover, for scientific writing, LLMs can help researchers draft documents, suggest improvements, and ensure adherence to specific formatting guidelines~\cite{aczel2023transparency, altmae2023artificial}. This not only saves time but also improves the clarity of scientific communication, enabling interdisciplinary teams to work together more effectively. \\
\noindent
\emph{\textbf{{Maths:}}} 
In addition to providing mathematical research and education support, LLMs can assist in solving mathematical problems by giving step-by-step explanations and guiding users through complex proofs and calculations. They can help identify errors in reasoning or computation and suggest corrections, serving as an invaluable tool for both learning and verification purposes~\cite{imani2023mathprompter, yuan2023scaling}. LLMs can be employed to check the validity of mathematical proofs, offering a preliminary filter before human review. While they are not a substitute for the meticulous work of mathematicians, they can help simplify the process of proof verification~\cite{yang2023leandojo, collins2023evaluating}. Moreover, LLMs enhance accessibility to mathematics by translating complex concepts and findings into understandable language for non-specialists~\cite{liu2023summary}, where the gap between theoretical mathematics and applied contexts such as physics, engineering, and economics can be bridged. \\
\noindent
\emph{\textbf{{Law:}}}
LLMs can assist with the thematic analysis of legal documents, including generating initial coding for datasets, identifying themes, and classifying data according to these themes. This collaborative effort between legal experts and LLMs has proved to be effective in analyzing legal texts such as court opinions on theft, improving both the efficiency and quality of the research~\cite{drapal2023using}. Additionally, LLMs have been evaluated for their ability to generate explanations of legal terms, focusing on improving factual accuracy and relevance by incorporating sentences from case law. By feeding relevant case law into the LLM, the augmented models can generate higher-quality explanations with less factually incorrect information~\cite{savelka2023explaining}. Moreover, LLMs can be trained with specialized domain knowledge to perform legal reasoning tasks~\cite{guha2023legalbench} and answer legal questions~\cite{cui2023chatlaw}. \\
\noindent
\emph{\textbf{{Finance:}}}
LLMs like BloombergGPT~\cite{bloomberggpt}, trained on extensive proprietary financial datasets, exhibit superior performance on financial tasks. This indicates the value of domain-specific training in creating LLMs that can more accurately understand and process industry-specific language and concepts. The introduction of FinGPT~\cite{yang2023fingpt} as an open-source model offers transparent and accessible resources to develop novel applications such as robo-advising, algorithmic trading, and low-code solutions, ultimately expanding the capabilities of financial services. Both BloombergGPT and FinGPT show the adaptability of LLMs to the financial domain, with the former showing the power of custom datasets and the latter emphasizing a data-centric approach and low-rank adaptation techniques for customization. Moreover, LLMs demonstrate an ability to break down complex financial tasks into actionable plans, enabling end-to-end solutions that were previously unfeasible with a single model~\cite{li2023large}. \\
\noindent
\emph{\textbf{{Robotics:}}}
In robotics research, LLMs have promising applications, such as enhancing human-robot interaction~\cite{zhang2023large, lykov2023llm,billing2023language, ye2023improved}, task planning~\cite{progprompt}, motion planning~\cite{ding2023task}, navigation~\cite{ding2023task, ding2023leveraging}, object manipulation~\cite{alphablock}, personalized robots~\cite{wu2023tidybot}, etc. LLMs enable robots to understand the environment effectively and generate plans to complete tasks collaboratively~\cite{saycan,PaLME}. They can facilitate continuous learning by allowing robots to access and integrate information from a wide range of sources, helping robots acquire new skills, adapt to changes, and refine their paths~\cite{hugginggpt,RAP,retroformer}.

\section{Challenges and Future Directions}
\label{sec:Challenges}
LLMs such as GPT-4 and its predecessors have significantly advanced natural language processing. Nevertheless, they also bring along a set of challenges. The computational cost, adversarial robustness, and interpretability are among the technical challenges that are intrinsic to these models. Furthermore, as these models are scaled up to handle more complex tasks or to operate in more complex or dynamic environments, new challenges in scalability, privacy, and real-time processing emerge. On the frontier of foundational research, integrating multi-modality and the effectiveness of transfer learning are being keenly explored. Additionally, the continuous learning aspect of these models, which aims to have models that can adapt to new information over time, presents a fresh set of challenges. These challenges not only underscore the technical intricacies involved but also highlight the broader impact and the future trajectory of LLMs in real-world applications. The following sections delve into these challenges, shedding light on the ongoing and potential efforts to address them. \\
\emph{\textbf{Computational Cost:}}
Training LLMs require extensive computational resources, which increases production costs and raises environmental concerns due to substantial energy consumption during large-scale training. Improved performance occurs as computational resources increase, but the rate of improvement gradually decreases when both the model and dataset size remain fixed, following the power law of diminishing returns~\cite{strubell2019energy}. \\
\emph{\textbf{Bias and Fairness:}}
LLMs can inherit and amplify societal biases in their training data. These biases can manifest in the model's outputs, leading to potential ethical and fairness issues~\cite{bender2021dangers}. \\
\emph{\textbf{Overfitting:}}
Although LLMs possess substantial learning capabilities, they are susceptible to overfitting noisy and peculiar patterns within their extensive training data. Consequently, this may cause them to generate illogical responses~\cite{zhang2021understanding}. The debate about Memorization vs.~Generalization in LLMs is about finding the right balance. Memorization allows the model to remember specific details from its training data, ensuring it can provide accurate answers to precise questions. However, generalization enables the model to make inferences and produce responses for inputs it has not seen before, which is essential for handling various real-world tasks. Striking the right balance is the challenge: too much memorization can lead to overfitting, making the model inflexible and struggling with new inputs~\cite{tanzer2021memorisation}. \\
\emph{\textbf{Economic and Research Inequality:}}
The high cost of training and deploying LLMs may make their development concentrated within well-funded organizations, potentially worsening economic and research inequalities in AI~\cite{west2019discriminating}. \\
\emph{\textbf{Reasoning and Planning:}}
Some reasoning and planning tasks, even as seemingly simple as common-sense planning, which humans find easy, remain well beyond the current capabilities of LLMs evaluated using an assessment framework. This is not entirely unexpected, considering that LLMs primarily generate text completions based on likelihood and offer no solid guarantees in terms of reasoning abilities~\cite{valmeekam2022large}. \\
\emph{\textbf{Hallucinations:}}
LLMs exhibit ``hallucinations", where they generate responses that, while sounding plausible, are incorrect or do not align with the provided information~\cite{zhang2023siren}. Hallucinations can be categorized into three categories.
\begin{itemize}
  \item Input-conflicting hallucination, wherein LLMs produce content that diverges from the input given by users.
  \item Context-conflicting hallucination, where LLMs generate content that contradicts information they have generated earlier.
  \item Fact-conflicting hallucination involves LLM's generation of content that does not align with established world knowledge.
\end{itemize} 

\noindent
\emph{\textbf{Prompt Engineering:}}
Prompts serve as inputs to LLMs, and their syntax and semantics play a crucial role in determining the model's output. The prompt variations, sometimes counter-intuitive to humans, can result in significant changes in model output and are addressed through prompt engineering, which involves designing natural language queries to guide LLMs responses effectively~\cite{webson2021prompt,prompt_eng_guide}. \\
%\emph{\textbf{Tokenization Limitations:}}
%Tokenizers bring about various difficulties, such as increased computational processing requirements, reliance on specific languages, management of new or uncommon words, limitations in vocabulary size, loss of information, and reduced human comprehensibility~\cite{kaddour2023challenges,chung2023unimax}. \\
%\emph{\textbf{Lack of Reproducibility:}}
%Parallelism distributes the workload across many computing nodes; therefore, the scheduling and communication are often nondeterministic, leading to reproducibility problems~\cite{recht2011hogwild}. \\
%\emph{\textbf{Lack of Evaluation Protocol:}}
%Absence of evaluation   \\
\emph{\textbf{Limited Knowledge:}}
Information acquired during pretraining is limited and may become obsolete after some time. Re-training the model using updated data is costly. To generate factually accurate responses, people use a retrieval augmentation pipeline~\cite{in_context_ralm}. However, pre-trained models are not trained with retrieval augmentation generation (RAG)~\cite{GPT-3,llama_2}; hence, adapting the training pipeline is necessary~\cite{retro,atlas}. \\
%\emph{\textbf{High Inference Latency:}}
%LLMs experience significant delays during inference for two primary reasons: (1) they can not be processed in parallel because the inference process happens one token at a time, and (2) they require a lot of memory due to their model size and the temporary data used during decoding, such as attention key and value tensors~\cite{pope2023efficiently,weng2023inference}. \\
\emph{\textbf{Safety and Controllability:}}
Using LLMs comes with the risk of generating harmful, misleading, or inappropriate content, whether by accident or when given specific prompts. Ensuring these models are safely utilized is a significant concern~\cite{shaikh2022second}. \\
\emph{\textbf{Security and Privacy:}}
LLMs are prone to leaking personal information and generating false, unethical, misaligned responses. Researchers have explored various security attacks, i.e., backdoor attacks, jailbreaking, prompt injection, and data poisoning, that lead to breaking LLMs security. Therefore, developing better defense mechanisms is essential to ensure LLMs are safe, reliable, and trustworthy for complex AI applications~\cite{security_and_privacy}.      \\
\emph{\textbf{Multi-Modality:}}
Multi-modal learning, where LLMs are trained on diverse data like text, images, and videos, aims to create models with richer understanding but faces challenges in data alignment, fusion strategies, and higher computational demands. \\
\emph{\textbf{Catastrophic Forgetting:}}
LLMs are often pre-trained on large datasets and then fine-tuned on domain-specific data, reducing training resources. However, they face issues like domain adaptation and catastrophic forgetting, which hinder the retention of original knowledge when learning new tasks. \\
\emph{\textbf{Adversarial Robustness:}}
Large Language Models (LLMs) have shown great capabilities in various tasks but are vulnerable to adversarial attacks, where slight, deliberate input alterations can mislead them. Especially with models like BERT, adversarial fine-tuning can enhance robustness, although it sometimes compromises generalization~\cite{liu2020adversarial}. As LLMs integrate more into complex systems, examining their security properties becomes crucial, given the emerging field of adversarial attacks on LLMs within trustworthy ML~\cite{shayegani2023survey}. This vulnerability is notable in safety-critical domains, necessitating robust adversarial evaluation tools to ensure LLM reliability~\cite{xu2023llm}. \\
\emph{\textbf{Interpretability and Explainability:}}
The \enquote{black-box} nature of LLMs poses challenges in understanding their decision-making, which is crucial for broader acceptance and trust, especially in sensitive domains. Despite their advanced capabilities, the lack of insight into their operation limits their effectiveness and trustworthiness~\cite{zhao2023explainability, huang2023large}. Efforts are being made to make LLMs more explainable to promote user trust and to ensure responsible AI usage. Understanding the logic behind LLMs' responses is essential for fostering trust and ensuring they align with human values and legal standards. \\
\emph{\textbf{Privacy Concerns:}}
Privacy concerns in Large Language Models (LLMs) have escalated with their growth in complexity and size, particularly around data sharing and potential misuse. There is a risk of malicious content creation, filter bypass, and data privacy issues, especially in e-commerce, where protecting customer privacy is crucial. If models are trained on private data, additional concerns arise if such models are made publicly available. LLMs tend to memorize phrases from their training sets, which an adversary could exploit to extract sensitive data, posing a threat to personal privacy~\cite{brown2022does, plant2022you}. \\
\emph{\textbf{Real-Time Processing:}}
Real-time processing in Large Language Models (LLMs) is pivotal for various applications, especially with the rising popularity of mobile AI applications and concerns regarding information security and privacy. However, LLMs often have hundreds of layers and millions of parameters, which impede real-time processing due to the high computational demands and limited weight storage on hardware platforms, particularly in edge computing environments~\cite{niu2020realtime}. While certain efforts like MobileBERT aim to reduce memory requirements, they still face substantial execution overhead due to the large number of model layers, leading to high inference latency. \\
\emph{\textbf{Long-Term Dependencies:}}
Large Language Models have shown considerable progress in understanding and generating text, yet they often struggle with preserving context and handling long-term dependencies, particularly in complex, multi-turn conversations or long documents. This limitation can lead to incoherent or irrelevant responses. \\
\emph{\textbf{Hardware Acceleration:}}
The growth of LLMs presents significant hardware challenges due to the increasing computational and memory demands associated with training and deploying these models. GPUs have played a crucial role in meeting the hardware requirements for training LLMs, with the networking industry also evolving to optimize hardware for training workloads. However, the growing size of LLMs, which has been outpacing hardware progress, makes model inference increasingly costly. Model quantization is a promising approach to bridge the widening gap between LLM size and hardware capacity~\cite{guo2023olive}. Although specialized hardware acceleration like GPUs or TPUs can significantly reduce the computational cost, making real-time applications more feasible, they may not fully resolve all limitations, necessitating further advancements in hardware technology. \\
\emph{\textbf{Regulatory and Ethical Frameworks:}}
The rapid advancements in artificial intelligence have given rise to sophisticated Large Language Models (LLMs) like OpenAI's GPT-4~\cite{GPT-4} and Google's Bard. These developments underscore the imperative for regulatory oversight to manage the ethical and social challenges accompanying LLMs' widespread use~\cite{mesko2023imperative}. For instance, LLMs can generate content that can be used positively or negatively, emphasizing the need for proactive ethical frameworks and policy measures to guide their responsible use and assign accountability for their outputs~\cite{zhang2023ethical}. Auditing is identified as a promising governance mechanism to ensure that AI systems, including LLMs, are designed and deployed ethically, legally, and technically robust~\cite{mokander2023auditing}. \\

\section{Conclusion}
\label{sec:conclusion}
This article has comprehensively reviewed the developments in LLMs. It contributes to summarizing significant findings of LLMs in the existing literature and provides a detailed analysis of the design aspects, including architectures, datasets, and training pipelines. We identified crucial architectural components and training strategies employed by different LLMs. These aspects are presented as summaries and discussions throughout the article. Moreover, we have discussed the performance differences of LLMs in zero-shot and few-shot settings, explored the impact of fine-tuning, and compared supervised and generalized models and encoder vs. decoder vs. encoder-decoder architectures. A comprehensive review of multi-modal LLMs, retrieval augmented LLMs, LLMs-powered agents, efficient LLMs, datasets, evaluation, applications, and challenges is also provided. This article is anticipated to serve as a valuable resource for researchers, offering insights into the recent advancements in LLMs and providing fundamental concepts and details to develop better LLMs. \\

\noindent
\textbf{Acknowledgement:}
The author/s would like to acknowledge the support received from Saudi Data and AI Authority (SDAIA) and King Fahd University of Petroleum and Minerals (KFUPM) under SDAIA-KFUPM Joint Research Center for Artificial Intelligence Grant No. JRC-AI-RFP-11.

\bibliographystyle{elsarticle-num}
\bibliography{refs}

\begin{thebibliography}{100}
\expandafter\ifx\csname url\endcsname\relax
  \def\url#1{\texttt{#1}}\fi
\expandafter\ifx\csname urlprefix\endcsname\relax\def\urlprefix{URL }\fi
\expandafter\ifx\csname href\endcsname\relax
  \def\href#1#2{#2} \def\path#1{#1}\fi

\bibitem{chernyavskiy2021transformers}
A.~Chernyavskiy, D.~Ilvovsky, P.~Nakov, Transformers:“the end of history” for natural language processing?, in: Machine Learning and Knowledge Discovery in Databases. Research Track: European Conference, ECML PKDD 2021, Bilbao, Spain, September 13--17, 2021, Proceedings, Part III 21, Springer, 2021, pp. 677--693.

\bibitem{wang2019superglue}
A.~Wang, Y.~Pruksachatkun, N.~Nangia, A.~Singh, J.~Michael, F.~Hill, O.~Levy, S.~Bowman, Superglue: A stickier benchmark for general-purpose language understanding systems, Advances in neural information processing systems 32 (2019).

\bibitem{adiwardana2020towards}
D.~Adiwardana, M.-T. Luong, D.~R. So, J.~Hall, N.~Fiedel, R.~Thoppilan, Z.~Yang, A.~Kulshreshtha, G.~Nemade, Y.~Lu, et~al., Towards a human-like open-domain chatbot, arXiv preprint arXiv:2001.09977 (2020).

\bibitem{y2022large}
B.~A. y~Arcas, Do large language models understand us?, Daedalus 151~(2) (2022) 183--197.

\bibitem{GPT-2}
A.~Radford, J.~Wu, R.~Child, D.~Luan, D.~Amodei, I.~Sutskever, et~al., Language models are unsupervised multitask learners, OpenAI blog 1~(8) (2019) 9.

\bibitem{GPT-3}
T.~Brown, B.~Mann, N.~Ryder, M.~Subbiah, J.~D. Kaplan, P.~Dhariwal, A.~Neelakantan, P.~Shyam, G.~Sastry, A.~Askell, et~al., Language models are few-shot learners, Advances in neural information processing systems 33 (2020) 1877--1901.

\bibitem{Bert}
J.~Devlin, M.-W. Chang, K.~Lee, K.~Toutanova, Bert: Pre-training of deep bidirectional transformers for language understanding, arXiv preprint arXiv:1810.04805 (2018).

\bibitem{ELMO}
M.~E. Peters, M.~Neumann, M.~Iyyer, M.~Gardner, C.~Clark, K.~Lee, L.~Zettlemoyer, Deep contextualized word representations, in: {NAACL-HLT}, Association for Computational Linguistics, 2018, pp. 2227--2237.

\bibitem{BART}
M.~Lewis, Y.~Liu, N.~Goyal, M.~Ghazvininejad, A.~Mohamed, O.~Levy, V.~Stoyanov, L.~Zettlemoyer, Bart: Denoising sequence-to-sequence pre-training for natural language generation, translation, and comprehension, arXiv preprint arXiv:1910.13461 (2019).

\bibitem{T5}
C.~Raffel, N.~Shazeer, A.~Roberts, K.~Lee, S.~Narang, M.~Matena, Y.~Zhou, W.~Li, P.~J. Liu, Exploring the limits of transfer learning with a unified text-to-text transformer, The Journal of Machine Learning Research 21~(1) (2020) 5485--5551.

\bibitem{mT5}
L.~Xue, N.~Constant, A.~Roberts, M.~Kale, R.~Al-Rfou, A.~Siddhant, A.~Barua, C.~Raffel, mt5: A massively multilingual pre-trained text-to-text transformer, arXiv preprint arXiv:2010.11934 (2020).

\bibitem{CPM-2}
Z.~Zhang, Y.~Gu, X.~Han, S.~Chen, C.~Xiao, Z.~Sun, Y.~Yao, F.~Qi, J.~Guan, P.~Ke, et~al., Cpm-2: Large-scale cost-effective pre-trained language models, AI Open 2 (2021) 216--224.

\bibitem{BLOOM}
T.~L. Scao, A.~Fan, C.~Akiki, E.~Pavlick, S.~Ili{\'c}, D.~Hesslow, R.~Castagn{\'e}, A.~S. Luccioni, F.~Yvon, M.~Gall{\'e}, et~al., Bloom: A 176b-parameter open-access multilingual language model, arXiv preprint arXiv:2211.05100 (2022).

\bibitem{OPT}
S.~Zhang, S.~Roller, N.~Goyal, M.~Artetxe, M.~Chen, S.~Chen, C.~Dewan, M.~Diab, X.~Li, X.~V. Lin, et~al., Opt: Open pre-trained transformer language models, arXiv preprint arXiv:2205.01068 (2022).

\bibitem{PaLM}
A.~Chowdhery, S.~Narang, J.~Devlin, M.~Bosma, G.~Mishra, A.~Roberts, P.~Barham, H.~W. Chung, C.~Sutton, S.~Gehrmann, et~al., Palm: Scaling language modeling with pathways, arXiv preprint arXiv:2204.02311 (2022).

\bibitem{Flan}
H.~W. Chung, L.~Hou, S.~Longpre, B.~Zoph, Y.~Tay, W.~Fedus, E.~Li, X.~Wang, M.~Dehghani, S.~Brahma, et~al., Scaling instruction-finetuned language models, arXiv preprint arXiv:2210.11416 (2022).

\bibitem{T0}
V.~Sanh, A.~Webson, C.~Raffel, S.~H. Bach, L.~Sutawika, Z.~Alyafeai, A.~Chaffin, A.~Stiegler, T.~L. Scao, A.~Raja, et~al., Multitask prompted training enables zero-shot task generalization, arXiv preprint arXiv:2110.08207 (2021).

\bibitem{Tk-INSTRUCT}
Y.~Wang, S.~Mishra, P.~Alipoormolabashi, Y.~Kordi, A.~Mirzaei, A.~Naik, A.~Ashok, A.~S. Dhanasekaran, A.~Arunkumar, D.~Stap, et~al., Super-naturalinstructions: Generalization via declarative instructions on 1600+ nlp tasks, in: Proceedings of the 2022 Conference on Empirical Methods in Natural Language Processing, 2022, pp. 5085--5109.

\bibitem{ft_self_instruct}
Y.~Wang, Y.~Kordi, S.~Mishra, A.~Liu, N.~A. Smith, D.~Khashabi, H.~Hajishirzi, Self-instruct: Aligning language model with self generated instructions, arXiv preprint arXiv:2212.10560 (2022).

\bibitem{instructgpt}
L.~Ouyang, J.~Wu, X.~Jiang, D.~Almeida, C.~Wainwright, P.~Mishkin, C.~Zhang, S.~Agarwal, K.~Slama, A.~Ray, et~al., Training language models to follow instructions with human feedback, Advances in Neural Information Processing Systems 35 (2022) 27730--27744.

\bibitem{llama_2}
H.~Touvron, L.~Martin, K.~Stone, P.~Albert, A.~Almahairi, Y.~Babaei, N.~Bashlykov, S.~Batra, P.~Bhargava, S.~Bhosale, et~al., Llama 2: Open foundation and fine-tuned chat models, arXiv preprint arXiv:2307.09288 (2023).

\bibitem{emergent_abilities}
J.~Wei, Y.~Tay, R.~Bommasani, C.~Raffel, B.~Zoph, S.~Borgeaud, D.~Yogatama, M.~Bosma, D.~Zhou, D.~Metzler, et~al., Emergent abilities of large language models, arXiv preprint arXiv:2206.07682 (2022).

\bibitem{emergent_analogical}
T.~Webb, K.~J. Holyoak, H.~Lu, Emergent analogical reasoning in large language models, Nature Human Behaviour 7~(9) (2023) 1526--1541.

\bibitem{emergent_research}
D.~A. Boiko, R.~MacKnight, G.~Gomes, Emergent autonomous scientific research capabilities of large language models, arXiv preprint arXiv:2304.05332 (2023).

\bibitem{atlas}
G.~Izacard, P.~Lewis, M.~Lomeli, L.~Hosseini, F.~Petroni, T.~Schick, J.~Dwivedi-Yu, A.~Joulin, S.~Riedel, E.~Grave, Few-shot learning with retrieval augmented language models, arXiv preprint arXiv:2208.03299 (2022).

\bibitem{PaLME}
D.~Driess, F.~Xia, M.~S. Sajjadi, C.~Lynch, A.~Chowdhery, B.~Ichter, A.~Wahid, J.~Tompson, Q.~Vuong, T.~Yu, et~al., Palm-e: An embodied multimodal language model, arXiv preprint arXiv:2303.03378 (2023).

\bibitem{talm}
A.~Parisi, Y.~Zhao, N.~Fiedel, Talm: Tool augmented language models, arXiv preprint arXiv:2205.12255 (2022).

\bibitem{zhang2023large}
B.~Zhang, H.~Soh, Large language models as zero-shot human models for human-robot interaction, arXiv preprint arXiv:2303.03548 (2023).

\bibitem{ye2023mplug}
Q.~Ye, H.~Xu, G.~Xu, J.~Ye, M.~Yan, Y.~Zhou, J.~Wang, A.~Hu, P.~Shi, Y.~Shi, et~al., mplug-owl: Modularization empowers large language models with multimodality, arXiv preprint arXiv:2304.14178 (2023).

\bibitem{wang2023visionllm}
W.~Wang, Z.~Chen, X.~Chen, J.~Wu, X.~Zhu, G.~Zeng, P.~Luo, T.~Lu, J.~Zhou, Y.~Qiao, et~al., Visionllm: Large language model is also an open-ended decoder for vision-centric tasks, arXiv preprint arXiv:2305.11175 (2023).

\bibitem{gpt4tools}
R.~Yang, L.~Song, Y.~Li, S.~Zhao, Y.~Ge, X.~Li, Y.~Shan, Gpt4tools: Teaching large language model to use tools via self-instruction, arXiv preprint arXiv:2305.18752 (2023).

\bibitem{prompt_eng_guide}
E.~Saravia, {Prompt Engineering Guide}, https://github.com/dair-ai/Prompt-Engineering-Guide (12 2022).

\bibitem{GLM-130B}
A.~Zeng, X.~Liu, Z.~Du, Z.~Wang, H.~Lai, M.~Ding, Z.~Yang, Y.~Xu, W.~Zheng, X.~Xia, et~al., Glm-130b: An open bilingual pre-trained model, arXiv preprint arXiv:2210.02414 (2022).

\bibitem{codet5+}
Y.~Wang, H.~Le, A.~D. Gotmare, N.~D. Bui, J.~Li, S.~C. Hoi, Codet5+: Open code large language models for code understanding and generation, arXiv preprint arXiv:2305.07922 (2023).

\bibitem{ernie3titan}
S.~Wang, Y.~Sun, Y.~Xiang, Z.~Wu, S.~Ding, W.~Gong, S.~Feng, J.~Shang, Y.~Zhao, C.~Pang, et~al., Ernie 3.0 titan: Exploring larger-scale knowledge enhanced pre-training for language understanding and generation, arXiv preprint arXiv:2112.12731 (2021).

\bibitem{Lib_DeepSpeed}
J.~Rasley, S.~Rajbhandari, O.~Ruwase, Y.~He, Deepspeed: System optimizations enable training deep learning models with over 100 billion parameters, in: Proceedings of the 26th ACM SIGKDD International Conference on Knowledge Discovery \& Data Mining, 2020, pp. 3505--3506.

\bibitem{ZeroOpt}
S.~Rajbhandari, J.~Rasley, O.~Ruwase, Y.~He, Zero: Memory optimizations toward training trillion parameter models, in: SC20: International Conference for High Performance Computing, Networking, Storage and Analysis, IEEE, 2020, pp. 1--16.

\bibitem{unified_peft}
J.~He, C.~Zhou, X.~Ma, T.~Berg-Kirkpatrick, G.~Neubig, Towards a unified view of parameter-efficient transfer learning, arXiv preprint arXiv:2110.04366 (2021).

\bibitem{LMAdapter_3}
Z.~Hu, Y.~Lan, L.~Wang, W.~Xu, E.-P. Lim, R.~K.-W. Lee, L.~Bing, S.~Poria, Llm-adapters: An adapter family for parameter-efficient fine-tuning of large language models, arXiv preprint arXiv:2304.01933 (2023).

\bibitem{Prompt_Tuning}
B.~Lester, R.~Al-Rfou, N.~Constant, The power of scale for parameter-efficient prompt tuning, arXiv preprint arXiv:2104.08691 (2021).

\bibitem{Prefix_Tuning}
X.~L. Li, P.~Liang, Prefix-tuning: Optimizing continuous prompts for generation, arXiv preprint arXiv:2101.00190 (2021).

\bibitem{llm_pruner}
X.~Ma, G.~Fang, X.~Wang, Llm-pruner: On the structural pruning of large language models, arXiv preprint arXiv:2305.11627 (2023).

\bibitem{contrastive_pruning}
R.~Xu, F.~Luo, C.~Wang, B.~Chang, J.~Huang, S.~Huang, F.~Huang, From dense to sparse: Contrastive pruning for better pre-trained language model compression, in: Proceedings of the AAAI Conference on Artificial Intelligence, Vol.~36, 2022, pp. 11547--11555.

\bibitem{SmoothQuant}
G.~Xiao, J.~Lin, M.~Seznec, H.~Wu, J.~Demouth, S.~Han, Smoothquant: Accurate and efficient post-training quantization for large language models, in: {ICML}, Vol. 202 of Proceedings of Machine Learning Research, {PMLR}, 2023, pp. 38087--38099.

\bibitem{compression_PLM_quant}
C.~Tao, L.~Hou, W.~Zhang, L.~Shang, X.~Jiang, Q.~Liu, P.~Luo, N.~Wong, Compression of generative pre-trained language models via quantization, arXiv preprint arXiv:2203.10705 (2022).

\bibitem{giraffe}
A.~Pal, D.~Karkhanis, M.~Roberts, S.~Dooley, A.~Sundararajan, S.~Naidu, Giraffe: Adventures in expanding context lengths in llms, arXiv preprint arXiv:2308.10882 (2023).

\bibitem{yarn}
B.~Peng, J.~Quesnelle, H.~Fan, E.~Shippole, Yarn: Efficient context window extension of large language models, arXiv preprint arXiv:2309.00071 (2023).

\bibitem{longt5}
M.~Guo, J.~Ainslie, D.~Uthus, S.~Ontanon, J.~Ni, Y.-H. Sung, Y.~Yang, Longt5: Efficient text-to-text transformer for long sequences, arXiv preprint arXiv:2112.07916 (2021).

\bibitem{pe_extending}
S.~Chen, S.~Wong, L.~Chen, Y.~Tian, Extending context window of large language models via positional interpolation, arXiv preprint arXiv:2306.15595 (2023).

\bibitem{Survey_LLM}
W.~X. Zhao, K.~Zhou, J.~Li, T.~Tang, X.~Wang, Y.~Hou, Y.~Min, B.~Zhang, J.~Zhang, Z.~Dong, et~al., A survey of large language models, arXiv preprint arXiv:2303.18223 (2023).

\bibitem{survey_smaller_LLMs_1}
U.~Naseem, I.~Razzak, S.~K. Khan, M.~Prasad, A comprehensive survey on word representation models: From classical to state-of-the-art word representation language models, Transactions on Asian and Low-Resource Language Information Processing 20~(5) (2021) 1--35.

\bibitem{survey_smaller_LLMs_2}
B.~Min, H.~Ross, E.~Sulem, A.~P.~B. Veyseh, T.~H. Nguyen, O.~Sainz, E.~Agirre, I.~Heinz, D.~Roth, Recent advances in natural language processing via large pre-trained language models: A survey, arXiv preprint arXiv:2111.01243 (2021).

\bibitem{survey_1}
C.~Zhou, Q.~Li, C.~Li, J.~Yu, Y.~Liu, G.~Wang, K.~Zhang, C.~Ji, Q.~Yan, L.~He, et~al., A comprehensive survey on pretrained foundation models: A history from bert to chatgpt, arXiv preprint arXiv:2302.09419 (2023).

\bibitem{survey_incontext_learning}
Q.~Dong, L.~Li, D.~Dai, C.~Zheng, Z.~Wu, B.~Chang, X.~Sun, J.~Xu, Z.~Sui, A survey for in-context learning, arXiv preprint arXiv:2301.00234 (2022).

\bibitem{survey_reasoning}
J.~Huang, K.~C.-C. Chang, Towards reasoning in large language models: A survey, arXiv preprint arXiv:2212.10403 (2022).

\bibitem{survey_aligning}
Y.~Wang, W.~Zhong, L.~Li, F.~Mi, X.~Zeng, W.~Huang, L.~Shang, X.~Jiang, Q.~Liu, Aligning large language models with human: A survey, arXiv preprint arXiv:2307.12966 (2023).

\bibitem{survey_compression}
X.~Zhu, J.~Li, Y.~Liu, C.~Ma, W.~Wang, A survey on model compression for large language models, arXiv preprint arXiv:2308.07633 (2023).

\bibitem{survey_multimodal_llm}
S.~Yin, C.~Fu, S.~Zhao, K.~Li, X.~Sun, T.~Xu, E.~Chen, A survey on multimodal large language models, arXiv preprint arXiv:2306.13549 (2023).

\bibitem{webster1992tokenization}
J.~J. Webster, C.~Kit, Tokenization as the initial phase in nlp, in: COLING 1992 volume 4: The 14th international conference on computational linguistics, 1992.

\bibitem{unigramLM}
T.~Kudo, Subword regularization: Improving neural network translation models with multiple subword candidates, in: Proceedings of the 56th Annual Meeting of the Association for Computational Linguistics (Volume 1: Long Papers), 2018, pp. 66--75.

\bibitem{bpe}
R.~Sennrich, B.~Haddow, A.~Birch, Neural machine translation of rare words with subword units, in: Proceedings of the 54th Annual Meeting of the Association for Computational Linguistics (Volume 1: Long Papers), 2016, pp. 1715--1725.

\bibitem{wordpiece}
M.~Schuster, K.~Nakajima, Japanese and korean voice search, in: 2012 IEEE international conference on acoustics, speech and signal processing (ICASSP), IEEE, 2012, pp. 5149--5152.

\bibitem{tokenizationsurvey}
S.~J. Mielke, Z.~Alyafeai, E.~Salesky, C.~Raffel, M.~Dey, M.~Gall{\'e}, A.~Raja, C.~Si, W.~Y. Lee, B.~Sagot, et~al., Between words and characters: A brief history of open-vocabulary modeling and tokenization in nlp, arXiv preprint arXiv:2112.10508 (2021).

\bibitem{Transformers}
A.~Vaswani, N.~Shazeer, N.~Parmar, J.~Uszkoreit, L.~Jones, A.~N. Gomez, {\L}.~Kaiser, I.~Polosukhin, Attention is all you need, Advances in neural information processing systems 30 (2017).

\bibitem{alibi}
O.~Press, N.~Smith, M.~Lewis, \href{https://openreview.net/forum?id=R8sQPpGCv0}{Train short, test long: Attention with linear biases enables input length extrapolation}, in: International Conference on Learning Representations, 2022.
\newline\urlprefix\url{https://openreview.net/forum?id=R8sQPpGCv0}

\bibitem{su2021roformer}
J.~Su, Y.~Lu, S.~Pan, A.~Murtadha, B.~Wen, Y.~Liu, Roformer: Enhanced transformer with rotary position embedding, arXiv preprint arXiv:2104.09864 (2021).

\bibitem{sparse_transformer}
R.~Child, S.~Gray, A.~Radford, I.~Sutskever, Generating long sequences with sparse transformers, arXiv preprint arXiv:1904.10509 (2019).

\bibitem{flashattention}
T.~Dao, D.~Fu, S.~Ermon, A.~Rudra, C.~R{\'e}, Flashattention: Fast and memory-efficient exact attention with io-awareness, Advances in Neural Information Processing Systems 35 (2022) 16344--16359.

\bibitem{activationfunction}
K.~Hornik, M.~Stinchcombe, H.~White, Multilayer feedforward networks are universal approximators, Neural networks 2~(5) (1989) 359--366.

\bibitem{relu}
V.~Nair, G.~E. Hinton, Rectified linear units improve restricted boltzmann machines, in: Proceedings of the 27th international conference on machine learning (ICML-10), 2010, pp. 807--814.

\bibitem{gelu}
D.~Hendrycks, K.~Gimpel, Gaussian error linear units (gelus), arXiv preprint arXiv:1606.08415 (2016).

\bibitem{srivastava2014dropout}
N.~Srivastava, G.~Hinton, A.~Krizhevsky, I.~Sutskever, R.~Salakhutdinov, Dropout: a simple way to prevent neural networks from overfitting, The journal of machine learning research 15~(1) (2014) 1929--1958.

\bibitem{krueger2016zoneout}
D.~Krueger, T.~Maharaj, J.~Kram{\'a}r, M.~Pezeshki, N.~Ballas, N.~R. Ke, A.~Goyal, Y.~Bengio, A.~Courville, C.~Pal, Zoneout: Regularizing rnns by randomly preserving hidden activations, arXiv preprint arXiv:1606.01305 (2016).

\bibitem{shazeer2020glu}
N.~Shazeer, Glu variants improve transformer, arXiv preprint arXiv:2002.05202 (2020).

\bibitem{glu}
Y.~N. Dauphin, A.~Fan, M.~Auli, D.~Grangier, Language modeling with gated convolutional networks, in: International conference on machine learning, PMLR, 2017, pp. 933--941.

\bibitem{layernorm}
J.~L. Ba, J.~R. Kiros, G.~E. Hinton, Layer normalization, arXiv preprint arXiv:1607.06450 (2016).

\bibitem{rmsnorm}
B.~Zhang, R.~Sennrich, Root mean square layer normalization, Advances in Neural Information Processing Systems 32 (2019).

\bibitem{preLN}
A.~Baevski, M.~Auli, Adaptive input representations for neural language modeling, arXiv preprint arXiv:1809.10853 (2018).

\bibitem{deepnorm}
H.~Wang, S.~Ma, L.~Dong, S.~Huang, D.~Zhang, F.~Wei, Deepnet: Scaling transformers to 1,000 layers, arXiv preprint arXiv:2203.00555 (2022).

\bibitem{Lib_Megatron}
M.~Shoeybi, M.~Patwary, R.~Puri, P.~LeGresley, J.~Casper, B.~Catanzaro, Megatron-lm: Training multi-billion parameter language models using model parallelism, arXiv preprint arXiv:1909.08053 (2019).

\bibitem{Lib_Bmtrain}
\href{https://github.com/OpenBMB/BMTrain}{"bmtrain: Efficient training for big models."}.
\newline\urlprefix\url{https://github.com/OpenBMB/BMTrain}

\bibitem{Lib_Transformers}
T.~Wolf, L.~Debut, V.~Sanh, J.~Chaumond, C.~Delangue, A.~Moi, P.~Cistac, T.~Rault, R.~Louf, M.~Funtowicz, et~al., Transformers: State-of-the-art natural language processing, in: Proceedings of the 2020 conference on empirical methods in natural language processing: system demonstrations, 2020, pp. 38--45.

\bibitem{Lib_Jax}
J.~Bradbury, R.~Frostig, P.~Hawkins, M.~J. Johnson, C.~Leary, D.~Maclaurin, G.~Necula, A.~Paszke, J.~VanderPlas, S.~Wanderman-Milne, et~al., Jax: composable transformations of python+ numpy programs (2018).

\bibitem{Lib_Colossal}
S.~Li, J.~Fang, Z.~Bian, H.~Liu, Y.~Liu, H.~Huang, B.~Wang, Y.~You, Colossal-ai: A unified deep learning system for large-scale parallel training, arXiv preprint arXiv:2110.14883 (2021).

\bibitem{Lib_Fastmoe}
J.~He, J.~Qiu, A.~Zeng, Z.~Yang, J.~Zhai, J.~Tang, Fastmoe: A fast mixture-of-expert training system, arXiv preprint arXiv:2103.13262 (2021).

\bibitem{Lib_Mindspore}
L.~Huawei Technologies~Co., Huawei mindspore ai development framework, in: Artificial Intelligence Technology, Springer, 2022, pp. 137--162.

\bibitem{Lib_Pytorch}
A.~Paszke, S.~Gross, F.~Massa, A.~Lerer, J.~Bradbury, G.~Chanan, T.~Killeen, Z.~Lin, N.~Gimelshein, L.~Antiga, et~al., Pytorch: An imperative style, high-performance deep learning library, Advances in neural information processing systems 32 (2019).

\bibitem{Lib_Tensorflow}
M.~Abadi, P.~Barham, J.~Chen, Z.~Chen, A.~Davis, J.~Dean, M.~Devin, S.~Ghemawat, G.~Irving, M.~Isard, et~al., Tensorflow: a system for large-scale machine learning., in: Osdi, Vol.~16, Savannah, GA, USA, 2016, pp. 265--283.

\bibitem{Lib_Mxnet}
T.~Chen, M.~Li, Y.~Li, M.~Lin, N.~Wang, M.~Wang, T.~Xiao, B.~Xu, C.~Zhang, Z.~Zhang, Mxnet: A flexible and efficient machine learning library for heterogeneous distributed systems, arXiv preprint arXiv:1512.01274 (2015).

\bibitem{fedus2022switch}
W.~Fedus, B.~Zoph, N.~Shazeer, Switch transformers: Scaling to trillion parameter models with simple and efficient sparsity, The Journal of Machine Learning Research 23~(1) (2022) 5232--5270.

\bibitem{du2022glam}
N.~Du, Y.~Huang, A.~M. Dai, S.~Tong, D.~Lepikhin, Y.~Xu, M.~Krikun, Y.~Zhou, A.~W. Yu, O.~Firat, et~al., Glam: Efficient scaling of language models with mixture-of-experts, in: International Conference on Machine Learning, PMLR, 2022, pp. 5547--5569.

\bibitem{PanGu_sigma}
X.~Ren, P.~Zhou, X.~Meng, X.~Huang, Y.~Wang, W.~Wang, P.~Li, X.~Zhang, A.~Podolskiy, G.~Arshinov, et~al., Pangu-$\sum$: Towards trillion parameter language model with sparse heterogeneous computing, arXiv preprint arXiv:2303.10845 (2023).

\bibitem{LLM_Objectives}
T.~Wang, A.~Roberts, D.~Hesslow, T.~Le~Scao, H.~W. Chung, I.~Beltagy, J.~Launay, C.~Raffel, What language model architecture and pretraining objective works best for zero-shot generalization?, in: International Conference on Machine Learning, PMLR, 2022, pp. 22964--22984.

\bibitem{Unified_LM}
L.~Dong, N.~Yang, W.~Wang, F.~Wei, X.~Liu, Y.~Wang, J.~Gao, M.~Zhou, H.-W. Hon, Unified language model pre-training for natural language understanding and generation, Advances in neural information processing systems 32 (2019).

\bibitem{kaplan_scaling}
J.~Kaplan, S.~McCandlish, T.~Henighan, T.~B. Brown, B.~Chess, R.~Child, S.~Gray, A.~Radford, J.~Wu, D.~Amodei, Scaling laws for neural language models, arXiv preprint arXiv:2001.08361 (2020).

\bibitem{chinchilla}
J.~Hoffmann, S.~Borgeaud, A.~Mensch, E.~Buchatskaya, T.~Cai, E.~Rutherford, D.~d.~L. Casas, L.~A. Hendricks, J.~Welbl, A.~Clark, et~al., Training compute-optimal large language models, arXiv preprint arXiv:2203.15556 (2022).

\bibitem{OPT_IML}
S.~Iyer, X.~V. Lin, R.~Pasunuru, T.~Mihaylov, D.~Simig, P.~Yu, K.~Shuster, T.~Wang, Q.~Liu, P.~S. Koura, et~al., Opt-iml: Scaling language model instruction meta learning through the lens of generalization, arXiv preprint arXiv:2212.12017 (2022).

\bibitem{self_align}
Z.~Sun, Y.~Shen, Q.~Zhou, H.~Zhang, Z.~Chen, D.~Cox, Y.~Yang, C.~Gan, Principle-driven self-alignment of language models from scratch with minimal human supervision, arXiv preprint arXiv:2305.03047 (2023).

\bibitem{askell2021general}
A.~Askell, Y.~Bai, A.~Chen, D.~Drain, D.~Ganguli, T.~Henighan, A.~Jones, N.~Joseph, B.~Mann, N.~DasSarma, et~al., A general language assistant as a laboratory for alignment, arXiv preprint arXiv:2112.00861 (2021).

\bibitem{ziegler2019fine}
D.~M. Ziegler, N.~Stiennon, J.~Wu, T.~B. Brown, A.~Radford, D.~Amodei, P.~Christiano, G.~Irving, Fine-tuning language models from human preferences, arXiv preprint arXiv:1909.08593 (2019).

\bibitem{cot_collection}
S.~Kim, S.~J. Joo, D.~Kim, J.~Jang, S.~Ye, J.~Shin, M.~Seo, The cot collection: Improving zero-shot and few-shot learning of language models via chain-of-thought fine-tuning, arXiv preprint arXiv:2305.14045 (2023).

\bibitem{zero_to_hero}
Q.~Liu, F.~Zhou, Z.~Jiang, L.~Dou, M.~Lin, From zero to hero: Examining the power of symbolic tasks in instruction tuning, arXiv preprint arXiv:2304.07995 (2023).

\bibitem{wei2022chain}
J.~Wei, X.~Wang, D.~Schuurmans, M.~Bosma, F.~Xia, E.~Chi, Q.~V. Le, D.~Zhou, et~al., Chain-of-thought prompting elicits reasoning in large language models, Advances in Neural Information Processing Systems 35 (2022) 24824--24837.

\bibitem{self_consistency}
X.~Wang, J.~Wei, D.~Schuurmans, Q.~Le, E.~Chi, S.~Narang, A.~Chowdhery, D.~Zhou, Self-consistency improves chain of thought reasoning in language models, arXiv preprint arXiv:2203.11171 (2022).

\bibitem{ToT}
S.~Yao, D.~Yu, J.~Zhao, I.~Shafran, T.~L. Griffiths, Y.~Cao, K.~Narasimhan, Tree of thoughts: Deliberate problem solving with large language models, arXiv preprint arXiv:2305.10601 (2023).

\bibitem{LMAdapter}
N.~Houlsby, A.~Giurgiu, S.~Jastrzebski, B.~Morrone, Q.~De~Laroussilhe, A.~Gesmundo, M.~Attariyan, S.~Gelly, Parameter-efficient transfer learning for nlp, in: International Conference on Machine Learning, PMLR, 2019, pp. 2790--2799.

\bibitem{batch_size_selec}
S.~McCandlish, J.~Kaplan, D.~Amodei, O.~D. Team, An empirical model of large-batch training, arXiv preprint arXiv:1812.06162 (2018).

\bibitem{PanGU_alpha}
W.~Zeng, X.~Ren, T.~Su, H.~Wang, Y.~Liao, Z.~Wang, X.~Jiang, Z.~Yang, K.~Wang, X.~Zhang, et~al., Pangu-$\alpha$ : Large-scale autoregressive pretrained chinese language models with auto-parallel computation, arXiv preprint arXiv:2104.12369 (2021).

\bibitem{WuDaoCorpus}
S.~Yuan, H.~Zhao, Z.~Du, M.~Ding, X.~Liu, Y.~Cen, X.~Zou, Z.~Yang, J.~Tang, Wudaocorpora: A super large-scale chinese corpora for pre-training language models, AI Open 2 (2021) 65--68.

\bibitem{ernie3}
Y.~Sun, S.~Wang, S.~Feng, S.~Ding, C.~Pang, J.~Shang, J.~Liu, X.~Chen, Y.~Zhao, Y.~Lu, et~al., Ernie 3.0: Large-scale knowledge enhanced pre-training for language understanding and generation, arXiv preprint arXiv:2107.02137 (2021).

\bibitem{dai2019transformer}
Z.~Dai, Z.~Yang, Y.~Yang, J.~Carbonell, Q.~V. Le, R.~Salakhutdinov, Transformer-xl: Attentive language models beyond a fixed-length context, arXiv preprint arXiv:1901.02860 (2019).

\bibitem{lieber2021jurassic}
O.~Lieber, O.~Sharir, B.~Lenz, Y.~Shoham, Jurassic-1: Technical details and evaluation, White Paper. AI21 Labs 1 (2021).

\bibitem{limit_to_depth}
Y.~Levine, N.~Wies, O.~Sharir, H.~Bata, A.~Shashua, Limits to depth efficiencies of self-attention, Advances in Neural Information Processing Systems 33 (2020) 22640--22651.

\bibitem{hyperclova}
B.~Kim, H.~Kim, S.-W. Lee, G.~Lee, D.~Kwak, D.~H. Jeon, S.~Park, S.~Kim, S.~Kim, D.~Seo, et~al., What changes can large-scale language models bring? intensive study on hyperclova: Billions-scale korean generative pretrained transformers, arXiv preprint arXiv:2109.04650 (2021).

\bibitem{wu2021yuan}
S.~Wu, X.~Zhao, T.~Yu, R.~Zhang, C.~Shen, H.~Liu, F.~Li, H.~Zhu, J.~Luo, L.~Xu, et~al., Yuan 1.0: Large-scale pre-trained language model in zero-shot and few-shot learning, arXiv preprint arXiv:2110.04725 (2021).

\bibitem{gopher}
J.~W. Rae, S.~Borgeaud, T.~Cai, K.~Millican, J.~Hoffmann, F.~Song, J.~Aslanides, S.~Henderson, R.~Ring, S.~Young, et~al., Scaling language models: Methods, analysis \& insights from training gopher, arXiv preprint arXiv:2112.11446 (2021).

\bibitem{mtnlg}
S.~Smith, M.~Patwary, B.~Norick, P.~LeGresley, S.~Rajbhandari, J.~Casper, Z.~Liu, S.~Prabhumoye, G.~Zerveas, V.~Korthikanti, et~al., Using deepspeed and megatron to train megatron-turing nlg 530b, a large-scale generative language model, arXiv preprint arXiv:2201.11990 (2022).

\bibitem{GPT_NeoX}
S.~Black, S.~Biderman, E.~Hallahan, Q.~Anthony, L.~Gao, L.~Golding, H.~He, C.~Leahy, K.~McDonell, J.~Phang, et~al., Gpt-neox-20b: An open-source autoregressive language model, arXiv preprint arXiv:2204.06745 (2022).

\bibitem{GPT_J_6B}
W.~Ben, K.~Aran, Gpt-j-6b: A 6 billion parameter autoregressive language model (2021).

\bibitem{Mixed_Precision}
P.~Micikevicius, S.~Narang, J.~Alben, G.~Diamos, E.~Elsen, D.~Garcia, B.~Ginsburg, M.~Houston, O.~Kuchaiev, G.~Venkatesh, et~al., Mixed precision training, arXiv preprint arXiv:1710.03740 (2017).

\bibitem{shazeer2017outrageously}
N.~Shazeer, A.~Mirhoseini, K.~Maziarz, A.~Davis, Q.~Le, G.~Hinton, J.~Dean, Outrageously large neural networks: The sparsely-gated mixture-of-experts layer, arXiv preprint arXiv:1701.06538 (2017).

\bibitem{soltan2022alexatm}
S.~Soltan, S.~Ananthakrishnan, J.~FitzGerald, R.~Gupta, W.~Hamza, H.~Khan, C.~Peris, S.~Rawls, A.~Rosenbaum, A.~Rumshisky, et~al., Alexatm 20b: Few-shot learning using a large-scale multilingual seq2seq model, arXiv preprint arXiv:2208.01448 (2022).

\bibitem{palm_2}
R.~Anil, A.~M. Dai, O.~Firat, M.~Johnson, D.~Lepikhin, A.~Passos, S.~Shakeri, E.~Taropa, P.~Bailey, Z.~Chen, et~al., Palm 2 technical report, arXiv preprint arXiv:2305.10403 (2023).

\bibitem{U-PaLM}
Y.~Tay, J.~Wei, H.~W. Chung, V.~Q. Tran, D.~R. So, S.~Shakeri, X.~Garcia, H.~S. Zheng, J.~Rao, A.~Chowdhery, et~al., Transcending scaling laws with 0.1\% extra compute, arXiv preprint arXiv:2210.11399 (2022).

\bibitem{UL2}
Y.~Tay, M.~Dehghani, V.~Q. Tran, X.~Garcia, J.~Wei, X.~Wang, H.~W. Chung, D.~Bahri, T.~Schuster, S.~Zheng, et~al., Ul2: Unifying language learning paradigms, in: The Eleventh International Conference on Learning Representations, 2022.

\bibitem{GLM}
Z.~Du, Y.~Qian, X.~Liu, M.~Ding, J.~Qiu, Z.~Yang, J.~Tang, Glm: General language model pretraining with autoregressive blank infilling, in: Proceedings of the 60th Annual Meeting of the Association for Computational Linguistics (Volume 1: Long Papers), 2022, pp. 320--335.

\bibitem{touvron2023llama}
H.~Touvron, T.~Lavril, G.~Izacard, X.~Martinet, M.-A. Lachaux, T.~Lacroix, B.~Rozi{\`e}re, N.~Goyal, E.~Hambro, F.~Azhar, et~al., Llama: Open and efficient foundation language models, arXiv preprint arXiv:2302.13971 (2023).

\bibitem{self_att_reduced_mem}
M.~N. Rabe, C.~Staats, Self-attention does not need o(n$^2$) memory, arXiv preprint arXiv:2112.05682 (2021).

\bibitem{reducing_act_recompute}
V.~A. Korthikanti, J.~Casper, S.~Lym, L.~McAfee, M.~Andersch, M.~Shoeybi, B.~Catanzaro, Reducing activation recomputation in large transformer models, Proceedings of Machine Learning and Systems 5 (2023).

\bibitem{llama3}
A.~Dubey, A.~Jauhri, A.~Pandey, A.~Kadian, A.~Al-Dahle, A.~Letman, A.~Mathur, A.~Schelten, A.~Yang, A.~Fan, et~al., The llama 3 herd of models, arXiv preprint arXiv:2407.21783 (2024).

\bibitem{mixtral}
\url{https://mistral.ai/news/mixtral-8x22b/}.

\bibitem{snowflake_arctic}
\url{https://github.com/Snowflake-Labs/snowflake-arctic}.

\bibitem{grok_1}
\url{https://github.com/xai-org/grok-1}.

\bibitem{grok_15}
\url{https://x.ai/blog/grok-1.5}.

\bibitem{gemini}
G.~Team, R.~Anil, S.~Borgeaud, Y.~Wu, J.-B. Alayrac, J.~Yu, R.~Soricut, J.~Schalkwyk, A.~M. Dai, A.~Hauth, et~al., Gemini: a family of highly capable multimodal models, arXiv preprint arXiv:2312.11805 (2023).

\bibitem{gemini_15}
M.~Reid, N.~Savinov, D.~Teplyashin, D.~Lepikhin, T.~Lillicrap, J.-b. Alayrac, R.~Soricut, A.~Lazaridou, O.~Firat, J.~Schrittwieser, et~al., Gemini 1.5: Unlocking multimodal understanding across millions of tokens of context, arXiv preprint arXiv:2403.05530 (2024).

\bibitem{nemotron4}
B.~Adler, N.~Agarwal, A.~Aithal, D.~H. Anh, P.~Bhattacharya, A.~Brundyn, J.~Casper, B.~Catanzaro, S.~Clay, J.~Cohen, et~al., Nemotron-4 340b technical report, arXiv preprint arXiv:2406.11704 (2024).

\bibitem{deepseek}
X.~Bi, D.~Chen, G.~Chen, S.~Chen, D.~Dai, C.~Deng, H.~Ding, K.~Dong, Q.~Du, Z.~Fu, et~al., Deepseek llm: Scaling open-source language models with longtermism, arXiv preprint arXiv:2401.02954 (2024).

\bibitem{deepseek_v2}
DeepSeek{-}AI, A.~Liu, B.~Feng, B.~Wang, B.~Wang, B.~Liu, C.~Zhao, C.~Deng, C.~Ruan, D.~Dai, D.~Guo, D.~Yang, D.~Chen, D.~Ji, E.~Li, F.~Lin, F.~Luo, G.~Hao, G.~Chen, G.~Li, H.~Zhang, H.~Xu, H.~Yang, H.~Zhang, H.~Ding, H.~Xin, H.~Gao, H.~Li, H.~Qu, J.~L. Cai, J.~Liang, J.~Guo, J.~Ni, J.~Li, J.~Chen, J.~Yuan, J.~Qiu, J.~Song, K.~Dong, K.~Gao, K.~Guan, L.~Wang, L.~Zhang, L.~Xu, L.~Xia, L.~Zhao, L.~Zhang, M.~Li, M.~Wang, M.~Zhang, M.~Zhang, M.~Tang, M.~Li, N.~Tian, P.~Huang, P.~Wang, P.~Zhang, Q.~Zhu, Q.~Chen, Q.~Du, R.~J. Chen, R.~L. Jin, R.~Ge, R.~Pan, R.~Xu, R.~Chen, S.~S. Li, S.~Lu, S.~Zhou, S.~Chen, S.~Wu, S.~Ye, S.~Ma, S.~Wang, S.~Zhou, S.~Yu, S.~Zhou, S.~Zheng, T.~Wang, T.~Pei, T.~Yuan, T.~Sun, W.~L. Xiao, W.~Zeng, W.~An, W.~Liu, W.~Liang, W.~Gao, W.~Zhang, X.~Q. Li, X.~Jin, X.~Wang, X.~Bi, X.~Liu, X.~Wang, X.~Shen, X.~Chen, X.~Chen, X.~Nie, X.~Sun, Deepseek-v2: {A} strong, economical, and efficient mixture-of-experts language model, CoRR abs/2405.04434 (2024).

\bibitem{CodeGen}
E.~Nijkamp, B.~Pang, H.~Hayashi, L.~Tu, H.~Wang, Y.~Zhou, S.~Savarese, C.~Xiong, Codegen: An open large language model for code with multi-turn program synthesis, arXiv preprint arXiv:2203.13474 (2022).

\bibitem{codex}
M.~Chen, J.~Tworek, H.~Jun, Q.~Yuan, H.~P. d.~O. Pinto, J.~Kaplan, H.~Edwards, Y.~Burda, N.~Joseph, G.~Brockman, et~al., Evaluating large language models trained on code, arXiv preprint arXiv:2107.03374 (2021).

\bibitem{li2022competition}
Y.~Li, D.~Choi, J.~Chung, N.~Kushman, J.~Schrittwieser, R.~Leblond, T.~Eccles, J.~Keeling, F.~Gimeno, A.~Dal~Lago, et~al., Competition-level code generation with alphacode, Science 378~(6624) (2022) 1092--1097.

\bibitem{shazeer2019fast}
N.~Shazeer, Fast transformer decoding: One write-head is all you need, arXiv preprint arXiv:1911.02150 (2019).

\bibitem{pang2020text}
R.~Y. Pang, H.~He, Text generation by learning from demonstrations, arXiv preprint arXiv:2009.07839 (2020).

\bibitem{dabre2020softmax}
R.~Dabre, A.~Fujita, Softmax tempering for training neural machine translation models, arXiv preprint arXiv:2009.09372 (2020).

\bibitem{codet5}
Y.~Wang, W.~Wang, S.~Joty, S.~C. Hoi, Codet5: Identifier-aware unified pre-trained encoder-decoder models for code understanding and generation, arXiv preprint arXiv:2109.00859 (2021).

\bibitem{starcoder}
R.~Li, L.~B. Allal, Y.~Zi, N.~Muennighoff, D.~Kocetkov, C.~Mou, M.~Marone, C.~Akiki, J.~Li, J.~Chim, et~al., Starcoder: may the source be with you!, arXiv preprint arXiv:2305.06161 (2023).

\bibitem{galactica}
R.~Taylor, M.~Kardas, G.~Cucurull, T.~Scialom, A.~Hartshorn, E.~Saravia, A.~Poulton, V.~Kerkez, R.~Stojnic, Galactica: A large language model for science, arXiv preprint arXiv:2211.09085 (2022).

\bibitem{fairscale}
{FairScale authors}, Fairscale: A general purpose modular pytorch library for high performance and large scale training, \url{https://github.com/facebookresearch/fairscale} (2021).

\bibitem{thoppilan2022lamda}
R.~Thoppilan, D.~De~Freitas, J.~Hall, N.~Shazeer, A.~Kulshreshtha, H.-T. Cheng, A.~Jin, T.~Bos, L.~Baker, Y.~Du, et~al., Lamda: Language models for dialog applications, arXiv preprint arXiv:2201.08239 (2022).

\bibitem{bloomberggpt}
S.~Wu, O.~Irsoy, S.~Lu, V.~Dabravolski, M.~Dredze, S.~Gehrmann, P.~Kambadur, D.~Rosenberg, G.~Mann, Bloomberggpt: A large language model for finance, arXiv preprint arXiv:2303.17564 (2023).

\bibitem{xuanyuan}
X.~Zhang, Q.~Yang, D.~Xu, Xuanyuan 2.0: A large chinese financial chat model with hundreds of billions parameters, arXiv preprint arXiv:2305.12002 (2023).

\bibitem{Mesh_Transformer_JAX}
W.~Ben, Mesh-transformer-jax: Model-parallel implementation of transformer language model with jax (2021).

\bibitem{mT0andBLOOMZ}
N.~Muennighoff, T.~Wang, L.~Sutawika, A.~Roberts, S.~Biderman, T.~L. Scao, M.~S. Bari, S.~Shen, Z.-X. Yong, H.~Schoelkopf, et~al., Crosslingual generalization through multitask finetuning, arXiv preprint arXiv:2211.01786 (2022).

\bibitem{dynosaur}
D.~Yin, X.~Liu, F.~Yin, M.~Zhong, H.~Bansal, J.~Han, K.-W. Chang, Dynosaur: A dynamic growth paradigm for instruction-tuning data curation, arXiv preprint arXiv:2305.14327 (2023).

\bibitem{gao2023llama}
P.~Gao, J.~Han, R.~Zhang, Z.~Lin, S.~Geng, A.~Zhou, W.~Zhang, P.~Lu, C.~He, X.~Yue, et~al., Llama-adapter v2: Parameter-efficient visual instruction model, arXiv preprint arXiv:2304.15010 (2023).

\bibitem{GPT-4}
Openai. gpt-4 technical report (2023).

\bibitem{alpaca}
R.~Taori, I.~Gulrajani, T.~Zhang, Y.~Dubois, X.~Li, C.~Guestrin, P.~Liang, T.~B. Hashimoto, Stanford alpaca: An instruction-following llama model, \url{https://github.com/tatsu-lab/stanford_alpaca} (2023).

\bibitem{vicuna}
W.-L. Chiang, Z.~Li, Z.~Lin, Y.~Sheng, Z.~Wu, H.~Zhang, L.~Zheng, S.~Zhuang, Y.~Zhuang, J.~E. Gonzalez, I.~Stoica, E.~P. Xing, \href{https://lmsys.org/blog/2023-03-30-vicuna/}{Vicuna: An open-source chatbot impressing gpt-4 with 90\%* chatgpt quality} (March 2023).
\newline\urlprefix\url{https://lmsys.org/blog/2023-03-30-vicuna/}

\bibitem{llama_gpt_4}
B.~Peng, C.~Li, P.~He, M.~Galley, J.~Gao, Instruction tuning with gpt-4, arXiv preprint arXiv:2304.03277 (2023).

\bibitem{goat}
T.~Liu, B.~K.~H. Low, Goat: Fine-tuned llama outperforms gpt-4 on arithmetic tasks, arXiv preprint arXiv:2305.14201 (2023).

\bibitem{huatuo}
H.~Wang, C.~Liu, N.~Xi, Z.~Qiang, S.~Zhao, B.~Qin, T.~Liu, Huatuo: Tuning llama model with chinese medical knowledge, arXiv preprint arXiv:2304.06975 (2023).

\bibitem{wizardlm}
C.~Xu, Q.~Sun, K.~Zheng, X.~Geng, P.~Zhao, J.~Feng, C.~Tao, D.~Jiang, Wizardlm: Empowering large language models to follow complex instructions, arXiv preprint arXiv:2304.12244 (2023).

\bibitem{wizardcoder}
Z.~Luo, C.~Xu, P.~Zhao, Q.~Sun, X.~Geng, W.~Hu, C.~Tao, J.~Ma, Q.~Lin, D.~Jiang, Wizardcoder: Empowering code large language models with evol-instruct, arXiv preprint arXiv:2306.08568 (2023).

\bibitem{gopher_cite}
J.~Menick, M.~Trebacz, V.~Mikulik, J.~Aslanides, F.~Song, M.~Chadwick, M.~Glaese, S.~Young, L.~Campbell-Gillingham, G.~Irving, et~al., Teaching language models to support answers with verified quotes, arXiv preprint arXiv:2203.11147 (2022).

\bibitem{nakano2021webgpt}
R.~Nakano, J.~Hilton, S.~Balaji, J.~Wu, L.~Ouyang, C.~Kim, C.~Hesse, S.~Jain, V.~Kosaraju, W.~Saunders, et~al., Webgpt: Browser-assisted question-answering with human feedback, arXiv preprint arXiv:2112.09332 (2021).

\bibitem{sparrow}
A.~Glaese, N.~McAleese, M.~Tr{\k{e}}bacz, J.~Aslanides, V.~Firoiu, T.~Ewalds, M.~Rauh, L.~Weidinger, M.~Chadwick, P.~Thacker, et~al., Improving alignment of dialogue agents via targeted human judgements, arXiv preprint arXiv:2209.14375 (2022).

\bibitem{DPO}
R.~Rafailov, A.~Sharma, E.~Mitchell, S.~Ermon, C.~D. Manning, C.~Finn, Direct preference optimization: Your language model is secretly a reward model, arXiv preprint arXiv:2305.18290 (2023).

\bibitem{raft}
H.~Dong, W.~Xiong, D.~Goyal, R.~Pan, S.~Diao, J.~Zhang, K.~Shum, T.~Zhang, Raft: Reward ranked finetuning for generative foundation model alignment, arXiv preprint arXiv:2304.06767 (2023).

\bibitem{rrhf}
Z.~Yuan, H.~Yuan, C.~Tan, W.~Wang, S.~Huang, F.~Huang, Rrhf: Rank responses to align language models with human feedback without tears, arXiv preprint arXiv:2304.05302 (2023).

\bibitem{PRO}
F.~Song, B.~Yu, M.~Li, H.~Yu, F.~Huang, Y.~Li, H.~Wang, Preference ranking optimization for human alignment, arXiv preprint arXiv:2306.17492 (2023).

\bibitem{CoH}
H.~Liu, C.~Sferrazza, P.~Abbeel, Languages are rewards: Hindsight finetuning using human feedback, arXiv preprint arXiv:2302.02676 (2023).

\bibitem{constitutional_ai}
Y.~Bai, S.~Kadavath, S.~Kundu, A.~Askell, J.~Kernion, A.~Jones, A.~Chen, A.~Goldie, A.~Mirhoseini, C.~McKinnon, et~al., Constitutional ai: Harmlessness from ai feedback, arXiv preprint arXiv:2212.08073 (2022).

\bibitem{alpacafarm}
Y.~Dubois, X.~Li, R.~Taori, T.~Zhang, I.~Gulrajani, J.~Ba, C.~Guestrin, P.~Liang, T.~B. Hashimoto, Alpacafarm: A simulation framework for methods that learn from human feedback, arXiv preprint arXiv:2305.14387 (2023).

\bibitem{align_with_prompts}
C.~Si, Z.~Gan, Z.~Yang, S.~Wang, J.~Wang, J.~Boyd-Graber, L.~Wang, Prompting gpt-3 to be reliable, arXiv preprint arXiv:2210.09150 (2022).

\bibitem{aling_with_prompts_2}
D.~Ganguli, A.~Askell, N.~Schiefer, T.~Liao, K.~Luko{\v{s}}i{\=u}t{\.e}, A.~Chen, A.~Goldie, A.~Mirhoseini, C.~Olsson, D.~Hernandez, et~al., The capacity for moral self-correction in large language models, arXiv preprint arXiv:2302.07459 (2023).

\bibitem{jailbroken}
A.~Wei, N.~Haghtalab, J.~Steinhardt, Jailbroken: How does llm safety training fail?, arXiv preprint arXiv:2307.02483 (2023).

\bibitem{red_team_lessons_learned}
D.~Ganguli, L.~Lovitt, J.~Kernion, A.~Askell, Y.~Bai, S.~Kadavath, B.~Mann, E.~Perez, N.~Schiefer, K.~Ndousse, et~al., Red teaming language models to reduce harms: Methods, scaling behaviors, and lessons learned, arXiv preprint arXiv:2209.07858 (2022).

\bibitem{red_team_explore}
S.~Casper, J.~Lin, J.~Kwon, G.~Culp, D.~Hadfield-Menell, Explore, establish, exploit: Red teaming language models from scratch, arXiv preprint arXiv:2306.09442 (2023).

\bibitem{red_team_language_models}
E.~Perez, S.~Huang, F.~Song, T.~Cai, R.~Ring, J.~Aslanides, A.~Glaese, N.~McAleese, G.~Irving, Red teaming language models with language models, arXiv preprint arXiv:2202.03286 (2022).

\bibitem{cont_learn}
T.~Scialom, T.~Chakrabarty, S.~Muresan, Fine-tuned language models are continual learners, in: Proceedings of the 2022 Conference on Empirical Methods in Natural Language Processing, 2022, pp. 6107--6122.

\bibitem{cont_learn_dont_stop_pretrain}
Z.~Shi, A.~Lipani, Don't stop pretraining? make prompt-based fine-tuning powerful learner, arXiv preprint arXiv:2305.01711 (2023).

\bibitem{sample_eff_inst_quick_learner}
H.~Gupta, S.~A. Sawant, S.~Mishra, M.~Nakamura, A.~Mitra, S.~Mashetty, C.~Baral, Instruction tuned models are quick learners, arXiv preprint arXiv:2306.05539 (2023).

\bibitem{sample_eff_maybe_less_data}
H.~Chen, Y.~Zhang, Q.~Zhang, H.~Yang, X.~Hu, X.~Ma, Y.~Yanggong, J.~Zhao, Maybe only 0.5\% data is needed: A preliminary exploration of low training data instruction tuning, arXiv preprint arXiv:2305.09246 (2023).

\bibitem{lima}
C.~Zhou, P.~Liu, P.~Xu, S.~Iyer, J.~Sun, Y.~Mao, X.~Ma, A.~Efrat, P.~Yu, L.~Yu, et~al., Lima: Less is more for alignment, arXiv preprint arXiv:2305.11206 (2023).

\bibitem{wt_lm_infinite}
C.~Han, Q.~Wang, W.~Xiong, Y.~Chen, H.~Ji, S.~Wang, Lm-infinite: Simple on-the-fly length generalization for large language models, arXiv preprint arXiv:2308.16137 (2023).

\bibitem{colt5}
J.~Ainslie, T.~Lei, M.~de~Jong, S.~Onta{\~n}{\'o}n, S.~Brahma, Y.~Zemlyanskiy, D.~Uthus, M.~Guo, J.~Lee-Thorp, Y.~Tay, et~al., Colt5: Faster long-range transformers with conditional computation, arXiv preprint arXiv:2303.09752 (2023).

\bibitem{longnet}
J.~Ding, S.~Ma, L.~Dong, X.~Zhang, S.~Huang, W.~Wang, F.~Wei, Longnet: Scaling transformers to 1,000,000,000 tokens, arXiv preprint arXiv:2307.02486 (2023).

\bibitem{longlora}
Y.~Chen, S.~Qian, H.~Tang, X.~Lai, Z.~Liu, S.~Han, J.~Jia, Longlora: Efficient fine-tuning of long-context large language models, arXiv preprint arXiv:2309.12307 (2023).

\bibitem{PCW}
N.~Ratner, Y.~Levine, Y.~Belinkov, O.~Ram, I.~Magar, O.~Abend, E.~Karpas, A.~Shashua, K.~Leyton-Brown, Y.~Shoham, Parallel context windows for large language models, in: Proceedings of the 61st Annual Meeting of the Association for Computational Linguistics (Volume 1: Long Papers), 2023, pp. 6383--6402.

\bibitem{mem_augmenting_long}
W.~Wang, L.~Dong, H.~Cheng, X.~Liu, X.~Yan, J.~Gao, F.~Wei, Augmenting language models with long-term memory, arXiv preprint arXiv:2306.07174 (2023).

\bibitem{mem_augmenting_long_conver}
X.~Xu, Z.~Gou, W.~Wu, Z.-Y. Niu, H.~Wu, H.~Wang, S.~Wang, Long time no see! open-domain conversation with long-term persona memory, arXiv preprint arXiv:2203.05797 (2022).

\bibitem{retro}
S.~Borgeaud, A.~Mensch, J.~Hoffmann, T.~Cai, E.~Rutherford, K.~Millican, G.~B. Van Den~Driessche, J.-B. Lespiau, B.~Damoc, A.~Clark, et~al., Improving language models by retrieving from trillions of tokens, in: International conference on machine learning, PMLR, 2022, pp. 2206--2240.

\bibitem{memorybank}
W.~Zhong, L.~Guo, Q.~Gao, Y.~Wang, Memorybank: Enhancing large language models with long-term memory, arXiv preprint arXiv:2305.10250 (2023).

\bibitem{reflexion}
N.~Shinn, F.~Cassano, B.~Labash, A.~Gopinath, K.~Narasimhan, S.~Yao, Reflexion: Language agents with verbal reinforcement learning, arXiv preprint arXiv:2303.11366 14 (2023).

\bibitem{chatdb}
C.~Hu, J.~Fu, C.~Du, S.~Luo, J.~Zhao, H.~Zhao, Chatdb: Augmenting llms with databases as their symbolic memory, arXiv preprint arXiv:2306.03901 (2023).

\bibitem{flare}
Z.~Jiang, F.~F. Xu, L.~Gao, Z.~Sun, Q.~Liu, J.~Dwivedi-Yu, Y.~Yang, J.~Callan, G.~Neubig, Active retrieval augmented generation, arXiv preprint arXiv:2305.06983 (2023).

\bibitem{in_context_ralm}
O.~Ram, Y.~Levine, I.~Dalmedigos, D.~Muhlgay, A.~Shashua, K.~Leyton-Brown, Y.~Shoham, In-context retrieval-augmented language models, arXiv preprint arXiv:2302.00083 (2023).

\bibitem{mot}
X.~Li, X.~Qiu, Mot: Pre-thinking and recalling enable chatgpt to self-improve with memory-of-thoughts, arXiv preprint arXiv:2305.05181 (2023).

\bibitem{rw_memory}
D.~Schuurmans, Memory augmented large language models are computationally universal, arXiv preprint arXiv:2301.04589 (2023).

\bibitem{retLLM}
A.~Modarressi, A.~Imani, M.~Fayyaz, H.~Sch{\"u}tze, Ret-llm: Towards a general read-write memory for large language models, arXiv preprint arXiv:2305.14322 (2023).

\bibitem{BM25}
S.~Robertson, H.~Zaragoza, et~al., The probabilistic relevance framework: Bm25 and beyond, Foundations and Trends{\textregistered} in Information Retrieval 3~(4) (2009) 333--389.

\bibitem{rationale_aug}
X.~Wang, J.~Wei, D.~Schuurmans, Q.~Le, E.~Chi, D.~Zhou, Rationale-augmented ensembles in language models, arXiv preprint arXiv:2207.00747 (2022).

\bibitem{repocoder}
F.~Zhang, B.~Chen, Y.~Zhang, J.~Liu, D.~Zan, Y.~Mao, J.-G. Lou, W.~Chen, Repocoder: Repository-level code completion through iterative retrieval and generation, arXiv preprint arXiv:2303.12570 (2023).

\bibitem{retro_study}
B.~Wang, W.~Ping, P.~Xu, L.~McAfee, Z.~Liu, M.~Shoeybi, Y.~Dong, O.~Kuchaiev, B.~Li, C.~Xiao, et~al., Shall we pretrain autoregressive language models with retrieval? a comprehensive study, arXiv preprint arXiv:2304.06762 (2023).

\bibitem{retrieve_icl_train}
L.~Wang, N.~Yang, F.~Wei, Learning to retrieve in-context examples for large language models, arXiv preprint arXiv:2307.07164 (2023).

\bibitem{retrieve_icl_GPT_3}
J.~Liu, D.~Shen, Y.~Zhang, B.~Dolan, L.~Carin, W.~Chen, What makes good in-context examples for gpt-$3 $?, arXiv preprint arXiv:2101.06804 (2021).

\bibitem{retrieve_prompts_learning}
O.~Rubin, J.~Herzig, J.~Berant, Learning to retrieve prompts for in-context learning, arXiv preprint arXiv:2112.08633 (2021).

\bibitem{replug}
W.~Shi, S.~Min, M.~Yasunaga, M.~Seo, R.~James, M.~Lewis, L.~Zettlemoyer, W.-t. Yih, Replug: Retrieval-augmented black-box language models, arXiv preprint arXiv:2301.12652 (2023).

\bibitem{rag_train_long}
O.~Rubin, J.~Berant, Long-range language modeling with self-retrieval, arXiv preprint arXiv:2306.13421 (2023).

\bibitem{realm}
K.~Guu, K.~Lee, Z.~Tung, P.~Pasupat, M.~Chang, Retrieval augmented language model pre-training, in: International conference on machine learning, PMLR, 2020, pp. 3929--3938.

\bibitem{fid}
S.~Hofst{\"a}tter, J.~Chen, K.~Raman, H.~Zamani, Fid-light: Efficient and effective retrieval-augmented text generation, in: Proceedings of the 46th International ACM SIGIR Conference on Research and Development in Information Retrieval, 2023, pp. 1437--1447.

\bibitem{internet_aug}
M.~Komeili, K.~Shuster, J.~Weston, Internet-augmented dialogue generation, arXiv preprint arXiv:2107.07566 (2021).

\bibitem{internet_aug_2}
A.~Lazaridou, E.~Gribovskaya, W.~Stokowiec, N.~Grigorev, Internet-augmented language models through few-shot prompting for open-domain question answering, arXiv preprint arXiv:2203.05115 (2022).

\bibitem{assistgpt}
D.~Gao, L.~Ji, L.~Zhou, K.~Q. Lin, J.~Chen, Z.~Fan, M.~Z. Shou, Assistgpt: A general multi-modal assistant that can plan, execute, inspect, and learn, arXiv preprint arXiv:2306.08640 (2023).

\bibitem{chameleon}
P.~Lu, B.~Peng, H.~Cheng, M.~Galley, K.-W. Chang, Y.~N. Wu, S.-C. Zhu, J.~Gao, Chameleon: Plug-and-play compositional reasoning with large language models, arXiv preprint arXiv:2304.09842 (2023).

\bibitem{ART}
B.~Paranjape, S.~Lundberg, S.~Singh, H.~Hajishirzi, L.~Zettlemoyer, M.~T. Ribeiro, Art: Automatic multi-step reasoning and tool-use for large language models, arXiv preprint arXiv:2303.09014 (2023).

\bibitem{tool_aug_doc}
C.-Y. Hsieh, S.-A. Chen, C.-L. Li, Y.~Fujii, A.~Ratner, C.-Y. Lee, R.~Krishna, T.~Pfister, Tool documentation enables zero-shot tool-usage with large language models, arXiv preprint arXiv:2308.00675 (2023).

\bibitem{RestGPT}
Y.~Song, W.~Xiong, D.~Zhu, C.~Li, K.~Wang, Y.~Tian, S.~Li, Restgpt: Connecting large language models with real-world applications via restful apis, arXiv preprint arXiv:2306.06624 (2023).

\bibitem{ToolkenGPT}
S.~Hao, T.~Liu, Z.~Wang, Z.~Hu, Toolkengpt: Augmenting frozen language models with massive tools via tool embeddings, arXiv preprint arXiv:2305.11554 (2023).

\bibitem{gorilla}
S.~G. Patil, T.~Zhang, X.~Wang, J.~E. Gonzalez, Gorilla: Large language model connected with massive apis, arXiv preprint arXiv:2305.15334 (2023).

\bibitem{tool_manipulation_cap}
Q.~Xu, F.~Hong, B.~Li, C.~Hu, Z.~Chen, J.~Zhang, On the tool manipulation capability of open-source large language models, arXiv preprint arXiv:2305.16504 (2023).

\bibitem{toolllm}
Y.~Qin, S.~Liang, Y.~Ye, K.~Zhu, L.~Yan, Y.~Lu, Y.~Lin, X.~Cong, X.~Tang, B.~Qian, et~al., Toolllm: Facilitating large language models to master 16000+ real-world apis, arXiv preprint arXiv:2307.16789 (2023).

\bibitem{hugginggpt}
Y.~Shen, K.~Song, X.~Tan, D.~Li, W.~Lu, Y.~Zhuang, Hugginggpt: Solving ai tasks with chatgpt and its friends in huggingface, arXiv preprint arXiv:2303.17580 (2023).

\bibitem{taskmatrix}
Y.~Liang, C.~Wu, T.~Song, W.~Wu, Y.~Xia, Y.~Liu, Y.~Ou, S.~Lu, L.~Ji, S.~Mao, et~al., Taskmatrix. ai: Completing tasks by connecting foundation models with millions of apis, arXiv preprint arXiv:2303.16434 (2023).

\bibitem{vipergpt}
D.~Sur{\'\i}s, S.~Menon, C.~Vondrick, Vipergpt: Visual inference via python execution for reasoning, arXiv preprint arXiv:2303.08128 (2023).

\bibitem{clippy}
A.~Maedche, S.~Morana, S.~Schacht, D.~Werth, J.~Krumeich, Advanced user assistance systems, Business \& Information Systems Engineering 58 (2016) 367--370.

\bibitem{deepblue}
M.~Campbell, A.~J. Hoane~Jr, F.-h. Hsu, Deep blue, Artificial intelligence 134~(1-2) (2002) 57--83.

\bibitem{metagpt}
S.~Hong, X.~Zheng, J.~Chen, Y.~Cheng, J.~Wang, C.~Zhang, Z.~Wang, S.~K.~S. Yau, Z.~Lin, L.~Zhou, et~al., Metagpt: Meta programming for multi-agent collaborative framework, arXiv preprint arXiv:2308.00352 (2023).

\bibitem{survey_agents}
Z.~Xi, W.~Chen, X.~Guo, W.~He, Y.~Ding, B.~Hong, M.~Zhang, J.~Wang, S.~Jin, E.~Zhou, et~al., The rise and potential of large language model based agents: A survey, arXiv preprint arXiv:2309.07864 (2023).

\bibitem{survey_agents_2}
L.~Wang, C.~Ma, X.~Feng, Z.~Zhang, H.~Yang, J.~Zhang, Z.~Chen, J.~Tang, X.~Chen, Y.~Lin, et~al., A survey on large language model based autonomous agents, arXiv preprint arXiv:2308.11432 (2023).

\bibitem{LLMs_zs_planners}
W.~Huang, P.~Abbeel, D.~Pathak, I.~Mordatch, Language models as zero-shot planners: Extracting actionable knowledge for embodied agents, in: International Conference on Machine Learning, PMLR, 2022, pp. 9118--9147.

\bibitem{RAP}
S.~Hao, Y.~Gu, H.~Ma, J.~J. Hong, Z.~Wang, D.~Z. Wang, Z.~Hu, Reasoning with language model is planning with world model, arXiv preprint arXiv:2305.14992 (2023).

\bibitem{retroformer}
W.~Yao, S.~Heinecke, J.~C. Niebles, Z.~Liu, Y.~Feng, L.~Xue, R.~Murthy, Z.~Chen, J.~Zhang, D.~Arpit, et~al., Retroformer: Retrospective large language agents with policy gradient optimization, arXiv preprint arXiv:2308.02151 (2023).

\bibitem{inner_monologue}
W.~Huang, F.~Xia, T.~Xiao, H.~Chan, J.~Liang, P.~Florence, A.~Zeng, J.~Tompson, I.~Mordatch, Y.~Chebotar, P.~Sermanet, T.~Jackson, N.~Brown, L.~Luu, S.~Levine, K.~Hausman, brian ichter, \href{https://openreview.net/forum?id=3R3Pz5i0tye}{Inner monologue: Embodied reasoning through planning with language models}, in: 6th Annual Conference on Robot Learning, 2022.
\newline\urlprefix\url{https://openreview.net/forum?id=3R3Pz5i0tye}

\bibitem{alphablock}
C.~Jin, W.~Tan, J.~Yang, B.~Liu, R.~Song, L.~Wang, J.~Fu, Alphablock: Embodied finetuning for vision-language reasoning in robot manipulation, arXiv preprint arXiv:2305.18898 (2023).

\bibitem{progprompt}
I.~Singh, V.~Blukis, A.~Mousavian, A.~Goyal, D.~Xu, J.~Tremblay, D.~Fox, J.~Thomason, A.~Garg, Progprompt: Generating situated robot task plans using large language models, in: 2023 IEEE International Conference on Robotics and Automation (ICRA), IEEE, 2023, pp. 11523--11530.

\bibitem{LLM_rewards}
W.~Yu, N.~Gileadi, C.~Fu, S.~Kirmani, K.-H. Lee, M.~G. Arenas, H.-T.~L. Chiang, T.~Erez, L.~Hasenclever, J.~Humplik, et~al., Language to rewards for robotic skill synthesis, arXiv preprint arXiv:2306.08647 (2023).

\bibitem{medagents}
X.~Tang, A.~Zou, Z.~Zhang, Y.~Zhao, X.~Zhang, A.~Cohan, M.~Gerstein, Medagents: Large language models as collaborators for zero-shot medical reasoning, arXiv preprint arXiv:2311.10537 (2023).

\bibitem{saycan}
A.~Brohan, Y.~Chebotar, C.~Finn, K.~Hausman, A.~Herzog, D.~Ho, J.~Ibarz, A.~Irpan, E.~Jang, R.~Julian, et~al., Do as i can, not as i say: Grounding language in robotic affordances, in: Conference on Robot Learning, PMLR, 2023, pp. 287--318.

\bibitem{ha2023scaling}
H.~Ha, P.~Florence, S.~Song, Scaling up and distilling down: Language-guided robot skill acquisition, arXiv preprint arXiv:2307.14535 (2023).

\bibitem{rajvanshi2023saynav}
A.~Rajvanshi, K.~Sikka, X.~Lin, B.~Lee, H.-P. Chiu, A.~Velasquez, Saynav: Grounding large language models for dynamic planning to navigation in new environments, arXiv preprint arXiv:2309.04077 (2023).

\bibitem{song2022llm}
C.~H. Song, J.~Wu, C.~Washington, B.~M. Sadler, W.-L. Chao, Y.~Su, Llm-planner: Few-shot grounded planning for embodied agents with large language models, arXiv preprint arXiv:2212.04088 (2022).

\bibitem{dorbala2023can}
V.~S. Dorbala, J.~F. Mullen~Jr, D.~Manocha, Can an embodied agent find your" cat-shaped mug"? llm-based zero-shot object navigation, arXiv preprint arXiv:2303.03480 (2023).

\bibitem{huang2023visual}
C.~Huang, O.~Mees, A.~Zeng, W.~Burgard, Visual language maps for robot navigation, in: 2023 IEEE International Conference on Robotics and Automation (ICRA), IEEE, 2023, pp. 10608--10615.

\bibitem{ding2023task}
Y.~Ding, X.~Zhang, C.~Paxton, S.~Zhang, Task and motion planning with large language models for object rearrangement, arXiv preprint arXiv:2303.06247 (2023).

\bibitem{Prompt_Tuning_2}
X.~Liu, Y.~Zheng, Z.~Du, M.~Ding, Y.~Qian, Z.~Yang, J.~Tang, Gpt understands, too, arXiv preprint arXiv:2103.10385 (2021).

\bibitem{revisiting_peft}
G.~Chen, F.~Liu, Z.~Meng, S.~Liang, Revisiting parameter-efficient tuning: Are we really there yet?, arXiv preprint arXiv:2202.07962 (2022).

\bibitem{adamix}
Y.~Wang, S.~Mukherjee, X.~Liu, J.~Gao, A.~H. Awadallah, J.~Gao, Adamix: Mixture-of-adapter for parameter-efficient tuning of large language models, arXiv preprint arXiv:2205.12410 1~(2) (2022) 4.

\bibitem{hu2021lora}
E.~J. Hu, Y.~Shen, P.~Wallis, Z.~Allen-Zhu, Y.~Li, S.~Wang, L.~Wang, W.~Chen, Lora: Low-rank adaptation of large language models, arXiv preprint arXiv:2106.09685 (2021).

\bibitem{p_tuning_v2}
X.~Liu, K.~Ji, Y.~Fu, W.~Tam, Z.~Du, Z.~Yang, J.~Tang, P-tuning: Prompt tuning can be comparable to fine-tuning across scales and tasks, in: Proceedings of the 60th Annual Meeting of the Association for Computational Linguistics (Volume 2: Short Papers), 2022, pp. 61--68.

\bibitem{progressive_prompts}
A.~Razdaibiedina, Y.~Mao, R.~Hou, M.~Khabsa, M.~Lewis, A.~Almahairi, Progressive prompts: Continual learning for language models, arXiv preprint arXiv:2301.12314 (2023).

\bibitem{adaptive_prefix}
Z.-R. Zhang, C.~Tan, H.~Xu, C.~Wang, J.~Huang, S.~Huang, Towards adaptive prefix tuning for parameter-efficient language model fine-tuning, arXiv preprint arXiv:2305.15212 (2023).

\bibitem{bitfit}
E.~B. Zaken, S.~Ravfogel, Y.~Goldberg, Bitfit: Simple parameter-efficient fine-tuning for transformer-based masked language-models, arXiv preprint arXiv:2106.10199 (2021).

\bibitem{8_bit_LLM}
T.~Dettmers, M.~Lewis, Y.~Belkada, L.~Zettlemoyer, Llm. int8 (): 8-bit matrix multiplication for transformers at scale, arXiv preprint arXiv:2208.07339 (2022).

\bibitem{OPTQ}
E.~Frantar, S.~Ashkboos, T.~Hoefler, D.~Alistarh, Gptq: Accurate post-training quantization for generative pre-trained transformers, arXiv preprint arXiv:2210.17323 (2022).

\bibitem{outlier_suppression}
X.~Wei, Y.~Zhang, Y.~Li, X.~Zhang, R.~Gong, J.~Guo, X.~Liu, Outlier suppression+: Accurate quantization of large language models by equivalent and optimal shifting and scaling, arXiv preprint arXiv:2304.09145 (2023).

\bibitem{OBC}
E.~Frantar, D.~Alistarh, Optimal brain compression: A framework for accurate post-training quantization and pruning, Advances in Neural Information Processing Systems 35 (2022) 4475--4488.

\bibitem{OWQ}
C.~Lee, J.~Jin, T.~Kim, H.~Kim, E.~Park, Owq: Lessons learned from activation outliers for weight quantization in large language models, arXiv preprint arXiv:2306.02272 (2023).

\bibitem{alphatuning}
S.~J. Kwon, J.~Kim, J.~Bae, K.~M. Yoo, J.-H. Kim, B.~Park, B.~Kim, J.-W. Ha, N.~Sung, D.~Lee, Alphatuning: Quantization-aware parameter-efficient adaptation of large-scale pre-trained language models, arXiv preprint arXiv:2210.03858 (2022).

\bibitem{qlora}
T.~Dettmers, A.~Pagnoni, A.~Holtzman, L.~Zettlemoyer, Qlora: Efficient finetuning of quantized llms, arXiv preprint arXiv:2305.14314 (2023).

\bibitem{llm_qat}
Z.~Liu, B.~Oguz, C.~Zhao, E.~Chang, P.~Stock, Y.~Mehdad, Y.~Shi, R.~Krishnamoorthi, V.~Chandra, Llm-qat: Data-free quantization aware training for large language models, arXiv preprint arXiv:2305.17888 (2023).

\bibitem{BCQ}
Y.~Guo, A.~Yao, H.~Zhao, Y.~Chen, Network sketching: Exploiting binary structure in deep cnns, in: Proceedings of the IEEE Conference on Computer Vision and Pattern Recognition, 2017, pp. 5955--5963.

\bibitem{PEQA}
J.~Kim, J.~H. Lee, S.~Kim, J.~Park, K.~M. Yoo, S.~J. Kwon, D.~Lee, Memory-efficient fine-tuning of compressed large language models via sub-4-bit integer quantization, arXiv preprint arXiv:2305.14152 (2023).

\bibitem{wanda}
M.~Sun, Z.~Liu, A.~Bair, J.~Z. Kolter, A simple and effective pruning approach for large language models, arXiv preprint arXiv:2306.11695 (2023).

\bibitem{structured_p_llms}
Z.~Wang, J.~Wohlwend, T.~Lei, Structured pruning of large language models, arXiv preprint arXiv:1910.04732 (2019).

\bibitem{OWL}
L.~Yin, Y.~Wu, Z.~Zhang, C.-Y. Hsieh, Y.~Wang, Y.~Jia, M.~Pechenizkiy, Y.~Liang, Z.~Wang, S.~Liu, Outlier weighed layerwise sparsity (owl): A missing secret sauce for pruning llms to high sparsity, arXiv preprint arXiv:2310.05175 (2023).

\bibitem{SIMPLE}
C.~Tao, L.~Hou, H.~Bai, J.~Wei, X.~Jiang, Q.~Liu, P.~Luo, N.~Wong, Structured pruning for efficient generative pre-trained language models, in: Findings of the Association for Computational Linguistics: ACL 2023, 2023, pp. 10880--10895.

\bibitem{alayrac2022flamingo}
J.-B. Alayrac, J.~Donahue, P.~Luc, A.~Miech, I.~Barr, Y.~Hasson, K.~Lenc, A.~Mensch, K.~Millican, M.~Reynolds, et~al., Flamingo: a visual language model for few-shot learning, Advances in Neural Information Processing Systems 35 (2022) 23716--23736.

\bibitem{li2023blip}
J.~Li, D.~Li, S.~Savarese, S.~Hoi, Blip-2: Bootstrapping language-image pre-training with frozen image encoders and large language models, arXiv preprint arXiv:2301.12597 (2023).

\bibitem{liu2023visual}
H.~Liu, C.~Li, Q.~Wu, Y.~J. Lee, Visual instruction tuning, arXiv preprint arXiv:2304.08485 (2023).

\bibitem{li2023videochat}
K.~Li, Y.~He, Y.~Wang, Y.~Li, W.~Wang, P.~Luo, Y.~Wang, L.~Wang, Y.~Qiao, Videochat: Chat-centric video understanding, arXiv preprint arXiv:2305.06355 (2023).

\bibitem{maaz2023video}
M.~Maaz, H.~Rasheed, S.~Khan, F.~S. Khan, Video-chatgpt: Towards detailed video understanding via large vision and language models, arXiv preprint arXiv:2306.05424 (2023).

\bibitem{zhang2023video}
H.~Zhang, X.~Li, L.~Bing, Video-llama: An instruction-tuned audio-visual language model for video understanding, arXiv preprint arXiv:2306.02858 (2023).

\bibitem{mei2023wavcaps}
X.~Mei, C.~Meng, H.~Liu, Q.~Kong, T.~Ko, C.~Zhao, M.~D. Plumbley, Y.~Zou, W.~Wang, Wavcaps: A chatgpt-assisted weakly-labelled audio captioning dataset for audio-language multimodal research, arXiv preprint arXiv:2303.17395 (2023).

\bibitem{lyu2023macaw}
C.~Lyu, M.~Wu, L.~Wang, X.~Huang, B.~Liu, Z.~Du, S.~Shi, Z.~Tu, Macaw-llm: Multi-modal language modeling with image, audio, video, and text integration, arXiv preprint arXiv:2306.09093 (2023).

\bibitem{zhu2023minigpt}
D.~Zhu, J.~Chen, X.~Shen, X.~Li, M.~Elhoseiny, Minigpt-4: Enhancing vision-language understanding with advanced large language models, arXiv preprint arXiv:2304.10592 (2023).

\bibitem{dosovitskiy2020image}
A.~Dosovitskiy, L.~Beyer, A.~Kolesnikov, D.~Weissenborn, X.~Zhai, T.~Unterthiner, M.~Dehghani, M.~Minderer, G.~Heigold, S.~Gelly, et~al., An image is worth 16x16 words: Transformers for image recognition at scale, arXiv preprint arXiv:2010.11929 (2020).

\bibitem{dai2023instructblip}
W.~Dai, J.~Li, D.~Li, A.~M.~H. Tiong, J.~Zhao, W.~Wang, B.~Li, P.~Fung, S.~Hoi, Instructblip: Towards general-purpose vision-language models with instruction tuning, arXiv preprint arXiv:2305.06500 (2023).

\bibitem{xu2022multiinstruct}
Z.~Xu, Y.~Shen, L.~Huang, Multiinstruct: Improving multi-modal zero-shot learning via instruction tuning, arXiv preprint arXiv:2212.10773 (2022).

\bibitem{zhao2023chatbridge}
Z.~Zhao, L.~Guo, T.~Yue, S.~Chen, S.~Shao, X.~Zhu, Z.~Yuan, J.~Liu, Chatbridge: Bridging modalities with large language model as a language catalyst, arXiv preprint arXiv:2305.16103 (2023).

\bibitem{li2023m}
L.~Li, Y.~Yin, S.~Li, L.~Chen, P.~Wang, S.~Ren, M.~Li, Y.~Yang, J.~Xu, X.~Sun, et~al., M3 it: A large-scale dataset towards multi-modal multilingual instruction tuning, arXiv preprint arXiv:2306.04387 (2023).

\bibitem{pi2023detgpt}
R.~Pi, J.~Gao, S.~Diao, R.~Pan, H.~Dong, J.~Zhang, L.~Yao, J.~Han, H.~Xu, L.~K.~T. Zhang, Detgpt: Detect what you need via reasoning, arXiv preprint arXiv:2305.14167 (2023).

\bibitem{luo2023cheap}
G.~Luo, Y.~Zhou, T.~Ren, S.~Chen, X.~Sun, R.~Ji, Cheap and quick: Efficient vision-language instruction tuning for large language models, arXiv preprint arXiv:2305.15023 (2023).

\bibitem{zhang2023llama}
R.~Zhang, J.~Han, A.~Zhou, X.~Hu, S.~Yan, P.~Lu, H.~Li, P.~Gao, Y.~Qiao, Llama-adapter: Efficient fine-tuning of language models with zero-init attention, arXiv preprint arXiv:2303.16199 (2023).

\bibitem{radford2023robust}
A.~Radford, J.~W. Kim, T.~Xu, G.~Brockman, C.~McLeavey, I.~Sutskever, Robust speech recognition via large-scale weak supervision, in: International Conference on Machine Learning, PMLR, 2023, pp. 28492--28518.

\bibitem{zhang2023multimodal}
Z.~Zhang, A.~Zhang, M.~Li, H.~Zhao, G.~Karypis, A.~Smola, Multimodal chain-of-thought reasoning in language models, arXiv preprint arXiv:2302.00923 (2023).

\bibitem{ge2023chain}
J.~Ge, H.~Luo, S.~Qian, Y.~Gan, J.~Fu, S.~Zhan, Chain of thought prompt tuning in vision language models, arXiv preprint arXiv:2304.07919 (2023).

\bibitem{wu2023visual}
C.~Wu, S.~Yin, W.~Qi, X.~Wang, Z.~Tang, N.~Duan, Visual chatgpt: Talking, drawing and editing with visual foundation models, arXiv preprint arXiv:2303.04671 (2023).

\bibitem{yang2023mm}
Z.~Yang, L.~Li, J.~Wang, K.~Lin, E.~Azarnasab, F.~Ahmed, Z.~Liu, C.~Liu, M.~Zeng, L.~Wang, Mm-react: Prompting chatgpt for multimodal reasoning and action, arXiv preprint arXiv:2303.11381 (2023).

\bibitem{wang2023caption}
T.~Wang, J.~Zhang, J.~Fei, Y.~Ge, H.~Zheng, Y.~Tang, Z.~Li, M.~Gao, S.~Zhao, Y.~Shan, et~al., Caption anything: Interactive image description with diverse multimodal controls, arXiv preprint arXiv:2305.02677 (2023).

\bibitem{zhu2022pointclip}
X.~Zhu, R.~Zhang, B.~He, Z.~Zeng, S.~Zhang, P.~Gao, Pointclip v2: Adapting clip for powerful 3d open-world learning, arXiv preprint arXiv:2211.11682 (2022).

\bibitem{gupta2023visual}
T.~Gupta, A.~Kembhavi, Visual programming: Compositional visual reasoning without training, in: Proceedings of the IEEE/CVF Conference on Computer Vision and Pattern Recognition, 2023, pp. 14953--14962.

\bibitem{gao2019dynamic}
P.~Gao, Z.~Jiang, H.~You, P.~Lu, S.~C. Hoi, X.~Wang, H.~Li, Dynamic fusion with intra-and inter-modality attention flow for visual question answering, in: Proceedings of the IEEE/CVF conference on computer vision and pattern recognition, 2019, pp. 6639--6648.

\bibitem{yu2019deep}
Z.~Yu, J.~Yu, Y.~Cui, D.~Tao, Q.~Tian, Deep modular co-attention networks for visual question answering, in: Proceedings of the IEEE/CVF conference on computer vision and pattern recognition, 2019, pp. 6281--6290.

\bibitem{you2023idealgpt}
H.~You, R.~Sun, Z.~Wang, L.~Chen, G.~Wang, H.~A. Ayyubi, K.-W. Chang, S.-F. Chang, Idealgpt: Iteratively decomposing vision and language reasoning via large language models, arXiv preprint arXiv:2305.14985 (2023).

\bibitem{zhang2023prompt}
R.~Zhang, X.~Hu, B.~Li, S.~Huang, H.~Deng, Y.~Qiao, P.~Gao, H.~Li, Prompt, generate, then cache: Cascade of foundation models makes strong few-shot learners, in: Proceedings of the IEEE/CVF Conference on Computer Vision and Pattern Recognition, 2023, pp. 15211--15222.

\bibitem{small_init}
T.~Q. Nguyen, J.~Salazar, Transformers without tears: Improving the normalization of self-attention, CoRR abs/1910.05895 (2019).

\bibitem{roberta}
Y.~Liu, M.~Ott, N.~Goyal, J.~Du, M.~Joshi, D.~Chen, O.~Levy, M.~Lewis, L.~Zettlemoyer, V.~Stoyanov, Roberta: A robustly optimized bert pretraining approach, arXiv preprint arXiv:1907.11692 (2019).

\bibitem{koala}
X.~Geng, A.~Gudibande, H.~Liu, E.~Wallace, P.~Abbeel, S.~Levine, D.~Song, \href{https://bair.berkeley.edu/blog/2023/04/03/koala/}{Koala: A dialogue model for academic research}, Blog post (April 2023).
\newline\urlprefix\url{https://bair.berkeley.edu/blog/2023/04/03/koala/}

\bibitem{gao2020pile}
L.~Gao, S.~Biderman, S.~Black, L.~Golding, T.~Hoppe, C.~Foster, J.~Phang, H.~He, A.~Thite, N.~Nabeshima, et~al., The pile: An 800gb dataset of diverse text for language modeling, arXiv preprint arXiv:2101.00027 (2020).

\bibitem{laurenccon2022bigscience}
H.~Lauren{\c{c}}on, L.~Saulnier, T.~Wang, C.~Akiki, A.~Villanova~del Moral, T.~Le~Scao, L.~Von~Werra, C.~Mou, E.~Gonz{\'a}lez~Ponferrada, H.~Nguyen, et~al., The bigscience roots corpus: A 1.6 tb composite multilingual dataset, Advances in Neural Information Processing Systems 35 (2022) 31809--31826.

\bibitem{Wikipedia}
\href{https://en.wikipedia.org/wiki/Main_Page}{Wikipedia}.
\newline\urlprefix\url{https://en.wikipedia.org/wiki/Main_Page}

\bibitem{redpajama}
{Together Computer}, \href{https://github.com/togethercomputer/RedPajama-Data}{Redpajama: An open source recipe to reproduce llama training dataset} (Apr. 2023).
\newline\urlprefix\url{https://github.com/togethercomputer/RedPajama-Data}

\bibitem{unnatural_inst}
O.~Honovich, T.~Scialom, O.~Levy, T.~Schick, Unnatural instructions: Tuning language models with (almost) no human labor, arXiv preprint arXiv:2212.09689 (2022).

\bibitem{hh_rlhf}
Y.~Bai, A.~Jones, K.~Ndousse, A.~Askell, A.~Chen, N.~DasSarma, D.~Drain, S.~Fort, D.~Ganguli, T.~Henighan, et~al., Training a helpful and harmless assistant with reinforcement learning from human feedback, arXiv preprint arXiv:2204.05862 (2022).

\bibitem{hendrycks2020measuring}
D.~Hendrycks, C.~Burns, S.~Basart, A.~Zou, M.~Mazeika, D.~Song, J.~Steinhardt, Measuring massive multitask language understanding, arXiv preprint arXiv:2009.03300 (2020).

\bibitem{srivastava2022beyond}
A.~Srivastava, A.~Rastogi, A.~Rao, A.~A.~M. Shoeb, A.~Abid, A.~Fisch, A.~R. Brown, A.~Santoro, A.~Gupta, A.~Garriga-Alonso, et~al., Beyond the imitation game: Quantifying and extrapolating the capabilities of language models, arXiv preprint arXiv:2206.04615 (2022).

\bibitem{wang2018glue}
A.~Wang, A.~Singh, J.~Michael, F.~Hill, O.~Levy, S.~R. Bowman, Glue: A multi-task benchmark and analysis platform for natural language understanding, arXiv preprint arXiv:1804.07461 (2018).

\bibitem{yao2021cuge}
Y.~Yao, Q.~Dong, J.~Guan, B.~Cao, Z.~Zhang, C.~Xiao, X.~Wang, F.~Qi, J.~Bao, J.~Nie, et~al., Cuge: A chinese language understanding and generation evaluation benchmark, arXiv preprint arXiv:2112.13610 (2021).

\bibitem{xu2020clue}
L.~Xu, H.~Hu, X.~Zhang, L.~Li, C.~Cao, Y.~Li, Y.~Xu, K.~Sun, D.~Yu, C.~Yu, et~al., Clue: A chinese language understanding evaluation benchmark, arXiv preprint arXiv:2004.05986 (2020).

\bibitem{xu2021fewclue}
L.~Xu, X.~Lu, C.~Yuan, X.~Zhang, H.~Xu, H.~Yuan, G.~Wei, X.~Pan, X.~Tian, L.~Qin, et~al., Fewclue: A chinese few-shot learning evaluation benchmark, arXiv preprint arXiv:2107.07498 (2021).

\bibitem{smith2020can}
E.~M. Smith, M.~Williamson, K.~Shuster, J.~Weston, Y.-L. Boureau, Can you put it all together: Evaluating conversational agents' ability to blend skills, arXiv preprint arXiv:2004.08449 (2020).

\bibitem{HELM}
P.~Liang, R.~Bommasani, T.~Lee, D.~Tsipras, D.~Soylu, M.~Yasunaga, Y.~Zhang, D.~Narayanan, Y.~Wu, A.~Kumar, et~al., Holistic evaluation of language models, arXiv preprint arXiv:2211.09110 (2022).

\bibitem{park2021klue}
S.~Park, J.~Moon, S.~Kim, W.~I. Cho, J.~Han, J.~Park, C.~Song, J.~Kim, Y.~Song, T.~Oh, et~al., Klue: Korean language understanding evaluation, arXiv preprint arXiv:2105.09680 (2021).

\bibitem{reddy2019coqa}
S.~Reddy, D.~Chen, C.~D. Manning, Coqa: A conversational question answering challenge, Transactions of the Association for Computational Linguistics 7 (2019) 249--266.

\bibitem{pilehvar2018wic}
M.~T. Pilehvar, J.~Camacho-Collados, Wic: 10,000 example pairs for evaluating context-sensitive representations, arXiv preprint arXiv:1808.09121 6 (2018).

\bibitem{merity2016pointer}
S.~Merity, C.~Xiong, J.~Bradbury, R.~Socher, Pointer sentinel mixture models, arXiv preprint arXiv:1609.07843 (2016).

\bibitem{rae2019compressive}
J.~W. Rae, A.~Potapenko, S.~M. Jayakumar, T.~P. Lillicrap, Compressive transformers for long-range sequence modelling, arXiv preprint arXiv:1911.05507 (2019).

\bibitem{liu2018lcqmc}
X.~Liu, Q.~Chen, C.~Deng, H.~Zeng, J.~Chen, D.~Li, B.~Tang, Lcqmc: A large-scale chinese question matching corpus, in: Proceedings of the 27th international conference on computational linguistics, 2018, pp. 1952--1962.

\bibitem{QQP}
S.~Iyer, N.~Dandekar, K.~Csernai, First quora dataset release: Question pairs, \url{https://quoradata.quora.com/First-Quora-Dataset-Release-Question-Pairs}.

\bibitem{rudinger2018gender}
R.~Rudinger, J.~Naradowsky, B.~Leonard, B.~Van~Durme, Gender bias in coreference resolution, arXiv preprint arXiv:1804.09301 (2018).

\bibitem{de2019commitmentbank}
M.-C. De~Marneffe, M.~Simons, J.~Tonhauser, The commitmentbank: Investigating projection in naturally occurring discourse, in: proceedings of Sinn und Bedeutung, Vol.~23, 2019, pp. 107--124.

\bibitem{li2019chinese}
Z.~Li, N.~Ding, Z.~Liu, H.~Zheng, Y.~Shen, Chinese relation extraction with multi-grained information and external linguistic knowledge, in: Proceedings of the 57th Annual Meeting of the Association for Computational Linguistics, 2019, pp. 4377--4386.

\bibitem{xu2017discourse}
J.~Xu, J.~Wen, X.~Sun, Q.~Su, A discourse-level named entity recognition and relation extraction dataset for chinese literature text, arXiv preprint arXiv:1711.07010 (2017).

\bibitem{chen2018bq}
J.~Chen, Q.~Chen, X.~Liu, H.~Yang, D.~Lu, B.~Tang, The bq corpus: A large-scale domain-specific chinese corpus for sentence semantic equivalence identification, in: Proceedings of the 2018 conference on empirical methods in natural language processing, 2018, pp. 4946--4951.

\bibitem{liu2018matching}
B.~Liu, D.~Niu, H.~Wei, J.~Lin, Y.~He, K.~Lai, Y.~Xu, Matching article pairs with graphical decomposition and convolutions, arXiv preprint arXiv:1802.07459 (2018).

\bibitem{li2016dataset}
P.~Li, W.~Li, Z.~He, X.~Wang, Y.~Cao, J.~Zhou, W.~Xu, Dataset and neural recurrent sequence labeling model for open-domain factoid question answering, arXiv preprint arXiv:1607.06275 (2016).

\bibitem{peng2015named}
N.~Peng, M.~Dredze, Named entity recognition for chinese social media with jointly trained embeddings, in: Proceedings of the 2015 conference on empirical methods in natural language processing, 2015, pp. 548--554.

\bibitem{ling2017program}
W.~Ling, D.~Yogatama, C.~Dyer, P.~Blunsom, Program induction by rationale generation: Learning to solve and explain algebraic word problems, arXiv preprint arXiv:1705.04146 (2017).

\bibitem{weischedel2011ontonotes}
R.~Weischedel, S.~Pradhan, L.~Ramshaw, M.~Palmer, N.~Xue, M.~Marcus, A.~Taylor, C.~Greenberg, E.~Hovy, R.~Belvin, et~al., Ontonotes release 4.0, LDC2011T03, Philadelphia, Penn.: Linguistic Data Consortium (2011).

\bibitem{vilares2019head}
D.~Vilares, C.~G{\'o}mez-Rodr{\'\i}guez, Head-qa: A healthcare dataset for complex reasoning, arXiv preprint arXiv:1906.04701 (2019).

\bibitem{blodgett2016demographic}
S.~L. Blodgett, L.~Green, B.~O'Connor, Demographic dialectal variation in social media: A case study of african-american english, arXiv preprint arXiv:1608.08868 (2016).

\bibitem{mostafazadeh2016corpus}
N.~Mostafazadeh, N.~Chambers, X.~He, D.~Parikh, D.~Batra, L.~Vanderwende, P.~Kohli, J.~Allen, A corpus and evaluation framework for deeper understanding of commonsense stories, arXiv preprint arXiv:1604.01696 (2016).

\bibitem{paperno2016lambada}
D.~Paperno, G.~Kruszewski, A.~Lazaridou, Q.~N. Pham, R.~Bernardi, S.~Pezzelle, M.~Baroni, G.~Boleda, R.~Fern{\'a}ndez, The lambada dataset: Word prediction requiring a broad discourse context, arXiv preprint arXiv:1606.06031 (2016).

\bibitem{hu2015lcsts}
B.~Hu, Q.~Chen, F.~Zhu, Lcsts: A large scale chinese short text summarization dataset, arXiv preprint arXiv:1506.05865 (2015).

\bibitem{shao2019long}
Z.~Shao, M.~Huang, J.~Wen, W.~Xu, X.~Zhu, Long and diverse text generation with planning-based hierarchical variational model, arXiv preprint arXiv:1908.06605 (2019).

\bibitem{novikova2017e2e}
J.~Novikova, O.~Du{\v{s}}ek, V.~Rieser, The e2e dataset: New challenges for end-to-end generation, arXiv preprint arXiv:1706.09254 (2017).

\bibitem{zheng2019chid}
C.~Zheng, M.~Huang, A.~Sun, Chid: A large-scale chinese idiom dataset for cloze test, arXiv preprint arXiv:1906.01265 (2019).

\bibitem{bisk2020piqa}
Y.~Bisk, R.~Zellers, J.~Gao, Y.~Choi, et~al., Piqa: Reasoning about physical commonsense in natural language, in: Proceedings of the AAAI conference on artificial intelligence, Vol.~34, 2020, pp. 7432--7439.

\bibitem{joshi2017triviaqa}
M.~Joshi, E.~Choi, D.~S. Weld, L.~Zettlemoyer, Triviaqa: A large scale distantly supervised challenge dataset for reading comprehension, arXiv preprint arXiv:1705.03551 (2017).

\bibitem{clark2018think}
P.~Clark, I.~Cowhey, O.~Etzioni, T.~Khot, A.~Sabharwal, C.~Schoenick, O.~Tafjord, Think you have solved question answering? try arc, the ai2 reasoning challenge, arXiv preprint arXiv:1803.05457 (2018).

\bibitem{aroca2021prost}
S.~Aroca-Ouellette, C.~Paik, A.~Roncone, K.~Kann, Prost: Physical reasoning of objects through space and time, arXiv preprint arXiv:2106.03634 (2021).

\bibitem{mihaylov2018can}
T.~Mihaylov, P.~Clark, T.~Khot, A.~Sabharwal, Can a suit of armor conduct electricity? a new dataset for open book question answering, arXiv preprint arXiv:1809.02789 (2018).

\bibitem{ferreira20202020}
T.~C. Ferreira, C.~Gardent, N.~Ilinykh, C.~Van Der~Lee, S.~Mille, D.~Moussallem, A.~Shimorina, The 2020 bilingual, bi-directional webnlg+ shared task overview and evaluation results (webnlg+ 2020), in: Proceedings of the 3rd International Workshop on Natural Language Generation from the Semantic Web (WebNLG+), 2020.

\bibitem{xu2021blow}
C.~Xu, W.~Zhou, T.~Ge, K.~Xu, J.~McAuley, F.~Wei, Blow the dog whistle: A chinese dataset for cant understanding with common sense and world knowledge, arXiv preprint arXiv:2104.02704 (2021).

\bibitem{lai2017race}
G.~Lai, Q.~Xie, H.~Liu, Y.~Yang, E.~Hovy, Race: Large-scale reading comprehension dataset from examinations, arXiv preprint arXiv:1704.04683 (2017).

\bibitem{choi2018quac}
E.~Choi, H.~He, M.~Iyyer, M.~Yatskar, W.-t. Yih, Y.~Choi, P.~Liang, L.~Zettlemoyer, Quac: Question answering in context, arXiv preprint arXiv:1808.07036 (2018).

\bibitem{geva2021did}
M.~Geva, D.~Khashabi, E.~Segal, T.~Khot, D.~Roth, J.~Berant, Did aristotle use a laptop? a question answering benchmark with implicit reasoning strategies, Transactions of the Association for Computational Linguistics 9 (2021) 346--361.

\bibitem{boyd2012besting}
J.~Boyd-Graber, B.~Satinoff, H.~He, H.~Daum{\'e}~III, Besting the quiz master: Crowdsourcing incremental classification games, in: Proceedings of the 2012 joint conference on empirical methods in natural language processing and computational natural language learning, 2012, pp. 1290--1301.

\bibitem{zhang2017chinese}
S.~Zhang, X.~Zhang, H.~Wang, J.~Cheng, P.~Li, Z.~Ding, Chinese medical question answer matching using end-to-end character-level multi-scale cnns, Applied Sciences 7~(8) (2017) 767.

\bibitem{zhang2018multi}
S.~Zhang, X.~Zhang, H.~Wang, L.~Guo, S.~Liu, Multi-scale attentive interaction networks for chinese medical question answer selection, IEEE Access 6 (2018) 74061--74071.

\bibitem{xu2020matinf}
C.~Xu, J.~Pei, H.~Wu, Y.~Liu, C.~Li, Matinf: A jointly labeled large-scale dataset for classification, question answering and summarization, arXiv preprint arXiv:2004.12302 (2020).

\bibitem{sakaguchi2021winogrande}
K.~Sakaguchi, R.~L. Bras, C.~Bhagavatula, Y.~Choi, Winogrande: An adversarial winograd schema challenge at scale, Communications of the ACM 64~(9) (2021) 99--106.

\bibitem{zellers2019hellaswag}
R.~Zellers, A.~Holtzman, Y.~Bisk, A.~Farhadi, Y.~Choi, Hellaswag: Can a machine really finish your sentence?, arXiv preprint arXiv:1905.07830 (2019).

\bibitem{roemmele2011choice}
M.~Roemmele, C.~A. Bejan, A.~S. Gordon, Choice of plausible alternatives: An evaluation of commonsense causal reasoning., in: AAAI spring symposium: logical formalizations of commonsense reasoning, 2011, pp. 90--95.

\bibitem{levesque2012winograd}
H.~Levesque, E.~Davis, L.~Morgenstern, The winograd schema challenge, in: Thirteenth international conference on the principles of knowledge representation and reasoning, 2012.

\bibitem{talmor2018commonsenseqa}
A.~Talmor, J.~Herzig, N.~Lourie, J.~Berant, Commonsenseqa: A question answering challenge targeting commonsense knowledge, arXiv preprint arXiv:1811.00937 (2018).

\bibitem{sap2019socialiqa}
M.~Sap, H.~Rashkin, D.~Chen, R.~LeBras, Y.~Choi, Socialiqa: Commonsense reasoning about social interactions, arXiv preprint arXiv:1904.09728 (2019).

\bibitem{sun2020investigating}
K.~Sun, D.~Yu, D.~Yu, C.~Cardie, Investigating prior knowledge for challenging chinese machine reading comprehension, Transactions of the Association for Computational Linguistics 8 (2020) 141--155.

\bibitem{zhang2018record}
S.~Zhang, X.~Liu, J.~Liu, J.~Gao, K.~Duh, B.~Van~Durme, Record: Bridging the gap between human and machine commonsense reading comprehension, arXiv preprint arXiv:1810.12885 (2018).

\bibitem{rajpurkar2016squad}
P.~Rajpurkar, J.~Zhang, K.~Lopyrev, P.~Liang, Squad: 100,000+ questions for machine comprehension of text, arXiv preprint arXiv:1606.05250 (2016).

\bibitem{clark2019boolq}
C.~Clark, K.~Lee, M.-W. Chang, T.~Kwiatkowski, M.~Collins, K.~Toutanova, Boolq: Exploring the surprising difficulty of natural yes/no questions, arXiv preprint arXiv:1905.10044 (2019).

\bibitem{rajpurkar2018know}
P.~Rajpurkar, R.~Jia, P.~Liang, Know what you don't know: Unanswerable questions for squad, arXiv preprint arXiv:1806.03822 (2018).

\bibitem{dua2019drop}
D.~Dua, Y.~Wang, P.~Dasigi, G.~Stanovsky, S.~Singh, M.~Gardner, Drop: A reading comprehension benchmark requiring discrete reasoning over paragraphs, arXiv preprint arXiv:1903.00161 (2019).

\bibitem{dagan2005pascal}
I.~Dagan, O.~Glickman, B.~Magnini, The pascal recognising textual entailment challenge, in: Machine learning challenges workshop, Springer, 2005, pp. 177--190.

\bibitem{chang2022webqa}
Y.~Chang, M.~Narang, H.~Suzuki, G.~Cao, J.~Gao, Y.~Bisk, Webqa: Multihop and multimodal qa, in: Proceedings of the IEEE/CVF Conference on Computer Vision and Pattern Recognition, 2022, pp. 16495--16504.

\bibitem{cui2017cmrc17}
Y.~Cui, T.~Liu, Z.~Chen, W.~Ma, S.~Wang, G.~Hu, Dataset for the first evaluation on chinese machine reading comprehension, arXiv preprint arXiv:1709.08299 (2017).

\bibitem{cui2018span}
Y.~Cui, T.~Liu, W.~Che, L.~Xiao, Z.~Chen, W.~Ma, S.~Wang, G.~Hu, A span-extraction dataset for chinese machine reading comprehension, arXiv preprint arXiv:1810.07366 (2018).

\bibitem{cui2020sentence}
Y.~Cui, T.~Liu, Z.~Yang, Z.~Chen, W.~Ma, W.~Che, S.~Wang, G.~Hu, A sentence cloze dataset for chinese machine reading comprehension, arXiv preprint arXiv:2004.03116 (2020).

\bibitem{li2018character}
Y.~Li, T.~Liu, D.~Li, Q.~Li, J.~Shi, Y.~Wang, Character-based bilstm-crf incorporating pos and dictionaries for chinese opinion target extraction, in: Asian Conference on Machine Learning, PMLR, 2018, pp. 518--533.

\bibitem{khashabi2018looking}
D.~Khashabi, S.~Chaturvedi, M.~Roth, S.~Upadhyay, D.~Roth, Looking beyond the surface: A challenge set for reading comprehension over multiple sentences, in: Proceedings of the 2018 Conference of the North American Chapter of the Association for Computational Linguistics: Human Language Technologies, Volume 1 (Long Papers), 2018, pp. 252--262.

\bibitem{kwiatkowski2019natural}
T.~Kwiatkowski, J.~Palomaki, O.~Redfield, M.~Collins, A.~Parikh, C.~Alberti, D.~Epstein, I.~Polosukhin, J.~Devlin, K.~Lee, et~al., Natural questions: a benchmark for question answering research, Transactions of the Association for Computational Linguistics 7 (2019) 453--466.

\bibitem{shao2018drcd}
C.~C. Shao, T.~Liu, Y.~Lai, Y.~Tseng, S.~Tsai, Drcd: A chinese machine reading comprehension dataset, arXiv preprint arXiv:1806.00920 (2018).

\bibitem{he2017dureader}
W.~He, K.~Liu, J.~Liu, Y.~Lyu, S.~Zhao, X.~Xiao, Y.~Liu, Y.~Wang, H.~Wu, Q.~She, et~al., Dureader: a chinese machine reading comprehension dataset from real-world applications, arXiv preprint arXiv:1711.05073 (2017).

\bibitem{tang2020dureaderrobust}
H.~Tang, J.~Liu, H.~Li, Y.~Hong, H.~Wu, H.~Wang, Dureaderrobust: A chinese dataset towards evaluating the robustness of machine reading comprehension models, arXiv preprint arXiv:2004.11142 (2020).

\bibitem{welbl2017crowdsourcing}
J.~Welbl, N.~F. Liu, M.~Gardner, Crowdsourcing multiple choice science questions, arXiv preprint arXiv:1707.06209 (2017).

\bibitem{xiong2017end}
C.~Xiong, Z.~Dai, J.~Callan, Z.~Liu, R.~Power, End-to-end neural ad-hoc ranking with kernel pooling, in: Proceedings of the 40th International ACM SIGIR conference on research and development in information retrieval, 2017, pp. 55--64.

\bibitem{penas2013qa4mre}
A.~Pe{\~n}as, E.~Hovy, P.~Forner, {\'A}.~Rodrigo, R.~Sutcliffe, R.~Morante, Qa4mre 2011-2013: Overview of question answering for machine reading evaluation, in: Information Access Evaluation. Multilinguality, Multimodality, and Visualization: 4th International Conference of the CLEF Initiative, CLEF 2013, Valencia, Spain, September 23-26, 2013. Proceedings 4, Springer, 2013, pp. 303--320.

\bibitem{lim2019korquad1}
S.~Lim, M.~Kim, J.~Lee, Korquad1. 0: Korean qa dataset for machine reading comprehension, arXiv preprint arXiv:1909.07005 (2019).

\bibitem{xiao2018cail2018}
C.~Xiao, H.~Zhong, Z.~Guo, C.~Tu, Z.~Liu, M.~Sun, Y.~Feng, X.~Han, Z.~Hu, H.~Wang, et~al., Cail2018: A large-scale legal dataset for judgment prediction, arXiv preprint arXiv:1807.02478 (2018).

\bibitem{hendrycks2021measuring}
D.~Hendrycks, S.~Basart, S.~Kadavath, M.~Mazeika, A.~Arora, E.~Guo, C.~Burns, S.~Puranik, H.~He, D.~Song, et~al., Measuring coding challenge competence with apps, arXiv preprint arXiv:2105.09938 (2021).

\bibitem{wang2017deep}
Y.~Wang, X.~Liu, S.~Shi, Deep neural solver for math word problems, in: Proceedings of the 2017 conference on empirical methods in natural language processing, 2017, pp. 845--854.

\bibitem{cobbe2021training}
K.~Cobbe, V.~Kosaraju, M.~Bavarian, M.~Chen, H.~Jun, L.~Kaiser, M.~Plappert, J.~Tworek, J.~Hilton, R.~Nakano, et~al., Training verifiers to solve math word problems, arXiv preprint arXiv:2110.14168 (2021).

\bibitem{MathQA}
J.~Austin, A.~Odena, M.~I. Nye, M.~Bosma, H.~Michalewski, D.~Dohan, E.~Jiang, C.~J. Cai, M.~Terry, Q.~V. Le, C.~Sutton, Program synthesis with large language models, CoRR abs/2108.07732 (2021).

\bibitem{shi2022language}
F.~Shi, M.~Suzgun, M.~Freitag, X.~Wang, S.~Srivats, S.~Vosoughi, H.~W. Chung, Y.~Tay, S.~Ruder, D.~Zhou, et~al., Language models are multilingual chain-of-thought reasoners, arXiv preprint arXiv:2210.03057 (2022).

\bibitem{roy2016solving}
S.~Roy, D.~Roth, Solving general arithmetic word problems, arXiv preprint arXiv:1608.01413 (2016).

\bibitem{ASDiv}
S.-Y. Miao, C.-C. Liang, K.-Y. Su, A diverse corpus for evaluating and developing english math word problem solvers, arXiv preprint arXiv:2106.15772 (2021).

\bibitem{koncel2016mawps}
R.~Koncel-Kedziorski, S.~Roy, A.~Amini, N.~Kushman, H.~Hajishirzi, Mawps: A math word problem repository, in: Proceedings of the 2016 conference of the north american chapter of the association for computational linguistics: human language technologies, 2016, pp. 1152--1157.

\bibitem{patel2021nlp}
A.~Patel, S.~Bhattamishra, N.~Goyal, Are nlp models really able to solve simple math word problems?, arXiv preprint arXiv:2103.07191 (2021).

\bibitem{ds_1000}
Y.~Lai, C.~Li, Y.~Wang, T.~Zhang, R.~Zhong, L.~Zettlemoyer, W.-t. Yih, D.~Fried, S.~Wang, T.~Yu, Ds-1000: A natural and reliable benchmark for data science code generation, in: International Conference on Machine Learning, PMLR, 2023, pp. 18319--18345.

\bibitem{austin2021program}
J.~Austin, A.~Odena, M.~Nye, M.~Bosma, H.~Michalewski, D.~Dohan, E.~Jiang, C.~Cai, M.~Terry, Q.~Le, et~al., Program synthesis with large language models, arXiv preprint arXiv:2108.07732 (2021).

\bibitem{nie2019adversarial}
Y.~Nie, A.~Williams, E.~Dinan, M.~Bansal, J.~Weston, D.~Kiela, Adversarial nli: A new benchmark for natural language understanding, arXiv preprint arXiv:1910.14599 (2019).

\bibitem{williams2017broad}
A.~Williams, N.~Nangia, S.~R. Bowman, A broad-coverage challenge corpus for sentence understanding through inference, arXiv preprint arXiv:1704.05426 (2017).

\bibitem{mccoy2019right}
R.~T. McCoy, E.~Pavlick, T.~Linzen, Right for the wrong reasons: Diagnosing syntactic heuristics in natural language inference, arXiv preprint arXiv:1902.01007 (2019).

\bibitem{liu2020logiqa}
J.~Liu, L.~Cui, H.~Liu, D.~Huang, Y.~Wang, Y.~Zhang, Logiqa: A challenge dataset for machine reading comprehension with logical reasoning, arXiv preprint arXiv:2007.08124 (2020).

\bibitem{lewis2019mlqa}
P.~Lewis, B.~O{\u{g}}uz, R.~Rinott, S.~Riedel, H.~Schwenk, Mlqa: Evaluating cross-lingual extractive question answering, arXiv preprint arXiv:1910.07475 (2019).

\bibitem{conneau2018xnli}
A.~Conneau, G.~Lample, R.~Rinott, A.~Williams, S.~R. Bowman, H.~Schwenk, V.~Stoyanov, Xnli: Evaluating cross-lingual sentence representations, arXiv preprint arXiv:1809.05053 (2018).

\bibitem{yang2019paws}
Y.~Yang, Y.~Zhang, C.~Tar, J.~Baldridge, Paws-x: A cross-lingual adversarial dataset for paraphrase identification, arXiv preprint arXiv:1908.11828 (2019).

\bibitem{narayan1808don}
S.~Narayan, S.~B. Cohen, M.~Lapata, Don’t give me the details, just the summary!, Topic-Aware Convolutional Neural Networks for Extreme Summarization. ArXiv, abs (1808).

\bibitem{ponti2020xcopa}
E.~M. Ponti, G.~Glava{\v{s}}, O.~Majewska, Q.~Liu, I.~Vuli{\'c}, A.~Korhonen, Xcopa: A multilingual dataset for causal commonsense reasoning, arXiv preprint arXiv:2005.00333 (2020).

\bibitem{tikhonov2021s}
A.~Tikhonov, M.~Ryabinin, It's all in the heads: Using attention heads as a baseline for cross-lingual transfer in commonsense reasoning, arXiv preprint arXiv:2106.12066 (2021).

\bibitem{clark2020tydi}
J.~H. Clark, E.~Choi, M.~Collins, D.~Garrette, T.~Kwiatkowski, V.~Nikolaev, J.~Palomaki, Tydi qa: A benchmark for information-seeking question answering in typologically diverse languages, Transactions of the Association for Computational Linguistics 8 (2020) 454--470.

\bibitem{scialom2020mlsum}
T.~Scialom, P.-A. Dray, S.~Lamprier, B.~Piwowarski, J.~Staiano, Mlsum: The multilingual summarization corpus, arXiv preprint arXiv:2004.14900 (2020).

\bibitem{lin2021truthfulqa}
S.~Lin, J.~Hilton, O.~Evans, Truthfulqa: Measuring how models mimic human falsehoods, arXiv preprint arXiv:2109.07958 (2021).

\bibitem{augenstein2019multifc}
I.~Augenstein, C.~Lioma, D.~Wang, L.~C. Lima, C.~Hansen, C.~Hansen, J.~G. Simonsen, Multifc: A real-world multi-domain dataset for evidence-based fact checking of claims, arXiv preprint arXiv:1909.03242 (2019).

\bibitem{thorne2018fever}
J.~Thorne, A.~Vlachos, C.~Christodoulopoulos, A.~Mittal, Fever: a large-scale dataset for fact extraction and verification, arXiv preprint arXiv:1803.05355 (2018).

\bibitem{mollas2020ethos}
I.~Mollas, Z.~Chrysopoulou, S.~Karlos, G.~Tsoumakas, Ethos: an online hate speech detection dataset, arXiv preprint arXiv:2006.08328 (2020).

\bibitem{nadeem2020stereoset}
M.~Nadeem, A.~Bethke, S.~Reddy, Stereoset: Measuring stereotypical bias in pretrained language models, arXiv preprint arXiv:2004.09456 (2020).

\bibitem{parrish2021bbq}
A.~Parrish, A.~Chen, N.~Nangia, V.~Padmakumar, J.~Phang, J.~Thompson, P.~M. Htut, S.~R. Bowman, Bbq: A hand-built bias benchmark for question answering, arXiv preprint arXiv:2110.08193 (2021).

\bibitem{zhao2018gender}
J.~Zhao, T.~Wang, M.~Yatskar, V.~Ordonez, K.-W. Chang, Gender bias in coreference resolution: Evaluation and debiasing methods, arXiv preprint arXiv:1804.06876 (2018).

\bibitem{nangia2020crows}
N.~Nangia, C.~Vania, R.~Bhalerao, S.~R. Bowman, Crows-pairs: A challenge dataset for measuring social biases in masked language models, arXiv preprint arXiv:2010.00133 (2020).

\bibitem{gehman2020realtoxicityprompts}
S.~Gehman, S.~Gururangan, M.~Sap, Y.~Choi, N.~A. Smith, Realtoxicityprompts: Evaluating neural toxic degeneration in language models, arXiv preprint arXiv:2009.11462 (2020).

\bibitem{borkan2019nuanced}
D.~Borkan, L.~Dixon, J.~Sorensen, N.~Thain, L.~Vasserman, Nuanced metrics for measuring unintended bias with real data for text classification, in: Companion proceedings of the 2019 world wide web conference, 2019, pp. 491--500.

\bibitem{bojar2016findings}
O.~Bojar, R.~Chatterjee, C.~Federmann, Y.~Graham, B.~Haddow, M.~Huck, A.~J. Yepes, P.~Koehn, V.~Logacheva, C.~Monz, et~al., Findings of the 2016 conference on machine translation, in: Proceedings of the First Conference on Machine Translation: Volume 2, Shared Task Papers, 2016, pp. 131--198.

\bibitem{loic2020findings}
B.~Lo{\"\i}c, B.~Magdalena, B.~Ond{\v{r}}ej, F.~Christian, G.~Yvette, G.~Roman, H.~Barry, H.~Matthias, J.~Eric, K.~Tom, et~al., Findings of the 2020 conference on machine translation (wmt20), in: Proceedings of the Fifth Conference on Machine Translation, Association for Computational Linguistics,, 2020, pp. 1--55.

\bibitem{li2021ccpm}
W.~Li, F.~Qi, M.~Sun, X.~Yi, J.~Zhang, Ccpm: A chinese classical poetry matching dataset, arXiv preprint arXiv:2106.01979 (2021).

\bibitem{dinan2018wizard}
E.~Dinan, S.~Roller, K.~Shuster, A.~Fan, M.~Auli, J.~Weston, Wizard of wikipedia: Knowledge-powered conversational agents, arXiv preprint arXiv:1811.01241 (2018).

\bibitem{rashkin2018towards}
H.~Rashkin, E.~M. Smith, M.~Li, Y.-L. Boureau, Towards empathetic open-domain conversation models: A new benchmark and dataset, arXiv preprint arXiv:1811.00207 (2018).

\bibitem{dinan2020second}
E.~Dinan, V.~Logacheva, V.~Malykh, A.~Miller, K.~Shuster, J.~Urbanek, D.~Kiela, A.~Szlam, I.~Serban, R.~Lowe, et~al., The second conversational intelligence challenge (convai2), in: The NeurIPS'18 Competition: From Machine Learning to Intelligent Conversations, Springer, 2020, pp. 187--208.

\bibitem{zhou2020kdconv}
H.~Zhou, C.~Zheng, K.~Huang, M.~Huang, X.~Zhu, Kdconv: A chinese multi-domain dialogue dataset towards multi-turn knowledge-driven conversation, arXiv preprint arXiv:2004.04100 (2020).

\bibitem{co2019iflytek}
L.~CO, Iflytek: a multiple categories chinese text classifier. competition official website (2019).

\bibitem{baumgartner2020pushshift}
J.~Baumgartner, S.~Zannettou, B.~Keegan, M.~Squire, J.~Blackburn, The pushshift reddit dataset, in: Proceedings of the international AAAI conference on web and social media, Vol.~14, 2020, pp. 830--839.

\bibitem{fan2019eli5}
A.~Fan, Y.~Jernite, E.~Perez, D.~Grangier, J.~Weston, M.~Auli, Eli5: Long form question answering, arXiv preprint arXiv:1907.09190 (2019).

\bibitem{wang2022benchmarking}
Y.~Wang, S.~Mishra, P.~Alipoormolabashi, Y.~Kordi, A.~Mirzaei, A.~Arunkumar, A.~Ashok, A.~S. Dhanasekaran, A.~Naik, D.~Stap, et~al., Benchmarking generalization via in-context instructions on 1,600+ language tasks, arXiv preprint arXiv:2204.07705 (2022).

\bibitem{xie2022unifiedskg}
T.~Xie, C.~H. Wu, P.~Shi, R.~Zhong, T.~Scholak, M.~Yasunaga, C.-S. Wu, M.~Zhong, P.~Yin, S.~I. Wang, et~al., Unifiedskg: Unifying and multi-tasking structured knowledge grounding with text-to-text language models, arXiv preprint arXiv:2201.05966 (2022).

\bibitem{ye2021crossfit}
Q.~Ye, B.~Y. Lin, X.~Ren, Crossfit: A few-shot learning challenge for cross-task generalization in nlp, arXiv preprint arXiv:2104.08835 (2021).

\bibitem{aribandi2021ext5}
V.~Aribandi, Y.~Tay, T.~Schuster, J.~Rao, H.~S. Zheng, S.~V. Mehta, H.~Zhuang, V.~Q. Tran, D.~Bahri, J.~Ni, et~al., Ext5: Towards extreme multi-task scaling for transfer learning, arXiv preprint arXiv:2111.10952 (2021).

\bibitem{williams2018multinli}
A.~Williams, N.~Nangia, S.~Bowman, \href{https://aclanthology.org/N18-1101}{A broad-coverage challenge corpus for sentence understanding through inference}, in: Proceedings of the 2018 Conference of the North {A}merican Chapter of the Association for Computational Linguistics: Human Language Technologies, Volume 1 (Long Papers), Association for Computational Linguistics, New Orleans, Louisiana, 2018, pp. 1112--1122.
\newblock \href {https://doi.org/10.18653/v1/N18-1101} {\path{doi:10.18653/v1/N18-1101}}.
\newline\urlprefix\url{https://aclanthology.org/N18-1101}

\bibitem{zhang2019pawss}
Y.~Zhang, J.~Baldridge, L.~He, \href{https://aclanthology.org/N19-1131}{{PAWS}: Paraphrase adversaries from word scrambling}, in: Proceedings of the 2019 Conference of the North {A}merican Chapter of the Association for Computational Linguistics: Human Language Technologies, Volume 1 (Long and Short Papers), Association for Computational Linguistics, Minneapolis, Minnesota, 2019, pp. 1298--1308.
\newblock \href {https://doi.org/10.18653/v1/N19-1131} {\path{doi:10.18653/v1/N19-1131}}.
\newline\urlprefix\url{https://aclanthology.org/N19-1131}

\bibitem{qin2023is}
C.~Qin, A.~Zhang, Z.~Zhang, J.~Chen, M.~Yasunaga, D.~Yang, \href{https://openreview.net/forum?id=u03xn1COsO}{Is chat{GPT} a general-purpose natural language processing task solver?}, in: The 2023 Conference on Empirical Methods in Natural Language Processing, 2023.
\newline\urlprefix\url{https://openreview.net/forum?id=u03xn1COsO}

\bibitem{hadi2023large}
M.~U. Hadi, R.~Qureshi, A.~Shah, M.~Irfan, A.~Zafar, M.~B. Shaikh, N.~Akhtar, J.~Wu, S.~Mirjalili, et~al., Large language models: a comprehensive survey of its applications, challenges, limitations, and future prospects, TechRxiv (2023).

\bibitem{dong2023towards}
X.~L. Dong, S.~Moon, Y.~E. Xu, K.~Malik, Z.~Yu, Towards next-generation intelligent assistants leveraging llm techniques, in: Proceedings of the 29th ACM SIGKDD Conference on Knowledge Discovery and Data Mining, 2023, pp. 5792--5793.

\bibitem{pandya2023automating}
K.~Pandya, M.~Holia, Automating customer service using langchain: Building custom open-source gpt chatbot for organizations, arXiv preprint arXiv:2310.05421 (2023).

\bibitem{li2023can}
J.~Li, B.~Hui, G.~Qu, B.~Li, J.~Yang, B.~Li, B.~Wang, B.~Qin, R.~Cao, R.~Geng, et~al., Can llm already serve as a database interface? a big bench for large-scale database grounded text-to-sqls, arXiv preprint arXiv:2305.03111 (2023).

\bibitem{rao2023evaluating}
A.~Rao, J.~Kim, M.~Kamineni, M.~Pang, W.~Lie, M.~D. Succi, Evaluating chatgpt as an adjunct for radiologic decision-making, medRxiv (2023) 2023--02.

\bibitem{benary2023leveraging}
M.~Benary, X.~D. Wang, M.~Schmidt, D.~Soll, G.~Hilfenhaus, M.~Nassir, C.~Sigler, M.~Kn{\"o}dler, U.~Keller, D.~Beule, et~al., Leveraging large language models for decision support in personalized oncology, JAMA Network Open 6~(11) (2023) e2343689--e2343689.

\bibitem{chiesa2023exploring}
C.~M. Chiesa-Estomba, J.~R. Lechien, L.~A. Vaira, A.~Brunet, G.~Cammaroto, M.~Mayo-Yanez, A.~Sanchez-Barrueco, C.~Saga-Gutierrez, Exploring the potential of chat-gpt as a supportive tool for sialendoscopy clinical decision making and patient information support, European Archives of Oto-Rhino-Laryngology (2023) 1--6.

\bibitem{montagna2023data}
S.~Montagna, S.~Ferretti, L.~C. Klopfenstein, A.~Florio, M.~F. Pengo, Data decentralisation of llm-based chatbot systems in chronic disease self-management, in: Proceedings of the 2023 ACM Conference on Information Technology for Social Good, 2023, pp. 205--212.

\bibitem{bill2023fine}
D.~Bill, T.~Eriksson, Fine-tuning a llm using reinforcement learning from human feedback for a therapy chatbot application (2023).

\bibitem{abbasian2023conversational}
M.~Abbasian, I.~Azimi, A.~M. Rahmani, R.~Jain, Conversational health agents: A personalized llm-powered agent framework, arXiv preprint arXiv:2310.02374 (2023).

\bibitem{lemley2023does}
K.~V. Lemley, Does chatgpt help us understand the medical literature?, Journal of the American Society of Nephrology (2023) 10--1681.

\bibitem{pal2023domain}
S.~Pal, M.~Bhattacharya, S.-S. Lee, C.~Chakraborty, A domain-specific next-generation large language model (llm) or chatgpt is required for biomedical engineering and research, Annals of Biomedical Engineering (2023) 1--4.

\bibitem{du2023calla}
Y.~Du, S.~Zhao, Y.~Chen, R.~Bai, J.~Liu, H.~Wu, H.~Wang, B.~Qin, The calla dataset: Probing llms' interactive knowledge acquisition from chinese medical literature, arXiv preprint arXiv:2309.04198 (2023).

\bibitem{abd2023large}
A.~Abd-Alrazaq, R.~AlSaad, D.~Alhuwail, A.~Ahmed, P.~M. Healy, S.~Latifi, S.~Aziz, R.~Damseh, S.~A. Alrazak, J.~Sheikh, et~al., Large language models in medical education: Opportunities, challenges, and future directions, JMIR Medical Education 9~(1) (2023) e48291.

\bibitem{mbakwe2023chatgpt}
A.~B. Mbakwe, I.~Lourentzou, L.~A. Celi, O.~J. Mechanic, A.~Dagan, Chatgpt passing usmle shines a spotlight on the flaws of medical education (2023).

\bibitem{ahn2023impending}
S.~Ahn, The impending impacts of large language models on medical education, Korean Journal of Medical Education 35~(1) (2023) 103.

\bibitem{waisberg2023large}
E.~Waisberg, J.~Ong, M.~Masalkhi, A.~G. Lee, Large language model (llm)-driven chatbots for neuro-ophthalmic medical education, Eye (2023) 1--3.

\bibitem{deiana2023artificial}
G.~Deiana, M.~Dettori, A.~Arghittu, A.~Azara, G.~Gabutti, P.~Castiglia, Artificial intelligence and public health: Evaluating chatgpt responses to vaccination myths and misconceptions, Vaccines 11~(7) (2023) 1217.

\bibitem{de2023chatgpt}
L.~De~Angelis, F.~Baglivo, G.~Arzilli, G.~P. Privitera, P.~Ferragina, A.~E. Tozzi, C.~Rizzo, Chatgpt and the rise of large language models: the new ai-driven infodemic threat in public health, Frontiers in Public Health 11 (2023) 1166120.

\bibitem{rane2023contribution}
N.~L. Rane, A.~Tawde, S.~P. Choudhary, J.~Rane, Contribution and performance of chatgpt and other large language models (llm) for scientific and research advancements: a double-edged sword, International Research Journal of Modernization in Engineering Technology and Science 5~(10) (2023) 875--899.

\bibitem{dai2023can}
W.~Dai, J.~Lin, H.~Jin, T.~Li, Y.-S. Tsai, D.~Ga{\v{s}}evi{\'c}, G.~Chen, Can large language models provide feedback to students? a case study on chatgpt, in: 2023 IEEE International Conference on Advanced Learning Technologies (ICALT), IEEE, 2023, pp. 323--325.

\bibitem{kasneci2023chatgpt}
E.~Kasneci, K.~Se{\ss}ler, S.~K{\"u}chemann, M.~Bannert, D.~Dementieva, F.~Fischer, U.~Gasser, G.~Groh, S.~G{\"u}nnemann, E.~H{\"u}llermeier, et~al., Chatgpt for good? on opportunities and challenges of large language models for education, Learning and individual differences 103 (2023) 102274.

\bibitem{rane2023enhancing}
N.~Rane, Enhancing the quality of teaching and learning through chatgpt and similar large language models: Challenges, future prospects, and ethical considerations in education, Future Prospects, and Ethical Considerations in Education (September 15, 2023) (2023).

\bibitem{young2023investigating}
J.~C. Young, M.~Shishido, Investigating openai’s chatgpt potentials in generating chatbot's dialogue for english as a foreign language learning, International Journal of Advanced Computer Science and Applications 14~(6) (2023).

\bibitem{irons2023exploring}
J.~Irons, C.~Mason, P.~Cooper, S.~Sidra, A.~Reeson, C.~Paris, Exploring the impacts of chatgpt on future scientific work, SocArXiv (2023).

\bibitem{schmidt2023using}
P.~G. Schmidt, A.~J. Meir, Using generative ai for literature searches and scholarly writing: Is the integrity of the scientific discourse in jeopardy?, arXiv preprint arXiv:2311.06981 (2023).

\bibitem{zheng2023large}
Y.~Zheng, H.~Y. Koh, J.~Ju, A.~T. Nguyen, L.~T. May, G.~I. Webb, S.~Pan, Large language models for scientific synthesis, inference and explanation, arXiv preprint arXiv:2310.07984 (2023).

\bibitem{aczel2023transparency}
B.~Aczel, E.-J. Wagenmakers, Transparency guidance for chatgpt usage in scientific writing, PsyArXiv (2023).

\bibitem{altmae2023artificial}
S.~Altm{\"a}e, A.~Sola-Leyva, A.~Salumets, Artificial intelligence in scientific writing: a friend or a foe?, Reproductive BioMedicine Online (2023).

\bibitem{imani2023mathprompter}
S.~Imani, L.~Du, H.~Shrivastava, Mathprompter: Mathematical reasoning using large language models, arXiv preprint arXiv:2303.05398 (2023).

\bibitem{yuan2023scaling}
Z.~Yuan, H.~Yuan, C.~Li, G.~Dong, C.~Tan, C.~Zhou, Scaling relationship on learning mathematical reasoning with large language models, arXiv preprint arXiv:2308.01825 (2023).

\bibitem{yang2023leandojo}
K.~Yang, A.~M. Swope, A.~Gu, R.~Chalamala, P.~Song, S.~Yu, S.~Godil, R.~Prenger, A.~Anandkumar, Leandojo: Theorem proving with retrieval-augmented language models, arXiv preprint arXiv:2306.15626 (2023).

\bibitem{collins2023evaluating}
K.~M. Collins, A.~Q. Jiang, S.~Frieder, L.~Wong, M.~Zilka, U.~Bhatt, T.~Lukasiewicz, Y.~Wu, J.~B. Tenenbaum, W.~Hart, et~al., Evaluating language models for mathematics through interactions, arXiv preprint arXiv:2306.01694 (2023).

\bibitem{liu2023summary}
Y.~Liu, T.~Han, S.~Ma, J.~Zhang, Y.~Yang, J.~Tian, H.~He, A.~Li, M.~He, Z.~Liu, et~al., Summary of chatgpt-related research and perspective towards the future of large language models, Meta-Radiology (2023) 100017.

\bibitem{drapal2023using}
J.~Dr{\'a}pal, H.~Westermann, J.~Savelka, Using large language models to support thematic analysis in empirical legal studies, arXiv preprint arXiv:2310.18729 (2023).

\bibitem{savelka2023explaining}
J.~Savelka, K.~D. Ashley, M.~A. Gray, H.~Westermann, H.~Xu, Explaining legal concepts with augmented large language models (gpt-4), arXiv preprint arXiv:2306.09525 (2023).

\bibitem{guha2023legalbench}
N.~Guha, J.~Nyarko, D.~E. Ho, C.~R{\'e}, A.~Chilton, A.~Narayana, A.~Chohlas-Wood, A.~Peters, B.~Waldon, D.~N. Rockmore, et~al., Legalbench: A collaboratively built benchmark for measuring legal reasoning in large language models, arXiv preprint arXiv:2308.11462 (2023).

\bibitem{cui2023chatlaw}
J.~Cui, Z.~Li, Y.~Yan, B.~Chen, L.~Yuan, Chatlaw: Open-source legal large language model with integrated external knowledge bases, arXiv preprint arXiv:2306.16092 (2023).

\bibitem{yang2023fingpt}
H.~Yang, X.-Y. Liu, C.~D. Wang, Fingpt: Open-source financial large language models, arXiv preprint arXiv:2306.06031 (2023).

\bibitem{li2023large}
Y.~Li, S.~Wang, H.~Ding, H.~Chen, Large language models in finance: A survey, in: Proceedings of the Fourth ACM International Conference on AI in Finance, 2023, pp. 374--382.

\bibitem{lykov2023llm}
A.~Lykov, D.~Tsetserukou, Llm-brain: Ai-driven fast generation of robot behaviour tree based on large language model, arXiv preprint arXiv:2305.19352 (2023).

\bibitem{billing2023language}
E.~Billing, J.~Ros{\'e}n, M.~Lamb, Language models for human-robot interaction, in: ACM/IEEE International Conference on Human-Robot Interaction, March 13--16, 2023, Stockholm, Sweden, ACM Digital Library, 2023, pp. 905--906.

\bibitem{ye2023improved}
Y.~Ye, H.~You, J.~Du, Improved trust in human-robot collaboration with chatgpt, IEEE Access (2023).

\bibitem{ding2023leveraging}
Y.~Ding, X.~Zhang, C.~Paxton, S.~Zhang, Leveraging commonsense knowledge from large language models for task and motion planning, in: RSS 2023 Workshop on Learning for Task and Motion Planning, 2023.

\bibitem{wu2023tidybot}
J.~Wu, R.~Antonova, A.~Kan, M.~Lepert, A.~Zeng, S.~Song, J.~Bohg, S.~Rusinkiewicz, T.~Funkhouser, Tidybot: Personalized robot assistance with large language models, arXiv preprint arXiv:2305.05658 (2023).

\bibitem{strubell2019energy}
E.~Strubell, A.~Ganesh, A.~McCallum, Energy and policy considerations for deep learning in nlp, arXiv preprint arXiv:1906.02243 (2019).

\bibitem{bender2021dangers}
E.~M. Bender, T.~Gebru, A.~McMillan-Major, S.~Shmitchell, On the dangers of stochastic parrots: Can language models be too big?, in: Proceedings of the 2021 ACM conference on fairness, accountability, and transparency, 2021, pp. 610--623.

\bibitem{zhang2021understanding}
C.~Zhang, S.~Bengio, M.~Hardt, B.~Recht, O.~Vinyals, Understanding deep learning (still) requires rethinking generalization, Communications of the ACM 64~(3) (2021) 107--115.

\bibitem{tanzer2021memorisation}
M.~T{\"a}nzer, S.~Ruder, M.~Rei, Memorisation versus generalisation in pre-trained language models, arXiv preprint arXiv:2105.00828 (2021).

\bibitem{west2019discriminating}
S.~M. West, M.~Whittaker, K.~Crawford, Discriminating systems, AI Now (2019) 1--33.

\bibitem{valmeekam2022large}
K.~Valmeekam, A.~Olmo, S.~Sreedharan, S.~Kambhampati, Large language models still can't plan (a benchmark for llms on planning and reasoning about change), arXiv preprint arXiv:2206.10498 (2022).

\bibitem{zhang2023siren}
Y.~Zhang, Y.~Li, L.~Cui, D.~Cai, L.~Liu, T.~Fu, X.~Huang, E.~Zhao, Y.~Zhang, Y.~Chen, et~al., Siren's song in the ai ocean: A survey on hallucination in large language models, arXiv preprint arXiv:2309.01219 (2023).

\bibitem{webson2021prompt}
A.~Webson, E.~Pavlick, Do prompt-based models really understand the meaning of their prompts?, arXiv preprint arXiv:2109.01247 (2021).

\bibitem{shaikh2022second}
O.~Shaikh, H.~Zhang, W.~Held, M.~Bernstein, D.~Yang, On second thought, let's not think step by step! bias and toxicity in zero-shot reasoning, arXiv preprint arXiv:2212.08061 (2022).

\bibitem{security_and_privacy}
B.~C. Das, M.~H. Amini, Y.~Wu, Security and privacy challenges of large language models: A survey, arXiv preprint arXiv:2402.00888 (2024).

\bibitem{liu2020adversarial}
X.~Liu, H.~Cheng, P.~He, W.~Chen, Y.~Wang, H.~Poon, J.~Gao, \href{https://www.microsoft.com/en-us/research/publication/adversarial-training-for-large-neural-language-models/}{Adversarial training for large neural language models}, ArXiv (April 2020).
\newline\urlprefix\url{https://www.microsoft.com/en-us/research/publication/adversarial-training-for-large-neural-language-models/}

\bibitem{shayegani2023survey}
E.~Shayegani, M.~A.~A. Mamun, Y.~Fu, P.~Zaree, Y.~Dong, N.~Abu-Ghazaleh, Survey of vulnerabilities in large language models revealed by adversarial attacks (2023).
\newblock \href {http://arxiv.org/abs/2310.10844} {\path{arXiv:2310.10844}}.

\bibitem{xu2023llm}
X.~Xu, K.~Kong, N.~Liu, L.~Cui, D.~Wang, J.~Zhang, M.~Kankanhalli, An llm can fool itself: A prompt-based adversarial attack (2023).
\newblock \href {http://arxiv.org/abs/2310.13345} {\path{arXiv:2310.13345}}.

\bibitem{zhao2023explainability}
H.~Zhao, H.~Chen, F.~Yang, N.~Liu, H.~Deng, H.~Cai, S.~Wang, D.~Yin, M.~Du, Explainability for large language models: A survey (2023).
\newblock \href {http://arxiv.org/abs/2309.01029} {\path{arXiv:2309.01029}}.

\bibitem{huang2023large}
S.~Huang, S.~Mamidanna, S.~Jangam, Y.~Zhou, L.~H. Gilpin, Can large language models explain themselves? a study of llm-generated self-explanations (2023).
\newblock \href {http://arxiv.org/abs/2310.11207} {\path{arXiv:2310.11207}}.

\bibitem{brown2022does}
H.~Brown, K.~Lee, F.~Mireshghallah, R.~Shokri, F.~Tram{\`e}r, What does it mean for a language model to preserve privacy?, in: Proceedings of the 2022 ACM Conference on Fairness, Accountability, and Transparency, 2022, pp. 2280--2292.

\bibitem{plant2022you}
R.~Plant, V.~Giuffrida, D.~Gkatzia, You are what you write: Preserving privacy in the era of large language models, arXiv preprint arXiv:2204.09391 (2022).

\bibitem{niu2020realtime}
W.~Niu, Z.~Kong, G.~Yuan, W.~Jiang, J.~Guan, C.~Ding, P.~Zhao, S.~Liu, B.~Ren, Y.~Wang, Real-time execution of large-scale language models on mobile (2020).
\newblock \href {http://arxiv.org/abs/2009.06823} {\path{arXiv:2009.06823}}.

\bibitem{guo2023olive}
C.~Guo, J.~Tang, W.~Hu, J.~Leng, C.~Zhang, F.~Yang, Y.~Liu, M.~Guo, Y.~Zhu, Olive: Accelerating large language models via hardware-friendly outlier-victim pair quantization, in: Proceedings of the 50th Annual International Symposium on Computer Architecture, 2023, pp. 1--15.

\bibitem{mesko2023imperative}
B.~Mesk{\'o}, E.~J. Topol, The imperative for regulatory oversight of large language models (or generative ai) in healthcare, npj Digital Medicine 6~(1) (2023) 120.

\bibitem{zhang2023ethical}
J.~Zhang, X.~Ji, Z.~Zhao, X.~Hei, K.-K.~R. Choo, Ethical considerations and policy implications for large language models: Guiding responsible development and deployment, arXiv preprint arXiv:2308.02678 (2023).

\bibitem{mokander2023auditing}
J.~M{\"o}kander, J.~Schuett, H.~R. Kirk, L.~Floridi, Auditing large language models: a three-layered approach, AI and Ethics (2023) 1--31.

\end{thebibliography}

\end{document}